%% file: main.tex
\documentclass[thesis]{mythesis}[2018/05/21]

\input{tex/frontmatter}
\input{packages}

\input{macros}

\chaptertitlefont{\normalfont\sffamily\Huge}

\begin{document}

\include{tex/intro}
\include{tex/paper_gnome}

\include{tex/paper_alps}
\include{tex/paper_caesar}

\include{tex/paper_grab}
\include{tex/paper_tar}

\include{tex/related}

\include{tex/conclusion}

\small
\bibliographystyle{plain}
\bibliography{reference}

\end{document}

%% file: tex/frontmatter.tex
\title{Efficient Pipelines for Vision-Based Context Sensing}

\author{Xiaochen Liu}

\hidecommitteepage 

\department{Mechanical and Aerospace Engineering}

\phdgrad{May, 2020}
\defensedate{May, 2020}
\acknowledgments{
I want to first express my greatest appreciation to my advisor, Professor Ramesh Govindan, for his support along my path pursuing the degree. He is very insightful in the domain of computer systems and networking, which helps me grow fast in doing research. The first two years was pretty tough to me because of lack of a big picture for the thesis and the challenging courses. Prof. Govindan helped me out by encouraging me to explore different ideas and giving me very useful feedback. Later on, as things rolled smoother, his high standard in research keeps me improving and the lessons learnt during the collaboration with him will benefit my whole career. He is the best advisor that I could ever imagine for my PhD life. None of the projects in this thesis could exist without his generous guidance. I would also thank my dissertation committee, Professor Barath Raghavan and Professor Bhaskar Krishnamachari, for their insightful comments and suggestions on improving the quality of the dissertation. 

Throughout my dissertation, it has been a great honor to work with many outstanding professors and researchers in top universities and labs: Gnome (2) and ALPS (3) are joint work with Suman Nath at Microsoft Research. Caesar (4) is a joint work with Oytun Ulutan and Professor B. S. Manjunath at University of California, Santa Barbara, and Kevin Chan from ARL. Grab (5) and TAR (6) are joint work with Yurong Jiang, Puneet Jain, and Kyu-Han Kim during my internship at HP Labs. I would also like to thank my fellow labmates Yitao Hu and Pradipta Ghosh, for the joint contribution to ALPS and Caesar. 

Last but not least, I want to thank my family and friends for their support these years: specially my beloved wife Olivia, my parents Hong and Songyan, and my grandparents Zuoxu, Hanjie, Baoling, and Ensheng.
}

\showabstract
\showacknowledgments

\showlistoffigures
\showlistoftables
\abstract{
Context awareness is an essential part of mobile and ubiquitous computing. Its goal is to unveil situational information about mobile users like locations and activities. The sensed context can enable many services like navigation, AR, and smarting shopping. Such context can be sensed in different ways including visual sensors. There  is  an  emergence  of  vision  sources  deployed  worldwide. The cameras could be installed on roadside, in-house, and on mobile platforms. This trend provides huge amount of vision data that could be used for context sensing. However, the vision data collection and analytics are still highly manual today. It is hard to deploy cameras at large scale for data collection. Organizing and labeling context from the data are also labor intensive. In recent years, advanced vision algorithms and deep neural networks are used to help analyze vision data. But this approach is limited by data quality, labeling effort, and dependency on hardware resources. In summary, there are three major challenges for today's vision-based context sensing systems: data collection and labeling at large scale, process large data volumes efficiently with limited hardware resources, and extract accurate context out of vision data. 

The thesis explores the design space that consists of three dimensions: sensing task, sensor types, and task locations. Our prior work explores several points in this design space. Specifically, we develop Gnome~\cite{liu2018gnome} for accurate outdoor localization. It leverages 2D and 3D information from Google Street View to mitigate GPS signal's error in different cities, and uses offloading and caching to work efficiently on stock phones in real time. For vision-only outdoor sensing, we present ALPS~\cite{hu2016alps} that applies optimized object detection algorithms to detects and localizes roadside objects captured in Google Street View. It optimizes the processing speed by adaptively downloading and processing images. For indoor scenarios, we designed TAR~\cite{liu2018tar} and Grab~\cite{grab} for tracking people and detecting their interactions with objects. Both projects fuse camera outputs with those of other sensors to achieve state-of-the-art performance. For outdoor behavior sensing, we develop Caesar~\cite{liu2019caesar} to abstract complex activities as graphs, and define customized activities using extensible vocabulary. Caesar also uses lazy evaluation optimizations to reduce GPU processing overhead and mobile energy consumption. 

The thesis makes contributions by (1) developing efficient and scalable solutions for different points in the design space of vision-based sensing tasks; (2) achieving state-of-the-art accuracy in those applications; (3) and developing guidelines for designing such sensing systems. 
}

%% file: packages.tex
\usepackage{booktabs} 
\usepackage{times}
\usepackage{algorithm}
\usepackage[noend]{algorithmic}
\usepackage{enumitem}
\usepackage{mathtools}
\usepackage{listings}
\usepackage{tabularx}
\usepackage{adjustbox}
\usepackage{array}
\usepackage{float}
\usepackage{graphicx}
\usepackage{epstopdf}
\usepackage{xcolor}
\usepackage{url}
\usepackage{xspace}
\usepackage{balance}
\usepackage{amsmath,amssymb,graphicx}
\usepackage[short]{optidef}
\usepackage{microtype}
\usepackage{sepfootnotes}
\usepackage[margin=5pt,font=small,labelfont={bf,rm}]{caption}
\usepackage{sectsty}

%% file: macros.tex
\newif\ifgmlegal
\gmlegalfalse

\definecolor{brown}{cmyk}{0,0.81,1,0.60}
\definecolor{magenta}{rgb}{0.4,0.7,0}
\definecolor{gray}{rgb}{0.5,0.5,0.5}
\definecolor{red}{rgb}{1,0,0}
\definecolor{green}{rgb}{0.5,0,0.5}
\definecolor{blue}{rgb}{0,0,1}
\definecolor{black}{rgb}{0,0,0}

\newcommand{\added}[1]{\textcolor{black}{#1}}
\newcommand{\changed}[1]{\textcolor{black}{#1}}

\newcommand{\newlyadded}[1]{\textcolor{black}{#1}}

\newcommand{\etc}{\emph{etc.}\xspace}
\newcommand{\ie}{\emph{i.e.,}\xspace}
\newcommand{\eg}{\emph{e.g.,}\xspace}
\newcommand{\etal}{\emph{et al.}\xspace}

\newcommand{\parab}[1]{\vspace*{0.5ex}\noindent\textbf{#1}}
\newcommand{\parae}[1]{\vspace*{0.5ex}\noindent\emph{#1}}

\newcommand{\chapref}[1]{Chapter~\ref{chap:#1}}

\newcommand{\figref}[1]{Figure~\ref{fig:#1}}
\newcommand{\tblref}[1]{Table~\ref{tbl:#1}}
\newcommand{\algref}[1]{Algorithm~(\ref{alg:#1})}

\newcommand{\alpsfig}[1]{Figure~\ref{fig:alps_#1}}
\newcommand{\caesarfig}[1]{Figure~\ref{fig:caesar_#1}}
\newcommand{\gnomefig}[1]{Figure~\ref{fig:gnome_#1}}
\newcommand{\grabfig}[1]{Figure~\ref{fig:grab_#1}}
\newcommand{\tarfig}[1]{Figure~\ref{fig:tar_#1}}

\newcommand{\losangeles}{Los Angeles\xspace}
\newcommand{\nyc}{New York\xspace}
\newcommand{\frankfurt}{Frankfurt\xspace}
\newcommand{\hongkong}{Hong Kong\xspace}

\def\setof#1{\left\{{\let\st\colon #1 }\right\}}

    \def\00{\mathbf{0}}

\newcommand{\eat}[1]{}



%% file: tex/intro.tex

\chapter{Introduction} \label{chap:intro}

Context awareness is an essential part of mobile and ubiquitous computing. Its goal is to unveil the situational information about mobile users like locations and activities. The sensed contexts can enable many services such as navigation, advertisements, AR/VR, and personal monitoring. Previous work leverages the sensor data (e.g., GPS, Bluetooth, gyroscope) to infer the context. However, those applications are limited by the hardware cost, sensor type, accuracy, battery, and the computing power on the device. These limitations prevent the development of easy-to-use and scalable solutions for context sensing. 

In recent years, there is an emergence of cameras deployed worldwide. Governments have installed denser surveillance networks for public monitoring. House and shop owners have set up more indoor cameras for safety and business insight purposes. Mobile platforms like phones, drones, and cars are also becoming more capable as video sources. 

Ubiquitous cameras can be used to sense context, but the data collection and analytics are still highly manual. For the data collection process, it is challenging to retrieve vision data from large scale and efficiently label the metadata (e.g. camera pose and location). The vision analytics are also heavy in labor. For instance, surveillance cameras are usually monitored by human beings, which is not scalable nor sensitive. Therefore, researchers have to spend a lot of effort on hiring volunteers, collecting data, and labeling the dataset. 

People start to leverage advanced algorithms and deep neural networks to help analyze vision data. The fast evolving hardware also supports more complex software to run on the cloud, edge servers, and mobile devices. However, such approach still faces challenges in practice. Vision-only solutions are not always accurate, especially when the image quality or the camera position is suboptimal, which reduces algorithm accuracy. Moreover, complex neural-network-based algorithms are compute-intensive and will not scale for tasks that have large scale of inputs and limited hardware resources. 


We summarize the above challenges of today's vision-based context sensing tasks as follows: (1) \textit{Scalability}: The vision data for context sensing can be hard to collect at large-scale; (2) \textit{Efficiency}: With the huge amount of data to process, sensing system with limited hardware resource will face long end-to-end latency and low processing speed; (3) \textit{Accuracy}: The vision-only and the sensor-only solution will sacrifice the accuracy due to their own limitations. 

\begin{figure}
\centering\includegraphics[width=0.7\columnwidth]{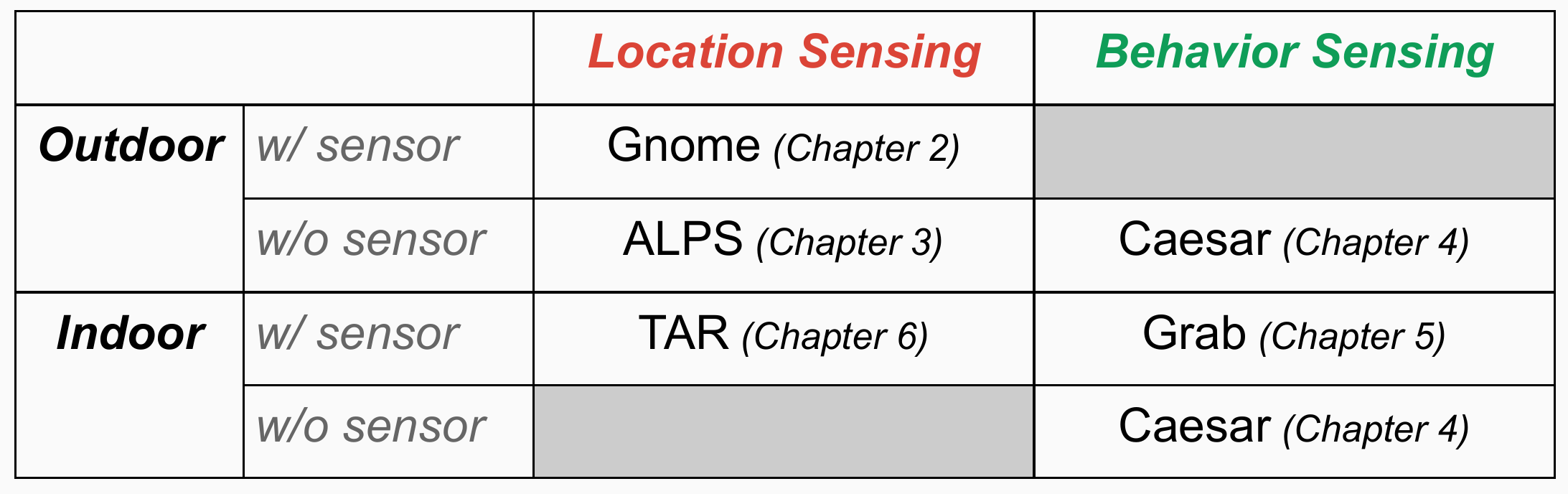}
\caption{\emph{The design space of vision-based context-sensing.}}
\label{fig:overview_table}
\end{figure}
\begin{figure}
\centering\includegraphics[width=0.7\columnwidth]{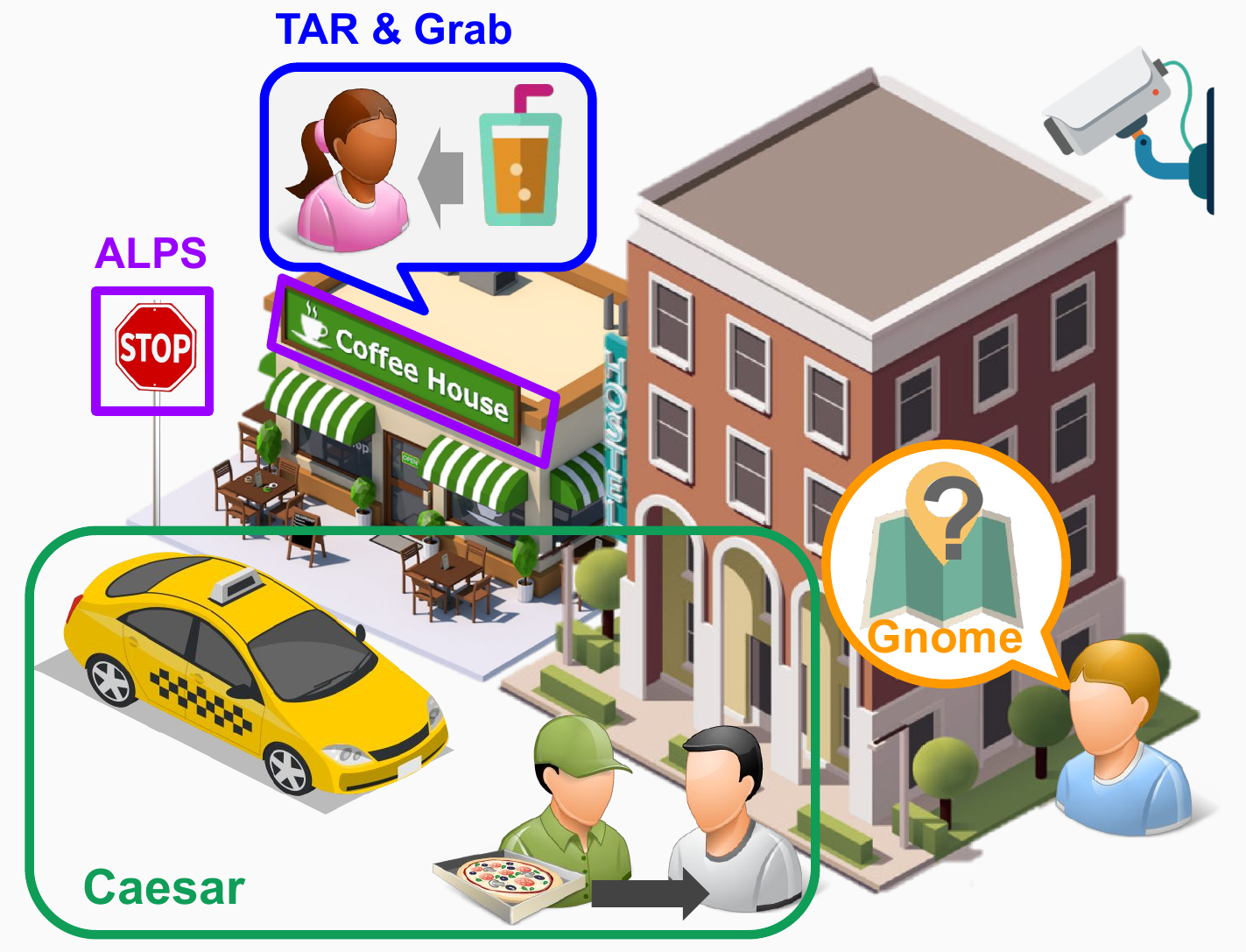}
\caption{\emph{Illustration of application scenarios of my work.}}
\label{fig:overview}
\end{figure}

The thesis explores the space of designs for vision-based context sensing. The space is defined by three dimensions. The first dimension is the sensing task. We find that most tasks are for at least one of the two purposes: localization, and detecting behaviors. Localization determines the spatial-temporal information of the target, and behavior sensing is to extract the target's identity and activity. The second dimension is the sensor type. We consider two classes of tasks: those that use vision sensors exclusively, and those that fuse vision with other sensors. The third dimension is the location of the task: indoor or outdoor. Indoor environment is usually more controllable in terms of sensor deployment and camera positions. The outdoor scenarios usually have much more data and the object movement is less constrained than indoor cases. With the above three dimensions, we could classify existing vision-based context sensing tasks into eight categories, which is shown in~\figref{overview_table}. Each of our prior projects focuses on the typical application scenario in that category to resolve the related challenges.~\figref{overview} illustrates our work in those context sensing scenarios. 

There are two slots marked in gray, which means that we did not explore the parts of the design space. The top-right part (outdoor, behavior sensing with sensor + vision) is not explored because the targets in those tasks are usually arbitrary, like pedestrians and cars on the street. This makes it hard to attach extra sensors to the targets to help improve the sensing accuracy. Moreover, wide-spread monitoring areas will prevent people deploying new sensors, and reusing existing surveillance cameras will be the best option. The bottom-left part (indoor, location sensing with vision-only) is skipped because our prior work (TAR) proves that, for indoor people tracking tasks, vision-only approaches have lower accuracy than vision+sensors. Therefore, the vision-only approach is not the best option for indoor location sensing. 


\paragraph{Outdoor person localization with vision and GPS sensors: Gnome~\cite{liu2018gnome}}

GPS signals suffer significant impairment in urban canyons because of limited line-of-sight to satellites and signal reflections. In Gnome, we focus on scalable and deployable techniques to reduce the impact of one specific impairment: reflected GPS signals from non-line-of-sight (NLOS) satellites. Specifically, we show how, using publicly available street-level imagery and off-the-shelf computer vision techniques, we can estimate the path inflation incurred by a reflected signal from a satellite. Using these path inflation estimates we develop techniques to estimate the most likely actual position given a set of satellite readings at some position. Finally, we develop optimizations for fast position estimation on modern smartphones. Using extensive experiments in the downtown area of several large cities, we find that our techniques can reduce positioning error by up to 55\% on average.

\paragraph{Outdoor landmark localization with vision: ALPS~\cite{hu2016alps}}

Ideally, every stationary object or entity in the built environment should be associated with position, so that applications can have precise spatial context about the environment surrounding a human. In ALPS, we take a step towards this ideal: by analyzing images from Google Street View that cover different perspectives of a given object and triangulating the location of the object, our system, ALPS, can discover and localize common landmarks at the scale of a city accurately and with high coverage. ALPS contains several novel techniques that help improve the accuracy, coverage, and scalability of localization. Evaluations of ALPS on many cities in the United States show that it can localize storefronts with a coverage higher than 90\% and a median error of 5 meters.

\paragraph{Outdoor person behavior recognition with vision: Caesar~\cite{liu2019caesar}}

Caesar is an edge computing based system for complex activity detection, which provides an extensible vocabulary of activities to allow users to specify complex actions in terms of spatial and temporal relationships between actors, objects, and activities. It converts these specifications to graphs, efficiently monitors camera feeds, partitions processing between cameras and the edge cluster, retrieves minimal information from cameras, carefully schedules neural network invocation, and efficiently matches specification graphs to the underlying data in order to detect complex activities. Our evaluations show that Caesar can reduce wireless bandwidth, on-board camera memory, and detection latency by an order of magnitude while achieving good precision and recall for all complex activities on a public multi-camera dataset.

\paragraph{Indoor shopper behavior recognition with vision and sensors: Grab~\cite{grab}}

Grab leverages existing infrastructure and devices to detect shopping behaviors to enable cashier-free shopping, which needs to accurately identify and track customers, and associate each shopper with items he or she retrieves from shelves. To do this, Grab uses a keypoint-based pose tracker as a building block for identification and tracking, develops robust feature-based face trackers, and algorithms for associating and tracking arm movements. It also uses a probabilistic framework to fuse readings from camera, weight and RFID sensors in order to accurately assess which shopper picks up which item. In experiments from a pilot deployment in a retail store, Grab can achieve over 90\% precision and recall even when 40\% of shopping actions are designed to confuse the system. Moreover, Grab has optimizations that help reduce investment in computing infrastructure four-fold.

\paragraph{Indoor person localization and tracking with vision and sensors: TAR~\cite{liu2018tar}}

TAR leverages widespread camera deployment and Bluetooth proximity information to accurately track and identify shoppers in the store. TAR is composed of four novel design components: (1) a deep neural network (DNN) based visual tracking, (2) a person trajectory estimation by using visual features and BLE proximity traces, (3) an identity matching and assignment to recognize person identity, and (4) a cross-camera calibration algorithm. TAR carefully combines these components to track and identify people in real-time across multiple non-overlapping cameras. It achieves 90\% accuracy in two different real-life deployments, which is 20\% better than the state-of-the-art solution.

\paragraph{Our Contributions}
Overall, the thesis makes these contributions: We develop efficient and scalable solutions (ALPS, Gnome, TAR, Caesar, and Grab) for different application scenarios in vision-based context sensing. In those works, we show that pure vision algorithms are not accurate enough in real surveillance scenarios, for which we can leverage low-cost sensors and cross-camera knowledge as complementary modules to achieve high accuracy. We also find that the worldwide street-level imagery can help scale vision algorithms to large scale. We analyze various trade-offs among accuracy, storage, latency, energy consumption, and cost. We build prototypes and conduct experiments in real scenarios, and prove that our solutions achieve state-of-the-art accuracy while being practical and scalable. We have developed guidelines for such vision-enabled context sensing systems as following: 
\begin{itemize}
  \item \textbf{Data Source and Platforms}: Leverage existing sensing infrastructure and imagery data source to scale the data collection coverage. Exploit different sensors' characteristics and leverage other sensors as complementary to vision data. 
  \item \textbf{Architecture Design}: Identify the bottleneck of the pipeline. Build the system around one high-level abstraction to keep accuracy and flexibility. Share the workload among all computation modules to improve efficiency. 
  \item \textbf{Pipeline Optimization}: Apply cheap vision algorithms to reduce the dependency for expensive hardware. Improve efficiency by exploiting software optimization tricks with data and task characteristics. 
\end{itemize}

\paragraph{Dissertation Outline}
In~\chapref{gnome}, we introduce Gnome, a practical approach to NLOS mitigation for GPS positioning in smartphones. In~\chapref{alps}, we introduce ALPS for accurate landmark positioning at city scales. In~\chapref{caesar}, we introduce Caesar for cross-camera complex activity recognition. In~\chapref{grab}, we introduce Grab, a cashier-free shopping system. In~\chapref{tar}, we introduce TAR that enables fine-grained tracking and targeted advertising. In~\chapref{related}, we present a comprehensive overview of related work in the literature. Finally, we summarize our work and discuss the guidelines in~\chapref{conclusion}.

%% file: tex/paper_gnome.tex
\chapter{Gnome: A Practical Approach to NLOS Mitigation for GPS Positioning in Smartphones}\label{chap:gnome}

\input{tex/gnome/intro}
\input{tex/gnome/motivation}
\input{tex/gnome/design}
\input{tex/gnome/eval}

%% file: tex/gnome/intro.tex
\section{Introduction}

Accurate positioning has proven to be an important driver for novel applications, including navigation, advertisement delivery, ride-sharing, and geolocation apps. While positioning systems generally work well in many places, positioning in urban canyons remains a significant challenge. Yet it is precisely in urban canyons in megacities that accurate positioning is most necessary. In these areas, smartphone usage is high, as is the density of places (storefronts, restaurants etc.), motivating the need for high positioning accuracy.

Over the last decade, several techniques have been used to improve positioning accuracy, many of which are applicable to urban canyons. \changed{Cell tower and Wi-Fi based localization~\cite{sen2012you,sen2013avoiding} enable smartphones to estimate their positions based on signals received from these wireless communication base stations. Map matching enables positioning systems to filter off-road location estimates of cars~\cite{bo2013smartloc, jiang2015carloc}. Dead-reckoning uses inertial sensors (accelerometers, and gyroscopes) to estimate travel distance, and thereby correct position estimates~\cite{jiang2015carloc}. Crowd-sourcing GPS readings~\cite{agadakos2017techu} or using differential GPS systems~\cite{hedgecock2013high,hedgecock2014accurate}  can also help improve GPS accuracy.} Despite these improvements, positioning errors in urban canyons can average 15m.

\changed{That is because these techniques do not tackle the \emph{fundamental} source of positioning error in urban canyons~\cite{GPSeBook,kaplan2005understanding}: non-line-of-sight (NLOS) satellite signals at GPS receivers}. GPS receivers use signals from four or more satellites to triangulate their positions. Specifically, each GPS receiver estimates the distance traveled by the signal from each visible satellite: this distance is called the satellite’s pseudorange. In an urban canyon, signals from some satellites can reach the receiver after being reflected from one or more buildings. This can inflate the pseudorange: a satellite may appear farther from the GPS receiver than it actually is. This path inflation can be tens or hundreds of meters, and can increase positioning error.

\paragraph{Contributions}

This chapter describes the design of techniques, and an associated system called Gnome, that revises GPS position estimates by compensating for path inflation due to NLOS satellite signals. Gnome can be used in many large cities in the world, and requires a few tens of milliseconds on a modern smartphone to compute revised position estimates. It does not require specialized hardware, nor does it require a phone to be rooted. In these senses, it is immediately deployable.

This chapter makes three contributions corresponding to three design challenges: (a) How to compute satellite path inflation? (b) How to revise position estimates? (c) How to perform these computations fast on a smartphone?

Gnome estimates path inflation using 3D models of the environment surrounding the GPS receiver’s position. \changed{While prior work on NLOS mitigation~\cite{peyraud2013non, kumar2014identifying,drevelle2012igps,sahmoudi2014deep,miura2015gps,hsu20163d, groves2011shadow,wang2015smartphone,adjrad2015enhancing} has used proprietary sources of 3D models}, we use a little known feature in Google Street View [@streetview] that provides depth information for planes (intuitively, each street-facing side of a building corresponds to a plane) surrounding the receiver’s position. This source of data makes Gnome widely usable, since these planes are available for many cities in North America, Europe and Asia. Unfortunately, these plane descriptions lack a crucial piece of information necessary for estimating path inflation: the height of building planes. Gnome’s first contribution is a novel algorithm for estimating building height from panoramic images provided by Street View. \added{Compared to prior work that determines building height from public data~\cite{guo2002snake,saito2016multiple}, or uses remote sensing radar data~\cite{soergel2009stereo,cellier2006building,brunnera2008building,brunner2008extraction}, our approach achieves higher coverage by virtue of using Street View data.}

To compute the path inflation correctly, Gnome needs to know the ground truth position. However, GPS receivers don’t, of course, provide this: they only provide satellite pseudoranges, and an estimated position. Gnome must therefore infer the position most likely to correspond to the observed satellite pseudoranges. To do this, Gnome’s second contribution is a technique to search candidate positions near the GPS location estimate, revise the candidate’s position by compensating for path inflation, and then determine the revised candidate position likely to be closest to the ground truth. \added{This contribution is inspired by, but different from the prior work that attempts to infer actual positions by simulating the satellite signal path~\cite{sahmoudi2014deep,miura2015gps,hsu20163d}, or by determining satellite visibility~\cite{groves2011shadow,wang2015smartphone,adjrad2015enhancing}}.

\changed{Gnome’s third contribution is to enable these computations to scale to smartphones, a capability that, to our knowledge, has not been demonstrated before.} To this end, it leverages the observation that 3D models of an environment are relatively static, so Gnome aggressively pre-computes, in the cloud, path inflation maps at each candidate position. These maps indicate the path inflation for each possible satellite position, and are loaded onto a smartphone. At runtime, Gnome simply needs to look up these maps, given the known positions of each satellite, to perform its pseudorange corrections. Gnome also scopes the search of candidate positions and hierarchically refines the search to reduce computation overhead.

Gnome differs from \cite{sahmoudi2014deep,miura2015gps,hsu20163d} in two ways. First, these approaches use proprietary 3D models for ray-tracing, which are not accessible in many cities. In contrast, Gnome leverages highly available Street View data for satellite signal tracing. Second, these approaches are offline while Gnome can compute location estimates in real-time on Android devices.

Our evaluations of Gnome in four major cities (Frankfurt, Hong Kong, Los Angeles and New York) reveal that Gnome can improve position accuracy in some scenarios by up to 55\% on average (or up to 8m on average). Gnome can process a position estimate on a smartphone in less than 80ms. It uses minimal additional battery capacity, and has modest storage requirements. Gnome’ cloud-based path inflation map pre-computation takes several hours for the downtown area of a major city, but these maps need only be computed once for areas with urban canyons in major cities. Finally, Gnome components each contribute significantly to its accuracy: height adjustment accounts for about 3m in error, and sparser candidate position selections also increase error by the same amount.

%% file: tex/gnome/motivation.tex
\section{Background, Motivation, and Approach}

\parab{How GPS works}
GPS is an instance of a Global Navigation Satellite System. It consists of 32 medium earth orbit satellites, and each satellite continuously broadcasts its position information and other metadata at an orbit of about 2x$10^{7}$ meters above the earth. The metadata specify various attributes of the signal such as the satellite position, timestamp, etc. Using these, the receiver computes, for each received signal, its \emph{pseudorange} or the signal’s travel distance, by multiplying light of speed with the signal’s propagation delay. With these pseudorange estimates, GPS uses three satellites’ position to trilaterate the receiver’s position in 3D coordinates. In practice, the receiver’s local clock is not accurate compared with satellite’s atomic clock, so GPS needs another satellite’s signal to estimate the receiving time. Thus, a GPS receiver must be able to receive signals from at least four satellites in order to fix its own position.

\parab{GPS Signal Impairments}
GPS signals undergo four\footnote{We have simplified this discussion. Additional sources of error can come from clock skews, receiver calibration errors and so forth \cite{GPSeBook}.} different types of impairments \cite{GPSeBook} that introduce errors in position estimates. The earth’s rotation between when the signal was transmitted and received can impact travel time, as can the Doppler effect due to the satellite’s velocity. Ionospheric and tropospheric delays caused by the earth’s atmosphere can inflate pseudoranges. Multipath transmissions, where the same signal is received directly from a satellite and via reflection, can cause constructive or destructive interference and introduce errors in the position fix. Finally, a receiver may receive a signal, via reflection, from an NLOS satellite.

Many modern receivers compensate for, either in hardware or software, the first two classes of errors. Specifically, GPS receivers can compensate for earth’s rotation and satellite Doppler effects. GPS signals also contain metadata that specify approximate corrections for atmospheric delays. Higher accuracy applications that need to eliminate such correlated errors (the atmospheric delay is correlated in the sense that two receivers within a few kilometers of each other are likely to see the same coordinated delays) can use either Differential GPS or Real-Time Kinematic GPS. Both of these approaches use base stations whose precise position is known a priori. Each base station can estimate correlated errors based on the difference between its position calculated from GPS, and its actual (known) position. It can then broadcast these corrections to nearby receivers, who can use these to update their position estimates.

Two other sources of error, multipath and NLOS reflections are not correlated, so different techniques must be used to overcome them. These sources of error are particularly severe in urban canyons~\cite{miura2015gps, groves2011shadow, wang2015smartphone}.

\parab{Urban Canyons}
To understand how NLOS reflections can impact positioning accuracy, consider \gnomefig{los_nlos}(a) in which a satellite is within line-of-sight (LOS) of a receiver, so the latter receives a direct signal. If a tall building blocks the LOS path, the satellite signal may still be received after being reflected, and causes NLOS reception (\gnomefig{los_nlos}(b)). Thus, depending on the environment and the receiver’s position, a satellite’s primary received signal can either be direct or reflected. In addition, the primary signal can itself be reflected (more than once), resulting in multipath receptions (\gnomefig{los_nlos}(c)).

To mitigate the impact of multipath, GPS receivers use multipath correctors (\cite{comp1998adaptive, zhang2004multipath, zhang2017multipath}) that use signal phase to distinguish (and filter out) the reflected signal from the primary signal. Modern receivers can reduce the impact of multipath errors to a few meters.

However, when the primary signal is reflected (i.e., the signal is from an NLOS satellite), the additional distance traveled by the signal due to the reflection can inflate the pseudorange estimate. The yellow lines in \gnomefig{los_nlos} represent reflected signal paths, which are longer than the primary paths shown in green. The difference in path length between these two signals can often be 100s of meters. Unfortunately, GPS receivers cannot reliably distinguish between reflected and direct signals, and this is the primary cause of positioning error in urban areas. This chapter focuses on NLOS mitigation for GPS positioning.

\begin{figure}
\centering\includegraphics[width=0.5\columnwidth]{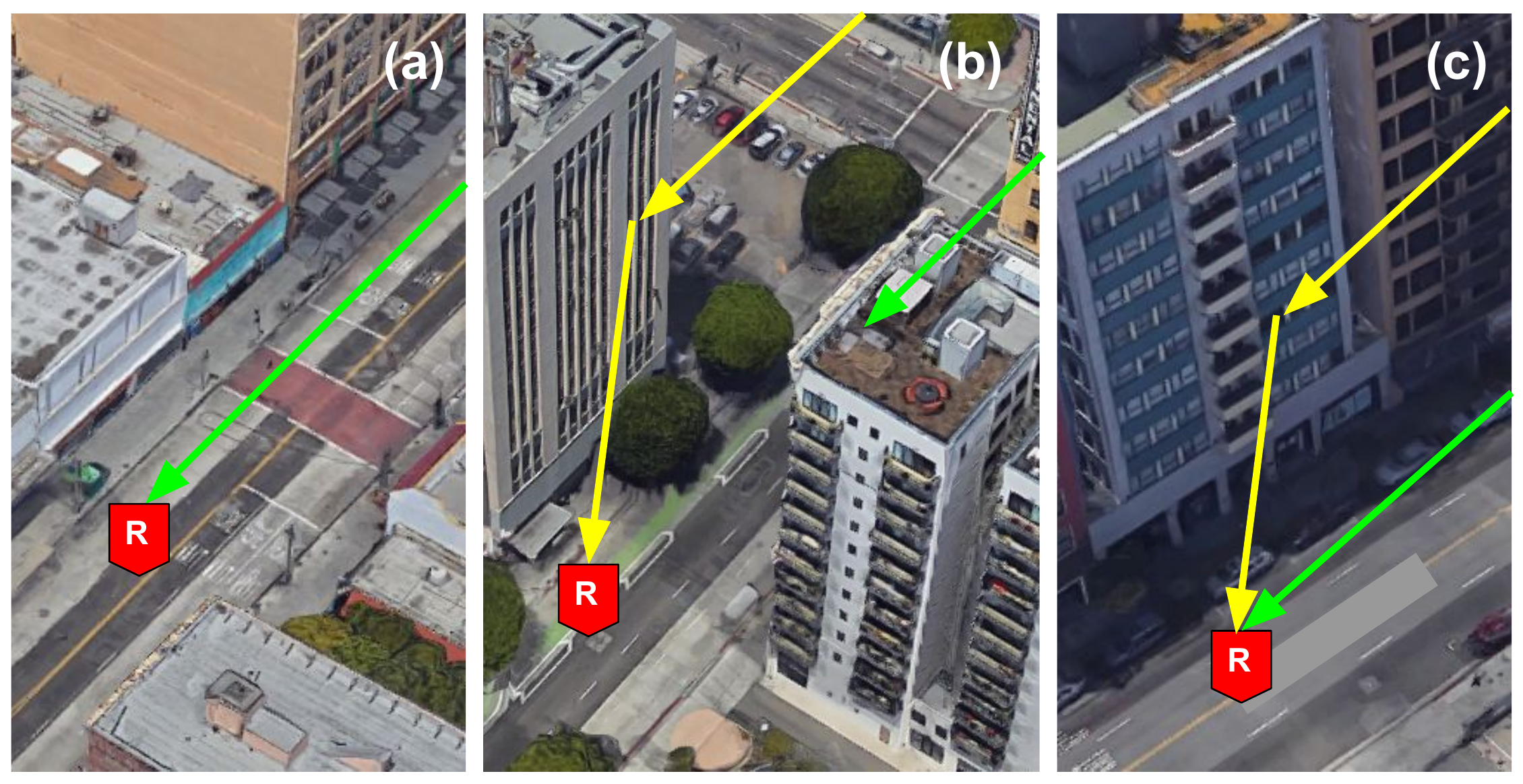}
\caption{\emph{(a) A line-of-sight (LOS) signal path. (b) A non-line-of-sight (NLOS) signal path. (c) Multipath}}
\label{fig:gnome_los_nlos}
\end{figure}

\parab{Alternative approaches}
To mitigate NLOS reception errors in urban canyons, smartphones use several techniques to augment position fixes. First, they use proximity to cellular base stations \cite{drane1998positioning} or Wi-Fi access points \cite{wifiAndroid} to refine their position estimates. In this approach, smartphones use multilateration of signals from nearby cell towers or Wi-Fi access points whose position is known a priori in order to estimate their own position. Despite this advance, positioning errors in urban areas can be upwards of 15m, as our experiments demonstrate in \gnomefig{phone_err}.

\begin{figure}
\centering\includegraphics[width=0.5\columnwidth]{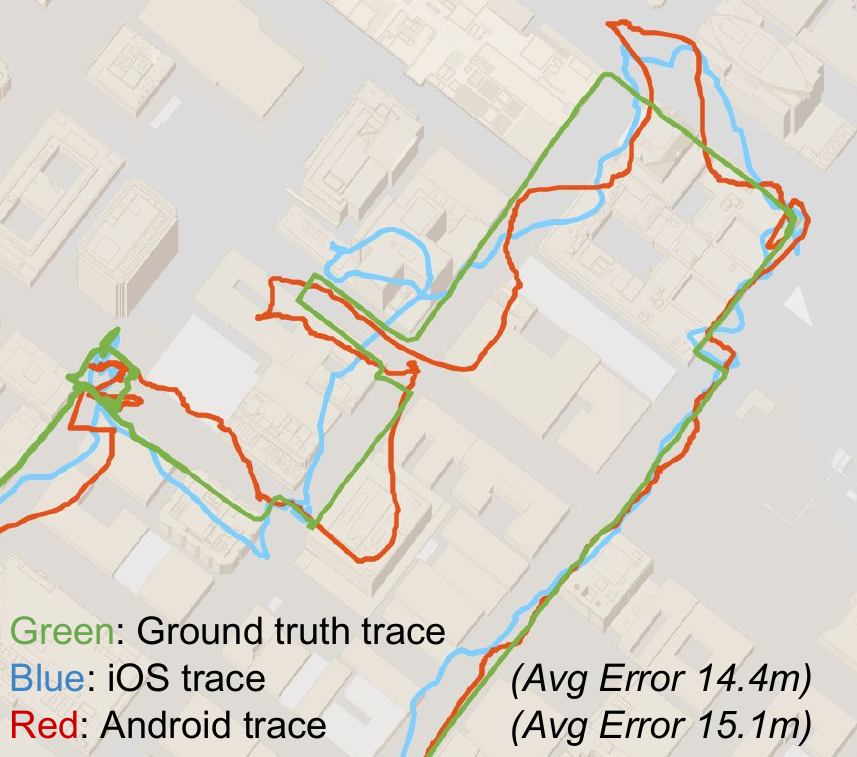}
\caption{\emph{An example of localization results in urban canyon on today’s smartphone platforms}}
\label{fig:gnome_phone_err}\vspace{-0.1in}
\end{figure}

Second, for positioning vehicles accurately, smartphones use map matching and dead-reckoning~\cite{jiang2015carloc} to augment position fixes. Map matching restricts candidate vehicle positions to street surfaces, and dead-reckoning uses vehicle speed estimates to update positions when GPS is unavailable or erroneous. While these techniques achieve good accuracy, they are not applicable to localizing pedestrians in urban settings.

Finally, as we have discussed above, approaches like Differential GPS and Real-time Kinematics assume correlated error within a radius of several  hundred meters or several kilometers. Errors due to NLOS receptions are not correlated over these large spatial scales, so these techniques cannot be applied in urban canyons.

\parab{Goal, Approach and Challenges}
The goal of this chapter is to develop a practical and deployable system for NLOS mitigation on smartphones. To be practical, such a system must not require proprietary sources of information. To be deployable, it must, in addition, be capable of correcting GPS readings efficiently on the smartphone itself.

Our approach is motivated by the following key insight: If we can determine the extra distance traveled by an NLOS signal, we can compensate for this extra distance, and recalculate the GPS location on the smartphone.

This insight poses three distinct challenges. The first is how to determine NLOS satellites and compute the extra travel distance? While satellite trajectories are known in advance, whether a satellite is within line-of-sight at a given location $L$ depends upon the portion of the sky visible at $L$, which in turn depends on the position and height of buildings around $L$. This latter information, also called the surface geometry at $L$ can be used to derive the set of surfaces that can possibly reflect satellite signals so that they are incident at $L$.

The second challenge is how to compensate for the extra distance traveled by an NLOS signal (we use the term path inflation to denote this extra distance) incident at a location $L$. This is a challenge because a smartphone cannot, in general, know the location $L$: it only has a potentially inaccurate estimate of $L$. To correctly compensate for path inflation, the smartphone has to determine that the location whose predicted reflected signal best explains the GPS signals observed at $L$.

The final challenge is to be able to perform these corrections on a smartphone. Determining the visibility mask and the surface geometry are significantly challenging both in terms of computing and storage, particularly at the scale of large downtown area of several square kilometers, and especially because these are functions of $L$ (i.e., each distinct location in an urban area has a distinct visibility mask and surface geometry). These computing and storage requirements are well beyond the capability of today’s smartphones.

In the next section, we describe the design of a system called Gnome that addresses all of these challenges, while providing significant performance improvements in positioning accuracy over today’s smartphones.

Prior work in this area has fallen into two categories: those that filter out the NLOS signal~\cite{tay2013weighting, kumar2014identifying, hsu2015nlos}, and those that compensate for the extra distance traveled~\cite{betaille2014enhance, hsu20163d, moreau2017fisheye, mirrorSV}. Gnome falls into the latter class, but is unique in addressing our deployability goals.

%% file: tex/gnome/design.tex
\section{Gnome Design}

In this section, we describe the design of Gnome. We begin by describing how Gnome addresses the challenges identified, then describe the individual components of Gnome.

\begin{figure}
\centering\includegraphics[width=0.6\columnwidth]{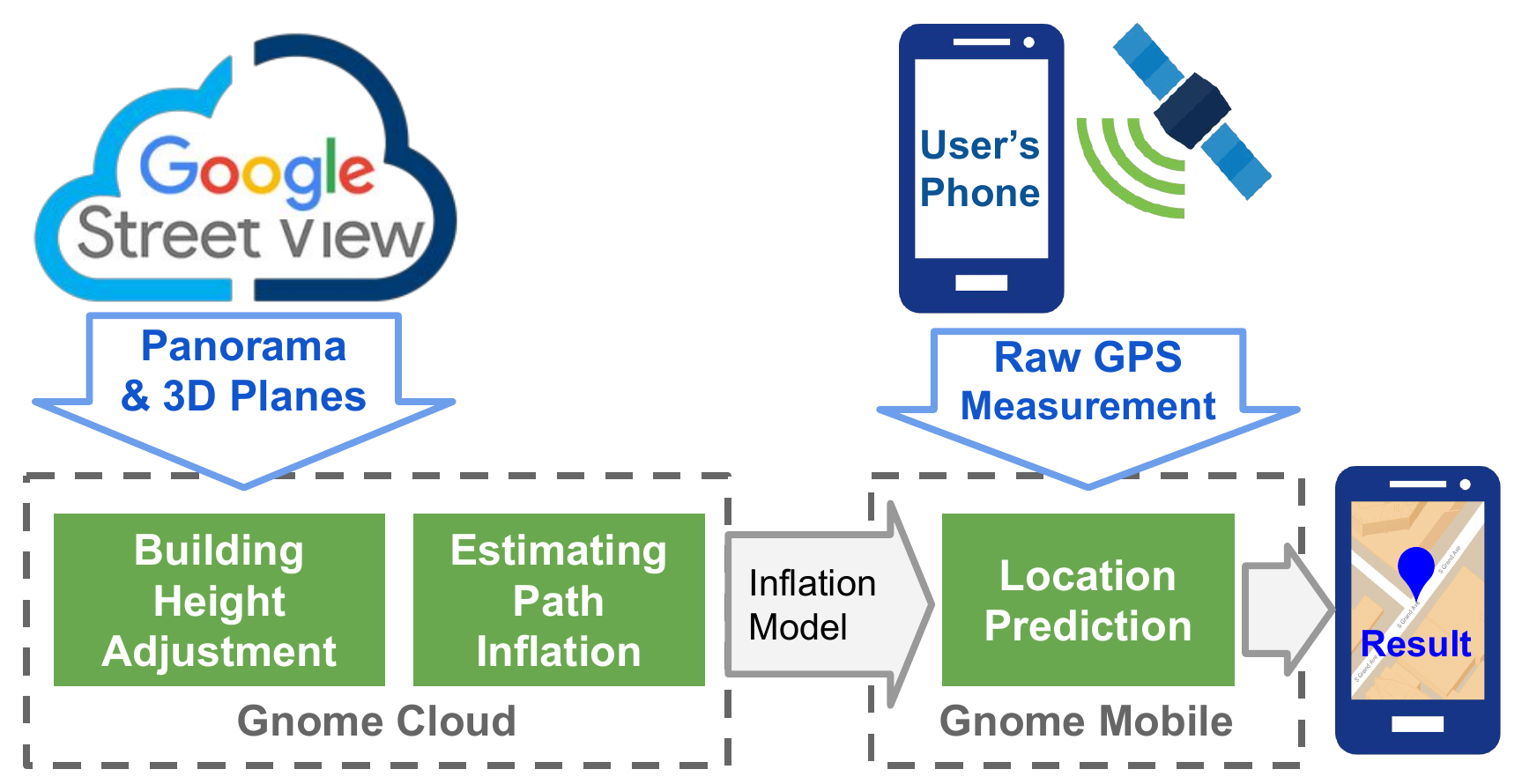}
\caption{\emph{Gnome workflow.}}
\label{fig:gnome_sys_arch}\vspace{-0.1in}
\end{figure}

\subsection{Overview}
As described above, Gnome detects whether a satellite is within line-of-sight or not, and for NLOS satellites, it estimates and compensates for the extra travel distance for the NLOS signal. Gnome is designed to perform all of these calculations entirely on a smartphone.

To determine whether, at a given location $L$, a satellite is NLOS or not, and to compute the path inflation, Gnome uses the satellite’s current position in the sky as well as the surface geometry around $L$. Specifically, with a 3-D model of the buildings surrounding $L$, Gnome can determine whether a satellite’s signal might have been reflected from any building by tracing signal paths from the satellite to $L$, and use that to compute the path inflation. There exist public services to precisely determine the position for satellite at a given time. Less well known is the fact that there also exist public sources for approximate surface geometry: specifically, Google Street View~\cite{streetview} provides both 2-D imagery of streets as well as 3-D models (as an undocumented feature) of streets for the downtown area of most large cities in the US, Europe, and Asia. These 3-D models are, however, incomplete: they lack building height information, which is crucial to trace reflected signals from satellites. In the section below, we describe how we use computer vision techniques to estimate a building’s height. The availability of public datasets with 3-D information makes Gnome widely applicable: prior work has relied on proprietary datasets, and so has not seen significant adoption.

To compensate for the NLOS path inflation on a smartphone, Gnome leverages the fact that modern mobile OSes expose important satellite signal metadata such as what satellite signals were received and their relative strength \cite{android_api}. Gnome uses this information. The metadata, however, does not inform Gnome whether the satellite was LOS or NLOS. To determine whether a satellite is NLOS at $L$, Gnome can use the derived surface geometry. Unfortunately, the GPS signal only gives Gnome an estimate of the true location $L$, so Gnome cannot know the exact path inflation. To address this challenge, Gnome searches within a neighborhood of the GPS-provided location estimate $L_{est}$ to find a candidate for the ground truth $L_c$ whose positioning error after path inflation adjustment is minimized.

Our third challenge is to enable Gnome to run entirely on a smartphone. Clearly, it is unrealistic to load models of surface geometry for every point in areas with urban canyons. We observe that, while satellite positions in the sky are time varying, the surface geometry at a given location $L$ is relatively static. So, we pre-compute the path inflation, on the cloud, of every point on the street or sidewalk from every possible location in the sky. As we show later, this scales well in downtown areas of large cities in the world because in those areas tall buildings limit the portion of the sky visible.

Gnome is implemented (\added{\gnomefig{sys_arch}}) as a library on smartphones (our current implementation runs on Android) which, given a GPS estimate and satellite visibility information, outputs a corrected location estimate. The library includes other optimizations that permit it to process GPS estimates within tens of milliseconds.

\subsection{Data sources}
Gnome uses three distinct sources of data. First, whenever Gnome needs a position fix, it uses the smartphone GPS API to obtain the following pieces of information:

\emph{Latitude, longitude, and error}: The latitude and longitude specify the estimated position, and the error specifies the position uncertainty (the actual position is within a circle centered at the estimated position, and with radius equal to the error).
\emph{Satellite metadata}: This information (often called NMEA data \cite{nmea}) includes each satellite’s azimuth and elevation, as well as the signal strength represented as the carrier-to-noise density, denoted $C/N_0$ ~\cite{cno}. \gnomefig{sv_pos} shows an example of satellite metadata obtained from a satellite during our experiments. Each square represents a satellite with the number as satellite ID. The color of the square indicates the satellite’s signal strength: green is very good ($C/N_0$ > 35), yellow is fair (25 < $C/N_0$ < 35), and red is bad ($C/N_0$ < 25). The blue line denotes the skyline for a particular street in our data. Notice that satellite number 6 which is NLOS with respect to the center still has good carrier-to-noise density, so this metric is not a good discriminator for NLOS satellites.
\emph{Propagation delay and pseudorange}: This contains, for each satellite, the estimated propagation delay and pseudorange for the received signal. This data is read from phone’s GPS module and has become accessible since a recent release of Android. This information is crucial to Gnome, as we shall describe later.

\begin{figure}
\centering\includegraphics[width=0.6\columnwidth]{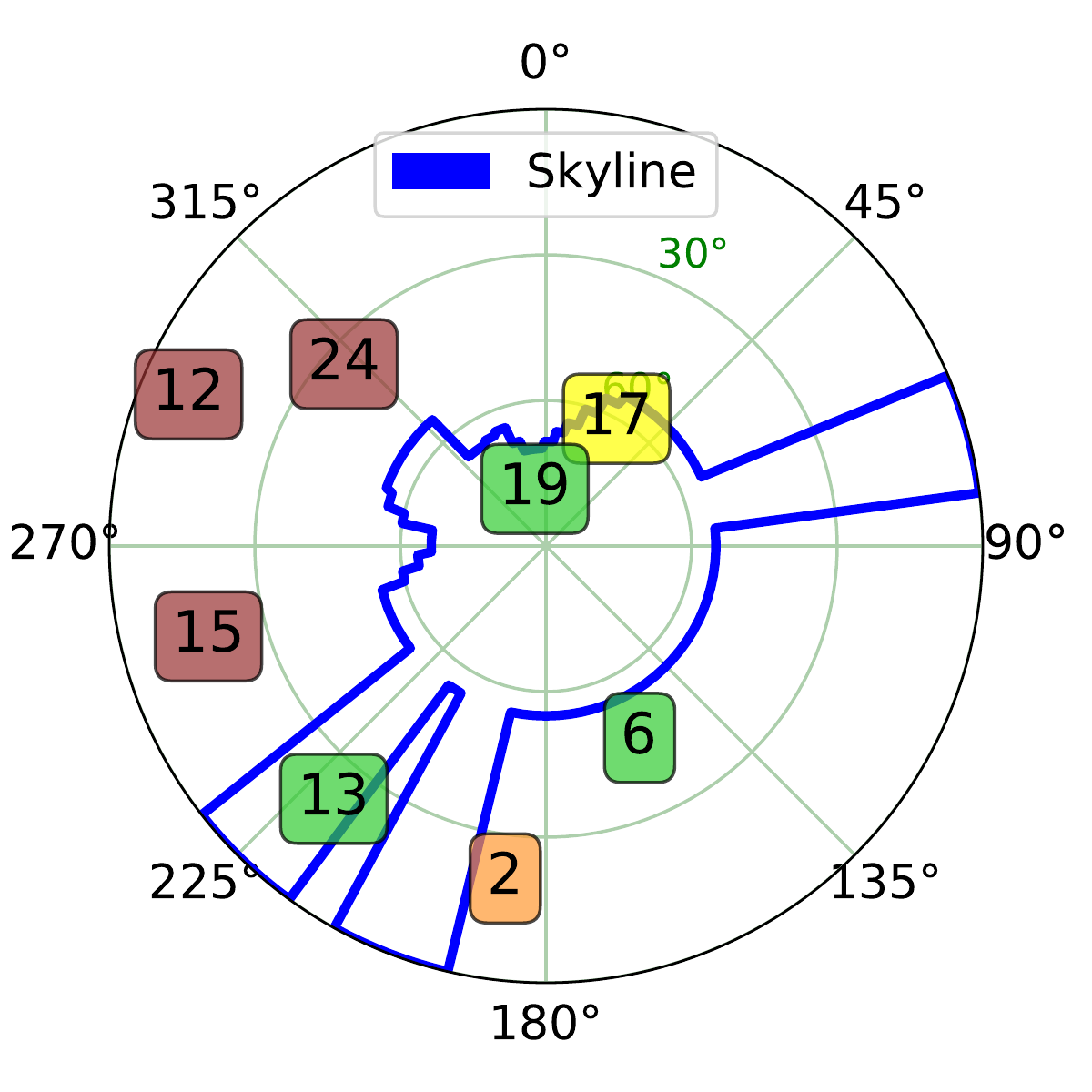}
\caption{\emph{The skyline and satellites locations seen by a receiver. The receiver is at the center of the circle and the squares represent satellites. The numbers represent satellite IDs and color signifies signal strength.}}
\label{fig:gnome_sv_pos}\vspace{-0.0in}
\end{figure}

Second, Gnome uses street-level imagery data available through Google Street View. This cloud service, when provided with a location $L$, returns an panoramic image around $L$.

Third, but most important, Gnome uses an approximate 3D model available through a separate Street View cloud service~\cite{3dmodel}. This service, given a location $L$ returns a 3-D model of all buildings or other structures around that point. This model is encoded (\gnomefig{height_diff} top) as a collection of planes, together with their depth (distance from $L$). Intuitively, each plane represents one surface of a building. The depth information is at the resolution of 0.7$^\circ$ in both azimuth and elevation, and has a maximum range of about 120m~\cite{anguelov2010google}.

Effectively, the 3D model describes the surface geometry around $L$, but it has one important limitation. The maximum height of a plane is 16m, a limitation arising from the range of the Street View scanning device. This limitation is critical for Gnome, because many buildings in urban canyons are an order of magnitude or more taller, and a good estimate of plane height is important for accurately determining satellite visibility. 

\added{Google Earth~\cite{google_earth} also provides a 3D map with models of buildings. As of this writing, extracting these 3D models is labor-intensive~\cite{model_from_google_earth} and does not scale to large cities. Moreover, its 3D models do not cover most countries in Asia, Europe, and South America, whereas Street View coverage is available in these continents. We leave it to future work to include 3D models extracted from Google Earth.}

\subsection{Estimating Building Height}

\begin{figure}
\centering\includegraphics[width=0.6\columnwidth]{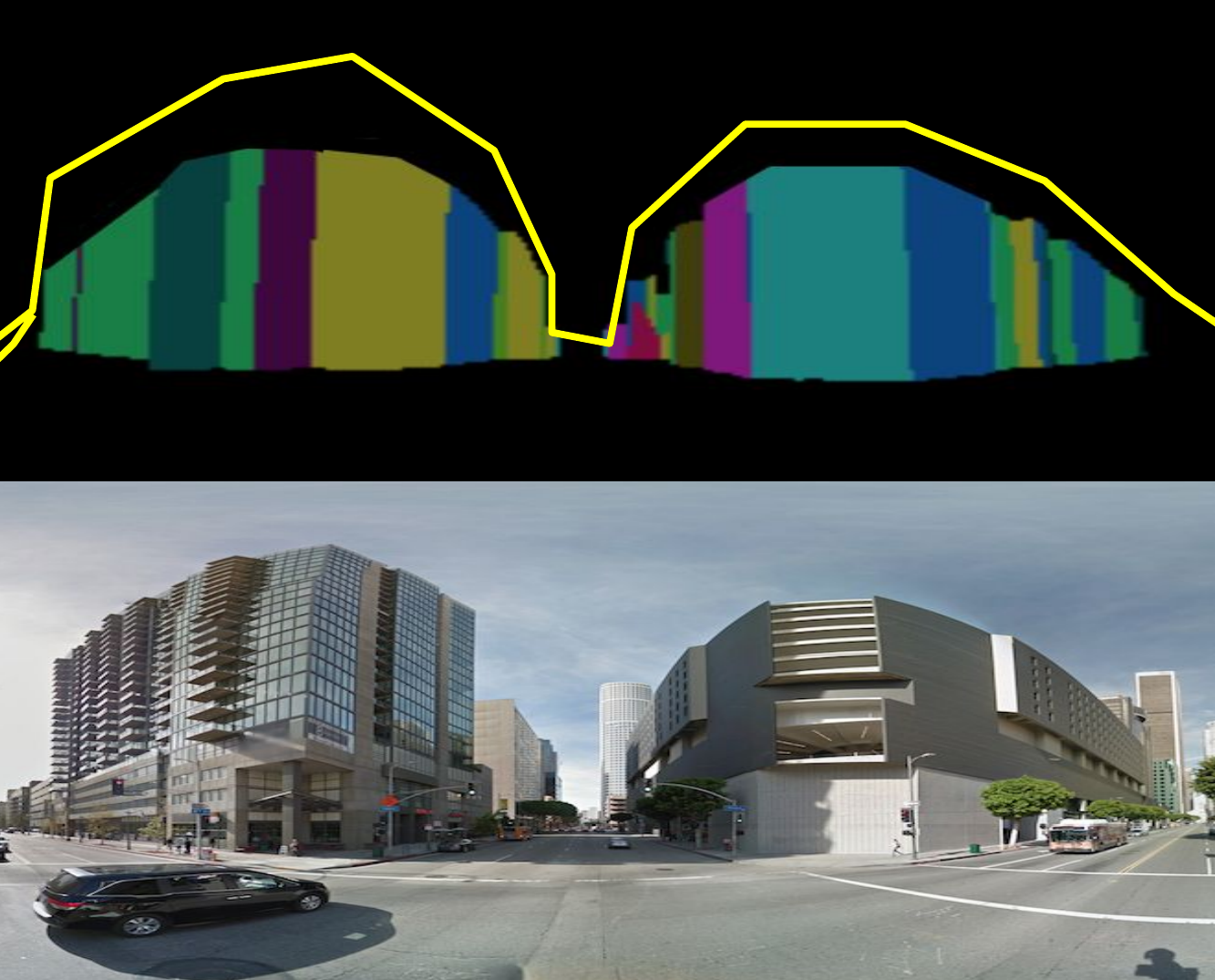}
\caption{\emph{\textbf{Upper}: the original depth information (each colored plane represents one surface of a building), together with the missing height information in yellow. \textbf{Lower}: the corresponding panoramic image}}
\label{fig:gnome_height_diff}\vspace{-0.1in}
\end{figure}

\gnomefig{height_diff} shows how the 3D model’s height differs from the actual height of a building: the yellow line above the planes is the actual height. To understand why obtaining height information for planes is crucial, consider a GPS receiver that is 15m away from a 30m building. All satellites behind the building with an elevation less than $\ang{63}$ will be blocked from the view of the receiver. However, with the planes we have from Street View, all satellites above $\ang{45}$ will still be LOS. Thus, our model may wrongly estimate an NLOS satellite as LOS, and may compensate for the path inflation where no such inflation exists.

\begin{figure}
\centering\includegraphics[width=0.95\columnwidth]{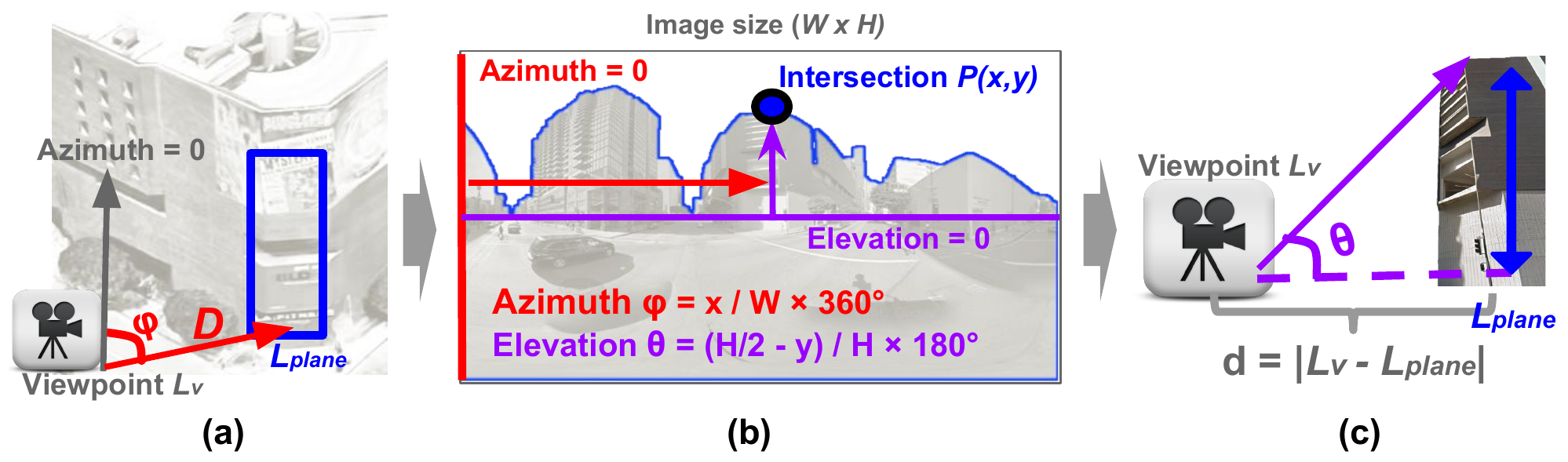}
\caption{\emph{Building height estimation. (a) shows how the vector $D$ is estimated from the surface geometry. (b) shows how the panoramic image can be used for skyline detection and $\theta$ estimation. (c) shows how building height is estimated using $D$ and $\theta$.}}
\label{fig:gnome_height_cal}\vspace{-0.0in}
\end{figure}

Gnome leverages Street View’s panoramic images to solve the issue. The basic idea is to estimate the building’s height by detecting skylines in the Street View images, and then extending each 3D plane to the estimated height. 
After extracting the surface geometry for a given location $L$, Gnome selects all planes whose reported height in the surface geometry is more than 13m: with high likelihood, such planes are likely to correspond to buildings higher than 15m. Next, for each such tall plane, Gnome computes its latitude and longitude (denoted by $L_{plane}$) on the map. It does this using the location $L$ and the relative location of the plane from $L$ (available as depth information in the 3D model). 

Gnome then selects a viewpoint $L_{v}$ near the plane and computes the vector $D$ from $L_{v}$ to $L_{plane}$ which is the centroid of the plane. This vector contains two pieces of information: (a) it specifies the heading of the plane relative to $L_{v}$, and (b) it specifies the distance from $L_{v}$ to the plane. Gnome then downloads the Street View image at $L_{v}$ and identifies the plane in the image using the heading information in $H$. \gnomefig{height_cal} shows an example of this calculation. \gnomefig{height_cal}(a) shows the satellite view at a given location $L$ and the heading vector for the plane (the blue box). The heading vector has azimuth $\phi$. \gnomefig{height_cal}(b) shows how Gnome uses $\phi$ to find the plane’s horizontal location $x$ in the corresponding panoramic image from Street View.

Now, Gnome runs a skyline detection algorithm on the image. Skyline detection demarcates the sky from other structures in an image. \gnomefig{height_vp} shows the output of skyline detection on some images. Intuitively, the part of the skyline that intersects with the plane delineates the actual height of the plane. Thus, in the three images on the right of \gnomefig{height_vp}, the intersection between the blue skyline and the red plane signifies the top of the building. Unfortunately, this intersection is visible in a two-dimensional image, whereas we need to augment the height of a plane in a 3D map.

\begin{figure}
\centering\includegraphics[width=0.5\columnwidth]{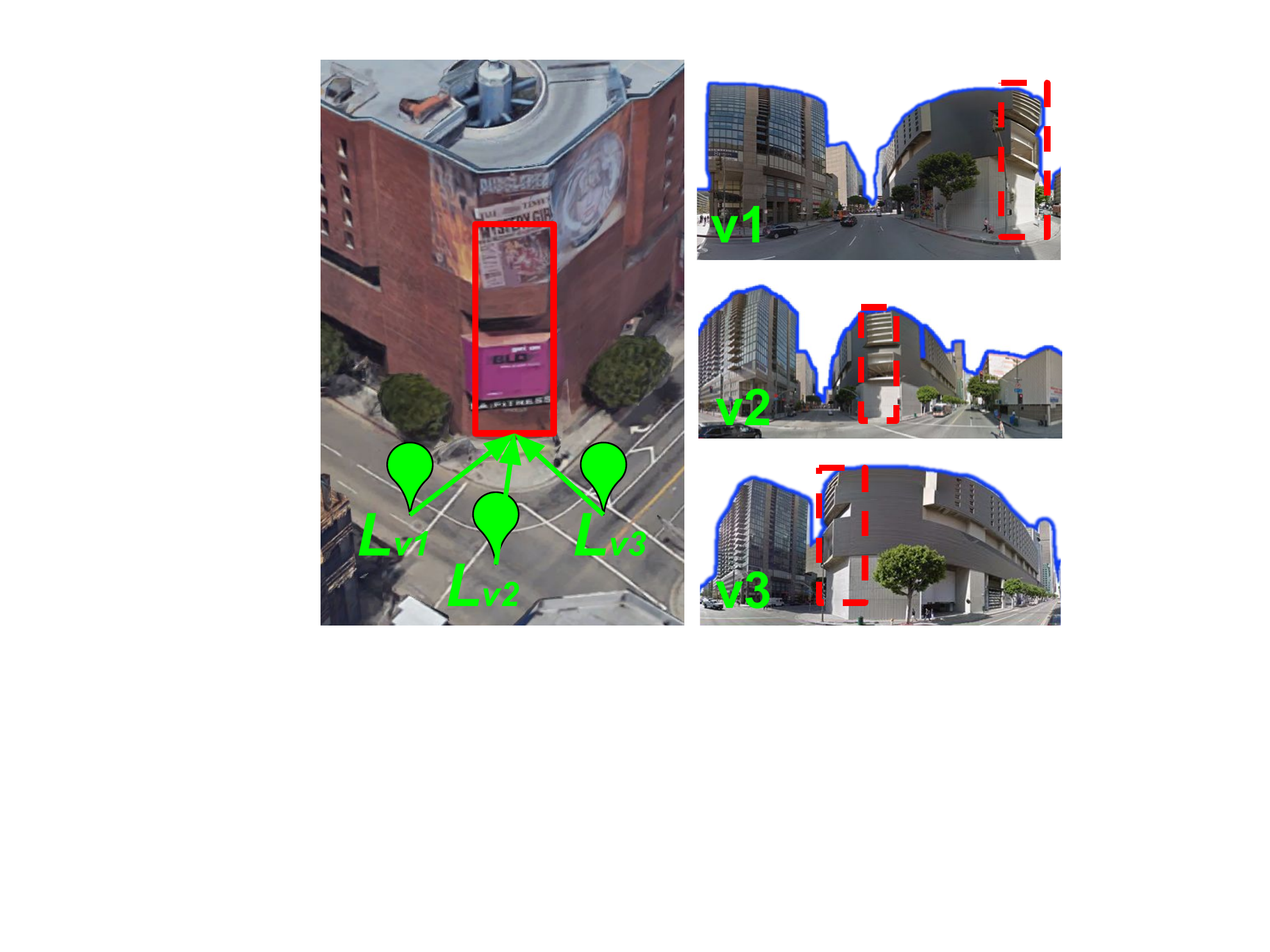}
\caption{\emph{For robust height estimation, Gnome uses three viewpoints to estimate height, then averages these estimates.}}
\label{fig:gnome_height_vp}
\end{figure}

We use simple geometry to solve this (\gnomefig{height_cal}(b)). Recall that the vector $D$ also encodes the distance $d$ from $L_{v}$ to the plane $L_{plane}$. To estimate the height, we need to estimate the angle of elevation of the intersection $\theta$. We estimate this using a property of Street View’s panorama images. Specifically, for a $W \times H$ Street View panorama image, a pixel at $(x,y)$ corresponds to a ray with azimuth $\frac{x}{W} \times \ang{360}$ and elevation $\frac{H/2 - y}{H} \times \ang{180}$. So, to estimate $\theta$, we first find the pixel(s) in the panoramic image corresponding to the intersection between the skyline and the plane. $\theta$ is then the elevation of that pixel, and we estimate the height of the building as $d \times tan(\theta)$. \gnomefig{height_res} shows the corrected height of all the planes in the surface geometry of one viewpoint.

In practice, because the target plane may be occluded by trees or other obstructions, we run the above procedure on three viewpoints (\gnomefig{height_vp}) near $L_{plane}$ and estimate the height of the building using the average of these three estimates.

At the end of this procedure, for a given location $L$, we will have an accurate surface geometry with better height estimates than those available with Google Street View.

\begin{figure}
\centering\includegraphics[width=0.6\columnwidth]{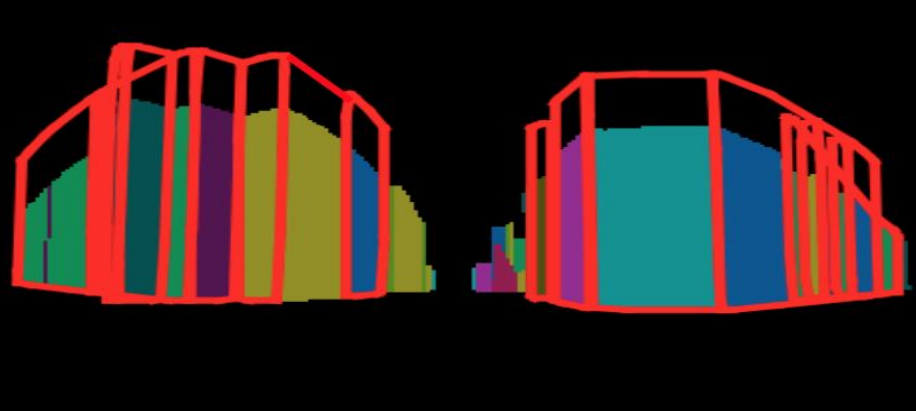}
\caption{\emph{The adjusted depth planes, augmented with the estimated height.}}
\label{fig:gnome_height_res}\vspace{-0.2in}
\end{figure}

\subsection{Estimating Path Inflation} 
To estimate the path inflation, Gnome uses ray tracing~\cite{ray_tracing}. Consider a satellite $S_i$ and a reflection plane $P_j$. As \gnomefig{reflection} shows, if the receiver is at position $R$, Gnome first computes its mirror point with respect to $P_j$, denoted by $R^\prime$. Then, it initializes a ray to $S_i$ starting from $R^\prime$ and intersects with $P_j$ at point $N_{i,j}$. For this ray to represent a valid reflection of a signal from $S_i$ to $R$ on plane $P_j$, three properties must hold. First, $N_{i,j}$ must be within the convex hull (or the boundary) of the plane $P_j$. Second, the ray $N_{i,j}$  to $S_i$ must not intersect any other plane. Finally, the ray from $N_{i,j}$ to $R$ must not intersect any other plane. 

For each ray that represents a valid reflection, Gnome employs geometry to calculate the path inflation (shown by the solid red arrows in \gnomefig{reflection}). For a given plane, a given satellite and a given receiver position, there can be at most one reflected ray. For a given receiver $R$, Gnome repeats this computation for every pair of $S_i$ and $P_j$, and obtains a path inflation in each case. It also computes whether $S_i$ is within line-of-sight of $R$: this is true if the ray from $S_i$ to $R$ does not intersect any other plane.

\begin{figure}
\centering\includegraphics[width=0.6\columnwidth]{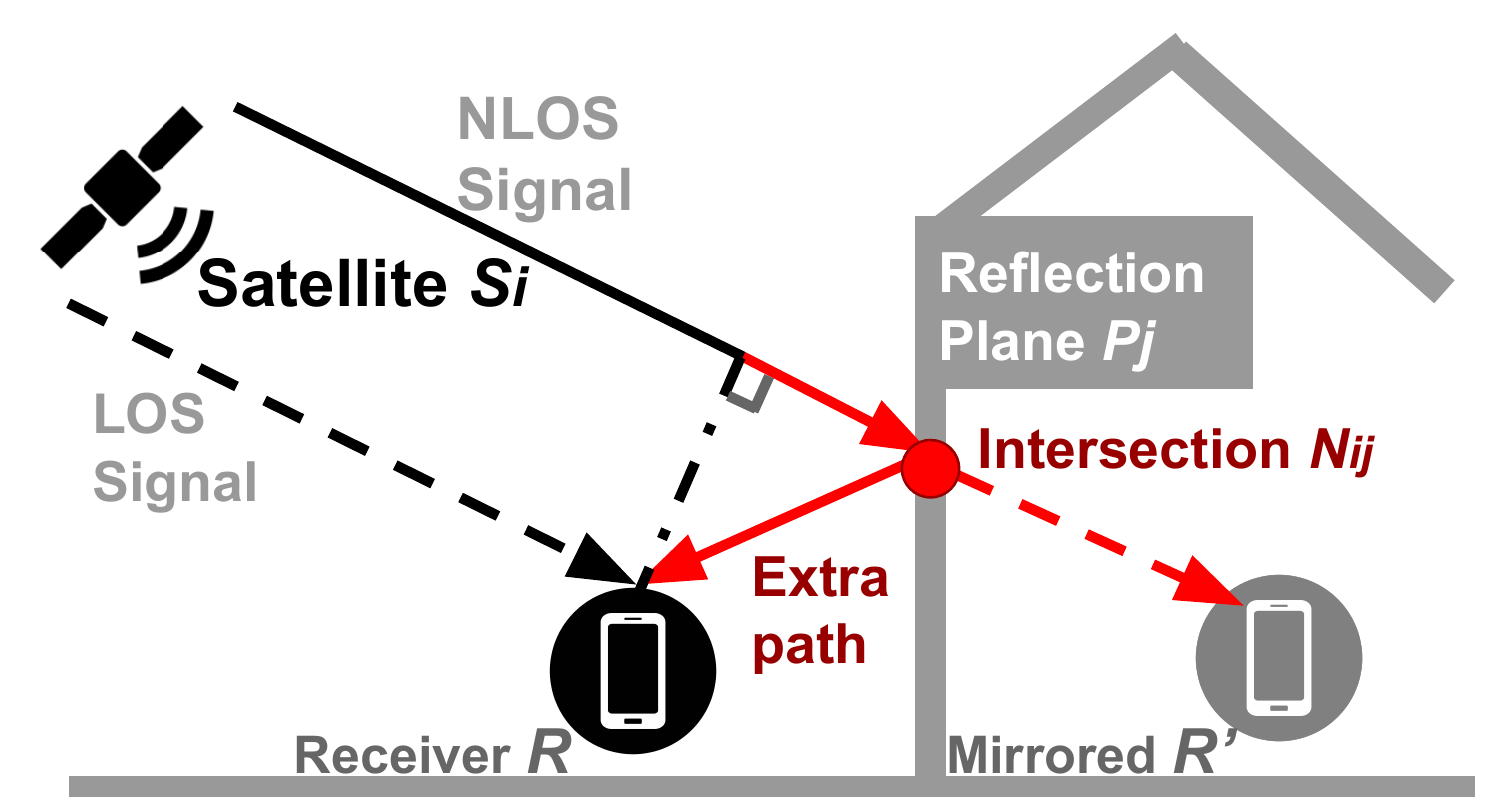}
\caption{\emph{An example of ray-tracing and path inflation calculation.}}
\label{fig:gnome_reflection}\vspace{-0.2in}
\end{figure}

These calculations can result in two possibilities for $S_i$, with respect to $R$. If the $S_i$ is within LOS of $R$, it will likely be the case that there may be one or more planes which provide valid reflections of the signal from $S_i$ to $R$. In this case, Gnome\ ignores these reflected signals, since they constitute multipath, and modern GPS receivers have multipath rejection capabilities. Thus, when there is a LOS path, path inflation is always assumed to be zero.

The second possibility is that $S_i$ is not within line-of-sight of $R$. In this case, $R$ can, in theory, receive multiple NLOS reflections from different planes. In practice, however, because of the geometry of buildings and streets, a receiver $R$ will often receive only one reflected signal. This is because building surfaces are either parallel or perpendicular to the street. If a street runs north-south, and a satellite is on the western sky, then, if the receiver is on the street, it will receive a reflection only from a plane on the east side of the street. However, in some cases, it is possible to receive more than one signal. In our example, if there is a gap between two tall buildings on the west side of the street, then it is possible for that signal to be reflected by a plane on one of those buildings perpendicular to the street (in addition to the reflection from the east side). In cases like these, Gnome computes path inflation using the plane nearer to $R$.

A signal from $S_i$ can, of course, be reflected off multiple planes before reaching $R$. In our example, the signal may first be reflected from a building on the east side, and then again from a building on the west side of the street. \added{If a signal can reach the receiver after a single reflection \textit{and} after a double reflection, the calculated pseudorange will be very close to the single-reflection trace because of multipath mitigation. If the signal can only reach the receiver after two reflections, the signal strength will be low (< 20dB) and will always be ignored in position computation~\cite{bilich2007mapping}. For this reason, Gnome only models single reflections to reduce computational complexity.} 

\subsection{Location Prediction}
In this section, we describe how Gnome uses the path inflation estimates to improve GPS positioning accuracy. Recall that the GPS receiver computes a position estimate from satellite metadata including pseudoranges for satellites. At a high level, Gnome\ subtracts the path inflation from the pseudorange for each satellite, then computes the new GPS position estimate.

However, the reported satellite pseudoranges correspond to the ground truth position of the receiver, which is not known! More precisely, let the estimated position be $L_e$ and the ground truth be $L_g$. The path inflation of satellite $S_i$ for these two points can be different. If we use the path inflation from $L_e$, but apply it to pseudoranges calculated at $L_g$, we will obtain incorrect position estimates.

\subsubsection{Searching Ground Truth Candidates}
To overcome this, Gnome selects several candidate positions within the vicinity of $L_e$ (we describe below how these candidate positions are chosen) and effectively tests whether a candidate location could be a viable candidate for $L_g$, the ground truth position. At each candidate $L_c$, it (a) reduces the pseudorange of each NLOS satellite by the computed path elevation and (b) recomputes the GPS position estimate with the revised pseudoranges. This gives a new position estimate ${L^\prime}_c$. Gnome chooses that ${L^\prime}_c$ (as the estimate for $L_g$) whose distance to its corresponding $L_c$ is least. \gnomefig{heatmap} shows the heatmap of the relative distances between $L_c$ and ${L^\prime}_c$ for candidates in a downtown area: notice how candidates closest to the ground truth have the low relative distances. In practice, for a reason described below, Gnome actually uses a voting strategy: it picks the five candidate positions with the lowest relative distance, clusters them, and uses the centroid of the cluster as the estimated position.

The intuition for this approach is as follows. When a candidate $L_c$ is close to (within a few meters of) the ground truth, its reflections are most likely to be correlated with the ground truth location $L_g$. In this case, the candidate’s estimated position ${L^\prime}_c$ will likely converge to the true ground truth position. Because $L_c$ is close to $L_g$, the distance between $L_c$ and ${L^\prime}_c$ will be small (e.g., candidate 1 in \gnomefig{localization}). However, when $L_c$ is far away from $L_g$, Gnome corrects the pseudoranges observed at $L_g$ with corrections appropriate for reflections observed at $L_c$. In this case, Gnome is likely to go astray since a LOS signal at $L_g$ may actually be an NLOS signal at the candidate position, or vice versa. In these cases, Gnome is likely to apply random corrections at the candidate positions (e.g., candidates 2 and 3 in \gnomefig{localization}). Because these corrections are random, the relative distances for distant candidate positions are unpredictable: they might range from small to large values. To filter randomly obtained small relative distances, Gnome uses the voting strategy described above. For example, in \gnomefig{heatmap}, the grid point at the bottom right of the heatmap shows up among the top three candidate positions with the lowest distance, for exactly this reason.

\begin{figure}
\centering\includegraphics[width=0.5\columnwidth]{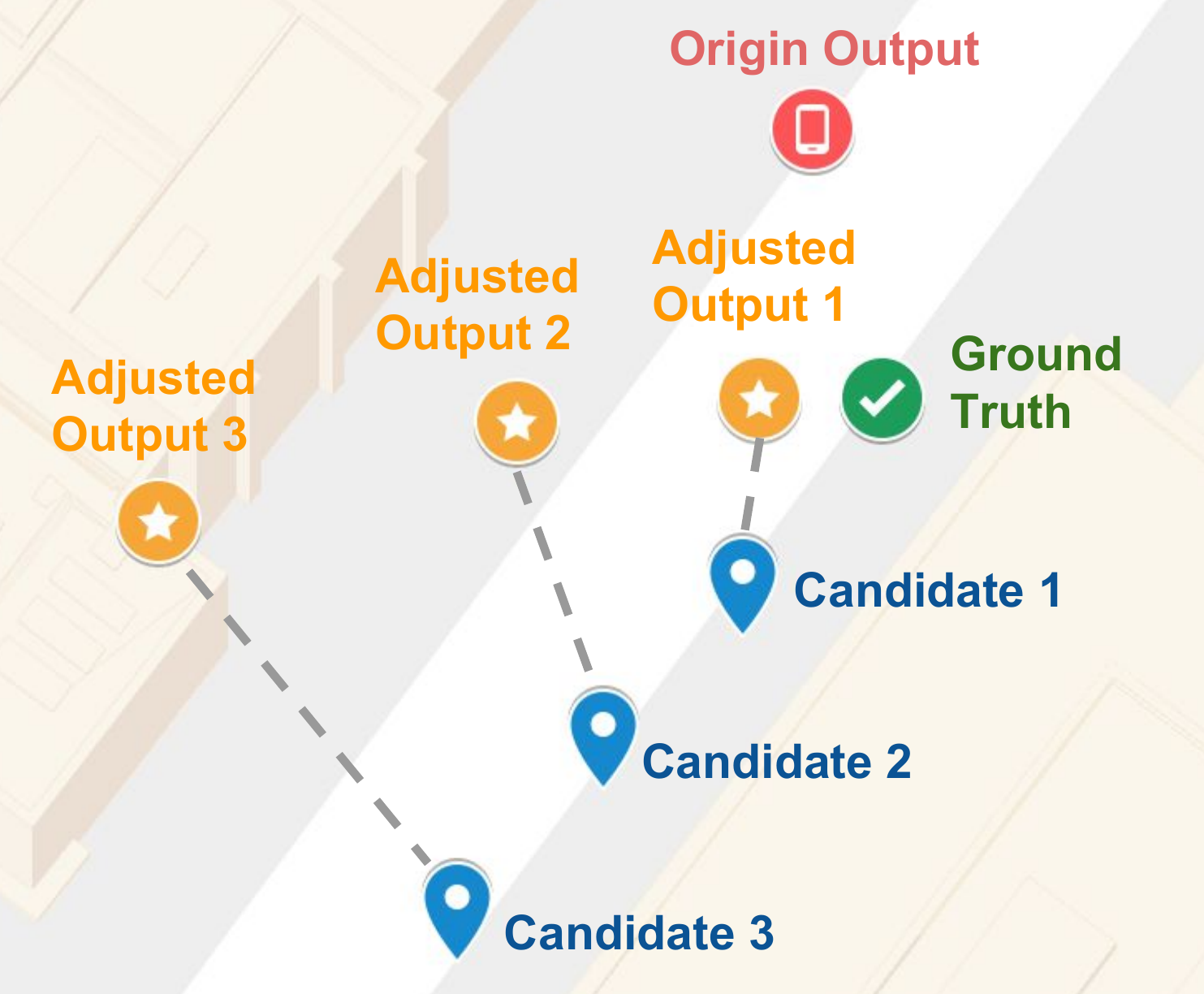}
\caption{\emph{After adjusting pseudoranges, candidate positions nearer the ground truth will have estimates that converge to the ground truth, while other candidate positions will have random corrections.}}
\label{fig:gnome_localization}\vspace{-0.2in}
\end{figure}

\subsubsection{Position Tracking}
As described until now, each predicted location is independent from the previous. However, most smartphone positioning applications require continuously tracking a user’s location. So, many navigation services, and, more generally, most localization algorithms for robots, drones or other mobile targets, smooth successive position estimates by using a Kalman filter. Gnome uses a similar technique to improve accuracy in tracking the smartphone user. Specifically, a Kalman filter treats the input state as an inaccurate observation and produces a statistically optimal estimate of the real system state. When Gnome computes a new location, its Kalman filter takes that location as input and outputs a revised estimate of the actual position. We refer the reader to~\cite{Kalman} for details on Kalman filtering.

\begin{figure}
\centering\includegraphics[width=0.5\columnwidth]{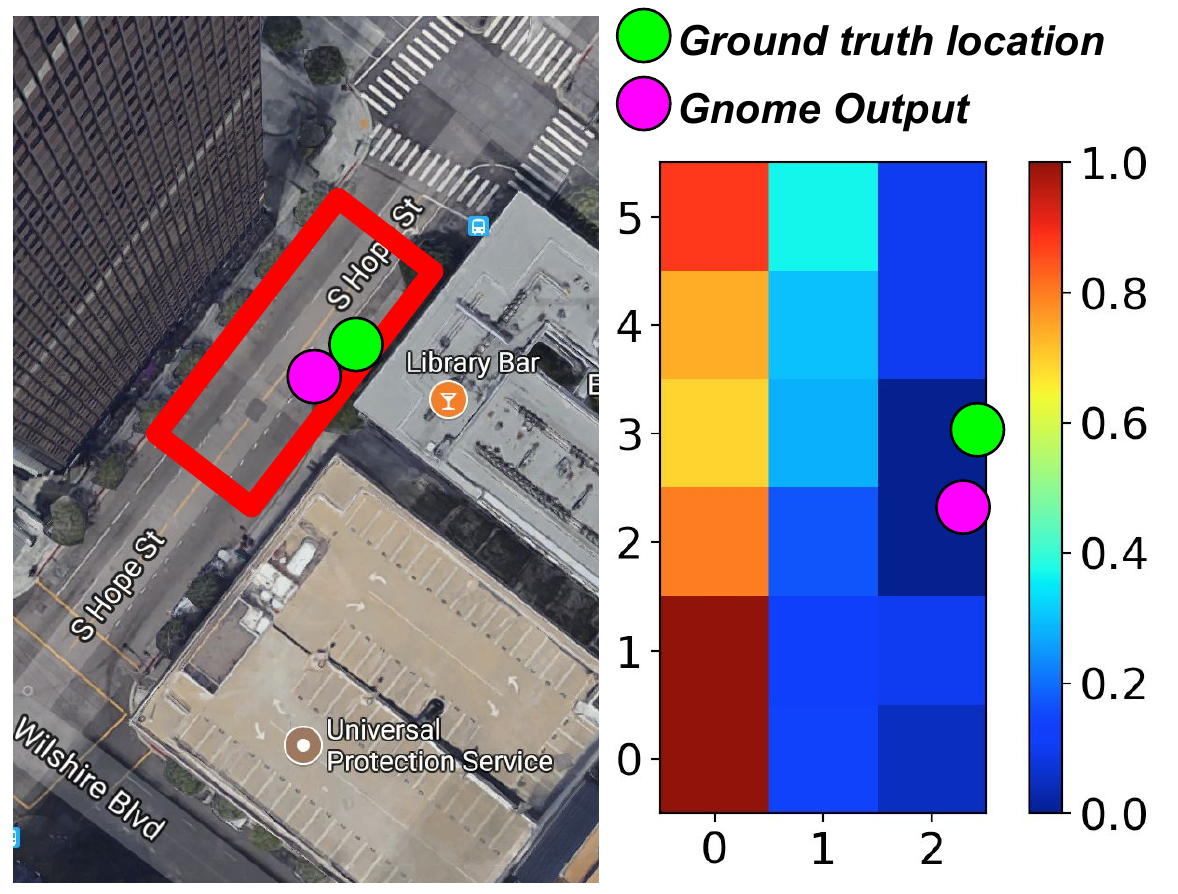}
\caption{\emph{Heatmap of the relative distance between candidate positions and the revised candidate positions. Candidates close to ground truth have small relative distances.}}
\label{fig:gnome_heatmap}\vspace{-0.2in}
\end{figure}

\subsection{Scaling Gnome}

As described, algorithms in Gnome to estimate building height and path inflation, as well as to predict location can be both compute and data intensive. 3D models can run into several tens of gigabytes, and ray-tracing is a computationally expensive operation. In this section, we describe how we architect Gnome to enable it to run on a smartphone. We use two techniques for this:  pre-processing in the cloud, and scoped refinement for candidate search.

\subsubsection{Preprocessing in the cloud}
Gnome preprocesses surface geometries in the cloud to produce path inflation maps. Because surface geometries are relatively static, these path-inflation maps can be computed once and reused by all users.

The input to this preprocessing step is a geographic area containing urban canyons (e.g., the downtown area of a large city). Given this input, Gnome first builds the 3D model of the entire area. To do this, it first downloads street positions and widths from OpenStreetMaps~\cite{osm}, and then retrieves Street View images and 3D models every 5m or so along every street in the area, similar to the technique used in~\cite{hu2016alps}. At each retrieval location, it augments its 3D model with the adjusted building height. At the end of this process, Gnome has a database of 3D models for each retrieval point.

In the second step of pre-processing, Gnome covers every street and sidewalk with a fine grid of candidate positions. These candidate positions are used in the location prediction algorithm. Our implementation uses a grid size of 2m $\times$ 2m. For each candidate position, Gnome pre-computes the path inflation for every possible satellite position. Specifically, it does this by using the pre-computed 3D models in the previous step and does ray tracing for every point (at a resolution of 1$^\circ$ in azimuth and elevation) in a hemisphere centered at the candidate position to determine the path inflation.

The output of these two steps is a path inflation map: for each candidate position, this map contains the path inflation from every possible satellite position. This map is pre-loaded onto the smartphone. The map captures the static reflective environment around the candidate position. This greatly simplifies the processing on the smartphone: when Gnome needs to adjust the pseudorange for satellite $S_i$ at candidate position $L_c$, it determines $S_i$’s location from the satellite metadata sent as part of the GPS, and uses that to determine the path inflation from $L_c$’s path inflation map.

\subsubsection{Scoped refinement for candidate search}
Even with path inflation maps, Gnome can require significant overhead on a smartphone because it has to search a potentially large number of candidate positions in its location prediction phase. Gnome first scopes the candidate positions to be within the error range reported by the GPS device --- recall that GPS receivers report a position estimate and an error radius. In urban canyons, however, the radius can be large and include hundreds or thousands of candidate positions. To further optimize search efficiency, we use a coarse-to-fine refinement strategy. We first consider candidate positions at a coarser granularity (e.g., one candidate position in every 8m $\times$ 8m grid) and select the best candidate. We then repeat the search on the finer spatial scale around the best candidate selected in the previous step. This reduces computational overhead by 20$\times$.

%% file: tex/gnome/eval.tex
\section{Evaluation}

Using a full-fledged implementation of Gnome, we evaluate its accuracy improvement in 4 major cities in North America, Europe and Asia. We also quantify the impact of individual design choices, and the overhead incurred in estimating building height, path inflation, and in location prediction.

\subsection{Methodology}

\subsubsection{Implementation}
Our implementation of Gnome has two components, one on Android and the other on the cloud and requires 2100 lines of code in total. The cloud-side component performs data retrieval (for both depth information and Street View images), height adjustment, and path inflation computations. The smartphone component pre-loads path inflation maps and performs pseudorange adjustments at each candidate position, recomputes the revised location estimate using the Ublox API, and performs voting to obtain the estimated location. 
\added{Currently, Gnome directly outputs its readings to a file. Without being rooted, Android does not permit Gnome to be run in the background by another app. Apps would have to incorporate its source code in order to use Gnome.}

\subsubsection{Metrics}
We measured several aspects of Gnome including: positioning accuracy measured as the average distance between estimated position and ground truth; processing latency of various components, both on the cloud and mobile; power consumption on the smartphone; and storage usage on the smartphone for the path inflation maps.

\subsubsection{Scenarios and Ground Truth collection}
To evaluate accuracy, we take measurements using Gnome in the downtown areas of four major cities in three continents: \changed{Los Angeles, New York, Frankfurt, and Hong Kong}. In most of these cities, we have measurements from a smartphone carried by a pedestrian. These devices include Huawei Mate10, Google Pixel, Samsung S8, and Samsung Note 8. In Los Angeles, we also have measurements from a smartphone on a vehicle, and from a smartphone on a stationary user. In the same city, we have measurements both on an Android device and an iPhone. Across our four cities, the total walking distance is 4.7km and the total driving distance is 9.3km. \changed{\gnomefig{exp_paths} shows the pedestrian traces collected in the four cities.}

\begin{figure}
\centering\includegraphics[width=0.95\columnwidth]{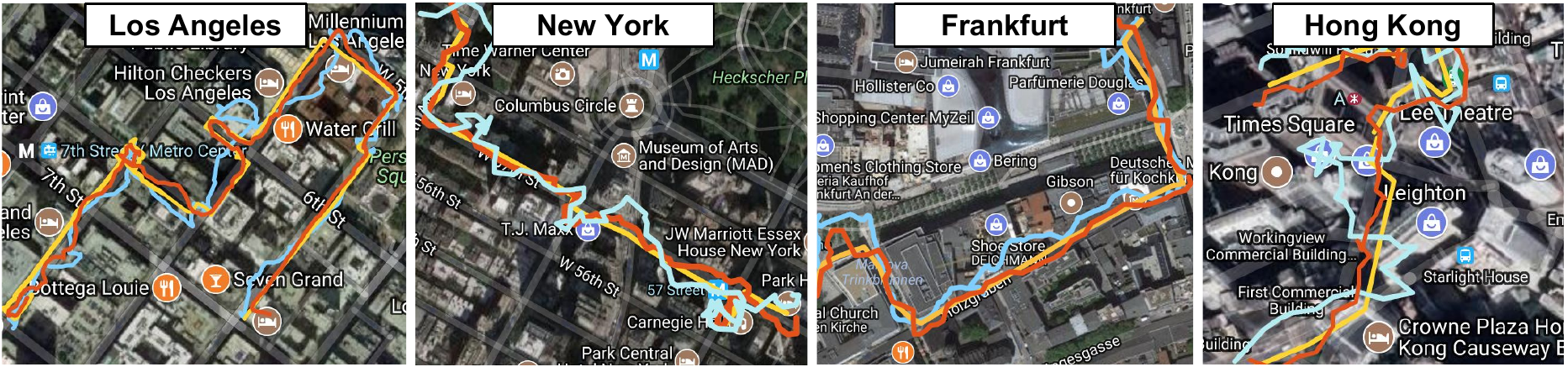}
\caption{\emph{The ground truth traces (yellow), Android output (blue), and Gnome output (red) in the four cities.}}
\label{fig:gnome_exp_paths}
\end{figure}

For the stationary user, we placed the phone at \changed{ten fixed known locations for 1 minute in each} and recorded the locations output by Gnome app. For the pedestrian experiment, tester $A$ walks while holding the phone running the Gnome app. Meanwhile, another tester $B$ follows $A$ and takes a video of $A$. We use the video to manually determine the ground truth position of $A$: we pinpoint this location manually on a map to determine the ground truth. For the driving experiment, we place the phone in car’s cup holder and drive in the target area. To collect the ground truth location of the car, we use a somewhat unusual technique: we attach a stereo camera ~\cite{stereo_cam} to the car and use the 3D car localization algorithm described in~\cite{qiu2017augmented}. The algorithm can estimate ground truth positions with sub-meter accuracy, sufficient for our purposes.

\subsection{Results}
Before we describe our results, we describe some statistics about satellite visibility and path inflation in our dataset. These results quantitatively motivate the need for Gnome, and also give some context for our results.

In our data, only 14.4\% of the observation points can see more than four LOS satellites, and only 27.5\% of all the received satellites are LOS satellites (\gnomefig{nlos_stat}). Thus, urban canyons in large cities contain significant dead spots for GPS signal reception. This also motivates our design decision to not omit NLOS satellites as other work has proposed: since four satellites are required for position fixes, omitting an NLOS satellite would render unusable nearly 85\% of our readings.

\begin{figure}
\centering\includegraphics[width=0.6\columnwidth]{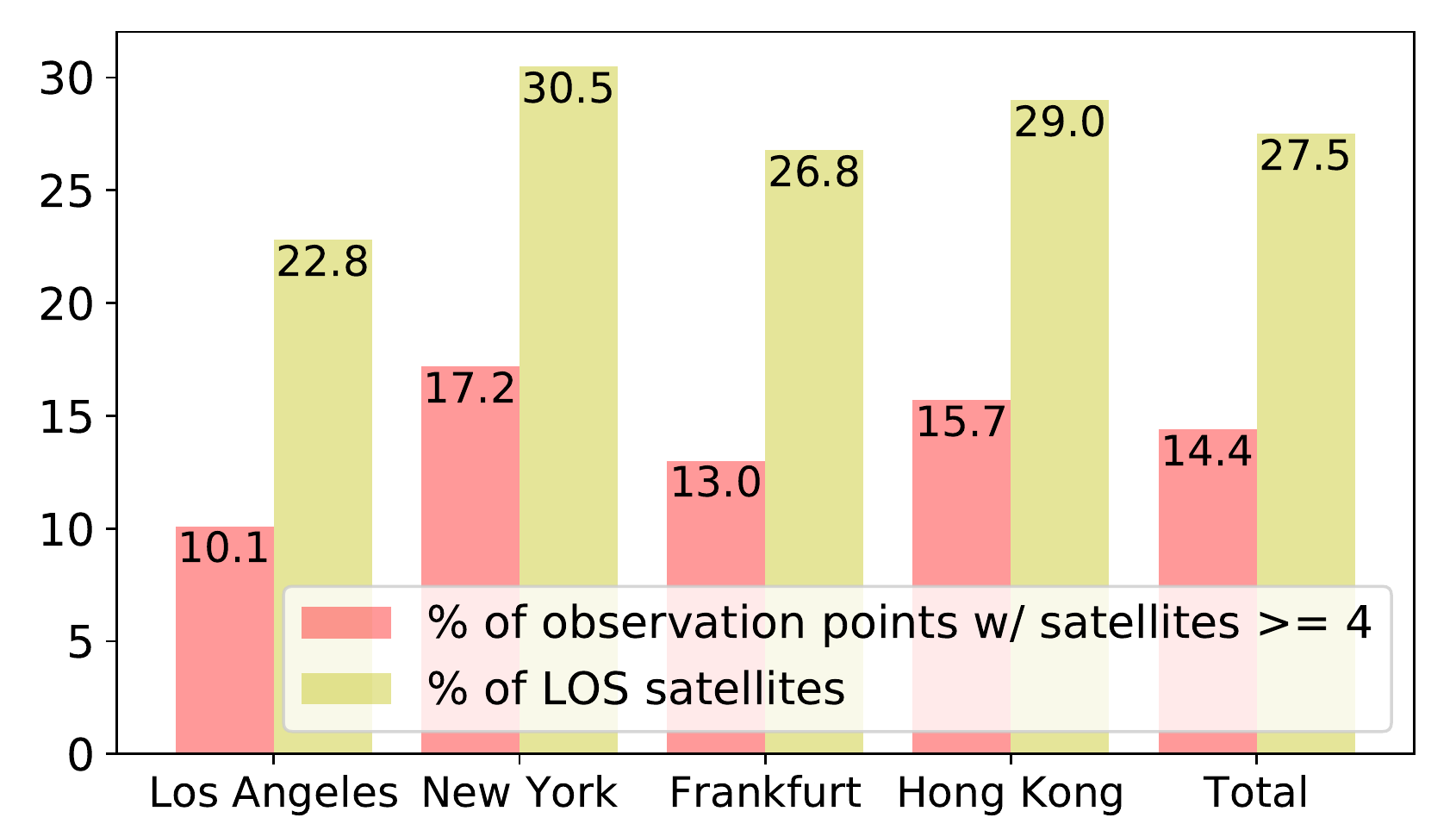}
\caption{\emph{Statistics of NLOS signals in the dataset}}
\label{fig:gnome_nlos_stat}
\end{figure}

\gnomefig{prange_cdf} shows the CDF of path inflation across the 4 cities in our dataset. Depending on the city, between 10\% and 15\% of the observations incur a path inflation of more than 50m. For two of the cities, 7-8\% incur a path inflation of over 100m \added{because the signal is reflected from a building plane that is far from the receiver and therefore leads to a large increase in pseudorange}.  This suggests that correcting these inflated paths can improve accuracy significantly, as we describe next.

\begin{figure}
\centering\includegraphics[width=0.5\columnwidth]{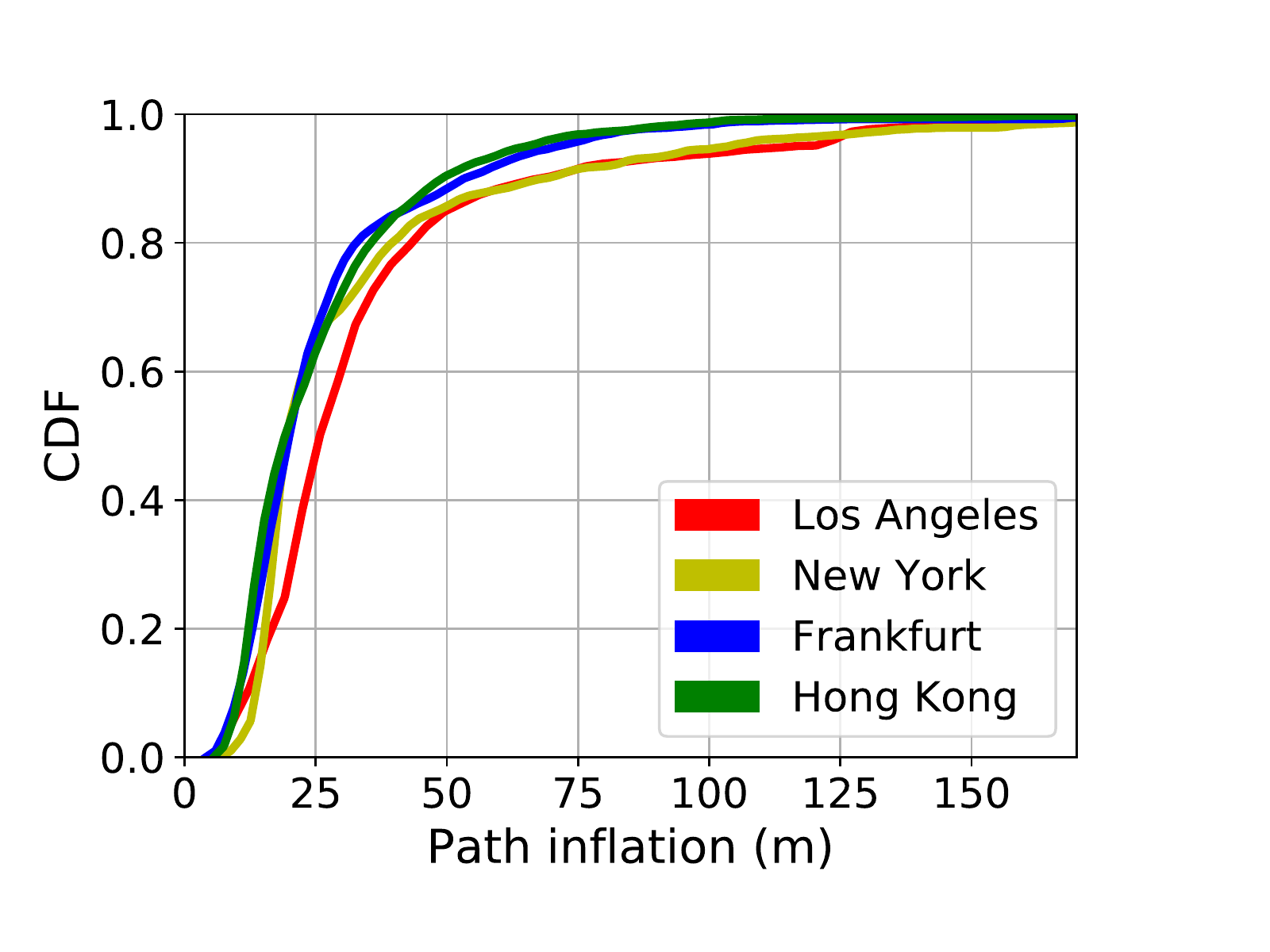}
\caption{\emph{Extra NLOS signal path distribution.}}
\label{fig:gnome_prange_cdf}
\end{figure}

\subsubsection{Accuracy}
\gnomefig{accuracy_bar} compares the average positioning error of Gnome, Android, and iPhone in four settings (stationary, walking, cycling, driving) for \losangeles. \added{We use the track recording app for Android~\cite{android_tracker} and iPhone~\cite{ios_tracker} as location recorders.} In the stationary setting, Gnome incurs an average error of less than 5m while Android and iOS incur more than twice that error. This setting directly quantifies the benefits of compensating for path inflation. More important, this shows how much of an improvement is still left on the table after all the optimizations that smartphones incorporate, include cell tower and Wi-Fi positioning, and map matching~\cite{mapmatching}.  The gains in the pedestrian setting are relatively high: Gnome incurs only a 6.4m error, while the iPhone’s error is the highest at 14.4m. In this setting, in addition to compensating for path inflation, Gnome also benefits from trajectory smoothing using Kalman filters.

Gnome performance while driving in \losangeles is comparable to that of Android and iOS. We conjecture that this is because today’s smartphones use dead-reckoning based on inertial sensors~\cite{deadreckoning}, and this appears to largely mask inaccuracy due to NLOS satellites. These benefits aren’t evident in our walking experiments, where the phone movement likely makes it difficult to dead-reckon using accelerometers and gyroscopes.

\begin{figure}
\centering\includegraphics[width=0.6\columnwidth]{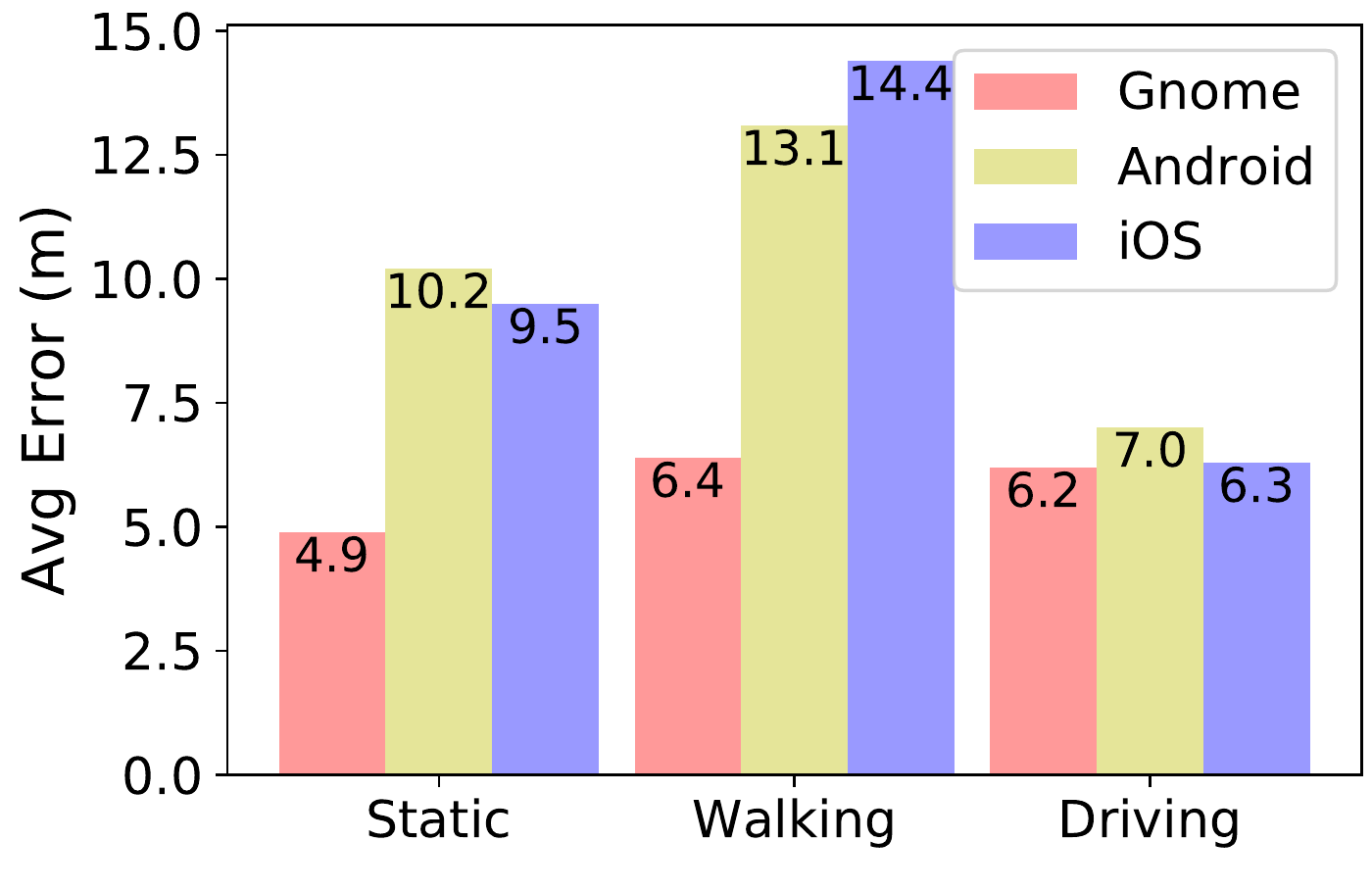}
\caption{\emph{Average positioning accuracy in different scenarios.}}
\label{fig:gnome_accuracy_bar}
\end{figure}

Across our other cities also (\gnomefig{accuracy_world}), Gnome consistently shows better performance than Android localization. In \nyc, Gnome obtains a 30\% reduction in error, in \frankfurt, a more than 38\% error reduction, and in \hongkong, a more than 40\% reduction. Equally important, this shows that the methodology of Gnome is generalizable. In obtaining these results, we did not have to modify the Gnome processing pipeline in any way. As described before, the Street View 3D models are available for many major cities across the world, which is the most crucial source of data for Gnome, so Gnome is applicable across a large part of the globe. 

\added{The maximum localization errors for Gnome for the four cities are 37m, 35m, 41m, and 29m respectively. In comparison, the maximum errors for Android are 51m, 42m, 45m, and 47m. These occur either because (a) the ground truth location is not within the error radius reported by the GPS receiver, so Gnome cannot generate candidate points near the ground truth location, or (b) because the error radius is too large and includes some non-ground-truth candidates where the distance from the candidate point to the pseudorange-adjusted position is short. The latter confuses Gnome’s voting mechanism of candidate selection. In future work, we plan to explore mitigations for these corner cases.}

\begin{figure}
\centering\includegraphics[width=0.6\columnwidth]{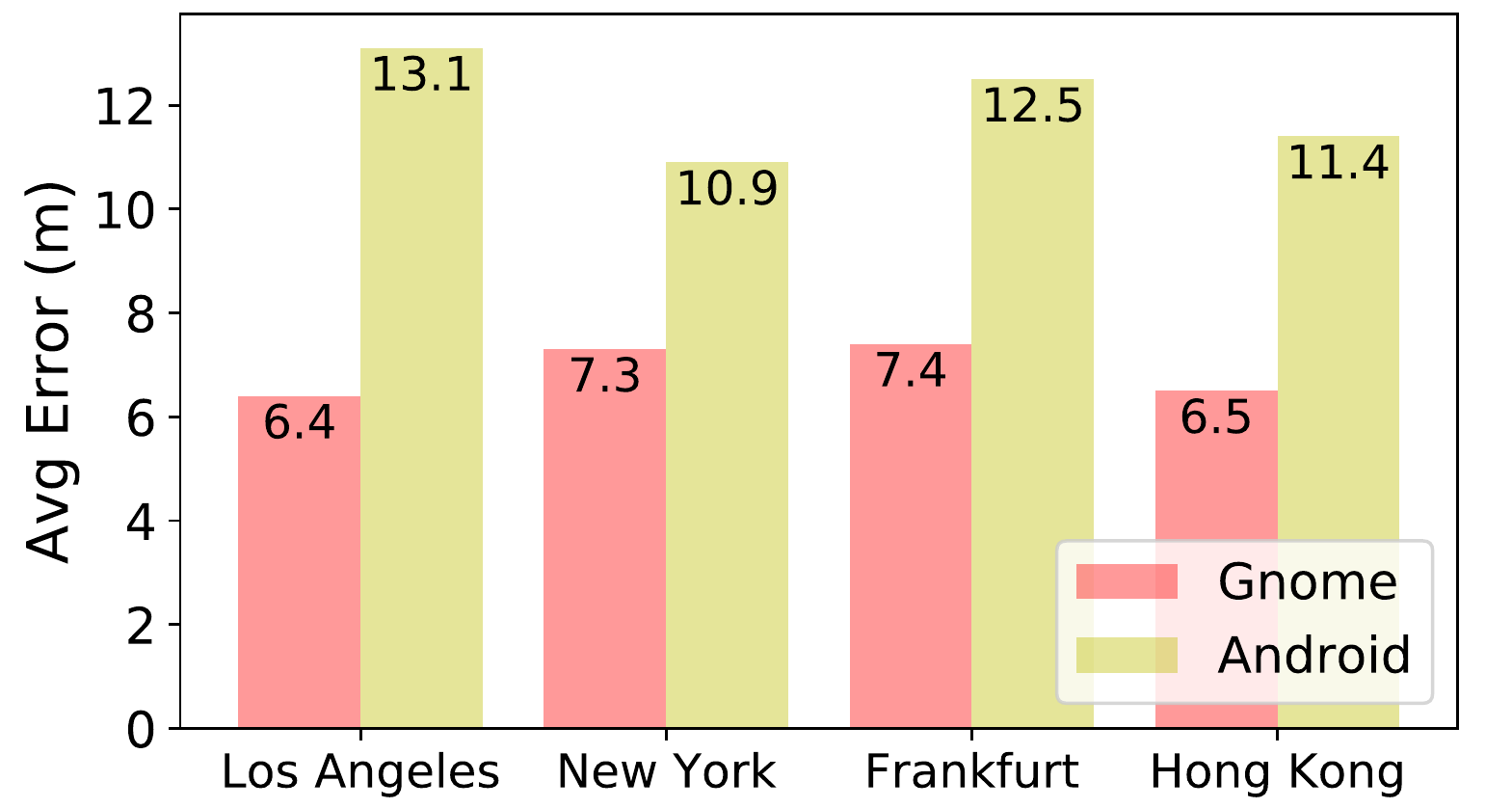}
\caption{\emph{Average accuracy in different cities.}}
\label{fig:gnome_accuracy_world}
\end{figure}

\subsubsection{Processing latency}
Gnome performs some computationally expensive vision and graphics algorithms. Fortunately, as we now show below, the latency of these computations is not on the critical path and much of the expensive computation happens on the cloud. (In this chapter, we have used the term “cloud” as a proxy for server-class computing resources. In fact, our experiments were carried out on a 12-core 2.4Ghz Xeon desktop with 32GB of memory and running Ubuntu 16.04). 

\begin{table}[htbp]
\centering
\begin{tabular}{|l|l|} 
\hline
\textbf{Module} & \textbf{Latency} \\ \hline
\textit{Mobile: Client-side positioning} & 77 ms/estimate \\ \hline
\textit{Cloud: Street View data retrieval} & 1.9 s/viewpoint  \\ \hline
\textit{Cloud: Height adjustment}  & 2.1 s/viewpoint \\ \hline
\textit{Cloud: Path inflation calculation}  & 28 s/candidate \\ \hline
\end{tabular}
\caption{Processing latency measurement}
\label{latency_table}
\end{table}

Table \ref{latency_table} summarizes the processing latency of several components of Gnome. The critical path of position estimation (which involves scoped refinement based candidate search) on the smartphone incurs only 77ms. \added{Thus, Gnome can support up to a 13Hz GPS sampling rate, which is faster than default location update rates (10 Hz) on both Android and iOS.}  Retrieval of depth data (a few KBs) and a panoramic image (about 450KB) for each viewpoint (recall that Gnome samples these at the granularity of 5m) takes a little under 2s, while adjusting the height of planes at each viewpoint takes an additional 2s. By far the most expensive operation (29s) is computing the path inflation for each candidate position: this requires ray tracing for all points in a hemisphere around the candidate position. This also explains why we only compute single reflections: considering multiple reflections would significantly increase the computational cost. The candidate positions are arranged in a 2m $\times$ 2m grid, and it takes about 39 minutes to compute the path inflation maps for a 1-km street, or about 17 hours for the downtown area of \losangeles which has 26.8km of roads. It is important to remember that Gnome’s path inflation maps are compute-once-use-often: a path inflation map for an urban canyon in a major city need only be computed once. \newlyadded{We choose the 2$\times$2 candidate scale because it achieves the best balance between the accuracy and runtime latency.}

\subsubsection{Power consumption}
In Android 7.0, the battery option in “Settings” provides detailed per-hour power reports for the top-5 highest power usage apps. Gnome is implemented as an app, so we obtain its power consumption by running it for \changed{three hours}. While doing so, we run the app in the foreground with screen brightness set to the lowest. \changed{We compare Gnome’s power consumption with that of the default Android location API (implemented as a simple app), and our results are averaged over multiple runs. During our experiment, Gnome app consumes 151 mAh and is 31\% higher than the default Android location API, which consumes 112 mAh. Most of the additional energy usage is attributable to the UBlox library that computes adjusted locations. All the four phones we tested have batteries larger than 2700mAh, so Gnome would deplete the battery by about 5.5\% every hour if used continuously. Our implementation is relatively unoptimized (Gnome is implemented in Python), and we plan to improve Gnome’s energy efficiency in future work.}

\subsubsection{Storage usage}
We have generated path inflation maps for the downtown area of \losangeles, a 3.9$km^2$ area. The total length of road is 26.8km and the average road width (road width is used for sampling candidate positions) is 21.3m. In this area, there are 2531 Street View viewpoints and Gnome generates 16390 candidate positions at a 2m $\times$ 2m granularity. The total size of the path inflation maps for \losangeles is 340 MB in compressed binary format. While this is significant, smartphone storage has been increasing in recent years, and we envision most users loading path inflation maps only for downtown areas of the city they live in.

\subsection{Evaluating Gnome components}
Each component of Gnome is crucial to its accuracy. We evaluate several components for \losangeles.

\subsubsection{Height adjustment}
In \losangeles, there are nearly 15,000 planes of which about 30\% need height adjustment. To understand how the height adjustment affects the final localization accuracy, we compute the accuracy when Gnome selects different random subsets of planes for which to perform height adjustment. In \gnomefig{height_stat}, the x-axis represents the fraction of planes for which height adjustment is performed. In our experiment, for each data point, we repeated the random selection five times, and the figure shows the maximum and minimum values for each data point. Height adjustment is responsible for up to a 3m reduction in error.

\begin{figure}
\centering\includegraphics[width=0.5\columnwidth]{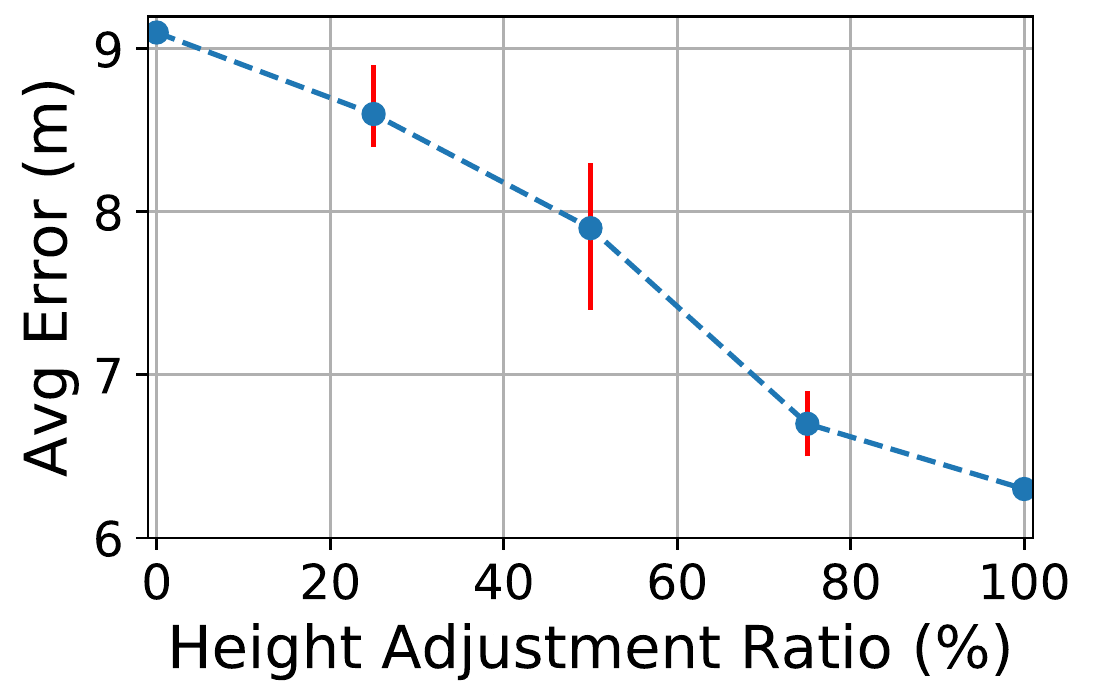}
\caption{\emph{Accuracy with different level of height adjustment.}}
\label{fig:gnome_height_stat}
\end{figure}

Finally, our height estimates themselves can be erroneous. We compared our estimated height with the actual height for \added{50 randomly selected buildings in \losangeles, whose heights are publicly available. Our height estimates are correct to within 5\% at the median and within 14\% at the 90th percentile.} Our position estimation accuracy is largely due to the fact that we are able to estimate heights of buildings quite accurately. \newlyadded{There are two causes for the height estimation error: (1) the inaccurate sky detection could recognize the building’s top as part of the sky, which makes the estimated height lower than the actual value, and (2) massive obstacles like trees and taller buildings behind the target plane will cause larger height measurement. The largest errors in both cases are 16\% and 23\%.}

\begin{figure}
\centering\includegraphics[width=0.5\columnwidth]{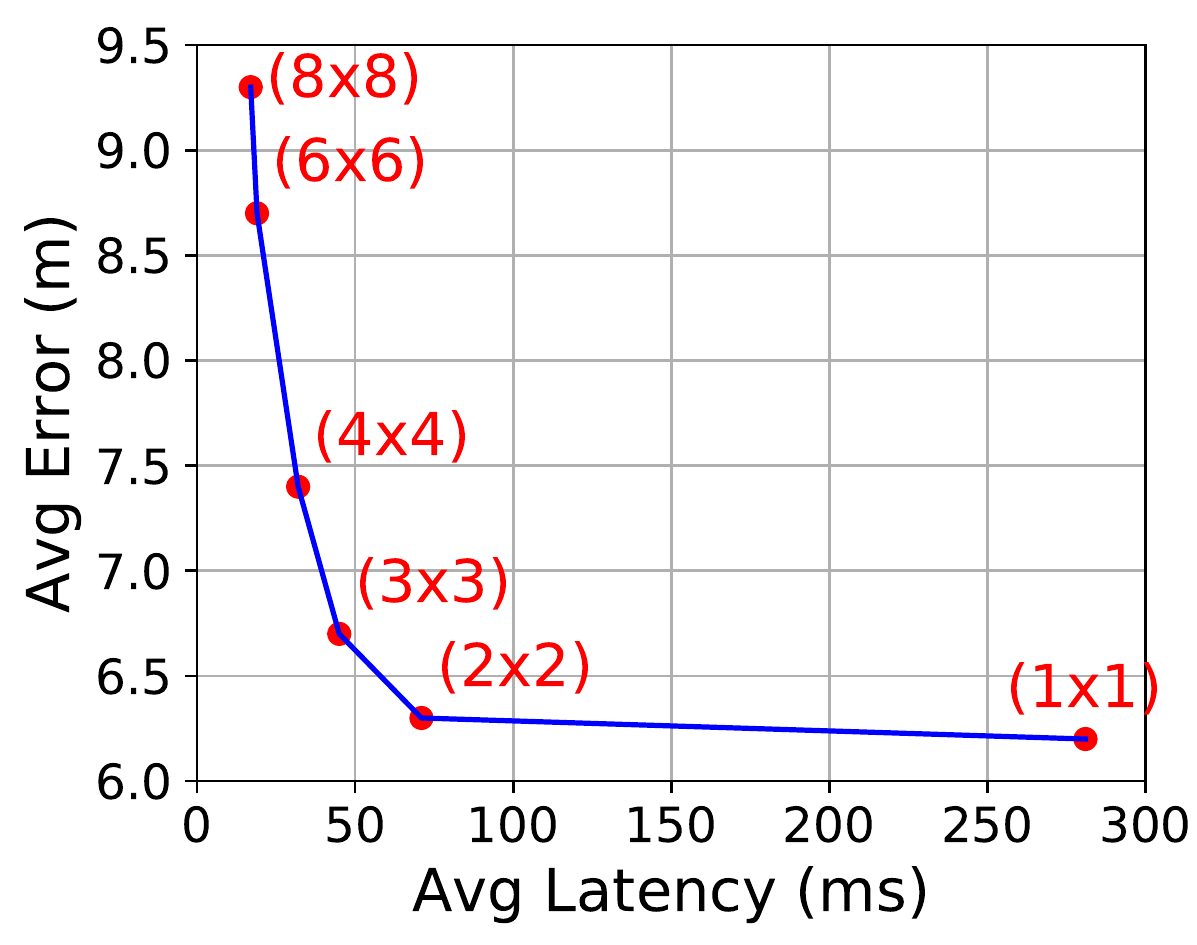}
\caption{\emph{Latency vs Accuracy with different grid sizes.}}
\label{fig:gnome_grid_size_2}
\end{figure}

\subsubsection{Candidate selection and ranking }
Candidate position granularity can also impact the error. Gnome uses a 2m $\times$ 2m grid. Using a coarser 8m $\times$ 8m grid would reduce storage requirements by a factor of 16 and could reduce the processing latency on the smartphone. \gnomefig{grid_size_2} captures the tradeoff between candidate granularity, accuracy and processing latency. As the figure shows, candidate selection granularity can significantly impact accuracy: an 8m $\times$ 8m grid would add almost 3m error to Gnome while reducing processing latency by about 50ms. Our choice of granularity is at the point of diminishing returns: a finer grid of 1m $\times$ 1m would more than triple processing latency while reducing the error by about 10cm.

\begin{figure}
\centering\includegraphics[width=0.5\columnwidth]{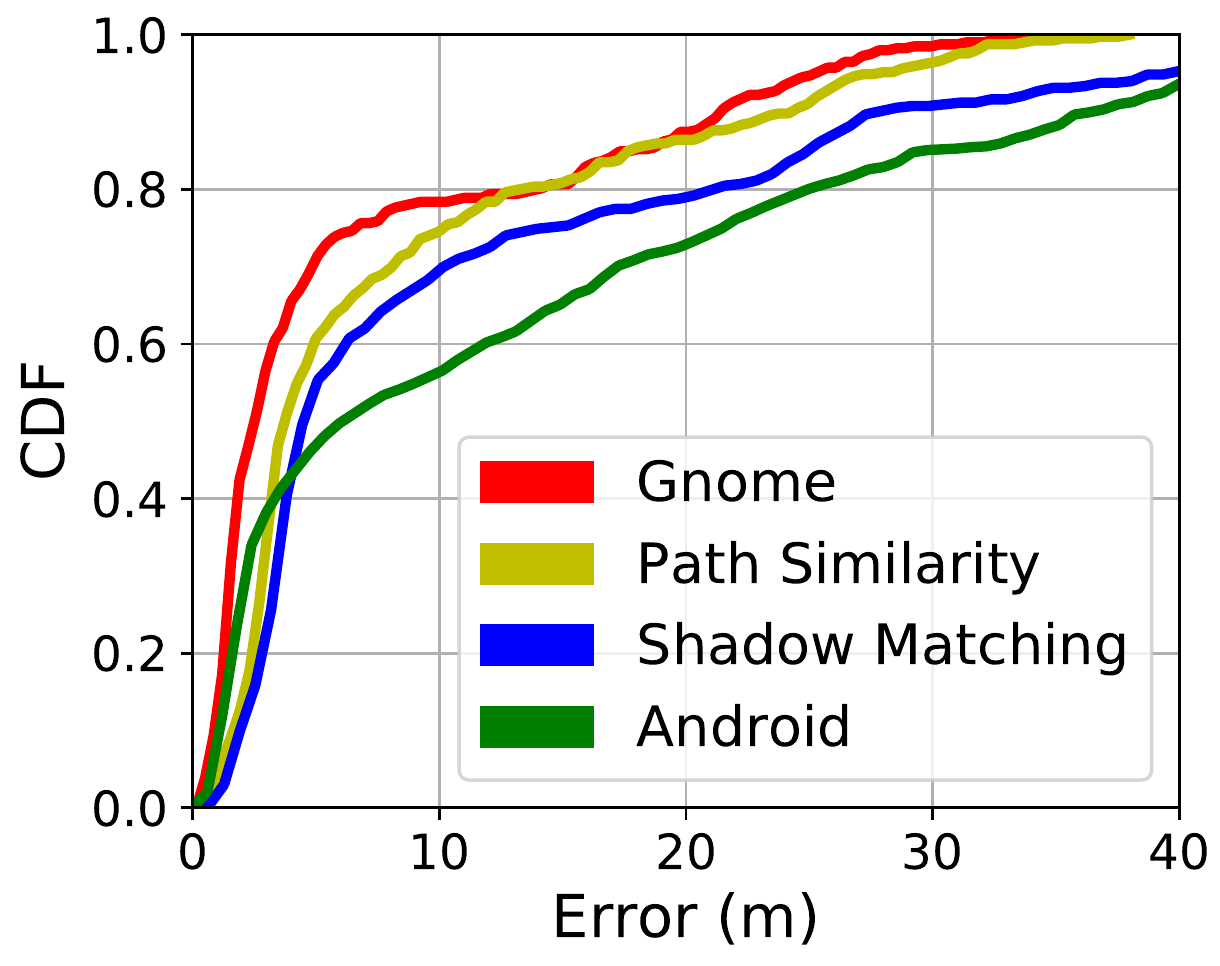}
\caption{\emph{Accuracy of different candidate ranking approaches.}}
\label{fig:gnome_err_comp}
\end{figure}

Position accuracy is also a function of the candidate search strategy. Prior work has considered two different approaches. The first~\cite{sahmoudi2014deep, miura2015gps, hsu20163d} is based on path similarity. This line of work uses ray-tracing to simulate the signal path and calculate the difference between the simulated one and the actual travel distance calculated by GPS module. The candidate whose path difference is least is selected as the output. The second approach is called shadow matching~\cite{groves2011shadow, wang2015smartphone, adjrad2015enhancing}. It uses satellite visibility as ranking indicator and assumes that NLOS signal always have worse carrier-to-noise density $C/N_0$ than LOS signal. It uses $C/N_0$ to classify each satellite’s visibility at the ground truth point and compares that with simulated (from 3D models) satellite visibility at each candidate point. The point with the highest similarity is the estimated position. \gnomefig{err_comp} shows the error distribution in \losangeles: we include the Android error distribution for calibration. While all approaches improve upon Android, Gnome is distributionally better than the other two approaches. The crucial difference between Gnome and path similarity is that Gnome revises the candidate positions using the path inflation maps, and this appears to give it some advantage. Gnome is better than shadow matching because carrier-to-noise-density is not a good predictor of NLOS signals. \added{In addition to better localization, Gnome is more practical than these prior approaches in three ways. First, these approaches use proprietary 3D models for ray-tracing, which may not be widely available for many cities. In Gnome, we use Google Street View which is available for most cities (as shown in our evaluation). Second, these approaches work offline and do not explore efficient online implementations. For instance, path simulation in~\cite{sahmoudi2014deep, miura2015gps, hsu20163d} can take up to ~ 2 sec on a desktop, rendering them impractical for mobile devices. Finally, these approaches rely on an external GPS receiver (UBlox) while Gnome is implemented on Android phones.}

\begin{figure}
\centering\includegraphics[width=0.5\columnwidth]{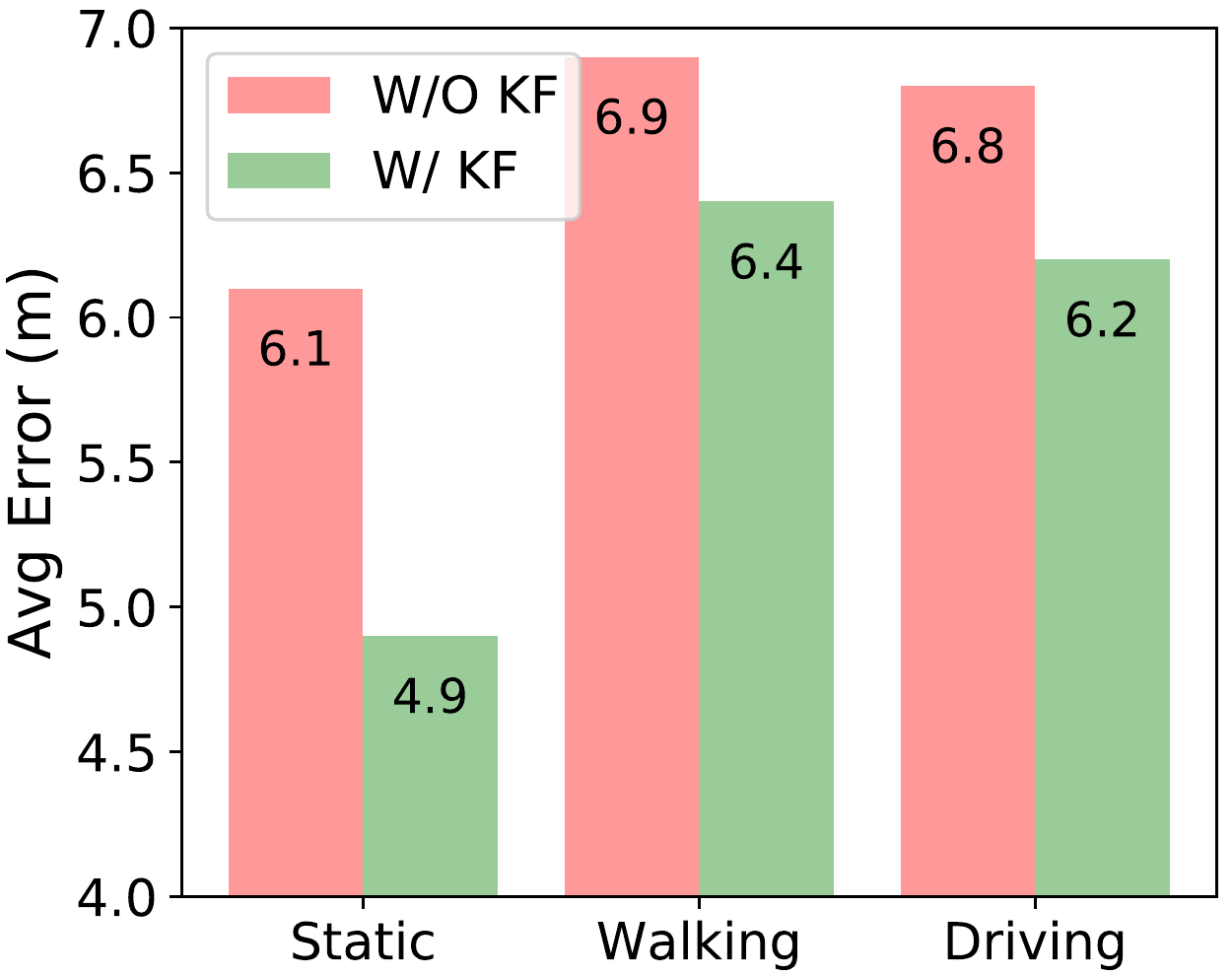}
\caption{\emph{Effectiveness of Kalman filter.}}
\label{fig:gnome_kalman_eval}
\end{figure}

\subsubsection{Kalman filter}
We also disabled the Kalman filter in Gnome to evaluate its contribution to the final accuracy. As \gnomefig{kalman_eval} shows, Kalman filter improves accuracy by 1.2m for the stationary case and about 0.5m in the other two scenarios. The image on the right shows how the Kalman filter results in a smoother trace closer to the ground truth.

%% file: tex/paper_alps.tex
\chapter{ALPS: Accurate Landmark Positioning at City Scales}\label{chap:alps}

\input{tex/alps/intro}
\input{tex/alps/motivation}
\input{tex/alps/design}

\input{tex/alps/eval}

%% file: tex/alps/intro.tex
\section{Introduction}

Context awareness is essential for ubiquitous computing, and prior work~\cite{Barometer, IndoorLoc} has studied automated methods to detect objects in the environment or determine their precise position. One type of object that has received relatively limited attention is the \textit{common landmark}, an easily recognizable outdoor object which can provide contextual cues. Examples of common landmarks include retail storefronts, signposts (stop signs, speed limits), and other structures (hydrants, street lights, light poles). These can help improve targeted advertising, vehicular safety, and the efficiency of city governments.

In this chapter, we explore the following problem: How can we \textit{automatically} collect an accurate database of the precise positions of common landmarks, at the scale of a large city or metropolitan area? The context aware applications described above require an accurate database that also has high coverage: imprecise locations, or spotty coverage, can diminish the utility of such applications. 

In this chapter, we discuss the design of a system called ALPS (Accurate Landmark Positioning at city Scales), which, given a set of landmark types of interest (\textit{e.g.}, Subway restaurant, stop sign, hydrant), and a geographic region, can enumerate and find the precise position of all instances of each landmark type within the geographic region. ALPS uses a novel combination of two key ideas. First, it uses \textit{image analysis} to find the position of a landmark, given a small number of images of the landmark from different perspectives. Second, it leverages recent efforts, like Google Street View~\cite{google_streetview}, that augment maps with visual documentation of street-side views, to obtain images of such landmarks. At a high-level, ALPS scours Google Street View for images, applies a state-of-the-art off-the-shelf object detector~\cite{yolo_paper} to detect landmarks in images, then triangulates the position of the landmarks using a standard least-squares formulation. On top of this approach, ALPS adds novel techniques that help the system \textit{scale} and improve its \textit{accuracy and coverage}.

\paragraph{Contributions} Our first contribution is techniques for scaling landmark positioning to large cities. Even a moderately sized city can have several million Street View images. If ALPS were to retrieving all images, it would incur two costs, both of which are scaling bottlenecks in ALPS: (1) the latency, network and server load cost of retrieving the images, and (2) the computational latency of applying object detection to the entire collection. ALPS optimizes these costs, without sacrificing coverage, using two key ideas. First, we observe that Street View has a finite resolution of a few meters, so it suffices to sample the geographic region at this resolution. Second, at each sampling point, we retrieve a small set of images, \textit{lazily} retrieving additional images for positioning only when a landmark has been detected in the retrieved set. In addition, the ALPS system can take location \textit{hints} to improve scalability: these hints specify where landmarks are likely to be found (\textit{e.g.}, at street corners), which helps narrow down the search space.

Our second contribution is techniques that improve accuracy and coverage. Object detectors can have false positives and false negatives. For an object detector, a false positive means that the detector detected a landmark in an image that doesn’t actually have the landmark. A false negative is when the detector didn’t detect the landmark in the image that actually does contain the landmark. ALPS can reduce false negatives by using multiple perspectives: if a landmark is not detected at a sampling point either because it is occluded or because of poor lighting conditions, ALPS tries to detect it in images retrieved at neighboring sampling points. To avoid false positives, when ALPS detects a landmark in an image, it retrieves zoomed in versions of that image and runs the object detector on them, using majority voting to increase detection confidence. Once ALPS has detected landmarks in images, it must resolve aliases (multiple images containing the same landmark). Resolving aliases is especially difficult for densely deployed landmarks like fire hydrants, since images from geographically nearby sampling points might contain different instances of hydrants. ALPS clusters images by position, then uses the relative bearing to the landmark to refine these clusters. Finally, ALPS uses least squares regression to estimate the position of the landmark; this enables it to be robust to position and orientation errors, as well as errors in the position of the landmark within the image as estimated by the object detector.

Our final contribution is an exploration of ALPS’ performance at the scale of a zip-code, and across several major cities. ALPS can cover over 92\% of Subway restaurants in several large cities and over 95\% of hydrants in one zip-code, and localize 93\% of Subways and 87\% of hydrants with an error less than 10 meters. Its localization accuracy is better than Google Places~\cite{googleplace_api} for over 85\% of the Subways in large cities. ALPS’s scaling optimizations can reduce the number of retrieved images by over a factor of 20, while sacrificing coverage only by 1-2\%. Its accuracy improvements are significant: for example, removing the bearing-based refinement (discussed above) can reduce coverage by half.

%% file: tex/alps/motivation.tex
\section{Motivation and Challenges}

\paragraph{Positioning Common Landmarks} 
Context awareness~\cite{context_awareness_1, context_awareness_2} is essential for ubiquitous computing since it can enable computing devices to reason about the built and natural environment surrounding a human, and provide appropriate services and capabilities. Much research has focused on automatically identifying various aspects of context~\cite{Barometer, IndoorLoc, indoorALPS, ahmad2018quicksketch}, such as places and locations where a human is or has been, the objects or people within the vicinity of the human and so forth, or using the detected landmarks to help position vehicles.

One form of context that can be useful for several kinds of outdoor ubiquitous computing applications is the landmark, an easily recognizable feature or object in the built environment. In colloquial usage, a landmark refers to a famous building or structure which is easily identifiable and can be used to give directions. In this chapter, we focus on common landmarks, which are objects that frequently occur in the environment, yet can provide contextual cues for ubiquitous computing applications. Examples of common landmarks include storefronts (\textit{e.g.}, fast food stores, convenience stores), signposts such as speed limits and stop signs, traffic lights, fire hydrants, and so forth. 

\paragraph{Potential Applications}
Knowing the type of a common landmark (henceforth, landmark) and its precise position (GPS coordinates), and augmenting maps with this information, can enable several applications. 

Autonomous cars \cite{auto_driving} and drones \cite{drone_control} both rely on visual imagery. Using cameras, they can detect command landmarks in their vicinity, and use the positions of those landmarks to improve estimates of their own position. Drones can also use the positions of common landmarks, like storefronts, for precise delivery.

Signposts can provide context for vehicular control or driver alerts. For example, using a vehicle’s position and a database of the position of speed limit signs \cite{mogelmose2012vision}, a car’s control system can either automatically regulate vehicle speed to within the speed limit, or warn drivers when they exceed the speed limit. Similarly, a vehicular control systems can use a stop sign position database to slow down a vehicle approaching a stop sign, or to warn drivers in danger of missing the stop sign. 

A database of automatically generated landmark positions can be an important component of a smart city \cite{smartcity}. Firefighters can respond faster to fires using a database of positions of fire hydrants \cite{firefighter_search_for_hydrant}. Cities can maintain inventories of their assets (street lights, hydrants, trees \cite{Wegner_2016_CVPR}, and signs \cite{balali2015multi} are example of city assets) \cite{need_hydrant_1, need_hydrant_2}; today, these inventories are generated and maintained manually. Finally, drivers can use a database of parking meter positions, or parking sign positions to augment the search for parking spaces \cite{mathur2010parknet}, especially in places where in-situ parking place occupancy sensors have not been installed \cite{dust-parking}.

Landmark locations can also improve context-aware customer behavior analysis \cite{lee2013understanding}. Landmark locations can augment place determination techniques \cite{CheckInside, PinPlace}. Indeed, a database of locations of retail storefronts can directly associate place names with locations. Furthermore, landmark locations, together with camera images taken by a user, can be used to more accurately localize the user itself than is possible with existing location services. This can be used in several ways. For example, merchants can use more precise position tracks of users to understand the customer shopping behavior. They can also use this positioning to target users more accurately with advertisements or coupons, enriching the shopping experience. 

Finally, landmark location databases can help provide navigation and context for visually impaired persons \cite{cities_unlocked, hara2013exploring}. This pre-computed database can be used by smart devices (\textit{e.g.} Google Glass) to narrate descriptions of surroundings (\textit{e.g.}, ``You are facing a post office and your destination is on its right, and there is a barbershop on its left.’’) to visually impaired users.

\paragraph{Challenges and Alternative Approaches}
An \textit{accurate} database which has high coverage of common landmark locations can enable these applications. High coverage is important because, for example, a missing stop sign can result in a missed warning. Moreover, if the database is complete for one part of a city, but non-existent for another, then it is not useful because applications cannot rely on this information being available.

To our knowledge, no such comprehensive public database exists today, and existing techniques for compiling the database can be inaccurate or have low coverage. Online maps (\textit{e.g.}, Google Places~\cite{googleplace_api} or Bing Maps~\cite{bing_maps_api}) contain approximate locations of some retail storefronts (discussed below). Each city, individually, is likely to have reasonably accurate databases of stores within the city, or city assets. In some cases, this information is public. For example, the city of Los Angeles has a list of fire hydrant locations~\cite{hydrant_la}, but not many other cities make such information available. Collecting this information from cities can be logistically difficult. For some common landmarks, like franchise storefronts, their franchiser makes available a list of franchisee addresses: for example, the list of Subway restaurants in a city can be obtained from `subway.com`. From this list, we can potentially derive locations through reverse geo-coding, but this approach doesn’t generalize to the other landmarks discussed above. Prior work has explored two other approaches to collecting this database: crowdsourcing~\cite{openstreetmap, yelp_api}, and image analysis~\cite{flickr_api}. The former approach relies on users to either explicitly (by uploading stop signs to OpenStreetMaps) or implicitly (by checking in on a social network) tag landmarks, but can be inaccurate due to user error, or have low coverage because not all common landmarks may be visited. Image analysis, using geo-tagged images from photo sharing sites, can also result in an incomplete database.

In this chapter, we ask the following question: \textit{is it possible to design a system to automatically compile, at the scale of a large metropolis, an accurate and high coverage database of landmark positions}? Such a system should, in addition to being accurate and having high coverage, be extensible to different types of landmarks, and scalable in its use of computing resources. In the rest of the chapter, we describe the design of a system called ALPS that satisfies these properties.

%% file: tex/alps/design.tex
\section{The Design of ALPS}

\subsection{Approach and Overview}

The input to ALPS is a landmark type (a chain restaurant, a stop sign, \textbf{etc.}) and a geographical region expressed either using a zip code or a city name. The output of ALPS is a list of GPS coordinates (or positions) at which the specified type of landmark may be found in the specified region. Users of ALPS can specify other optional inputs, discussed later.

ALPS localizes landmarks by analyzing images using the following idea. To localize a fire hydrant, for example, suppose we are given three images of the same fire hydrant, taken from three different perspectives, and the position and orientation of the camera when each image was taken is also known. Then, if we can detect the hydrant in each image using an object detector, then we can establish the bearing of the hydrant relative to each image. From the three bearings, we can triangulate the location of the hydrant. ALPS uses more complex variants of this idea to achieve accuracy, as discussed below.

To obtain such images, ALPS piggybacks on map-based visual documentation of city streets~\cite{streetview_api, bing_streetside_api}. Specifically, ALPS uses the imagery captured by Google’s Street View. The vehicles that capture Street View images have positioning and bearing sensors~\cite{understand_streetview}, and the Street View API permits a user to request an image taken at a given position and with a specified bearing. ALPS’s coverage is dictated in part by Street View’s coverage, and its completeness is a combination of its coverage, and the efficacy of its detection and localization algorithms.

Street View (and similar efforts) have large databases, and downloading and processing all images in a specified geographic region can take time, computing power, and network bandwidth. To scale to large geographic regions (\textbf{e.g.}, an entire zipcode or larger), ALPS employs novel techniques that (a) retrieve just sufficient images to ensure high coverage, (b) robustly detect the likely presence of the specified landmark, then (c) drill down and retrieve additional images in the vicinity to localize the landmarks.

Finally, users can easily extend ALPS to new landmark types, and specify additional scaling hints.

ALPS comprises two high-level capabilities (\alpsfig{system_model}): Seed location generation takes a landmark type specified by user as input, and generates a list of seed locations  where the landmarks might be located; and Landmark localization takes seed locations as input and generates landmark positions in the specified geographic region  as output.

In turn, seed location generation requires three conceptual capabilities: (1) base image retrieval which downloads a subset of all Street View images; (2) landmark detection that uses the state-of-the-art computer vision object detection \cite{yolo_paper} to detect and localize landmarks retrieved by base image retrieval, and applies additional filters to improve the accuracy of detection; (3) image clustering groups detected images that likely contain the same instance of the landmark. The result of these three steps is a small set of seed locations where the landmark is likely to be positioned, derived with minimal resources without compromising coverage.

Landmark localization reuses the landmark detection capability, but requires two additional capabilities: (1) adaptive image retrieval, which drills down at each seed location to retrieve as many images as necessary for localizing the object; (2) and a landmark positioning capability that uses least squares regression to triangulate the landmark position.

\begin{figure}
\centering\includegraphics[width=0.7\columnwidth]{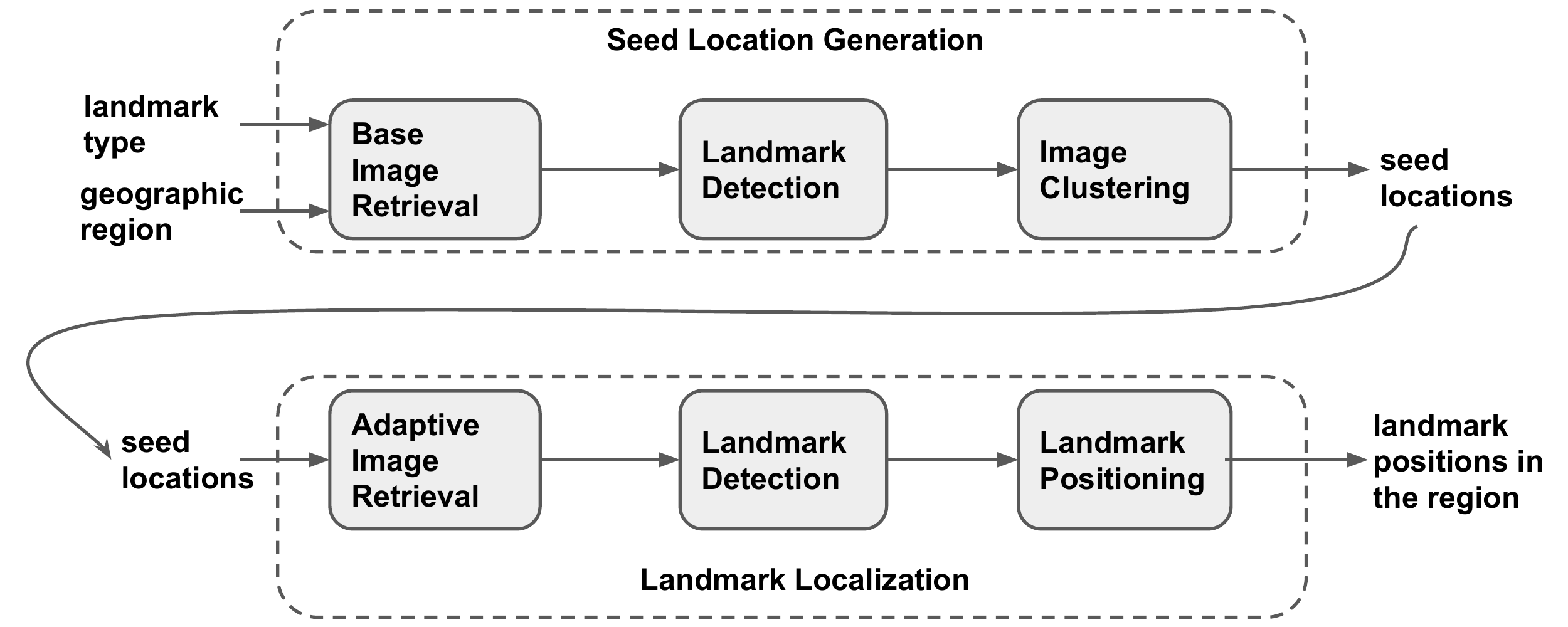}
\caption{\emph{ALPS Components.}}
\label{fig:alps_system_model}
\end{figure}

\subsection{Base Image Retrieval}

ALPS retrieves images from Street View, but does not retrieve all Street View images within the input geographic region. This brute-force retrieval does not scale, since even a small city like Mountain View can have more than 10 million images. Moreover, this approach is wasteful, since Street View’s resolution is finite: in \alpsfig{basic_image_database}(a), a Street View query for an image anywhere within the dotted circle will return the image taken from one of the points within that circle. 

ALPS scales better by retrieving as small a set of images as possible, without compromising coverage (\alpsfig{basic_image_database}(b)). It only retrieves two Street View images in two opposing directions perpendicular to the street, at intervals of $2r$ meters, where $2r$ is Street View’s resolution (from experiments, $r$ is around 4 meters). By using nominal lane~\cite{us_lane_widths} and sidewalk~\cite{sidewalk_guideline} widths, Street View’s default angle of view of $60^\circ$, it is easy to show using geometric calculations that successive 8 meter samples of Street View images have overlapping views, thereby ensuring visual coverage of the entire geographic region.

\begin{figure}
\centering\includegraphics[width=0.7\columnwidth]{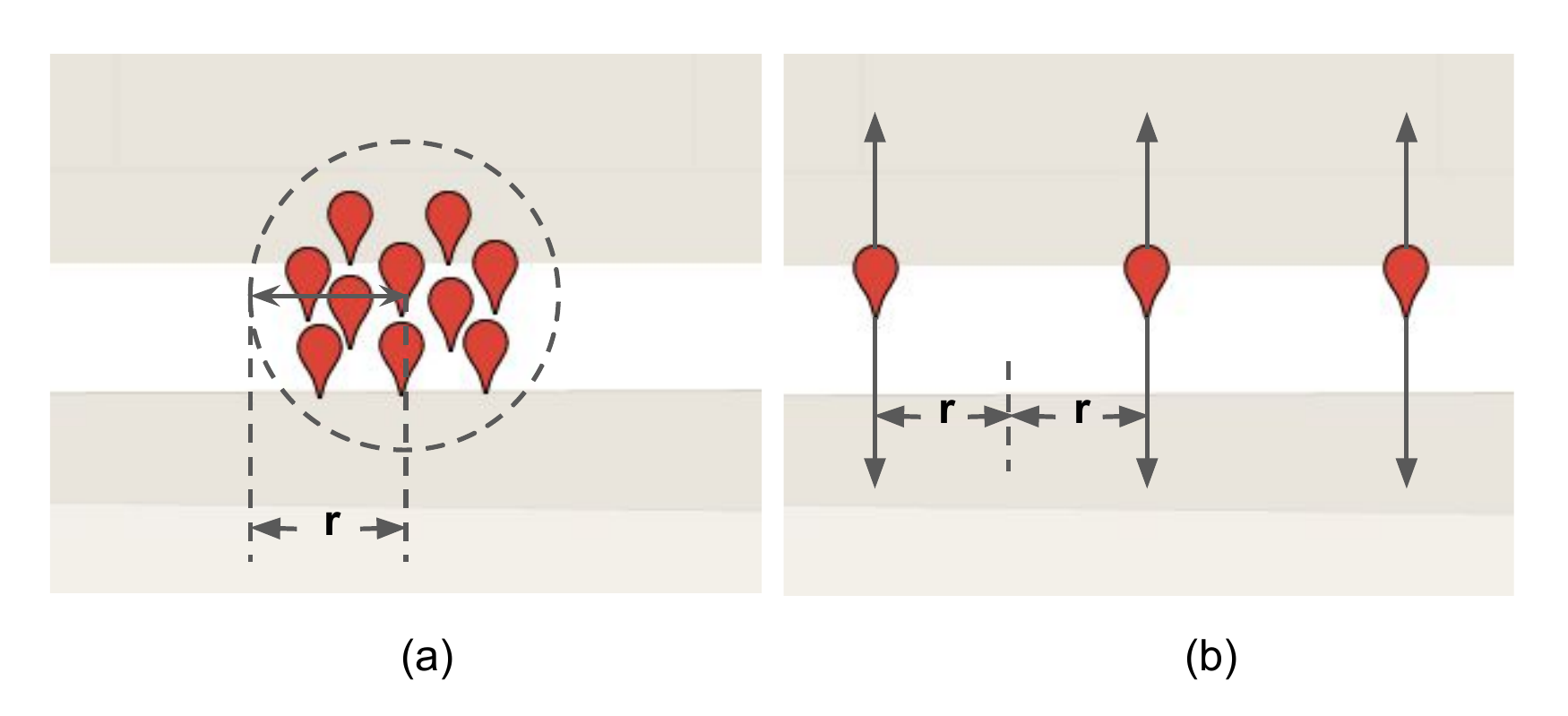}
\caption{\emph{Base Image Retrieval.}}
\label{fig:alps_basic_image_database}
\end{figure}

\subsection{Landmark Detection}

Given an image, this capability detects and localizes the landmark within the image. This is useful both for seed location generation, as well as for landmark localization, discussed earlier.

Recent advances \cite{imagenet, fast_rcnn} in deep learning techniques have enabled fast and accurate object detection and localization. We use a state-of-the-art object detector, called YOLO~\cite{yolo_paper}. YOLO uses a neural network to determine whether an object is present in an image, and also draws a bounding box around the part of the image where it believes the object to be (\emph{i.e.}, localizes the object in the image). YOLO needs to be trained, with a large number of training samples, to detect objects. We have trained YOLO to detect logos of several chain restaurants or national banks, as well as stop signs and fire hydrants. Users wishing to extend ALPS functionality to other landmark types can simply provide a neural network trained for that landmark.

Even the best object detection algorithms can have false positives and negatives~\cite{ilsvrc_15}. False positives occur when the detector mistakes other objects for the target landmark due to lighting conditions, or other reasons. False negatives can decrease the coverage and false positives can reduce positioning accuracy. In our experience, false negatives arise because YOLO cannot detect objects smaller than $50\times50$ pixels or objects that are blurred, partially obscured or in shadow, or visually indistinguishable from the background.

\begin{figure}
\centering\includegraphics[width=0.7\columnwidth]{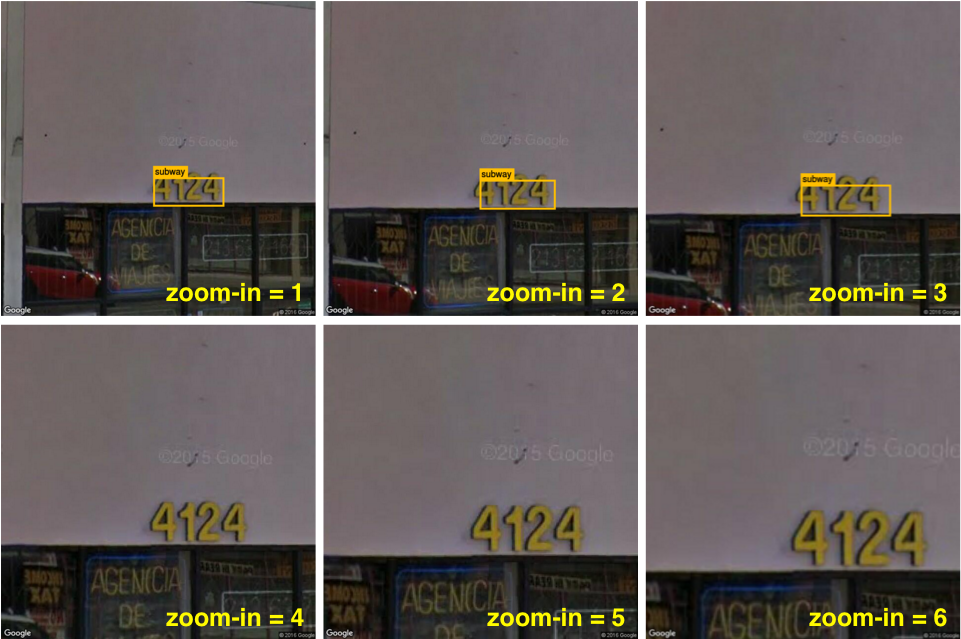}
\caption{\emph{How zooming-in can help eliminate false positives.}}
\label{fig:alps_zoom-in-filter}
\end{figure}

ALPS reduces false positives by using Street View’s support for retrieving images at different zoom levels. Recall that base image retrieval downloads two images at each sampling point. ALPS applies the landmark detector to each image: if the landmark is detected, ALPS retrieves six different versions of the corresponding Street View image each at different zoom levels. It determines the tilt and bearing for each of these zoomed images based on the detected landmark. ALPS then uses two criteria to mark the detection as a true positive: that YOLO should detect a landmark in a majority of the zoom levels, and that the size of the bounding box generated by YOLO is proportional to the zoom level. For example, in \alpsfig{zoom-in-filter}, YOLO incorrectly detected a residence number, when detecting a Subway logo, in the first three zoom levels (the first zoom level corresponds to the base image). After zooming in further, YOLO was unable to detect the Subway logo in the last 3 zoomed-in images. In this case, ALPS declares that the image does not contain a Subway logo, because the majority vote failed. We address false negatives in later steps.

\subsection{Image Clustering}

To generate seed locations, ALPS performs landmark detection on each image obtained by base image retrieval. However, two different images might contain the same landmark: the clustering step uses image position and orientation to cluster such images together. In some cases, this clustering can reduce the number of seed locations dramatically: in \alpsfig{clustering3}(a), 87 landmarks are detected in the geographic region shown, but a much smaller fraction of them represent unique landmarks (\alpsfig{clustering3}(b)).

\begin{figure}
\centering\includegraphics[width=0.7\columnwidth]{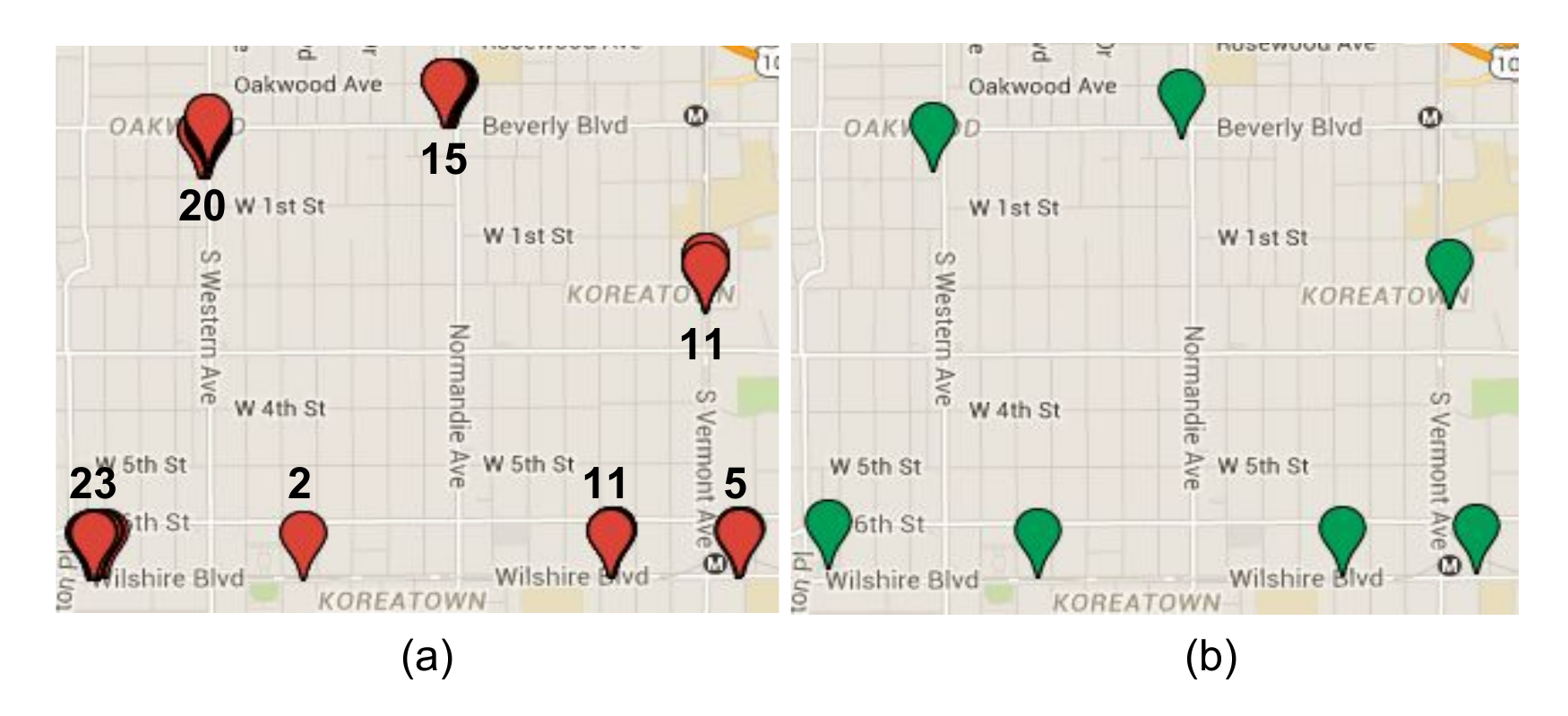}
\caption{\emph{Clustering can help determine which images contain the same landmark.}}
\label{fig:alps_clustering3}
\end{figure}

The input to clustering is the set of images from the base set in which a landmark has been detected. ALPS clusters this set by using the position and bearing associated with the image, in two steps: first, it clusters by position, then, within each cluster it distinguishes pairs of images whose bearing is inconsistent. 

To cluster by position, we use mean shift clustering~\cite{mean_shift}: (1) put all images into a candidate pool; (2) select a random image in the candidate pool as the center of a new cluster; (3) find all images within $R$ meters ($R$=50 in our implementation) of the cluster center, put these images into the cluster, and remove them from the candidate pool; (4) calculate the mean shift of the center of all nodes within the cluster, and if the center is not stable, go to step (3), otherwise go to step (2).

Clustering by position works well for landmarks likely to be geographically separated (\emph{e.g.}, a Subway restaurant), but not for landmarks (\emph{e.g.}, a fire hydrant) that can be physically close. In the latter case, clustering by position can reduce accuracy and coverage.

\begin{figure}
\centering\includegraphics[width=0.7\columnwidth]{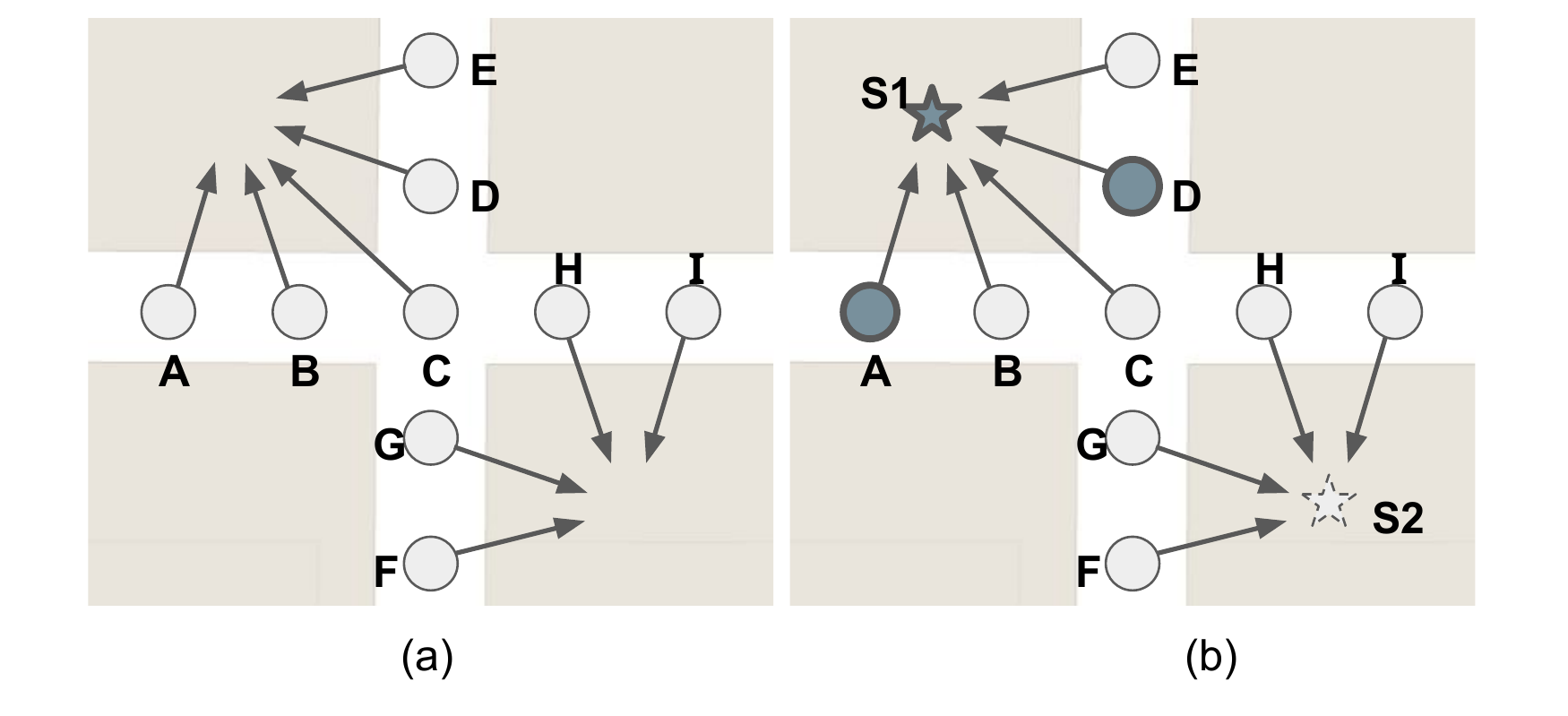}
\caption{\emph{Clustering by bearing is necessary to distinguish between two nearby landmarks.}}
\label{fig:alps_clustering4}
\end{figure}

To improve the accuracy of clustering, we use bearing information in the Street View images to refine clusters generated by position-based clustering. Our algorithm is inspired by the RANSAC~\cite{ransac} algorithm for outlier detection, and is best explained using an example. \alpsfig{clustering4}(a) shows an example where ALPS’s position-based clustering returns a cluster with 9 images $A$-$I$. In \alpsfig{clustering4}(b), images  $A$-$E$ and images $F$-$I$ see different landmarks. ALPS randomly picks two images $A$ and $D$, adds them to a new proto-cluster, and uses its positioning algorithm (described below) to find the approximate position of the landmark ($S1$) as determined from these images. It then determines which other images have a bearing consistent with the estimated position of $S1$. $H$’s bearing is inconsistent with $S1$, so it doesn’t belong to the new proto-cluster, but $B$’s bearing is. ALPS computes all possible proto-clusters in the original cluster, then picks the lowest-error large proto-cluster, outputs this as a refinement of the original cluster, removes these images from the original cluster, and repeats the process. In this way, images $A$-$E$ are first output as one cluster, and images $F$-$I$ as another.

Each cluster contains images that, modulo errors in position, bearing and location, contain the same landmark. ALPS next uses its positioning algorithm (discussed below) to generate a seed location for the landmark.

\begin{figure}
\centering\includegraphics[width=0.7\columnwidth]{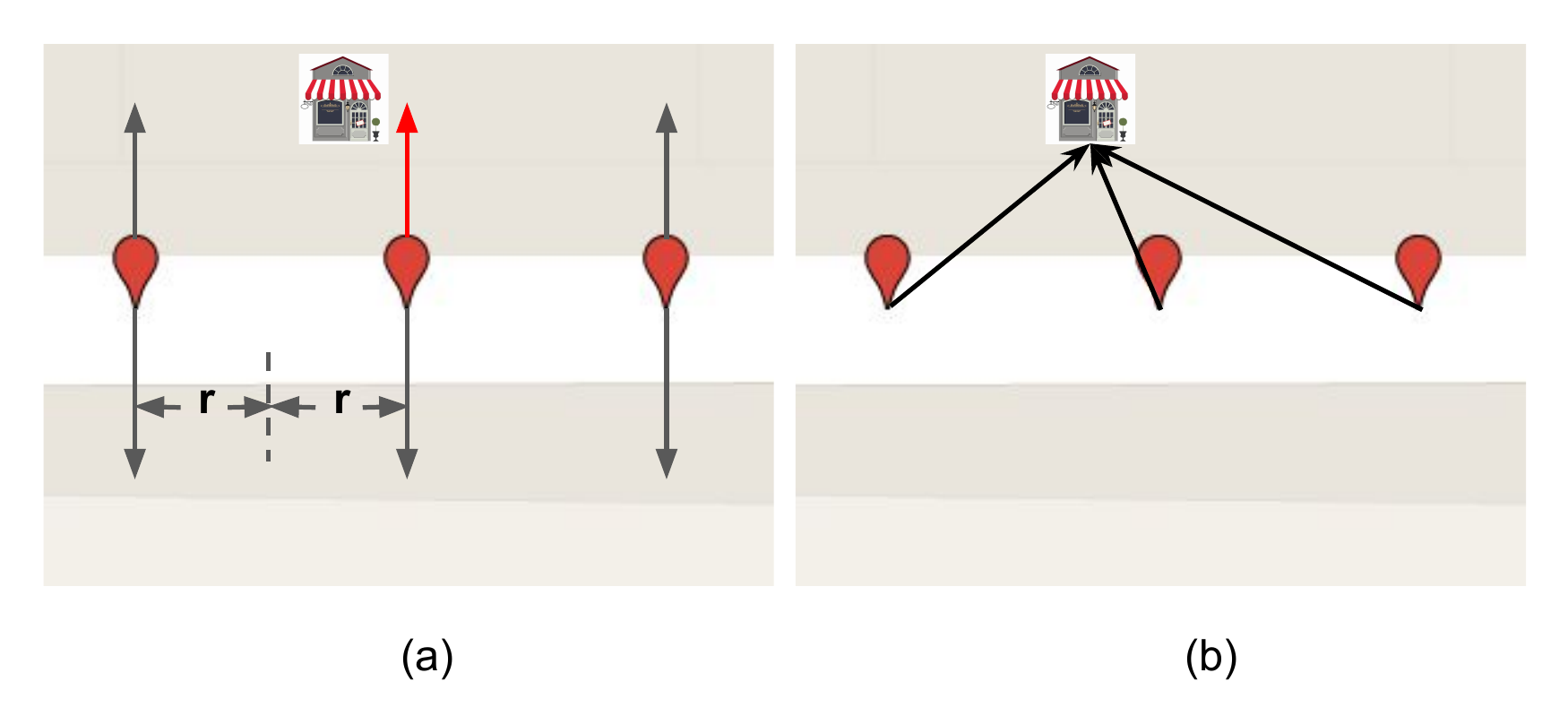}
\caption{\emph{Adaptive Image Retrieval.}}
\label{fig:alps_adaptive_image_downloader}
\end{figure}

\subsection{Adaptive Image Retrieval}

A seed location may not be precise, because the images used to compute it are taken perpendicular to the street (\alpsfig{adaptive_image_downloader}(a)). If the landmark is offset from the center of the image, errors in bearing and location can increase the error of the positioning algorithm. Moreover, the landmark detector may not be able to accurately draw the bounding box around a landmark that is a little off-center. Location accuracy can be improved by retrieving images whose bearing matches the heading from the sampling point to the seed location (so, the landmark is likely to be closer to the center of the image, \alpsfig{adaptive_image_downloader}(b)). (A seed location may not be precise also because a cluster may have too few images to triangulate the landmark position. We discuss below how we address this.)

To this end, we use an idea we call adaptive image retrieval: for each image in the cluster, we retrieve one additional image with the same position, but with a bearing directed towards the seed location ( \alpsfig{adaptive_image_downloader}(b)). At this stage, we also deal with false negatives. If a cluster has fewer than $k$ images ($k=4$ in our implementation), we retrieve one image each from neighboring sampling points with a heading towards the seed location, even if these points are not in the cluster. In these cases, the landmark detector may have missed the landmark because it was in a corner of the image; retrieving an image with a bearing towards the seed location may enable the detector to detect the landmark, and enable higher positioning accuracy because we have more perspectives.

\subsection{Landmark Positioning}

Precisely positioning a landmark from the images obtained using adaptive image retrieval is a central capability in ALPS. Prior work in robotics reconstructs the 3-D position of an object from multiple locations using three key steps~\cite{affine_reconstruction, slam}: (1) camera calibration with intrinsic and extrinsic parameters (camera 3-D coordinates, bearing direction, tilt angle, field of view, focus length, etc.); (2) feature matching with features extracted by algorithm like SIFT~\cite{sift}; (3) triangulation from multiple images using a method like singular value decomposition.

In our setting, these approaches don’t work too well: (1) Street View does not expose all the extrinsic and intrinsic camera parameters; (2) some of available parameters (GPS as 3-D coordinates, camera bearing) are noisy and erroneous, which may confound feature matching; (3) Street View images of a landmark are taken from different directions and may have differing light intensity, which can reduce  feature matching accuracy; (4) panoramic views in Street View can potentially increase accuracy, but there can be distortion at places in the panoramic views where images have been stitched together~\cite{understand_streetview}. 

Instead, ALPS (1) projects the landmark (\emph{e.g.}, a logo) onto a 2-dimensional plane to compute the relative bearing of the landmark and the camera, then (2) uses least squares regression to estimate the landmark position.

\parab{Estimating Relative Bearing}
ALPS projects the viewing directions onto a 2-D horizontal plane as shown in \alpsfig{calculate_relative_bearing}(a). $O$ represents the landmark in 3-dimensions, and $O^{’}$ represents the projected landmark on a 2-D horizontal plane. $C_i$ and its corresponding $C^{’}_i$ represent the camera locations in 3-D and 2-D respectively. Thus, $\vec{C_iO}$ is the relative bearing from camera $i$ to landmark $O$, and $\vec{C^{’}_iO}$ is its projection.

The landmark detector draws a bounding box around the pixels representing the landmark, and for positioning, we need to be able to estimate the relative bearing of the center of this bounding box relative to the bearing of the camera itself.
In \alpsfig{calculate_relative_bearing}(b),  line $\vec{AB}$  demarcates the (unknown) depth of the image and vector $\vec{C^{’}H}$ represents the bearing direction of camera, so $O^{"}$ is the image of $O^{’}$ on $\vec{AB}$. Our goal is to estimate $\angle{O^{’}C^{’}X}$ or $\angle{4}$, which is the bearing of the landmark relative to x-axis.

To do this, we need to estimate the following three variables: (1) the camera angle of view $\angle{AC^{’}B}$ or $\angle{1}$, which is the maximum viewing angle of the camera; (2) the camera bearing direction $\angle{HC^{’}X}$ or $\angle{2}$, which is the bearing direction of the camera when the image was taken; (3) the relative bearing direction of the landmark $\angle{O^{"}C^{’}D}$ or $\angle{3}$, which is the angle between the bearing direction of the camera and the bearing direction of the landmark.

$\angle{1}$ and $\angle{2}$ can be directly obtained from image metadata returned by Street View. \alpsfig{calculate_relative_bearing}(b) illustrates how to calculate $\angle{3}=\arctan(|\vec{DO^{"}}|/|\vec{DC^{’}}|)$. Landmark detection returns the image width in pixels and the pixel coordinates of the landmark bounding box. Thus, $|\vec{DO^{"}}|=|\vec{AO^{"}}|-\frac{1}{2}|\vec{AB}|$. Since $\tan(\frac{1}{2}\angle{1})=|\vec{AD}|/|\vec{DC^{’}}|$, we can calculate $|\vec{DC^{’}}|=\frac{1}{2}|\vec{AB}|\tan(\frac{1}{2}\angle{1})$. Then we derive $\angle{3}$ as $\arctan(|\vec{DO^{"}}|/|\vec{DC^{’}}|)$. Finally, we can calculate the bearing direction of the landmark: $\angle{4}=\angle{2}-\angle{3}$.

\begin{figure}
\centering\includegraphics[width=0.7\columnwidth]{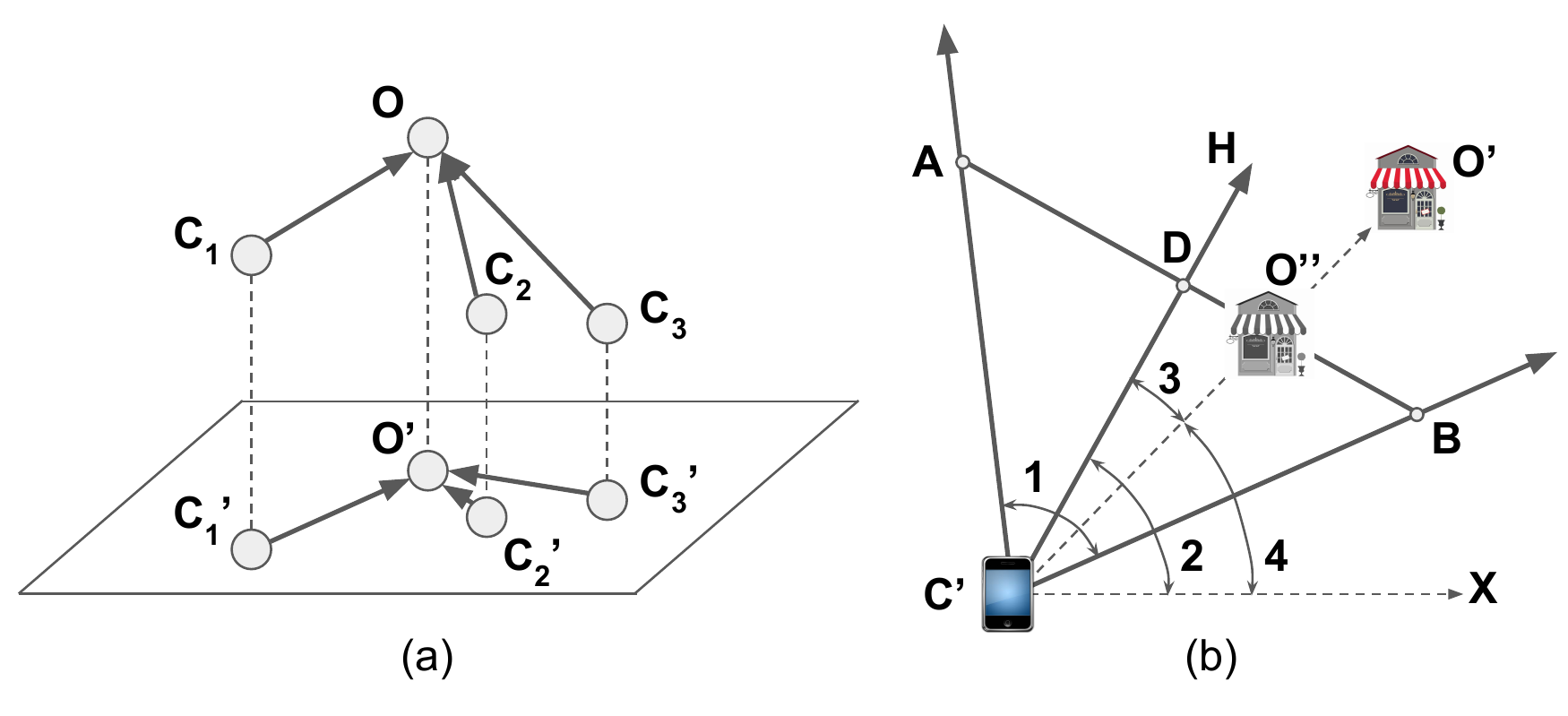}
\caption{\emph{Landmark Positioning.}}
\label{fig:alps_calculate_relative_bearing}
\end{figure}

\parab{Positioning using Least Squares Regression} 
For each cluster, using adaptive image retrieval, ALPS retrieves $N$ images for a landmark and can calculate the relative bearing of the landmark to the camera by executing landmark detection on each image.
Positioning the landmark then becomes an instance of the landmark localization problem~\cite{least_square, ls_book}, where we have to find the landmark location $P=[x_o,y_o]$ given $N$ distinct viewing locations $p_i =[x_i,y_i], i=1,2,\dots,N$ with corresponding bearing $\theta_i, i=1,2,\dots,N$, where $[x_i,y_i]$ is point $p_i$’s GPS coordinates in x-y plane. From first principles, we can write $\theta$ (or $\angle{4}$) as follows:

\vspace*{-3ex}
\begin{equation}
\tan(\theta_i)=\frac{\sin(\theta_i)}{\cos(\theta_i)}=\frac{y_o-y_i}{x_o-x_i}.
\label{ls_equation_1}
\end{equation}

Simplifying this equation and combining the equations for all images, we can write the following system of linear equations:

\vspace*{-3ex}
\begin{equation}
G\beta=h,
\label{ls_equation_3}
\end{equation}

where $\beta=[x_o, y_o]^T$ represents the landmark location, $G=[g_1, g_2, \dots, g_N]^T$, $g_i=[\sin(\theta_i)$, $-\cos(\theta_i)]$, $h=[h_1, h_2, \dots, h_N]^T$, $h_i=[\sin(\theta_i)x_i - \cos(\theta_i)y_i]$.

In this system of linear equations, there are two unknowns $x_o$ and $y_o$, but as many equations as images, resulting in an overdetermined system. However, many of the $\theta_i$s may be inaccurate because of errors in camera bearing, location, or landmark detection. To determine the most likely position, ALPS approximates $\hat{\beta}$ using least squares regression, which minimizes the squared residuals $S(\beta)=||G\beta-h||^2$, as output. If $G$ is full rank, the least squares solution of Equation \ref{ls_equation_3} is:

\vspace*{-3ex}
\begin{equation}
\hat{\beta}=\operatorname{arg\,min}(S(\beta))=(G^TG)^{-1}G^Th.
\label{ls_equation_5}
\end{equation}

\subsection{Putting it All Together}
Given a landmark type and a geographic region, ALPS first retrieves a base set of images for the complete region, which ensures coverage. On each image in this set, it applies landmark detection, retrieving zoomed-in versions of the image if necessary to obtain higher confidence in the detection and reduce false positives. It applies position and bearing based clustering on the images where a landmark was detected. Each resulting cluster defines a seed location, where the landmark might be.

At each seed location, ALPS adaptively retrieves additional images, runs landmark detection on each image again to find the bearing of the landmark relative to the camera for each image, and uses these bearings to formulate a system of linear equations whose least squares approximation represents the position of the landmark.

\subsection{Flexibility}

ALPS is flexible enough to support extensions that add to its functionality, or improve its scalability.

\textbf{New landmark types} Users can add to ALPS’s library of landmark types by simply training a neural network to detect that type of landmark. No other component of the system needs to be modified.

\textbf{Seed location hints} To scale ALPS better, users can specify seed location hints in two forms. ALPS can take a list of addresses and generate seed locations from this using reverse geo-coding. ALPS also takes spatial constraints that restrict base image retrieval to sampling points satisfying these constraints. For example, fire hydrants usually can be seen at or near street corners, or on a street midway between two cross-streets. Therefore, to specify such constraints, ALPS provides users with a simple language with 4 primitives: at\_corner (only at street corners), midway (at the midpoint between two cross-streets, searching\_radius (search within a radius of the points specified by other constraints), and lower\_image (the landmark like a fire hydrant only appears in the lower part of the image). More spatial constraints may be required for other landmarks: we have left this to future work.

%% file: tex/alps/eval.tex
\section{Evaluation}
In this section, we evaluate the coverage, accuracy, scalability and flexibility of ALPS on two different types of landmarks: Subway restaurants, and fire hydrants. 

\subsection{Methodology}

\textbf{Implementation and Experiments} We implemented ALPS in C++ and Python and accessed Street View images using Google’s API~\cite{streetview_api}. Our implementation is 2708 lines of code.\footnote{Available at \url{https://github.com/USC-NSL/ALPS}} All experiments described in the chapter are run on a single server with an Intel Xeon CPU at 2.70GHz, 32GB RAM, and one Nvidia GTX Titan X GPU inside. Below, we discuss the feasibility of parallelizing ALPS’s computations across multiple servers.

\textbf{Dataset} We evaluate ALPS using images for several geographic regions across five cities of the United States. In some of our experiments, we use seed location hints to understand the coverage and accuracy at larger scales. 

\textbf{Ground Truth}
For both landmark types we evaluate, getting ground truth locations is not easy because no accurate position databases exist for these. So, we manually collected ground truth locations for these as follows. For Subway restaurants, we obtained street addresses for each restaurant within the geographic region from the chain’s website~\cite{subway}. For fire hydrants, there exists an ArcGIS visualization of the fire hydrants in 2 zipcodes~\cite{hydrant_la} (as an aside, such visualizations are not broadly available for other geographic locations, and even these do not reveal exact position of the landmark). From these, we obtained the approximate location for each instance of the landmark. Using this approximate location, we first manually viewed the landmark on Street View, added a pinpoint on Google Maps at the location where we observed the landmark to be, then extracted the GPS coordinate of that pinpoint. This GPS coordinate represents the ground truth location for that instance. 

We validated of this method by collecting measurements at 30 of these landmarks using a high accuracy GPS receiver~\cite{ublox}. The 90th percentile error between our manual labeling and the GPS receiver is 6 meters. In the three cases where the error was high, we noticed that a sunshade obstructed our view of the sky, so the GPS receiver is likely to have obtained an incorrect position fix.

\textbf{Metrics} To measure the performance of ALPS, we use three metrics. Coverage is measured as the fraction of landmarks discovered by ALPS from the ground truth (this measures recall of our algorithms). Where appropriate, we also discuss the false positive rate of ALPS, which can be used to determine ALPS’s precision. The accuracy of ALPS is measured by its positioning error, the distance between ALPS’s position and ground truth. For scalability, we quantify the processing speed of each module in ALPS and the number of retrieved images. (We use the latter as a proxy for the time to retrieve images, which can depend on various factors like the Street View image download quota (25,000 images per day per user \cite{streetview_api}) and access bandwidth that can vary significantly).

\subsection{Coverage and Accuracy}

To understand ALPS’s coverage and accuracy, we applied ALPS to the zip-code 90004 whose area is 4 sq. km., to localize both Subway restaurants and fire hydrants. To understand ALPS’s performance at larger scales, we used seed location hints to run ALPS at the scale of large cities in the US.

\textbf{Zip-code 90004} Table \ref{no_seed_coverage} shows the coverage of the two landmark types across the entire zip-code. There are seven Subways in this region and ALPS discovers all of them, with no false positives. Table \ref{subway_no_seed} shows that ALPS localizes all Subways within 6 meters, with a median error of 4.7 meters. By contrast, the error of the GPS coordinates obtained from Google Places is over 10 meters for each Subway and nearly 60 meters in one case. Thus, at the scale of a single zip-code, ALPS can have high coverage and accuracy for this type of landmark. 

Fire hydrants are much harder to cover because they are smaller in shape, lower in position so can be occluded, can blend into the background or be conflated with other objects. As Table \ref{no_seed_coverage} shows, ALPS finds 262 out of 330 fire hydrants for an 79.4\% coverage. Of the ones that ALPS missed, about 16 were not visible to the naked eye in any Street View image, so no image analysis technique could have detected these. In 12 of these 16, the hydrant was occluded by a parked vehicle (which is illegal, \alpsfig{combinedSV} b)) in the Street View image, and the remaining 4 simply did not exist in the locations indicated in~\cite{hydrant_la}. Excluding these, ALPS’s coverage increases to about 83.4\%. ALPS can also position these hydrants accurately. \alpsfig{cdf_err_no_seed} shows the cumulative distribution function (CDF) of errors of ALPS for fire hydrants in 90004. It can localize 87\% of the hydrants within 10 meters, and its median error is 4.96 meters.

We then manually inspected the remaining 52 fire hydrants visible to the human eye but not discovered by ALPS. In 6 of these cases, the fire hydrant was occluded by a car in the image downloaded by base image retrieval: a brute-force image retrieval technique would have discovered these (see below). The remaining 46 missed hydrants fell roughly evenly into two categories. First, 24 of them were missed because of shortcomings of the object detector we use. In these cases, even though the base image retrieval downloaded images with hydrants in them, the detector did not recognize the hydrant in any of the images either because of lighting conditions (\emph{e.g.}, hydrant under the shade of a tree, \alpsfig{combinedSV} a)), or the hydrant was blurred in the image. The remaining 22 false negatives occurred because of failures in the positioning algorithm. This requires multiple perspectives (multiple images) to triangulate the hydrant, but in these cases, ALPS couldn’t obtain enough perspectives either because of detector failures or occlusions. Finally, the 21 false positives were caused by the object detector misidentifying other objects (such as a bollard, \alpsfig{combinedSV} c)) as hydrants. Future improvements to object detection, or better training and parametrization of the object detector, can reduce both false positives and false negatives. We have left this to future work.

Finally, both false positives and negatives in ALPS can be eliminated by using competing services like Bing Streetside~\cite{bing_streetside_api} which may capture images when a landmark is not occluded, or under different lighting conditions, or from perspectives that eliminate false positive detections. To evaluate this idea, we downloaded images from Bing Streetside near fire hydrants that were not detected using Google Street View. By combining both image sources, ALPS detected 300 out of 314 visible fire hydrants, resulting in 95.5\% coverage in this area.

\begin{figure}
\centering\includegraphics[width=0.7\columnwidth]{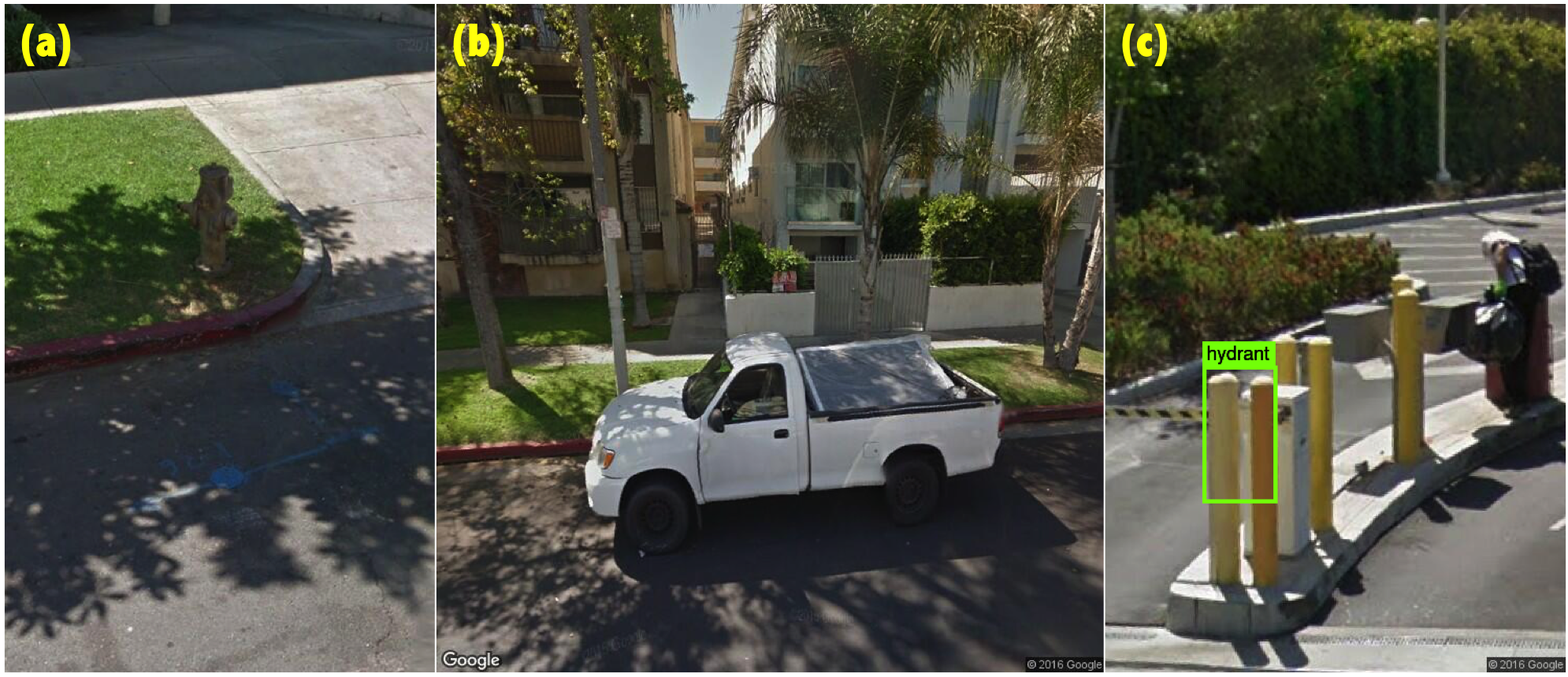}
\caption{\emph{\textbf{(a)} Hydrant occluded by a parked vehicle. \textbf{(b)} Detection failure because hydrant is under the shade of a tree. \textbf{(c)} False positive detection of bollard as hydrant.}}
\label{fig:alps_combinedSV}
\end{figure}

\begin{table}
\small
\centering
\begin{tabular}{|l|l|l|l|l|}
\hline
\textit{\textbf{Type}} & \textbf{\# landmark} & \textbf{\# visible} & \textbf{\# ALPS} & \textbf{Coverage} \\ \hline
Subway            & 7                       & 7                              & 7                        & 100\%                         \\ \hline
Hydrant           & 330                        & 314                               & 262                         & 83.4\%                      \\ \hline
\end{tabular}
\caption{\emph{Coverage of ALPS}}
\label{no_seed_coverage}
\end{table}

\textbf{City-Scale Positioning} To understand the efficacy of localizing logos, like that of Subway, over larger scales, we evaluated ALPS on larger geographic areas on the scale of an entire city. At these scales, ALPS will work, but we did not wish to abuse the Street View service and download large image sets. So, we explored city-scale positioning performance by feeding seed-location hints in the form of addresses for Subway restaurants, obtained from the chain’s web page.  

Table \ref{seed_coverage} shows the coverage with seed locations in different areas. Across these five cities, ALPS achieves more than 92\% coverage. With seed location hints, ALPS does not perform base image retrieval, so errors arise for other reasons. We manually analyzed the causes for errors in these five cities. In all cities, the invisible Subways were inside a building or plaza, so image analysis could not have located them. The missed Subways in Los Angeles, Mountain View, San Diego and Redmond were either because: (a) the logo detector failed to detect the logo in any images (because the image was partly or completely occluded), or (b) the positioning algorithm did not, in some clusters, have enough perspectives to localize.

ALPS does not exhibit false positives for Subway restaurants. For hydrants, all false positives arise because the landmark detector mis-detected other objects as hydrants. The Subway sign is distinctive enough that, even though the landmark detector did have some false positives, these were weeded out by the rest of the ALPS pipeline.

At city-scales also, the accuracy of ALPS is high. \alpsfig{cdf_err_seed} shows the CDF of errors of ALPS and Google Places locations for all of the Subways (we exclude the Subways that are not visible in any Street View image). ALPS can localize 93\% of the Subways within 10 meters, and its median error is 4.95 meters while the median error from Google places is 10.17 meters. Moreover, for 87\% of the Subways, Google Places has a higher error than ALPS in positioning. These differences might be important for high-precision applications like drone-based delivery.

\begin{table*}
\small
\centering
\begin{tabular}{|l|l|l|l|l|l|}
\hline
\textit{\textbf{City}} & \textbf{\# Subway} & \textbf{\# Visible} & \textbf{\# ALPS} & \textbf{Coverage} & \textbf{Median error(m)} \\ \hline
Los Angeles            & 123                       & 118                              & 115                        & 97\%           & 4.8     \\ \hline
Mountain View         & 38                        & 26                               & 24                         & 92\%              & 5.1    \\ \hline
San Francisco           & 49                        & 39                               & 39                         & 100\%           & 4.2    \\ \hline
San Diego         	    & 57                        & 44                               & 41                         & 93\%              & 5.0    \\ \hline
Redmond           	     & 31                        & 25                               & 24                         & 96\%              & 4.8    \\ \hline
\end{tabular}
\caption{\emph{Coverage with Seed Locations}}
\label{seed_coverage}
\end{table*}

\begin{table*}[]
\small
\centering

\begin{tabular}{|l|l|l|l|l|l|l|l|}
\hline
Subway \#           & 1     & 2     & 3     & 4     & 5     & 6     & 7     \\ \hline
Error of Google (m) & 10.06 & 11.53 & 14.10 & 30.38 & 59.48 & 16.60 & 14.90 \\ \hline
Error of ALPS (m)    & 2.03  & 4.53  & 6.78  & 2.93  & 7.39  & 5.94  & 3.33  \\ \hline
\end{tabular}
\caption{\emph{Error of ALPS and Google for localizing Subways}}
\label{subway_no_seed}
\end{table*}

 \begin{figure*}[!ht]
 \centering
 \begin{minipage}{0.32\linewidth}
\centering
\includegraphics[width=0.95\textwidth]{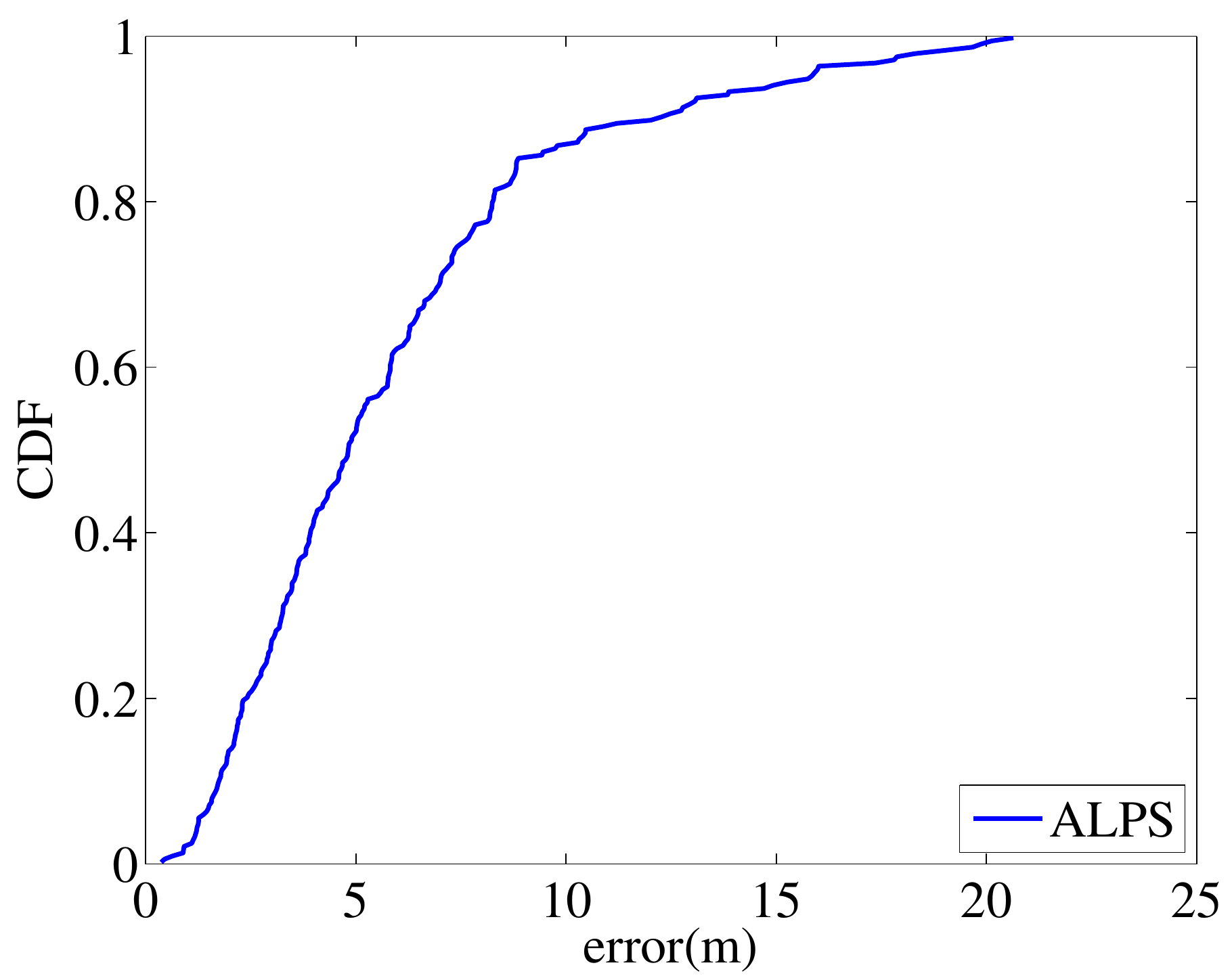}
\caption{\emph{Distribution of position errors for hydrants in 90004 zip-code}}
\label{fig:alps_cdf_err_no_seed}
  \end{minipage}
\hfill
 \begin{minipage}{0.32\linewidth}
\centering
\includegraphics[width=0.95\textwidth]{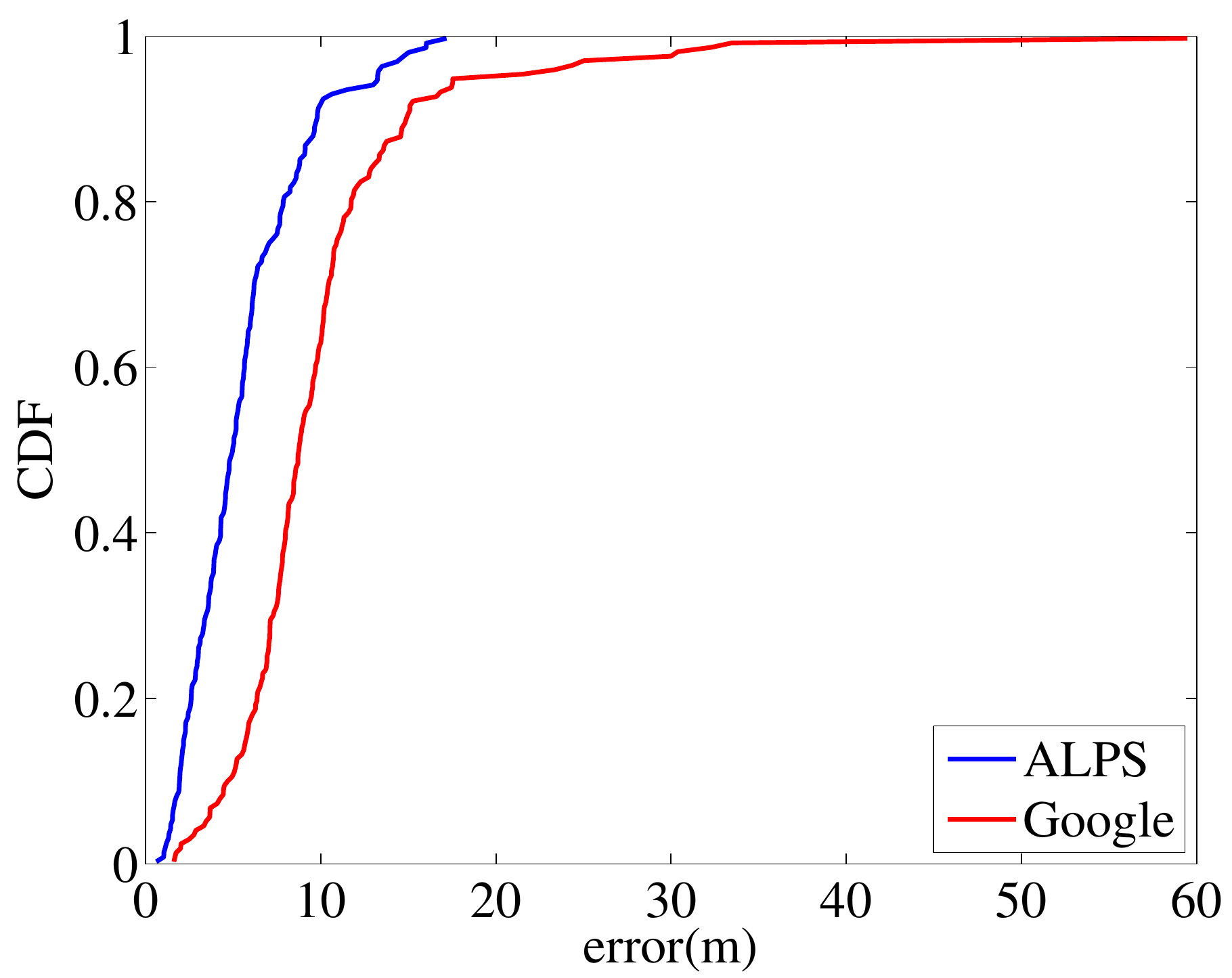}
    \caption{\emph{Distribution of errors for Subways in five cities}}
    \label{fig:alps_cdf_err_seed}
  \end{minipage}
\hfill
 \begin{minipage}{0.32\linewidth}
\centering
\includegraphics[width=0.95\textwidth]{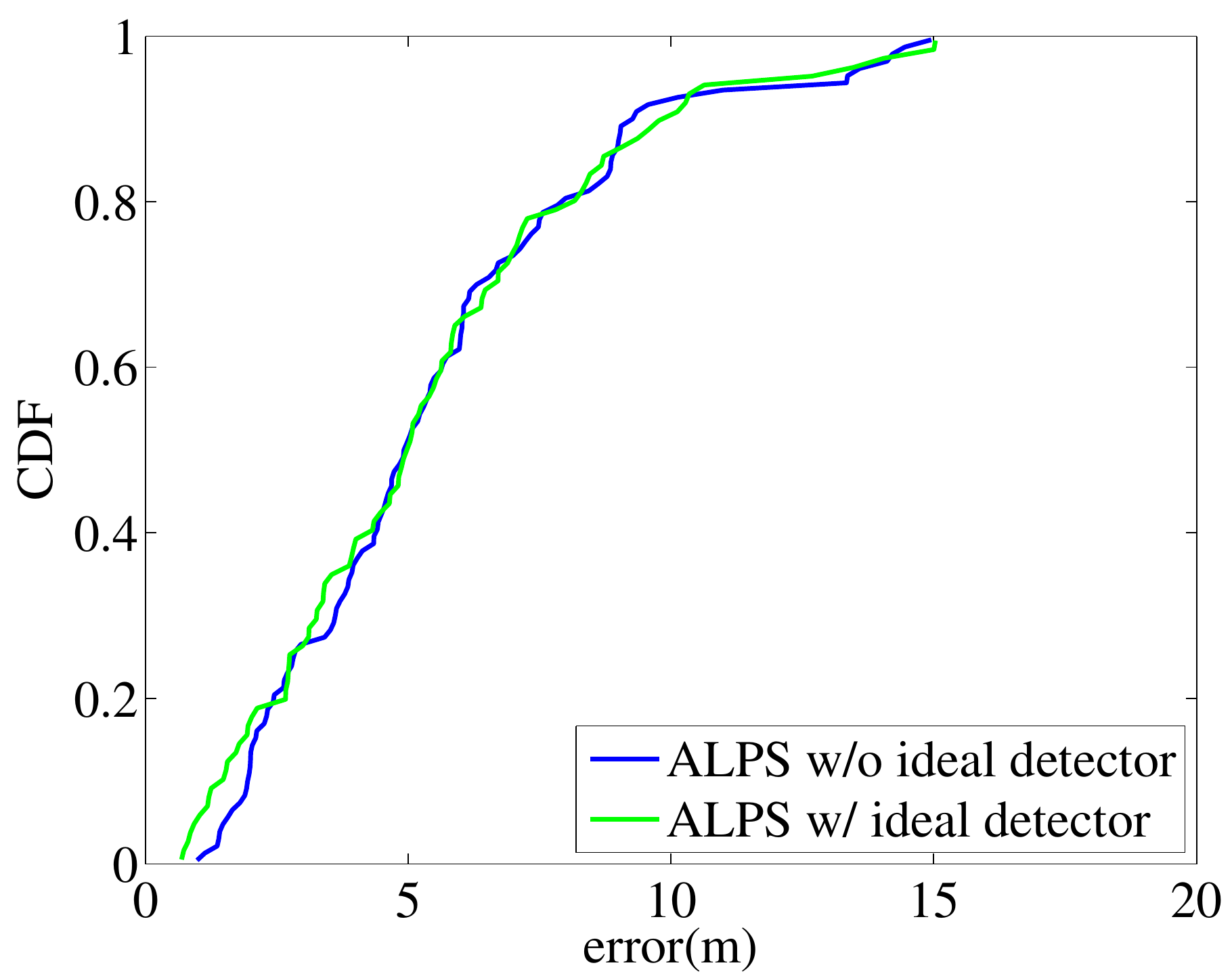}
\caption{\emph{Distribution of position errors for ALPS on Subway w/ and w/o ideal detector in Los Angeles}}
\label{fig:alps_cdf_comp_with_eye_label}
  \end{minipage}
\end{figure*}

\subsection{Scalability: Bottlenecks and Optimizations}

\begin{table}[]
\small
\centering
\begin{tabular}{|l|l|l|l|l}
\cline{1-4}
Module   & Base Retrieval     & Base Detection     & Cluster  &  \\ \cline{1-4}
Time (s) & 3528               & 8741               & 0.749       &  \\ \cline{1-4} 
Module   & Adaptive Retrieval & Adaptive Detection & Positioning  \\ \cline{1-4}
Time (s) & 715                & 1771               & 0.095       &  \\ \cline{1-4}
\end{tabular}
\caption{\emph{Processing time of each module}}
\label{processing_time}
\end{table}

\textbf{Processing Time} To understand scaling bottlenecks in ALPS, Table \ref{processing_time} breaks down the time taken by each component for the 90004 zip-code experiment (for both Subways and hydrants). In this experiment, base image retrieval, which retrieved nearly 150 thousand images, was performed only once (since that component is agnostic to the type of landmark being detected). Every other component was invoked once for each type of landmark.

Of the various components, clustering and positioning are extremely cheap. ALPS thus has two bottlenecks. The first is image retrieval, which justifies our optimization of this component (we discuss this more below). The second bottleneck is running the landmark detector. On average, it takes \textit{59 milliseconds} for the landmark detector to run detection on an image, regardless of whether the image contains the landmark or not. However, because we process over 150 thousand images, these times become significant. (Landmark detection is performed both on the base images to determine clusters, and on adaptively retrieved images for positioning, hence the two numbers in the table). Faster GPUs can reduce this time.

Fortunately, ALPS can be scaled to larger regions by parallelizing its computations across multiple servers. Many of its components are trivially parallelizable, including base image retrieval which can be parallelized by partitioning the geographic region, and adaptive image retrieval and positioning which can be partitioned across seed location. Only clustering might not be amenable to parallelization, but clustering is very fast. We have left an implementation of this parallelization to future work.

\textbf{The Benefit of Adaptive Retrieval} Instead of ALPS’s two phase (basic and adaptive) retrieval strategy, we could have adopted two other strategies: (a) a naive strategy which downloads images at very fine spatial scales of 1 meter, (b) a one phase strategy which downloads 6 images, each with a $60^\circ$ viewing angle so ALPS can have high visual coverage. For the 90004 zip-code experiment, the naive strategy retrieves 24$\times$ more images than ALPS’s two-phase strategy, while one-phase retrieves about 3$\times$ as many. The retrieval times are roughly proportional to the number of images retrieved, so ALPS’s optimizations provide significant gains. These gains come at a very small loss in coverage: one-phase has 1.91\% higher coverage than two-phase for hydrants mostly because the former has more perspectives: for example, hydrants that were occluded in the base images can be seen in one-phase images.

\textbf{Seed Location Hints} We have already seen that seed location hints helped us scale ALPS to large cities. These hints provide similar trade-offs as adaptive retrieval: significantly fewer images to download at the expense of slightly lower coverage. For hydrants in 90004, using hints that tell ALPS to look at street corners or mid-way between intersections and in the lower half of the image enabled ALPS to retrieve 3$\times$ fewer images, while only detecting 5\% fewer hydrants.

\subsection{Accuracy and Coverage Optimizations}

\textbf{Object Detection Techniques} The accuracy of the object detector is central to ALPS.
We evaluated the recall, precision, and processing time of several different object detection approaches: YOLO, HoG+SVM, and keypoint matching~\cite{keypoint_matching} with SIFT~\cite{sift} features. For HoG+SVM, we trained LIBSVM~\cite{libsvm} with HoG~\cite{hog} features and a linear kernel. Table \ref{different_cv} shows that YOLO outperforms the other two approaches in both recall and precision for recognizing the Subway logo. YOLO also has the fastest processing time due to GPU acceleration. 

\begin{table}[]
\small
\centering
\begin{tabular}{|l|l|l|l|}
\hline
                         & \textbf{YOLO} & \textbf{HoG+SVM} & \textbf{SIFT} \\ \hline
\textit{Precision}       & 85.1\%        & 74.2\%           & 63.7\%        \\ \hline
\textit{Recall}          & 87.4\%        & 80.5\%           & 40.6\%        \\ \hline
\textit{Speed (sec/img)} & 0.059        & 0.32             & 0.65          \\ \hline
\end{tabular}
\caption{\emph{Evaluation of different object detection methods}}
\label{different_cv}
\end{table}

\textbf{Street View Zoom} ALPS uses zoomed Street View images to increase detection accuracy. To quantify this, after using zoomed in images the landmark detector had a precision of 96.2\% and a recall of 86.8\%. In comparison, using only YOLO without zoomed in images had a precision of 85.1\% and recall of 87.4\%. 

To understand how the object detector affects the accuracy of ALPS, we manually labeled the position of the Subway logo in all the images in the dataset of Subways in LA. We thus emulated an object detector with 100\% precision and recall. This ideal detector finds the three missing Subways (by design), but with position accuracy comparable to YOLO (\alpsfig{cdf_comp_with_eye_label}).

\textbf{Importance of Bearing-based Clustering} We used the fire hydrant dataset to understand the incremental benefit of bearing-based cluster refinement. Without this refinement, ALPS can only localize 141 fire hydrants of 314 visible ones, while the refinement increases coverage by nearly 2$\times$ to 262 hydrants. Moreover, without bearing-based refinement, position errors can be large (in one case, as large as 80 meters) because different hydrants can be grouped into one cluster.

%% file: tex/paper_caesar.tex
\chapter{Caesar: Cross-Camera Complex Activity Recognition}\label{chap:caesar}

\input{tex/caesar/intro}
\input{tex/caesar/motivation}
\input{tex/caesar/design}
\input{tex/caesar/eval}

%% file: tex/caesar/intro.tex
\section{Introduction}

Being able to automatically detect activities occurring in the view of
a single camera is an important challenge in machine vision. The
availability of action data sets~\cite{ava,virat} has enabled the use
of deep learning for this problem. Deep neural networks (DNNs) can
detect what we call \textit{atomic actions} occurring within a single
camera. Examples of atomic actions include ``talking on the phone'',
``talking to someone else'', ``walking'' \etc

Prior to the advent of neural networks, activity detection relied on
inferring spatial and temporal relationships between objects. For
example, consider the activity ``getting into a car'', which involves
a person walking towards the car, then disappearing from the camera
view. \textit{Rules} that specify spatial and temporal relationships
can express this sequence of actions, and a detection system can
evaluate these rules to detect such activities.

In this chapter, we consider the next frontier in activity detection
research, exploring the \textit{near real-time} detection of
\textit{complex activities} potentially occurring across
\textit{multiple cameras}. A complex activity comprises two or more
atomic actions, some of which may play out in one camera and some in
another: \eg a person gets into a car in one camera, then gets out of
the car in another camera and hands off a bag to a person.

We take a pragmatic, systems view of the problem, and ask: given a
collection of (possibly wireless) surveillance cameras, what
architecture and algorithms should an end-to-end system incorporate to
provide \textit{accurate} and \textit{scalable} complex activity
detection?

Future cameras are likely to be wireless and incorporate onboard GPUs. However, activity detection using DNNs is
too resource intensive for embedded GPUs on these cameras. Moreover,
because complex activities may occur across multiple cameras, another
device may need to aggregate detections at individual cameras. An
\emph{edge cluster} at a cable head end or a cellular base station is
ideal for our setting: GPUs on this edge cluster can process videos
from multiple cameras with low detection latency because the edge
cluster is topologically close to the cameras (\caesarfig{high_level}).

Even with this architecture, complex activity detection poses several
challenges: (a) How to specify complex activities occurring across
multiple cameras? (b) How to partition the processing of the videos
between compute resources available on the camera and the edge
cluster? (c) How to reduce the wireless bandwidth requirement between
the camera and the edge cluster? (d) How to scale processing on the
edge cluster in order to multiplex multiple cameras on a single
cluster while still being able to process cameras in near real-time?

\parab{Contributions.}
In addressing these challenges, Caesar makes three important
contributions.

First, it adopts a \emph{hybrid} approach to complex activity
detection where some parts of the complex activity use DNNs, while
others are rule-based. This architectural choice is unavoidable: in the
foreseeable future, purely DNN-based complex activity detection is
unlikely, since training data for such complex activities is hard to
come by. Moreover, a hybrid approach permits evolution of complex
activity descriptions: as training data becomes available over time,
it may be possible to train DNNs to detect more atomic actions.

Second, to support this evolution, Caesar defines a language to describe
complex activities. In this language, a complex
activity consists of a sequence of \textit{clauses} linked together by
temporal relationships. A clause can either express a spatial
relationship, or an atomic action. Caesar users can express multiple
complex activities of interest, and Caesar can process camera feeds in
near real-time to identify these complex activities.

Third, Caesar incorporates a \textit{graph matching} algorithm that
efficiently matches camera feeds to complex activity descriptions. This algorithm leverages these descriptions to
optimize wireless network bandwidth and edge cluster scaling. To
optimize wireless network bandwidth, it performs object detection on
the camera, then, at the edge cluster, lazily retrieves images
associated with the detected objects only when needed (\eg to identify
whether an object has appeared in another camera). To scale the edge
cluster computation, it lazily invokes the action detection DNNs (the
computational bottleneck) only when necessary.

\changed{
Using a publicly available multi-camera data set, and an implementation of Caesar on an edge cluster, we show that, compared to a strawman approach which does not incorporate our optimizations, Caesar has 1-2 orders of magnitude lower detection latency and requires an order of magnitude less on-board camera memory (to support lazy retrieval of images).
Caesar's graph matching algorithm works perfectly, and its accuracy is only limited by the DNNs we use for action detection and re-identification (determining whether two human images belong to the same person).}

\added{While prior work has explored the single-camera action detection~\cite{ulutan2018actor, shou2018online, zhao2017temporal}, tracking of people across multiple overlapping cameras~\cite{xu2017cross,nithin2017globality,solera2016tracking} and non-overlapping cameras~\cite{ristani2018features, tesfaye2017multi,chen2017equalized}, to our knowledge, no prior work has explored a near real-time hybrid system for multi-camera complex activity detection.}

%% file: tex/caesar/motivation.tex
\section{Background and Motivation}

\parab{Goal and requirements.} Caesar detects \textit{complex
  activities} across \emph{multiple non-overlapping cameras}. It must
support \textit{accurate}, \textit{efficient}, \textit{near real-time}
detection while permitting \textit{hybrid} activity specifications. In
this section, we discuss the goal and these requirements in greater
detail.

\begin{figure}[!ht]
\centering\includegraphics[width=0.8\columnwidth]{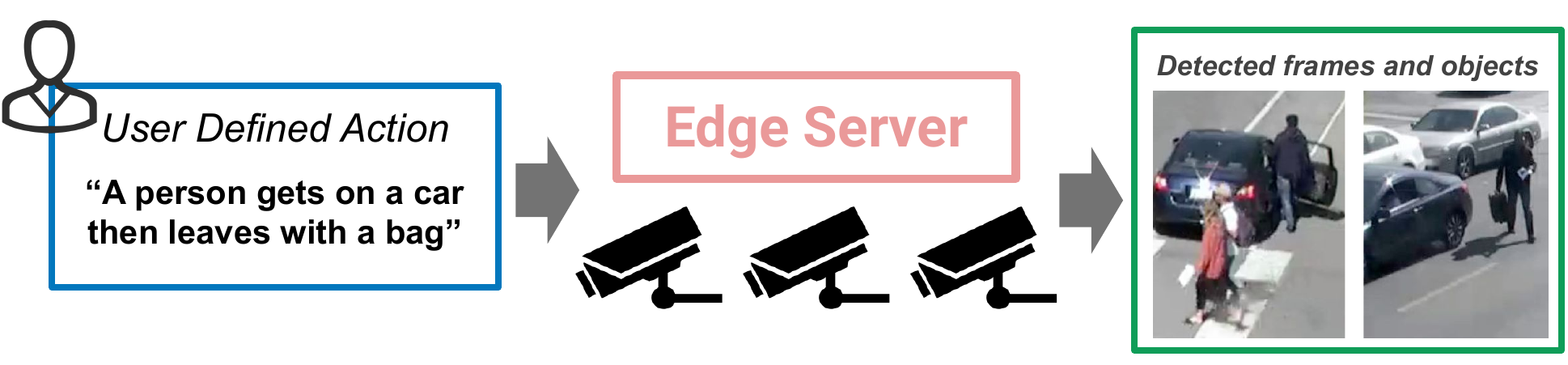}
\caption{\emph{The high-level concept of a complex activity detection system: the user defines the rule then the system monitors incoming videos and outputs the matched frames.}}
\label{fig:caesar_high_level}
\end{figure}

\parab{Atomic and complex activities.} An \textit{atomic} activity is
one that can be succinctly described by a single word label or short
phrase, such as ``walking'', ``talking'', ``using a phone''. In this
chapter, we assume that atomic activities can be entirely
captured on a single camera.

A \textit{complex} activity (i) involves multiple atomic activities
(ii) related in time (\eg one occurs before or after another), space
(\eg two atomic activities occur near each other), or in the set of
participants (\eg the same person takes part in two atomic
activities), and (iii) can span multiple cameras whose views do not
overlap. An example of a complex activity is: ``A person walking while
talking on the phone in one camera, and the same person talking to
another person at a different camera a short while later''. This
statement expresses temporal relationships between activities
occurring in two cameras (``a short while later'') and spatial
relationships between participants (``talking to another person'').

\parab{Applications of complex activity detection.} Increasingly,
cities are installing surveillance cameras on light poles or mobile
platforms like police cars and drones. However, manually monitoring
all cameras is labor intensive given the large number of
cameras~\cite{whats_wrong_surveillance}, so today's surveillance
systems can only deter crimes and enable forensic analysis. They
cannot anticipate events as they unfold in near real time. A recent
study~\cite{haelterman2016crime} shows that such anticipation is
possible: many crimes share common signatures such as ``a group of
people walking together late at night'' or ``a person getting out of a
car and dropping something''. Automated systems to identify
these signatures will likely increase the effectiveness of
surveillance systems.

The retail industry can also use complex activity detection. Today, shop owners install cameras to prevent theft and to track consumer behavior. A complex activity detection system can track customer purchases and browsing habits, providing valuable behavioral analytics to improve sales and design theft countermeasures.

\parab{Caesar architecture.} \caesarfig{high_level} depicts the high-level functional architecture of Caesar. Today, video processing and activity detection are well beyond the capabilities of mobile devices or embedded processors on cameras. So Caesar will need to leverage \textit{edge computing}, in which these devices offload video processing to a nearby server cluster. This cluster is a convenient rendezvous point for correlating data from non-overlapping cameras.


\parab{Caesar requirements.} \changed{Caesar should process videos with \emph{high throughput} and \emph{low end-to-end} latency. Throughput, or the rate at which it can process frames, can impact Caesar's accuracy and can determine if it is able to keep up with the video source. Typical surveillance applications process 20 frames per second. The end-to-end latency, which is the time between when a complex activity occurs and when Caesar reports it, must be low to permit fast near real-time response to developing situations. In some settings, such as large outdoor events in locations with minimal infrastructure~\cite{BurningMan}, video capture devices might be un-tethered so Caesar should \emph{conserve wireless bandwidth} when possible. To do this, Caesar can leverage significant on-board compute infrastructure: over the past year, companies have announced plans to develop surveillance cameras with onboard GPUs~\cite{dawn_smart}. Since edge cluster usage is likely to incur cost (in the same way as cloud usage), Caesar should \emph{scale} well: it should maximize the number of cameras that can be concurrently processed on a given set of resources. Finally, Caesar should have high precision and recall detecting complex activities.}

\parab{The case for hybrid complex activity detection.}
Early work on activity detection used a \emph{rule-based}
approach~\cite{qi2017predicting, tani2014events}. A rule codifies
relationships between actors (people); rule specifications can use
ontologies~\cite{tani2014events} or And-Or
Graphs~\cite{qi2017predicting}. Activity detection algorithms 
match these rule specifications to actors and objects detected in a
video.

More recent approaches are data-driven~\cite{ulutan2018actor, shou2018online, zhao2017temporal}, and train deep neural nets (DNNs) to detect activities. These approaches extract \textit{tubes} (sequences of bounding boxes) from video feeds; these tubes contain the actor performing an activity, as well as the surrounding context. They are then fed into a DNN trained on one or more action data sets (\eg AVA~\cite{ava}, UCF101~\cite{ucf101}, and VIRAT~\cite{virat}), which output the label associated with the activity. Other work~\cite{mettes2017spatial} has used a slightly different approach. It learns rules as relationships between actors and objects from training data, then applies these rules to match objects and actors detected in a video feed.

While data-driven approaches are preferable over rule-based ones
because they can generalize better, complex activity detection cannot
use purely data-driven approaches. By definition, a complex activity
comprises individual actions combined together. Because there can be
combinatorially many complex activities from a given set of individual
activities, and because data-driven approaches require large amounts
of training data, it will likely \emph{be infeasible to train neural
networks for all possible complex activities of interest}.

Thus, in this chapter, we explore a hybrid approach in which rules,
based on an extensible vocabulary, describe complex activities. The
vocabulary can include atomic actions: \eg ``talking on a phone'', or
``walking a dog''. Using this vocabulary, Caesar users can define a rule
for ``walking a dog while talking on the phone''. Then, Caesar can
detect a more complex activity over this new atomic action: ``walking
a dog while talking on the phone, then checking the postbox for mail
before entering a doorway''. (For brevity of description, a rule can,
in turn, use other rules in its definition.)

\parab{Challenges.} Caesar uses hybrid complex activity detection to
process feeds in near real-time while satisfying the requirements
described above. To do this, it must determine:
(a) How to specify complex activities across multiple
  non-overlapping cameras?
(b) How to optimize the use of edge compute resources to permit the
  system to scale to multiple cameras?
(c) How to conserve wireless bandwidth by leveraging on-board GPUs
  near the camera?

%% file: tex/caesar/design.tex
\section{Caesar Design}

\begin{figure}
\centering\includegraphics[width=0.65\columnwidth]{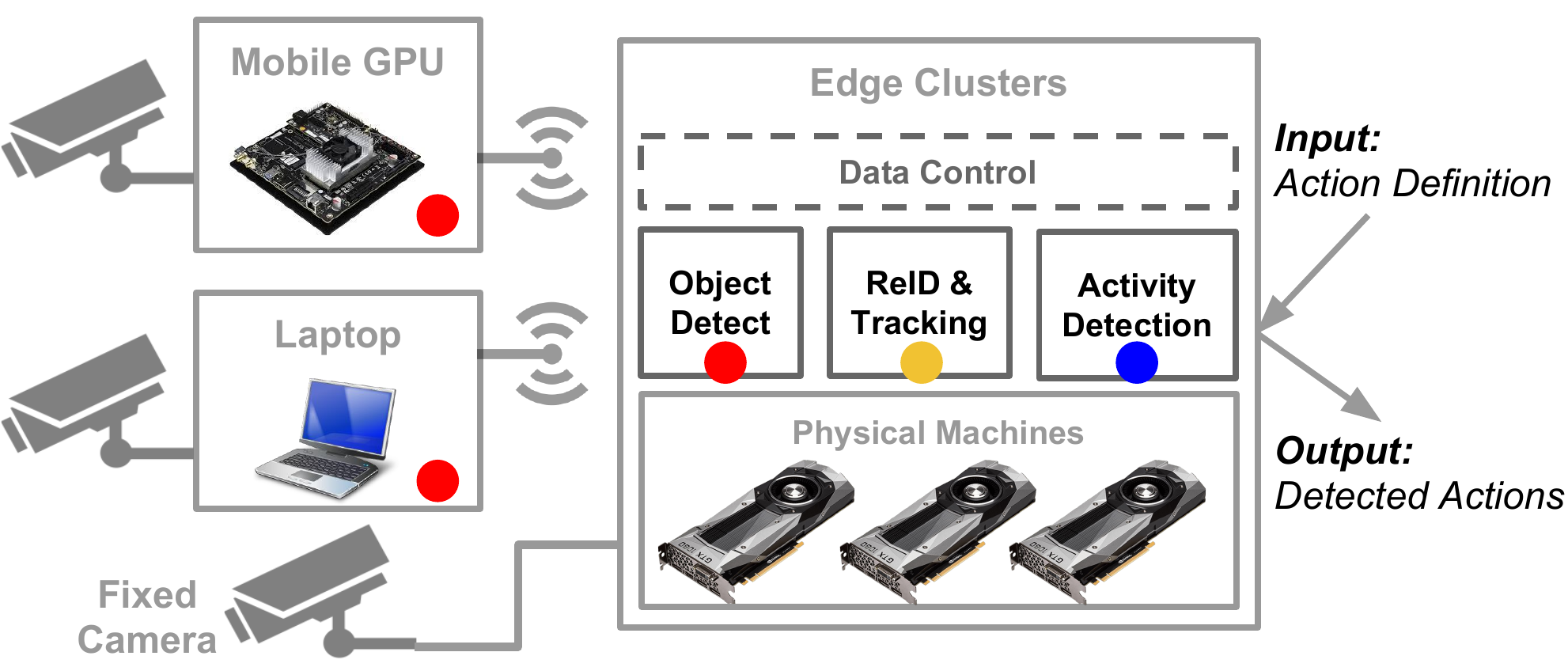}
\caption{\emph{The high-level design of Caesar. Dots with different colors represent different DNN modules for specific tasks.}}
\label{fig:caesar_design_high_level}
\end{figure}

In Caesar, users first specify one or more rules that describe complex
activities (\caesarfig{act_def}): this rule definition
language includes elements such as objects, actors, and actions, as
well as spatial and temporal relationships between them.

Cameras generate video feeds, and Caesar processes these using a three-stage pipeline (\caesarfig{design_high_level}). In the \textit{object detection} stage, Caesar generates bounding boxes of actors and objects seen in each frame. For wireless cameras, Caesar can leverage on board mobile GPUs to run object detection on the device; subsequent stages must run on the edge cluster. \added{The input to, and output of, object detection is the same regardless of whether it runs on the mobile device or the edge cluster}. A \textit{re-identification and tracking} module processes these bounding boxes. It (a) extracts \emph{tubes} for actors and objects by tracking them across multiple frames and (b) determines whether actors in different cameras represent the same person. Finally, a \emph{graph matching and lazy action detection} module determines: (a) whether the relationships between actor and object tubes match pre-defined rules for complex activities and (b) when and where to invoke DNNs to detect actions to complete rule matches. Table \ref{table:data_type} shows the three modules' data format.

\caesarfig{sys_out} shows an example of Caesar's output for a single camera. It annotates the live camera feed with detected activities. In this snapshot, two activities are visible: one is a person who was using a phone in another camera, another is a person exiting a car. Our demonstration video\footnote{Caesar's demo video: https://vimeo.com/330176833} shows Caesar's outputs for multiple concurrent camera feeds.

Caesar meets the requirements and challenges described in as follows: it processes streams continuously,
so can detect events in near-real time; it incorporates robustness
optimizations for tracking, re-identification, and graph matching to
ensure accuracy; it scales by lazily detecting actions, thereby
minimizing DNN invocation.

\begin{table}
\resizebox{0.9\linewidth}{!}{
\begin{tabular}{|l|l|l|}
\hline
& \textbf{Input} & \textbf{Output} \\ \hline
\textit{\begin{tabular}[c]{@{}l@{}}Object\\ Detection\end{tabular}}  & Image & Object Bounding Boxes \\ \hline
\textit{Track \& ReID}                                               & \begin{tabular}[c]{@{}l@{}}Object Bounding Boxes\\ Image\end{tabular} & Object TrackID  \\ \hline
\textit{\begin{tabular}[c]{@{}l@{}}Action \\ Detection\end{tabular}} & \begin{tabular}[c]{@{}l@{}}Object Boxes \& TrackID\\ Image\end{tabular}  & Actions         \\ \hline
\end{tabular}}
\caption{\emph{Input and output content of each module in Caesar.}}
\label{table:data_type}
\end{table}

\begin{figure}
\centering\includegraphics[width=0.6\columnwidth]{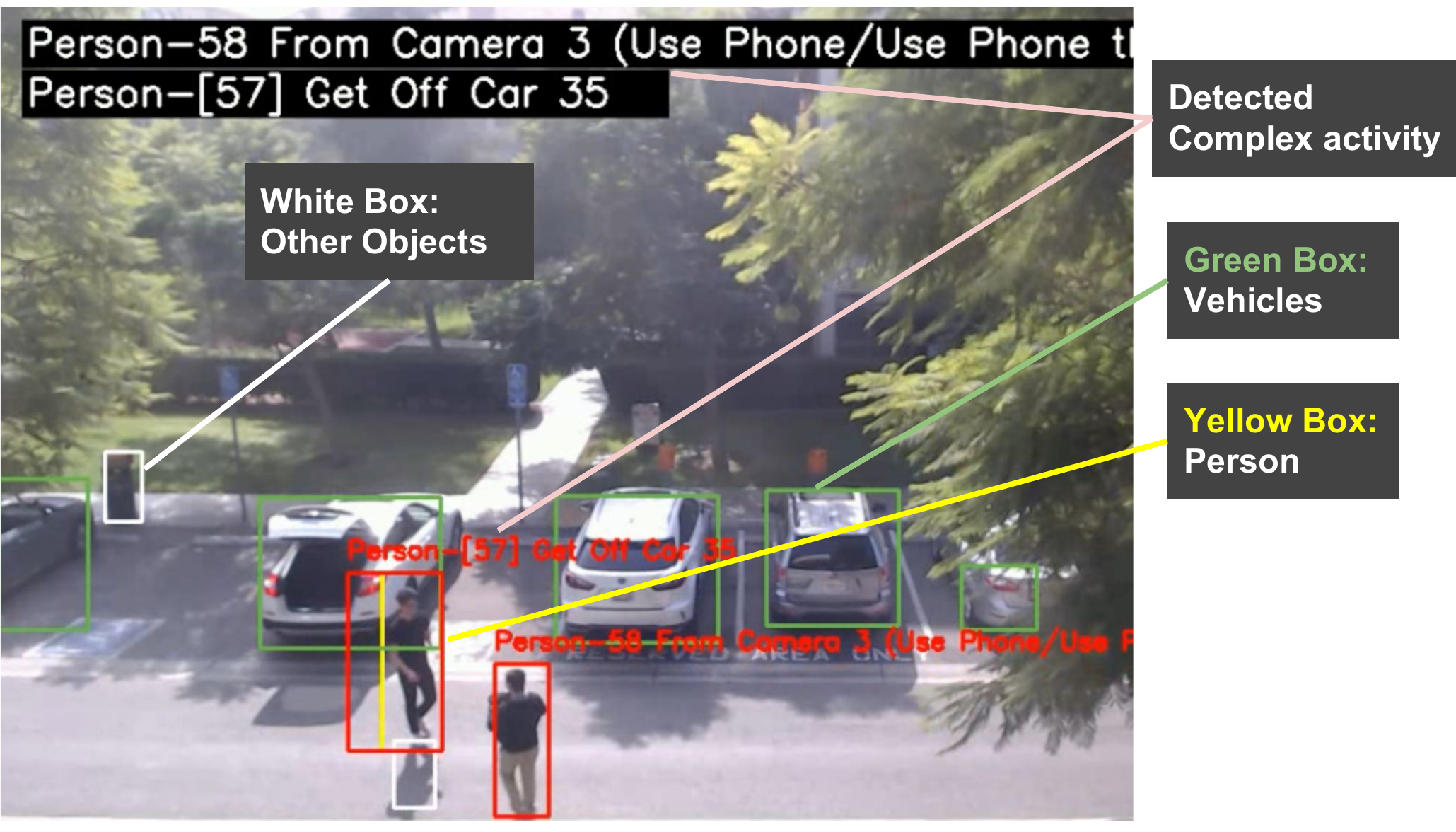}
\caption{\emph{The output of Caesar with annotations.}}
\label{fig:caesar_sys_out}
\end{figure}

\subsection{Rule Definition and Parsing}
\label{sec:rule-definition}

Caesar's first contribution is an extensible rule definition language.
Based on the observation that complex activity definitions specify
relationships in space and time between actors, objects, and/or atomic
actions (henceforth simply actions), the
language incorporates three different \emph{vocabularies}
(\caesarfig{vocabulary_elements}).

\parab{Vocabularies.} An \emph{element vocabulary} specifies the list
of actors or objects (\eg ``person'', ``bag'', ``bicycle'') and
actions (\eg ``talking on the phone''). As additional detectors for
atomic actions become available (\eg ``walking a dog'') from new DNNs
or new action definition rules, Caesar can incorporate corresponding
vocabulary extensions for these.

\begin{figure}
\centering\includegraphics[width=0.86\columnwidth]{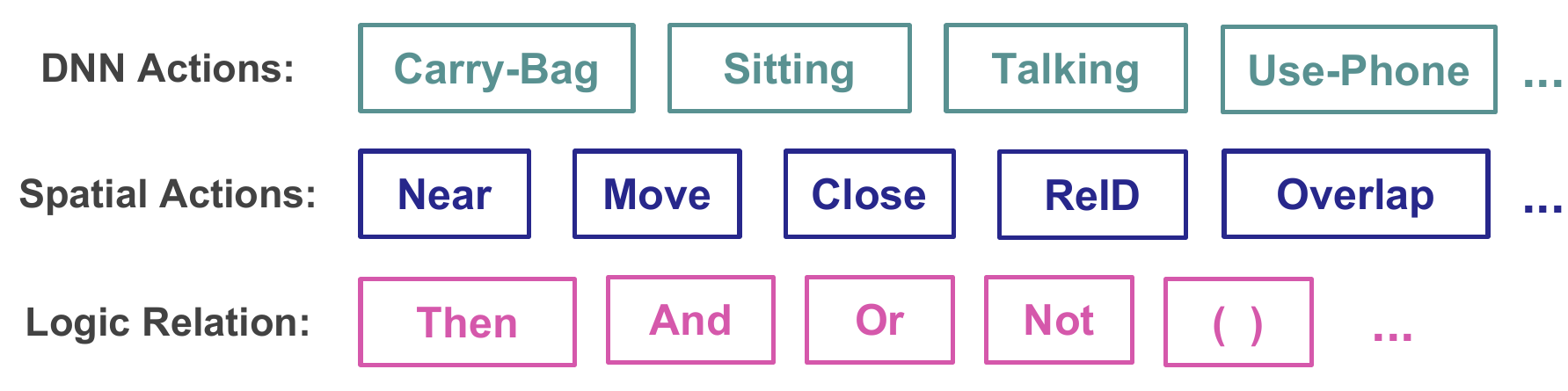}
\caption{\emph{Examples of the vocabulary elements.}}
\label{fig:caesar_vocabulary_elements}
\end{figure}


A \textit{spatial operator vocabulary} defines spatial relationships
between actors, objects, and actions. Spatial relationships use binary
operators such as ``near'' and ``approach''. For example, before a
person $p1$ can talk to $p2$, $p1$ must ``approach'' $p2$ and then
come ``near'' $p2$ (or vice versa). Unary operators such as ``stop''
or ``disappear'' specify the dispensation of participants or objects.
For example, after approaching $p2$, $p1$ must ``stop'' before he or
she can talk to $p2$. Another type of spatial operator is for
describing which camera an actor appears in. The operator
``re-identified'' specifies an actor recognized in a new camera. The
binary operator ``same-camera'' indicates that two actors are in the
same camera.

Finally, a \textit{temporal operator vocabulary} defines concurrent as
well as sequential activities. The binary operator ``then'' specifies
that one object or action is visible after another, ``and'' specifies
that two objects or actions are concurrently visible, while ``or''
specifies that they may be concurrently visible. The unary operator
``not'' specifies the absence of a corresponding object or action.


A complex activity definition contains three components
(\caesarfig{act_def}). The first is a unique name for the activity,
and the second is a set of variable names representing actors or
objects. For instance, $p1$ and $p2$ might represent two people, and
$c$ a car. The third is the definition of the complex activity in
terms of these variables. A complex activity definition is a sequence of
clauses, where each clause is either an action (\eg \texttt{p1
  use-phone}), or a unary or binary spatial operator (\eg \texttt{(p1
  close p2)}, or \texttt{(p1 move)}). Temporal operators link two
clauses, so a complex activity definition is a sequence of clauses
separated by temporal operators. \caesarfig{act_def} shows examples
of two complex actions, one to describe a person getting into a car,
and another to describe a person who is seen, in two different
cameras, talking on the phone while carrying a bag.

\begin{figure}
\centering\includegraphics[width=0.55\columnwidth]{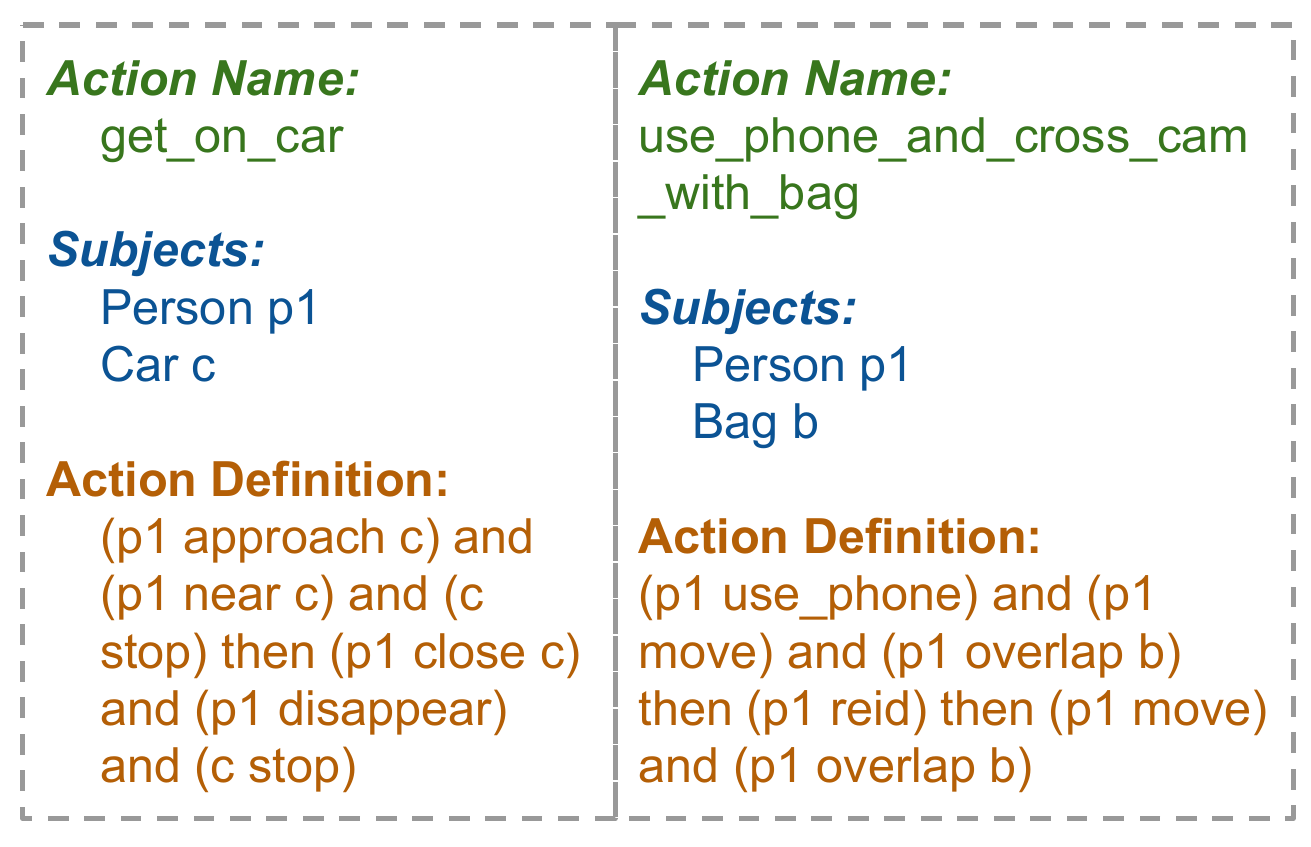}
\caption{\emph{Two examples of action definition using Caesar syntax.}}
\label{fig:caesar_act_def}
\end{figure}

\parab{The rule parser.}
Caesar parses each rule to extract an intermediate representation
suitable for matching. In Caesar, that representation is a directed
acyclic graph (or DAG), in which nodes are clauses and edges represent
temporal relationships.~\caesarfig{graph_sample} shows the parsed graphs
of the definition rules. At runtime, Caesar's graph matching component attempts to
match each complex activity DAG specification to the actors, objects,
and actions detected in the video feeds of multiple cameras.

\begin{figure}
\centering\includegraphics[width=0.65\columnwidth]{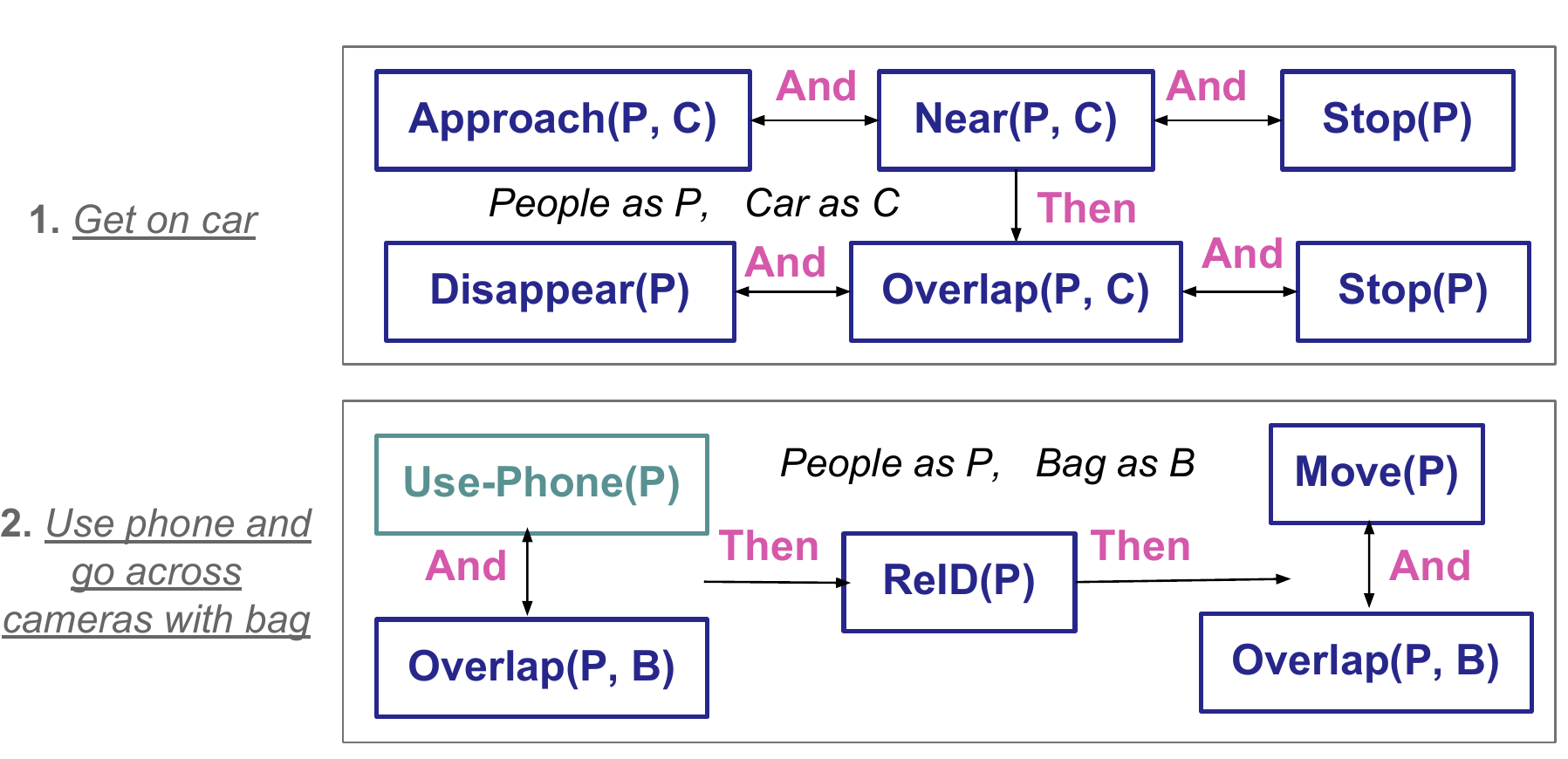}
\caption{\emph{Two examples of parsed complex activity graphs.}}
\label{fig:caesar_graph_sample}
\end{figure}

\subsection{Object Detection}
\label{sec:obj_detection}
\thispagestyle{empty}

\parab{On-camera object detection.} The first step in detecting a complex activity is detecting objects in frames. This component processes each frame, extracts a bounding box for each distinct object within the frame, and emits the box coordinates, the cropped image within the bounding box, and the object label. Today, DNNs like YOLO~\cite{yolov3} and SSD~\cite{ssd} can quickly and accurately detect objects. These detectors also have stripped-down versions that permit execution on a mobile device. Caesar allows a camera with on-board GPUs to run these object detectors locally. When this is not possible, Caesar schedules GPU execution on the edge cluster. (The next step in our pipeline involves re-identifying actors across multiple cameras, and cannot be easily executed on the mobile device).

\parab{Optimizing wireless bandwidth.} When the mobile device runs the
object detector, it may still be necessary to upload the cropped
images for each of the bounding boxes (in addition to the bounding box
coordinates and the labels). Surveillance cameras can see tens to
hundreds of people or cars per frame, so uploading images can be
bandwidth intensive. In Caesar, the mobile device maintains a cache of
recently seen images and the edge cluster \textit{lazily retrieves}
images from the mobile device to reduce this overhead.

Caesar is able to perform this optimization for two reasons. First, for
tracking and re-identification, not all images
might be necessary; for instance, if a person appears in 20 or 30
successive frames, Caesar might need only the cropped image of the
person from one of these frames for re-identification. Second, while
all images might be necessary for action detection, Caesar minimizes
invocation of the action detection module,
reducing the need for image transfer.

\begin{figure}
\centering\includegraphics[width=0.55\columnwidth]{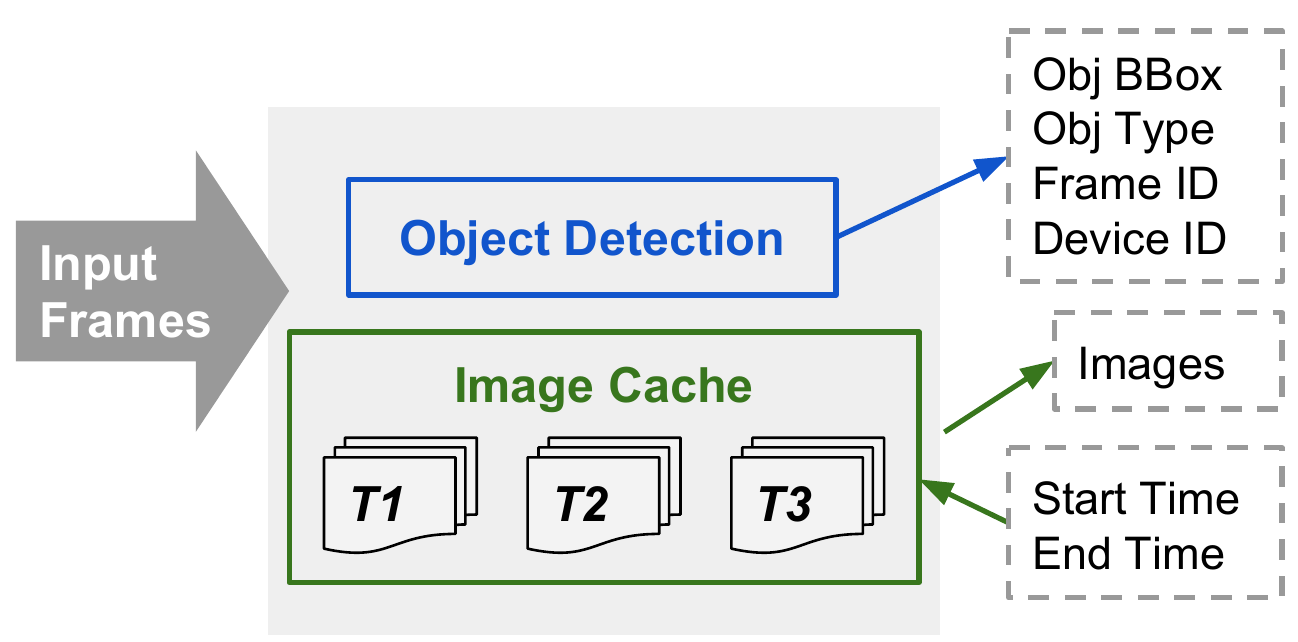}
\caption{\emph{Workflow of object detection on mobile device.}}
\label{fig:caesar_obj_det_crop}
\end{figure}

\subsection{Tracking and Re-Identification}
\label{sec:tracking}
\thispagestyle{empty}

\parab{The tube abstraction.} Caesar's expressivity in capturing complex
activities comes from the \textit{tube} abstraction. A tube is a
sequence of bounding boxes over successive frames that represent the
same object. As such, a tube has a distinct start and end time, and a
label associated with the object. Caesar's tracker
(\algref{tracking}) takes as input the sequence of bounding boxes
from the object detector, and assigns, to each bounding box a globally
unique \emph{tube ID}. In its rule definitions, Caesar detects spatial and temporal
relationships between tubes. Tubes also permit low overhead
re-identification, as we discuss below.
\begin{algorithm}[!ht]
\caption{Cross-Camera Tracking and Re-Identification}
\label{alg:tracking}
\begin{algorithmic}[1]
\STATE $\boldsymbol{INPUT}:\ list\ of\ bounding\ boxes$
\FOR{$each\ person\ box\ B\ \boldsymbol{in}\ bounding\ boxes$}
\STATE $ID_B\ =\ update\_tubes(existing\_tubes,\ B)$
\IF{$ID_B\ \boldsymbol{not\ in}\ existing\_tubes$}
\STATE $frame\ =\ get\_frame\_from\_camera()$
\STATE $F_B\ =\ get\_DNN\_feature(frame,\ B)$
\FOR{$ID_{local}\ \boldsymbol{in}\ local\_tubes$}
\IF{$feature\_dist(ID_{local},\ ID_B)$}
\STATE $ID_B\ =\ ID\_{local};\ Update(ID\_{local});\ \boldsymbol{return}$
\ENDIF
\ENDFOR
\FOR{$ID_{other}\ \boldsymbol{in}\ other\_tubes$}
\IF{$feature\_dist(ID_{other},\ ID_B)$}
\STATE $ID_B\ =\ ID\_{other};\ Update(ID\_{other});\ \boldsymbol{return}$
\ENDIF
\ENDFOR
\STATE $Update(ID\_B);$
\ENDIF
\ENDFOR
\end{algorithmic}
\end{algorithm}

\parab{Tracking.} The job of the tracking sub-component is to extract
tubes. This sub-component uses a state-of-the-art tracking algorithm 
called DeepSORT~\cite{deepsort} that runs on the edge server side (line 6,
\algref{tracking}). DeepSORT takes as input bounding box
positions and extracts features of the image within the bounding box. It then tracks the bounding boxes using Kalman filtering,
as well as judging the image feature and the intersection-over-union between successive bounding boxes.

Caesar receives bounding boxes from the object detector and passes it to
DeepSORT, which either associates the bounding box with an existing
tube or fails to do so. In the latter case, Caesar starts a new tube
with this bounding box. As it runs, Caesar's tracking component
continuously saves bounding boxes and their tube ID associations to a
distributed key-value store within the edge cluster, described below,
that enables fast tube matching in subsequent steps.

Caesar makes one important performance optimization. Normally, DeepSORT
needs to run a DNN for person re-identification features.
This is feasible when object detection runs on the edge cluster. However, when object detection runs on
the mobile device, feature extraction can require additional compute
and network resources, so Caesar relies entirely on DeepSORT's ability
to track using bounding box positions alone. This design choice permits Caesar to conserve
wireless network bandwidth by transmitting only bounding box
positions instead of uploading the whole frame.

\parab{Robust tracking.} When in the view of the camera, an object or
actor might be partially obscured. If this happens, the tracking
algorithm detects two distinct tubes. To be robust to partial
occlusions, Caesar retrieves the cropped image corresponding to the
first bounding box in the tube (line 5, \algref{tracking}). Then,
it applies a re-identification DNN (described below) to match this
tube with existing tubes detected in the local camera (lines 7-10,
\algref{tracking}). If it finds a match, Caesar uses simple
geometric checks (\eg bounding box continuity) before assigning the
same identifier to both tubes.

\parab{Cross-camera re-identification.} Cross camera
\textit{re-identification} is the ability to re-identify a person or
object between two cameras. Caesar uses an off-the-shelf
DNN~\cite{Re-ID} which, trained on a corpus of
images, outputs a feature vector that uniquely identifies the input
image. Two images belong to the same person if the distance between
the feature vectors is within a predefined threshold.

To perform re-identification, Caesar uses the image retrieved for robust
tracking, and searches a \emph{distributed key-value store} for a
matching tube from another camera.
Because the edge cluster can have multiple servers, and different
servers can process feeds from different cameras, Caesar uses a fast
in-memory distributed key-value store~\cite{redis} to save tubes.

Re-identification can incur a high false positive rate. To make it
more robust, we encode the camera topology~\cite{Qiu18c} in the
re-identification sub-component. In this topology, nodes are cameras,
and an edge exists between two cameras only if a person or a car can
go from one camera to another without entering the field of view of
any other \textit{non-overlapping} camera. Given this, when Caesar tries
to find a match for a tube seen at camera \textbf{A}, it applies the
re-identification DNN only to tubes at neighbors of \textbf{A} in the
topology. To scope this search, Caesar uses travel times between
cameras~\cite{Qiu18c}.

\subsection{Action Detection and Graph Matching}
\label{sec:act_matching}
\thispagestyle{empty}

In a given set of cameras, users may want to detect multiple complex
activities, and multiple instances of each activity can occur. Caesar's
rule parser generates an intermediate
graph representation for each complex activity, and the \textit{graph
  matching} component matches tubes to the graph in order to detect
when a complex activity has occurred. For reasons discussed below,
graph matching dynamically invokes atomic action detection, so we
describe these two components together in this section.

\parab{Node matching.} The first step in graph matching is to match
tubes to nodes in one or more graphs. Recall that a node in a graph
represents a clause that describes spatial actions or spatial
relationships. Nodes consist of a unary or binary operator, together
with the corresponding operands. For each operator, Caesar defines
algorithm to evaluate the operator.

\parae{Matching unary operators.}
For example, consider the clause
\texttt{stop c}, which determines whether the car \texttt{c} is
stationary. This is evaluated to true if the bounding box for
\texttt{c} has the same position in successive frames. Thus, a tube
belonging to a stationary car matches the node \texttt{stop c} in each
graph, and Caesar binds \texttt{c} to its tube ID.

Similarly, the unary operator \texttt{disappear} determines if its
operand is no longer visible in the camera. The algorithm to evaluate
this operator considers two scenarios: an object or person
disappearing (visible in one frame and not visible in the next) by (a)
entering the vehicle or building, or (b) leaving the camera's field of
view. When either of these happen, the object's tube matches the
corresponding node in a graph.

\parae{Matching binary operators.} For binary operators, node matching
is a little more involved, and we explain this using an example.
Consider the clause \texttt{p1 near p2}, which asks: is there a person
near another person? To evaluate this, Caesar checks each pair of person
tubes to see if there was any instant at which the corresponding
persons were close to each other. For this, it divides up each tube
into small chunks of duration $t$ (1 second in our implementation),
and checks for the \emph{proximity} of all bounding boxes pairwise in
each pair of chunks.

To determine proximity, Caesar uses the following metric. Consider two
bounding boxes $x$ and $y$. Let $d(x,y)$ be the smallest pixel
distance between the outer edges of the bounding box. Let $b(x)$
(respectively $b(y)$) be the largest dimension of bounding box $x$
(respectively, $y$). Then, if either $\frac{d(x,y)}{b(x)}$ or
$\frac{d(x,y)}{b(y)}$ is less than a fixed threshold $\delta$ we say
that the two bounding boxes are proximate to each other. Intuitively,
the measure defines proximity with respect to object dimensions: two
large objects can have a larger pixel distance between them than two
small objects, yet Caesar may declare the larger objects close to each
other, but not the smaller ones.


Finally, \texttt{p1 near p2} is true for two people tubes if there is
a chunk within those tubes in which a majority of bounding boxes are
proximate to each other. We use the majority test to be robust 
in bounding box determinations in the underlying object detector.

Caesar includes similar algorithms for other binary spatial operators.
For example, the matching algorithm for \texttt{p1 approaches p2} is a
slight variant of \texttt{p1 near p2}: in addition to the proximity
check, Caesar also detects whether bounding boxes in successive frames
decrease in distance just before they come near each other.

\parae{Time of match.} In all of these examples, a match occurs within
a specific time interval $(t_1,t_2)$. This time interval is crucial
for edge matching, as we discuss below.

\parab{Edge matching.} In Caesar's intermediate graph representation,
an edge represents a temporal constraint. We permit two types of
temporal relationships: concurrent (represented by \texttt{and} which
requires that two nodes must be concurrent, and \texttt{or} which
specifies that two nodes may be concurrent), and sequential (one node
occurs strictly after another).

To illustrate how edge matching works, consider the following example.
Suppose there are two matched nodes \texttt{a} and \texttt{b}. Each
node has a time interval associated with the match. Then \texttt{a}
and \texttt{b} are concurrent if their time intervals overlap.
Otherwise, \texttt{a then b} is true if \texttt{b}'s time interval is
strictly after \texttt{a}'s.

\parab{Detecting atomic actions.} Rule-based activity detection has
its limits. Consider the atomic action ``talking on the phone''. One
could specify this action using the rule \texttt{p1 near m}, where
\texttt{p1} represents a person and \texttt{m} represents a mobile
phone. Unfortunately, phones are often too small in
surveillance videos to be captured by object detectors. 
DNNs, when trained on a large number of samples of people using phones, 
can more effectively detect this atomic action.

\parae{Action matching.} For this reason, Caesar rules can include clauses matched by a DNN. For example, the clause \texttt{talking\_phone(p1)} tries to find a person tube by applying each tube to a DNN. \changed{For this, Caesar uses the DNN described in~\cite{ulutan2018actor}. We have trained this on Google's AVA~\cite{ava} dataset which includes 1-second video segments from movies and action annotations. The training process ensures that the resulting model can detect atomic actions in surveillance videos without additional fine-tuning; see~\cite{ulutan2018actor} for additional details. The model can recognize a handful of actions associated with person tubes, such as: ``talking on the phone'', ``sitting'', and ``opening a door''. For each person tube, it uses the Inflated 3D features~\cite{carreira2017quo} to extract features which represent the temporal dynamics of actions, and returns a list of possible action labels and associated confidence levels.} Given this, we say that a person tube matches \texttt{talking\_phone(p1)} if there is a video chunk in which ``talking on the phone'' has a higher confidence value than a fixed threshold $\tau$.

\thispagestyle{empty}

\parae{Efficiency considerations.} In Caesar's rule definition language,
an action clause is a node. Matching that node requires running the
DNN on every chunk of every tube. This is inefficient for two reasons.
The first is \textit{GPU inefficiency}: the DNN takes about 40~ms for
each chunk, so a person who appears in the video for 10~s would
require 0.4~s to process (each chunk is 1~s) unless Caesar
provisions multiple GPUs to evaluate chunks in parallel. The second is
\textit{network inefficiency}. To feed a person tube to the DNN, Caesar
would need to retrieve \emph{all} images for that person tube from the
mobile device.

\parae{Lazy action matching.} To address these inefficiencies, Caesar
matches actions lazily: it first tries to match all non-action clauses
in the graph, and only then tries to match actions. To understand how
this works, consider the rule definition \texttt{a then b then c},
where \texttt{a} and \texttt{c} are spatial clauses, and \texttt{b} is
a DNN-based clause. Now, suppose \texttt{a} occurs at time $t_1$ and
\texttt{c} at $t_2$, Caesar executes the DNN on all tubes that start
after $t_1$ and end before $t_2$ in order to determine if there is a
match. This addresses both the GPU and network inefficiency discussed
above, since the DNN executes fewer tubes and Caesar retrieves fewer
images.

\algref{graph_matching} shows the complete graph matching algorithm. The input contains the current frame number as timestamp and a list of positions of active tubes with their tube IDs and locations. If Caesar has not completed assembling a tube (\eg the tube's length is less than 1 second), it appends the tube images to the temporary tube videos and records the current bounding box locations (lines 2-3). When a tube is available, Caesar attempts to match the spatial clauses in one or more complex activity definition graphs (line 5). Once it determines a match, Caesar marks the vertex in that graph as done, and checks the all its neighbor nodes in the graph. If the neighbor is a DNN node, it adds a new entry to the DNN checklist of the graph, and moves on to its neighbors. The entry contains the tube ID, DNN action label, starting time, and the ending time. The starting time is the time when the node is first visited. The end time is the time when Caesar matches its next node. In our example above, when \texttt{a} is matched at timestamp $T1$, Caesar creates an entry for \texttt{b} in this graph, with the starting time as $T1$. When \texttt{c} is matched at $T2$, the algorithm adds $T2$ to \texttt{b}'s entry as the ending timestamp.

\begin{figure}
\centering\includegraphics[width=0.6\columnwidth]{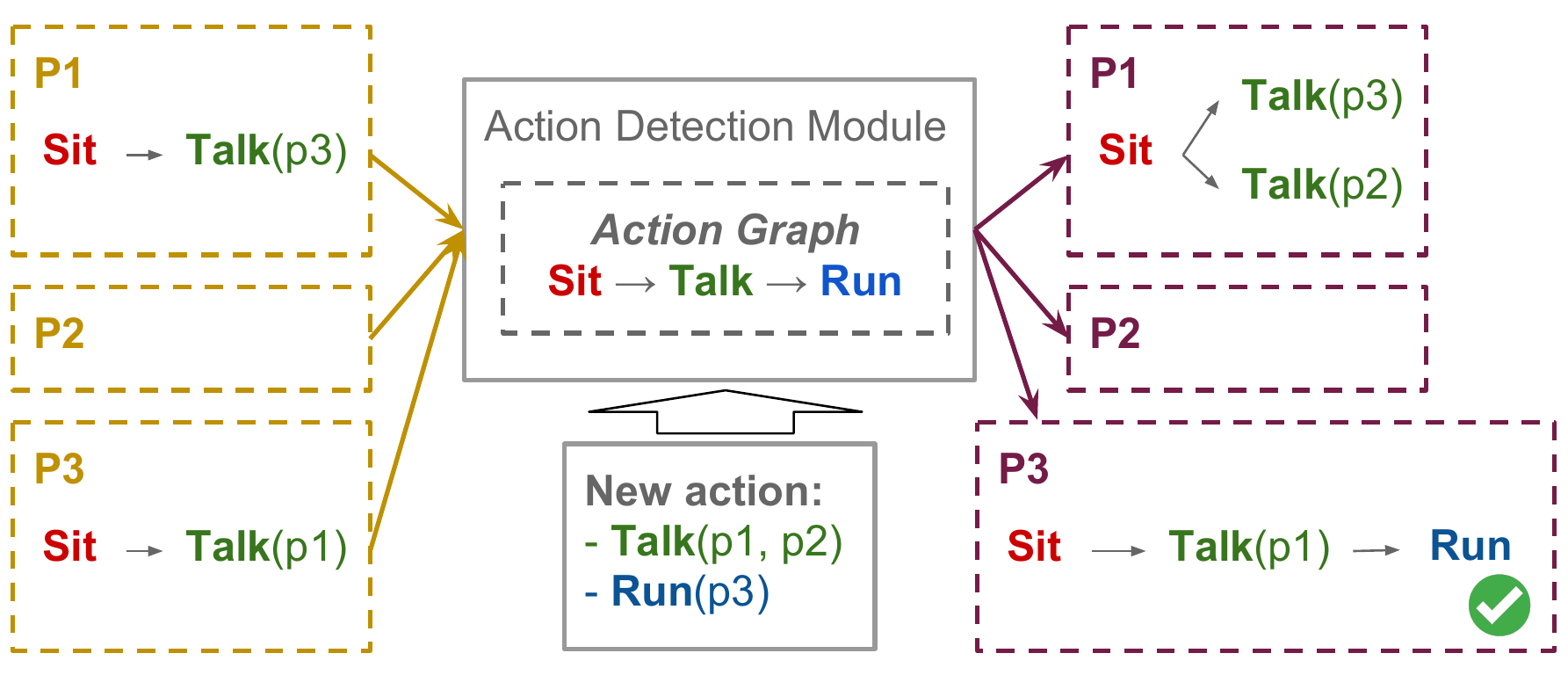}
\caption{\emph{An example of graph matching logic: the left three graphs are unfinished graphs of each tube, and the right three graphs are their updated graphs.}}
\label{fig:caesar_graph_matching}
\end{figure}

\begin{table}[]
\centering
\begin{tabular}{|l|l|}
\hline
\textit{\textbf{DNN}} & \textit{\textbf{Speed (FPS)}} \\ \hline
Object Detection~\cite{ssd,yolov3}      & 40$\sim$60                    \\ \hline
Tracking \& ReID~\cite{deepsort}      & 30$\sim$40                    \\ \hline
Action Detection~\cite{ulutan2018actor}      & 50$\sim$60 (per tube)         \\ \hline
\end{tabular}
\caption{\emph{The runtime frame rate of each DNN model used by Caesar (evaluated on a single desktop GPU~\cite{t2080}).}}
\label{table:dnn_speed}
\end{table}

\begin{algorithm}
\caption{Activity Detection with Selective DNN Activation}
\label{alg:graph_matching}
\begin{algorithmic}[1]
\STATE $\boldsymbol{INPUT}:\ incoming\ tube$
\IF{$tube\_cache\ \boldsymbol{not}\ full$}
\STATE $tube\_cache.add(tube);\ \boldsymbol{return}$
\ENDIF
\STATE $spatial\_acts\ =\ get\_spatial\_actions\ (tube\_cache)$
\FOR{$sa\ \boldsymbol{in}\ spatial\_acts$}
\FOR{$g\ \boldsymbol{in}\ tube\_graph\_mapping\ [sa.tube\_id]$}
\IF{$sa\ \boldsymbol{not\ in}\ g.next\_acts$}
\STATE $\boldsymbol{continue}$
\ENDIF
\IF{$g.has\_pending\_nn\_act$}
\STATE $nn\_acts\ =\ get\_nn\_actions\ (g.nn\_start,\ cur\_time())$
\IF{$g.pending\_nn\_act\ \boldsymbol{not\ in}\ nn\_acts$}
\STATE $\boldsymbol{continue}$
\ENDIF
\ENDIF
\STATE $g.next\_acts\ =\ sa.neighbors()$
\STATE $tube\_graph\_mapping.update()$
\IF{$g.last\_node\_matched$}
\STATE $add\ g\ to\ output\ activities$
\ENDIF
\ENDFOR
\ENDFOR
\end{algorithmic}
\end{algorithm}

%% file: tex/caesar/eval.tex
\section{Evaluation}

In this section, we evaluate Caesar's accuracy and scalability on a
publicly available multi-camera data set.

\subsection{Methodology}

\parab{Implementation and experiment setup.} Our experiments use an
implementation of Caesar which contains (a) object
detection and image caching on the camera, (b) tracking,
re-identification, action detection, and graph matching on the edge
cluster. Caesar's demo code is available at \textit{https://github.com/USC-NSL/Caesar}.

In our experiments, we use multiple cameras equipped 
with Nvidia's TX2 GPU boards~\cite{tx2}. Each platform
contains a CPU, a GPU, and 4~GB memory shared between the CPU and the
GPU. Caesar runs a DNN-based object detector,
SSD-MobilenetV2~\cite{sandler2018mobilenetv2}, on the camera. As
described earlier, Caesar caches the frames on the camera, as
well as cropped images of the detected bounding boxes. It sends box
coordinates to the edge cluster using RPC~\cite{grpc}. The server can
subsequently request a camera for additional frames, permitting
lazy retrieval.

A desktop with three Nvidia RTX 2080 GPUs~\cite{t2080} runs Caesar on
the server side. One of the GPUs runs the state-of-the-art online
re-identification DNN~\cite{Re-ID} and the other two execute the
action detection DNNs~\cite{ulutan2018actor}. Each action DNN instance
has its own input data queue so Caesar can load-balance action detection
invocations across these two for efficiency. We use Redis~\cite{redis}
as the in-memory key-value storage. Our implementation also includes
Flask~\cite{flask} web server that allows users to input complex
activity definitions, and visualize the results. 

\parab{DNN model selection.} The action detector and the
ReID DNN require 7.5~GB of memory, far more than the 4~GB available on
the camera. This is why, as described earlier, our mobile
device can only run DNN-based object detection. Among the available
models for object detection, we have evaluated four that fit within
the camera's GPU memory. \changed{Table~\ref{table:obj_dnn_choice} shows
the accuracy and speed (in frames per second) of these models on our evaluation dataset. Our experiments use SSD-Mobilenet2 because it has good accuracy
with high frame rate, which is crucial for Caesar because a higher frame
rate can lead to higher accuracy in detecting complex activities.}

\added{
We use~\cite{Re-ID} for re-identification  because it is lightweight
enough to not be the bottleneck. Other 
ReID models~\cite{sun2018beyond, zheng2019joint}  are more accurate than~\cite{Re-ID} on our dataset (tested offline), but are too slow (< 10~fps) to use in Caesar. 
For atomic actions, other options~\cite{DBLP:journals/corr/abs-1812-02707, sun2018actor} 
have slight higher accuracy than~\cite{ulutan2018actor} on the AVA dataset,
but are not publicly available yet and their performance is not reported.}

\parab{Dataset.} We use DukeMTMC~\cite{dukemtmc} for our evaluations.
It has videos recorded by eight non-overlapping surveillance cameras
on campus. The dataset also contains annotations for each person,
which gives us the ground truth of each person tube's position at any
time. We selected 30 minutes of those cameras' synchronized videos for
testing Caesar. There are 624 unique person IDs in that period, and each
person shows up in the view of 1.7 cameras on average. The average number
of people showing up in all cameras is 11.4, and the maximum is 69.

\parae{Atomic action ground truth.}
DukeMTMC was originally designed for testing cross-camera person
tracking and ReID, so it does not have any action-related ground
truth. Therefore, we labeled the atomic actions in each frame for
each person, using our current action vocabulary. Our action ground
truth contains the action label, timestamp, and the actor's person ID.
We labeled a total of 1,289 actions in all eight cameras. Besides the
atomic actions, we also labeled the ground truth traces of cars and
bikes in the videos. \caesarfig{cam_topo} shows the placement of
these cameras and a sample view from each camera.

\begin{figure}
\centering\includegraphics[width=0.6\columnwidth]{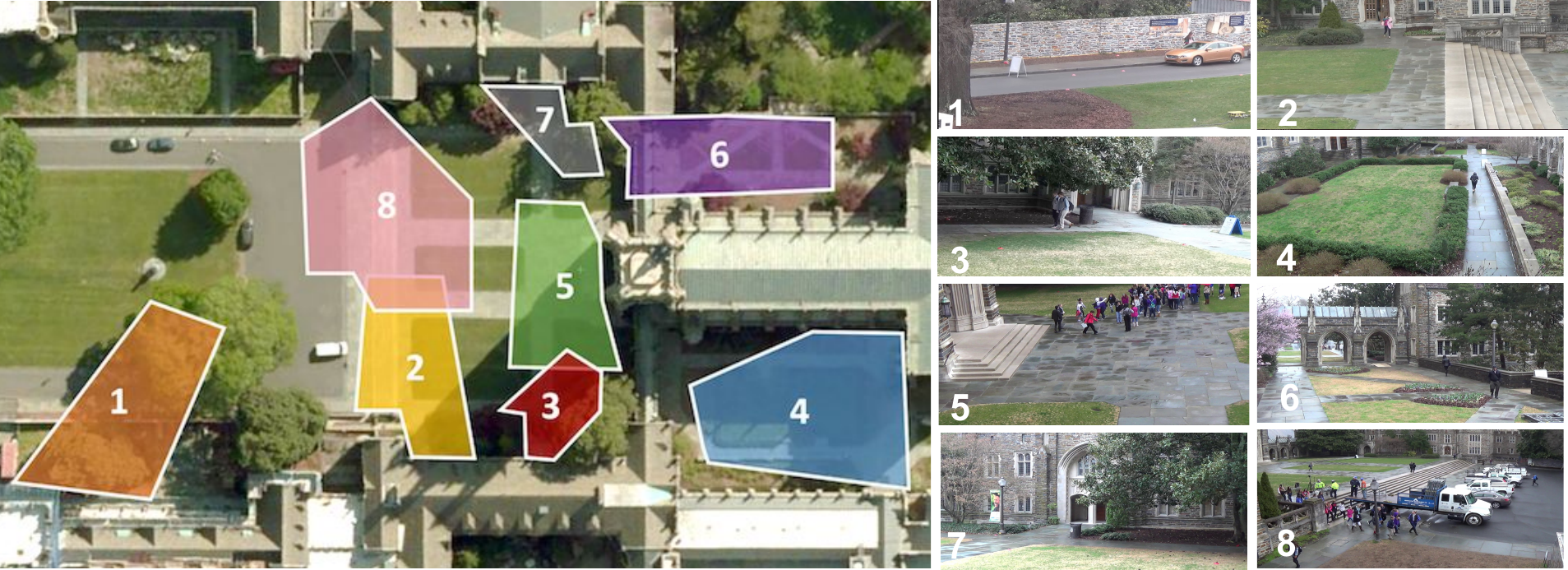}
\caption{\emph{Camera placement and the content of each camera.}}
\label{fig:caesar_cam_topo}
\end{figure}

\parae{Complex activity ground truth.} We manually identified 149
complex activities. There are seven different categories of these
complex activities as shown in Table~\ref{table:comp_act_list}. This
table also lists two other attributes of the complex activity type and
the dataset. The third column of the table shows the number of
instances in the data set of the corresponding complex activity,
broken down by how many of them are seen on a single camera vs.
multiple cameras. Thus, for the first complex activity, the entry
\textit{12/1} means that our ground-truth contains 12 instances that
are visible only on a single camera, and one that is visible across
multiple cameras.

These complex activities are of three kinds. \#1's clauses, labeled
``NN-only'', are all atomic actions detected using a NN. \#2 through
\#5, labeled ``Mixed'', have clauses that are either spatial or are
atomic actions. The last two, \#6 and \#7, labeled ``Spatial-only'',
have only spatial clauses.

\parab{Metrics.} We replay the eight videos at 20~fps to simulate the
realtime camera input on cameras. Caesar's server takes the input
from all mobile nodes, runs the action detection algorithm for a
graph, and outputs the result into logs. The results contain the
activity's name, timestamp, and the actor or object's bounding box
location when the whole action finishes. Then we compare the log with
the annotated ground truth. A \textit{true positive} is when the
detected activity matches the ground truth's complex activity label,
overlaps with the ground truth's tubes for the complex activity, and
has timestamp difference within a 2-second threshold. We report:
\textit{recall}, which is the fraction of the ground truth classified
as true positives, and \textit{precision}, which is the fraction of
true positives among all Caesar-detected complex activities. We also
evaluate the Caesar's scalability, as well as the impact of its
performance optimizations; we describe the corresponding metrics for
these later.

\begin{table}[]
\centering
\resizebox{0.8\linewidth}{!}{
\begin{tabular}{|l|l|l|}
\hline
\textit{\textbf{DNN}} & \textit{\textbf{Speed (FPS)}} & \textit{\textbf{Accuracy (mAP)}} \\ \hline
SSD~\cite{ssd}                   & 3.7                           & 91                               \\ \hline
YOLOv3~\cite{yolov3}                & 4.1                           & 88                               \\ \hline
TinyYOLO~\cite{redmon2016yolo9000}              & 8.5                          & 84                               \\ \hline
SSD-MobileNetv2~\cite{sandler2018mobilenetv2}       & 11.2                            & 83                               \\ \hline
\end{tabular}}
\caption{\emph{Speed and accuracy of different DNNs on mobile GPU.}}
\label{table:obj_dnn_choice}
\end{table}

\begin{figure}[t]
  \begin{minipage}{0.32\linewidth}
    \centerline{\includegraphics[width=0.99\columnwidth]{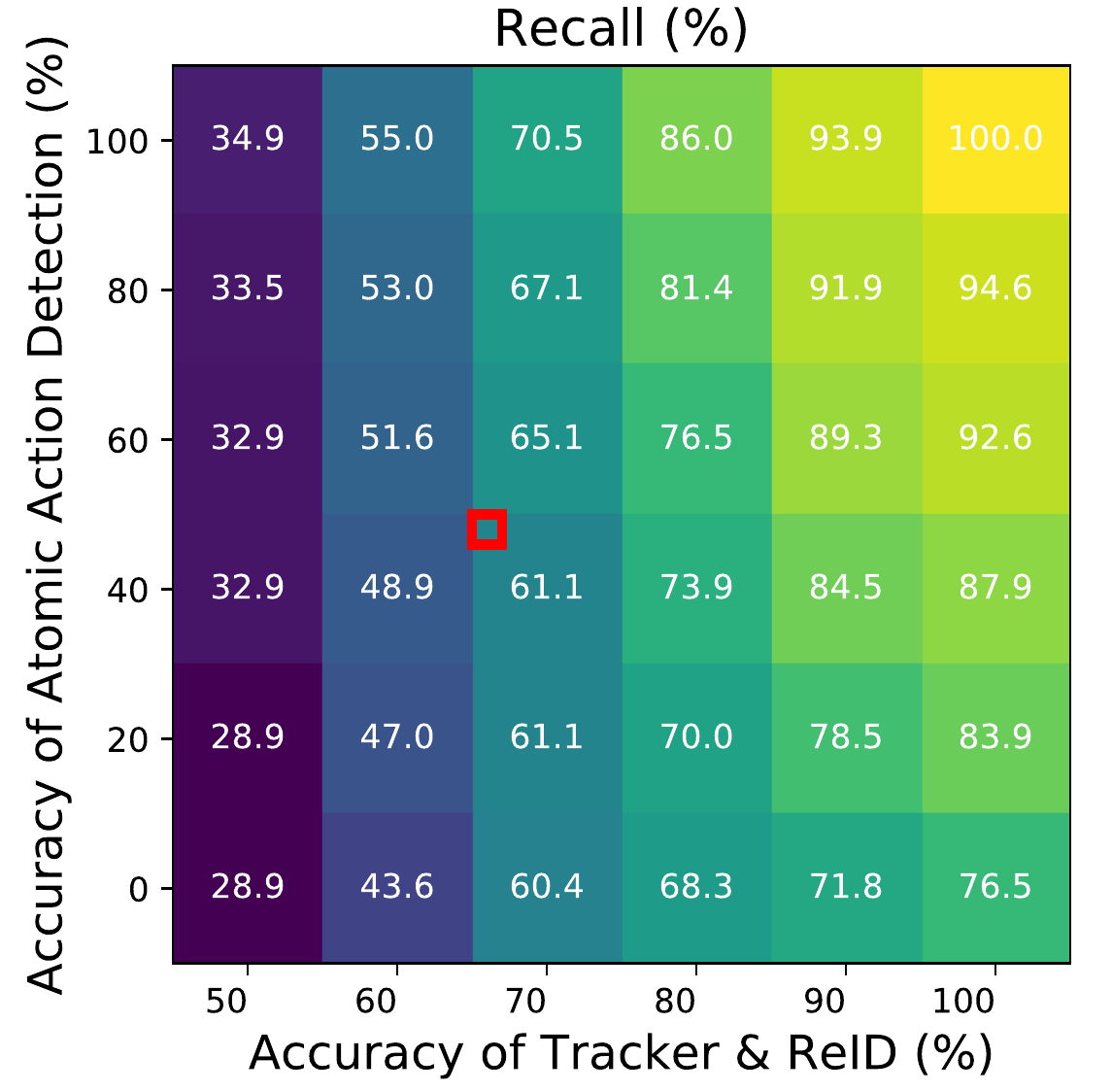}}
    \centerline{(a)}
  \end{minipage}
    \begin{minipage}{0.32\linewidth}
    \centerline{\includegraphics[width=0.99\columnwidth]{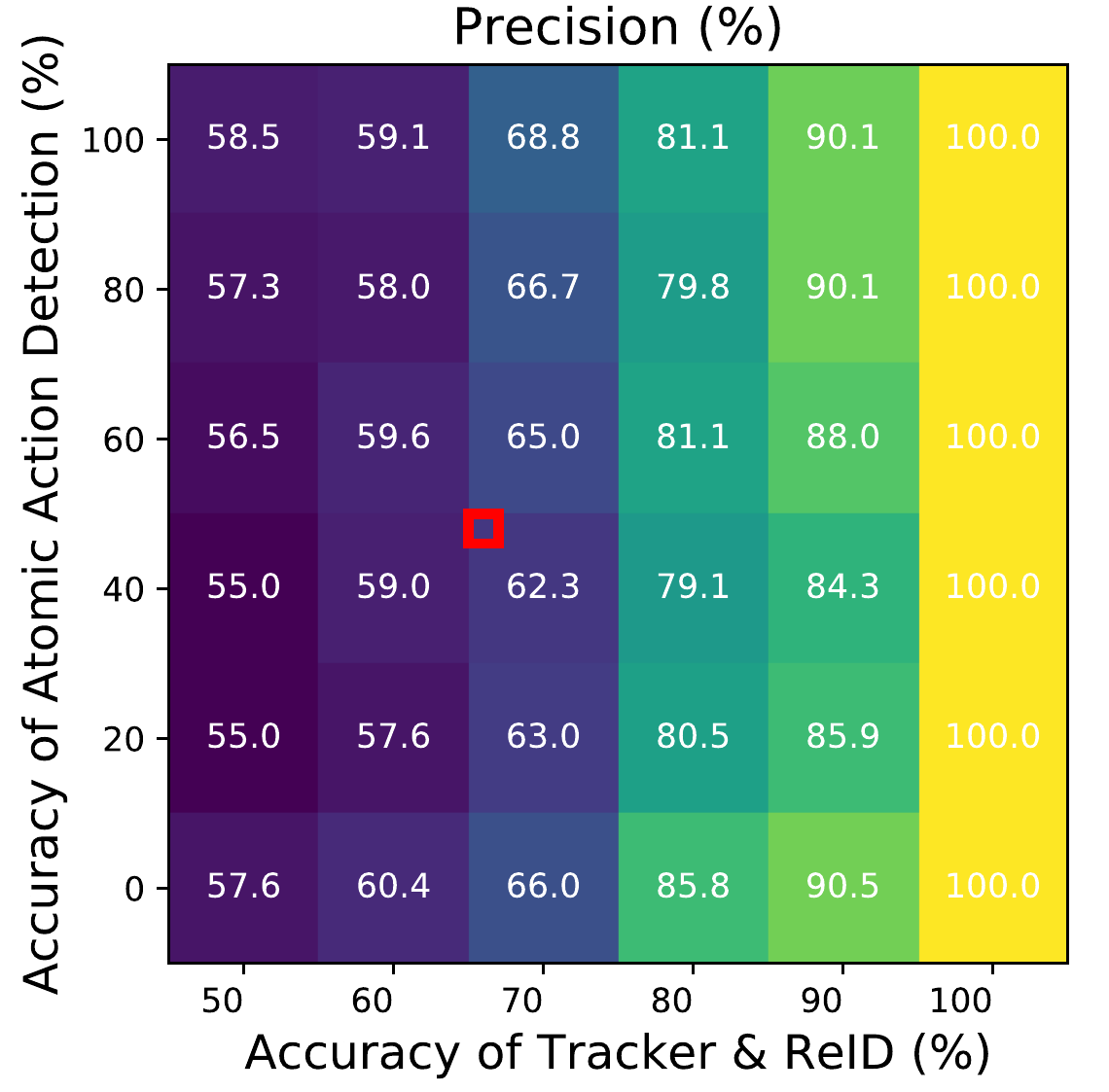}}
    \centerline{(b)}
  \end{minipage}
    \begin{minipage}{0.32\linewidth}
    \centerline{\includegraphics[width=0.99\columnwidth]{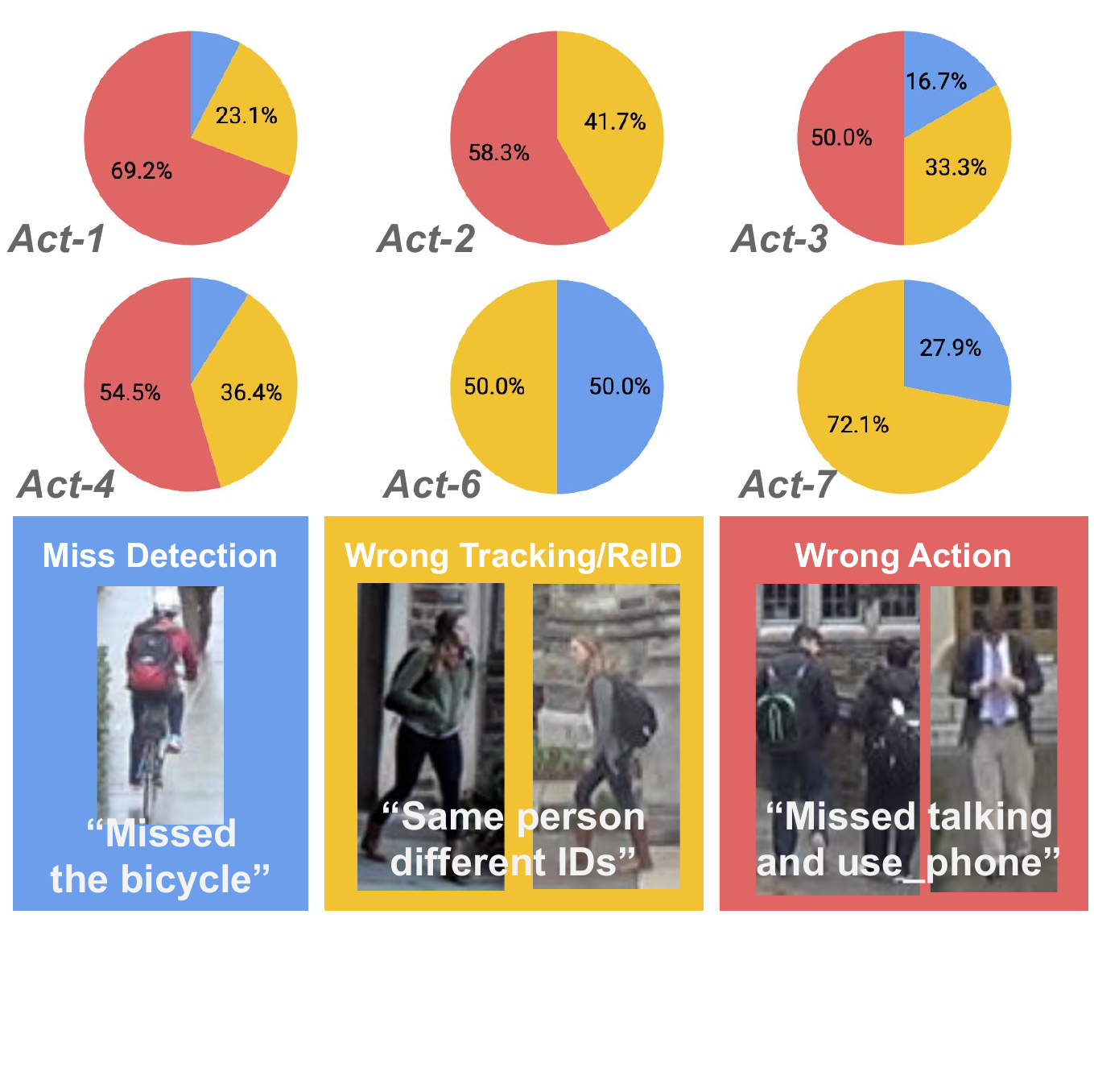}}
    \centerline{(c)}
  \end{minipage}
  \caption{\emph{Caesar's (a) recall rate and (b) precision rate with different action detection and tracker accuracy. (c) The statistics and sample images of failures in all complex activities.}}
  \label{fig:caesar_eval_all}
\end{figure}

\begin{table}[]
\resizebox{\linewidth}{!}{
\begin{tabular}{|l|l|l|l|}
\hline
\textit{\textbf{ID}} & \textit{\textbf{Complex Activity}}  & \textit{\textbf{\begin{tabular}[c]{@{}l@{}}\# of Samples  \\ (Single/Multi)\end{tabular}}} & \textit{\textbf{Type}} \\ \hline
1                    & Use phone then talk              & 12 / 1                                                                                       & {\color[HTML]{6200C9} NN-only}      \\ \hline
2                    & Stand, use phone then open door  & 9 / 2                                                                                        & {\color[HTML]{329A9D} Mixed}          \\ \hline
3                    & Approach and give stuff          & 10 / 0                                                                                       & {\color[HTML]{329A9D} Mixed}          \\ \hline
4                    & Walk together then stop and talk & 6 / 2                                                                                        & {\color[HTML]{329A9D} Mixed}          \\ \hline
5                    & Load stuff and get on car        & 2 / 0                                                                                        & {\color[HTML]{329A9D} Mixed}          \\ \hline
6                    & Ride with bag in two cams        & 0 / 8                                                                                        & {\color[HTML]{ABB000} Spatial-only} \\ \hline
7                    & Walk together in two cams        & 0 / 97                                                                                       & {\color[HTML]{ABB000} Spatial-only} \\ \hline
\end{tabular}}
\caption{\emph{Summary of labeled complex activities}}
\label{table:comp_act_list}
\end{table}

\subsection{Accuracy}

\parab{Overall.} Table~\ref{table:overall_accuracy} shows the recall and
precision of all complex activities. \#1 (using the phone and then
talking to a person) and \#4 (walking together then stopping to talk)
have the lowest recall at 46.2\% and the lowest precision at 36.4\%.
At the other end, \#5's two instances achieve 100\% recall and
precision. Across all complex activities, Caesar has a recall of \%61.0
and a precision of \%59.5 precision.

\parab{Understanding the accuracy results.} 
Our results show that most NN-only and Mixed activities have
lower position and recall than those in the Spatial-only category.
Recall that Caesar uses off-the-shelf neural networks for action
detection and re-identification. This suggests that the action
detection DNN, used in the first two categories but not in the third,
is the larger source of detection failures than the re-identification DNN. Indeed,
the reported mean average precision for these two models are
respectively 45\% and 65\% in our dataset.

We expect the overall accuracy of complex activity detection to
increase in the future for two reasons. We use off-the-shelf networks
for these activities that are not customized for this camera. There is
significant evidence that customization can improve the accuracy of
neural networks~\cite{Satyam} especially for surveillance cameras
since their viewing angles are often different from the images used
for training these networks. Furthermore, these are highly active
research areas, so with time we can expect improvements in accuracy.

Two natural questions arise: (a) as these neural networks improve, how
will the overall accuracy increase? and (b) to what extent does Caesar's
graph matching algorithm contribute to detection error? We address
both of these questions in the following analysis.

\parab{Projected accuracy improvements.} 
We evaluate Caesar with the tracker and the action detector at different accuracy levels.
To do this, for each of these components, we vary the
accuracy $p$ (expressed as a percentage) as follows. We re-run our
experiment, but instead of running the DNNs when Caesar invokes the
re-identification and action detection components, we return the ground truth
\textit{for that DNN} (which we also have) $p$\% of the time, else
return an incorrect answer. By varying $p$, we can effectively
simulate the behavior of the system as DNN accuracy improves. 
\added{When $p$ is 100\%, the re-identification DNN works perfectly and
Caesar always gets the correct person ID in the same camera and across cameras, so tracking is 100\% accurate. Similarly, when the action DNN has 100\% accuracy, it always captures every atomic action correctly for each person.}
We then compare the end-to-end result with the complex activity ground-truth
to explore precision and accuracy.

\caesarfig{eval_all}(a) and~\caesarfig{eval_all}(b) visualize the
recall and precision of Caesar with different accuracy in the tracker
and the atomic action detector. In both of these figures, the red box
represents Caesar's current performance (displayed for context).

The first important observation from these graphs is that, when the
action detection and re-identification DNNs are perfect, Caesar achieves 100\%
precision and recall. This means that the rest of the system,
  which is a key contribution of the chapter, works perfectly; this
includes matching the spatial clauses, and the temporal relationships,
while lazily retrieving frames and bounding box contents from the
camera, and using shared key-value store to help perform
matching across multiple cameras.

The second observation from this graph is that the re-identification
DNN's accuracy affects overall performance more than that of the
action detector. \changed{Recall that the re-identification DNN tracks
people both within the same camera and across cameras}. 
There are two reasons for why it affects performance
more than the action detector. \changed{The first is the dependency between
re-identification, action detection, and graph matching. If the
re-identification is incorrect, then regardless of whether action
detection is correct or not, matching will fail since those actions do not belong to the same tube. Thus, when
re-identification accuracy increases, correct action detection will
boost overall accuracy.} This is also why, in
\caesarfig{eval_all}(b), the overall precision is perfect even
when the action detector is less than perfect. Second, 70\% (105)
samples of complex activities in the current dataset are Spatial-only,
which relies more heavily on re-identification,  making the
effectiveness of that DNN more prominent.

From those two figures, for a complex activity
detector with $>$90\% in recall and precision, the object
detector/tracker must have $>$90\% accuracy and the action detector
should have $>$80\% accuracy. We observe that object detectors
(which have been the topic of active research longer than activity
detection and re-identification) have started to approach 90\%
accuracy in recent years.

Finally, as an aside, note that the precision projections are
non-monotonic (\caesarfig{eval_all}(b)): for a given accuracy of
the Re-ID, precision is non-monotonic with respect to accuracy of
atomic action detection. This results from the dependence observed
earlier: if Re-ID is wrong, then even if action detection is accurate,
Caesar will miss the complex activity.


\begin{table}[]
\centering
\begin{tabular}{|l|l|l|l|l|l|l|l|}
\hline
\textit{\textbf{Action ID}}      & 1    & 2    & 3    & 4    & 5   & 6   & 7    \\ \hline
\textit{\textbf{Recall (\%)}}    & 46.2 & 54.5 & 60   & 50   & 100 & 50  & 65.3 \\ \hline
\textit{\textbf{Precision (\%)}} & 43.8 & 41.7 & 42.8 & 36.4 & 100 & 100 & 63   \\ \hline
\end{tabular}
\caption{\emph{Caesar's recall and precision on the seven complex activities shown in Table~\ref{table:comp_act_list}.}}
\label{table:overall_accuracy}
\end{table}

\thispagestyle{empty}

\parab{Failure analysis.}
To get more insight into these macro-level observations, we examined
all the failure cases, including false positives and false negatives;
\caesarfig{eval_all}(c) shows the error breakdown for each activity (\#5
is not listed because it does not have an error case). 

The errors fall into three categories. First, the object detection DNN
is not perfect and can miss the boxes of some actors or objects such
as bags and bicycles, affecting tube construction and subsequent
re-identification. \added{This performance is  worse in videos
with rapid changes in the ratio of object size to image scale, large within-class 
variations of natural object classes, and background clutter~\cite{Tank2018}.}
\caesarfig{eval_all}(c) shows the
object detector missing a frontal view of a bicycle.

Re-ID either within a single camera, or across multiple cameras, is
error-prone. Within a camera, a person may be temporarily occluded.
Tracking occluded people is difficult especially in a crowded scene. Existing
tracking algorithms and Re-ID DNNs have not completely solved the
challenge, and result in many \textit{ID-switches} within a
single camera. Similarly, Re-ID can fail across cameras even with our
robustness enhancements which encode camera topology. \changed{This occurs
because of different lighting conditions, changes in pose between
different cameras, different percentage of occlusions, and different sets of detectable discriminative features~\cite{su2016deep,bak2012human}}. An incorrect re-identification results in Caesar
missing all subsequent atomic actions performed by a person.

Action detection is the third major cause of detection failures.
\changed{Blurry frames, occlusions, and an insufficient number of frames in a
  tube can all cause high error rates for the action DNN, resulting from incorrect
  labeling ~\cite{poppe2010survey}}. As described
earlier, errors in other modules can propagate to action detection:
object detection failures can result in shorter tubes for action
detection; the same is true of re-identification failures within a single camera.

Object detection failure is the least important factor, although it
affects \#6 because it requires detecting a
bicycle. For the graphs that require DNN-generated atomic actions, the
action detection error is more influential than tracking. In the
Spatial-Only cases, the tracking issue is the major cause of errors.

\begin{figure}[t]
  \begin{minipage}{0.32\linewidth}
    \centerline{\includegraphics[width=0.99\columnwidth]{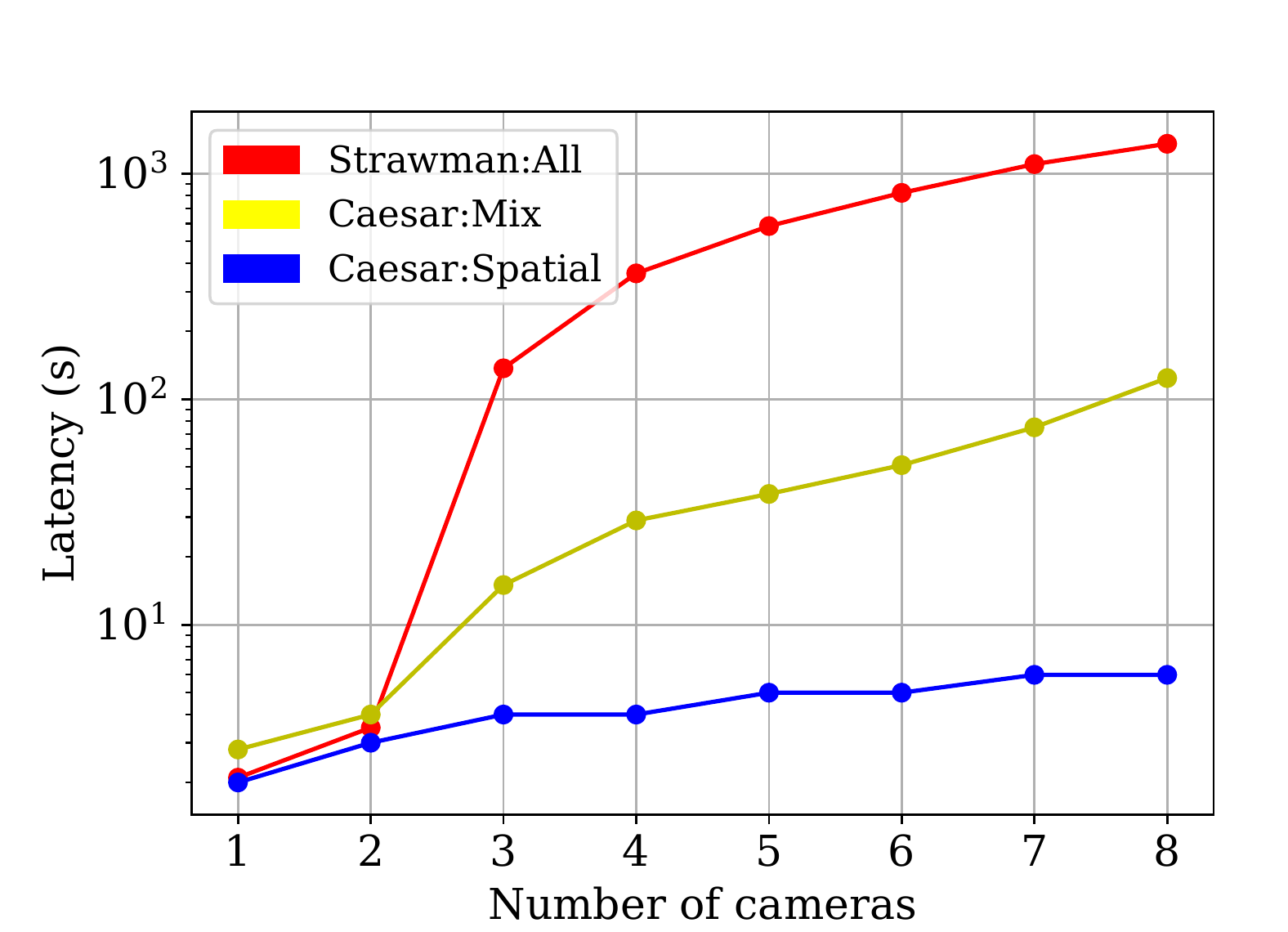}}
    \centerline{(a)}
    \label{fig:caesar_latency_camera}
  \end{minipage}
    \begin{minipage}{0.32\linewidth}
    \centerline{\includegraphics[width=0.99\columnwidth]{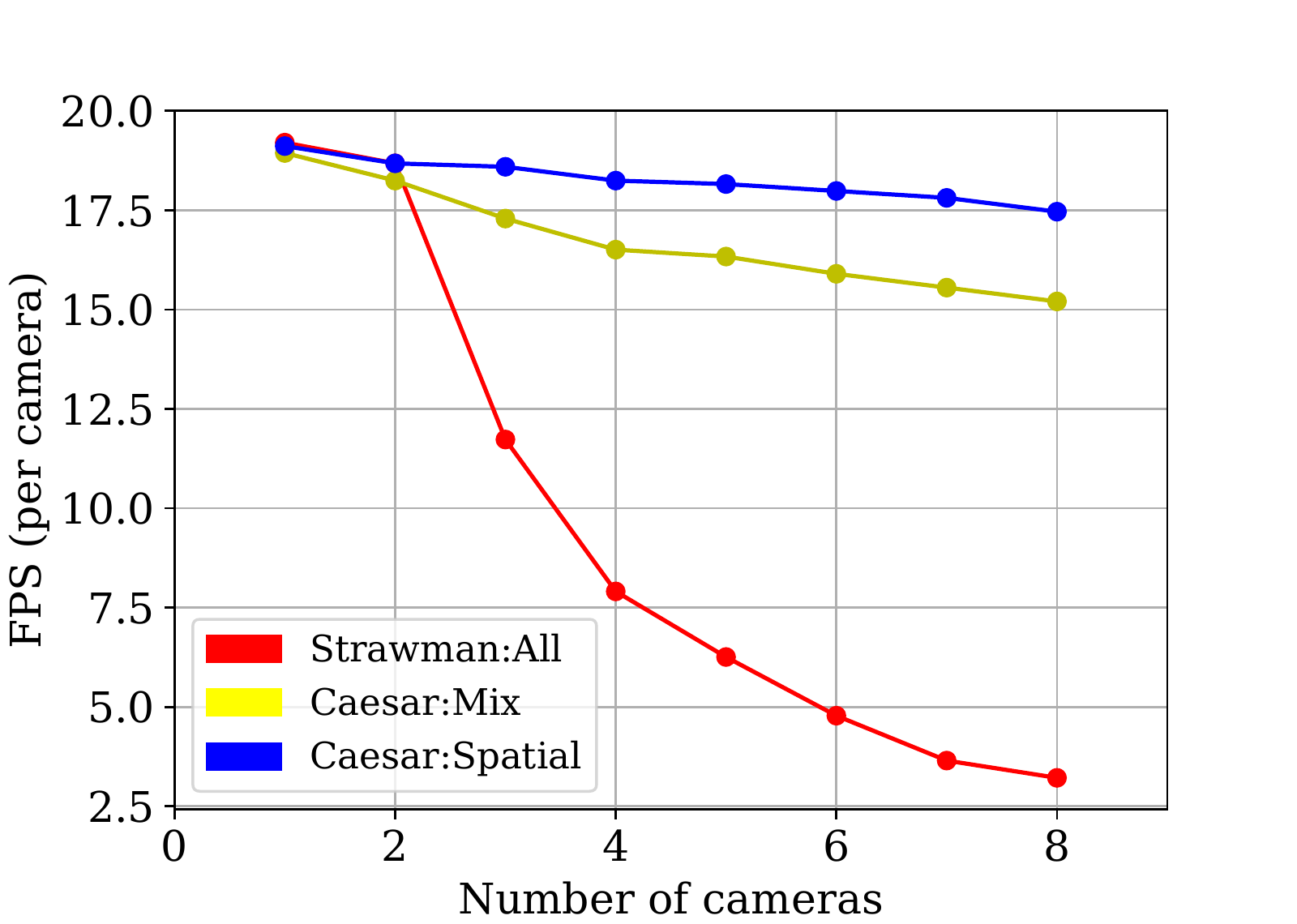}}
    \centerline{(b)}
    \label{fig:caesar_fps_camera}
  \end{minipage}
    \begin{minipage}{0.32\linewidth}
    \centerline{\includegraphics[width=0.99\columnwidth]{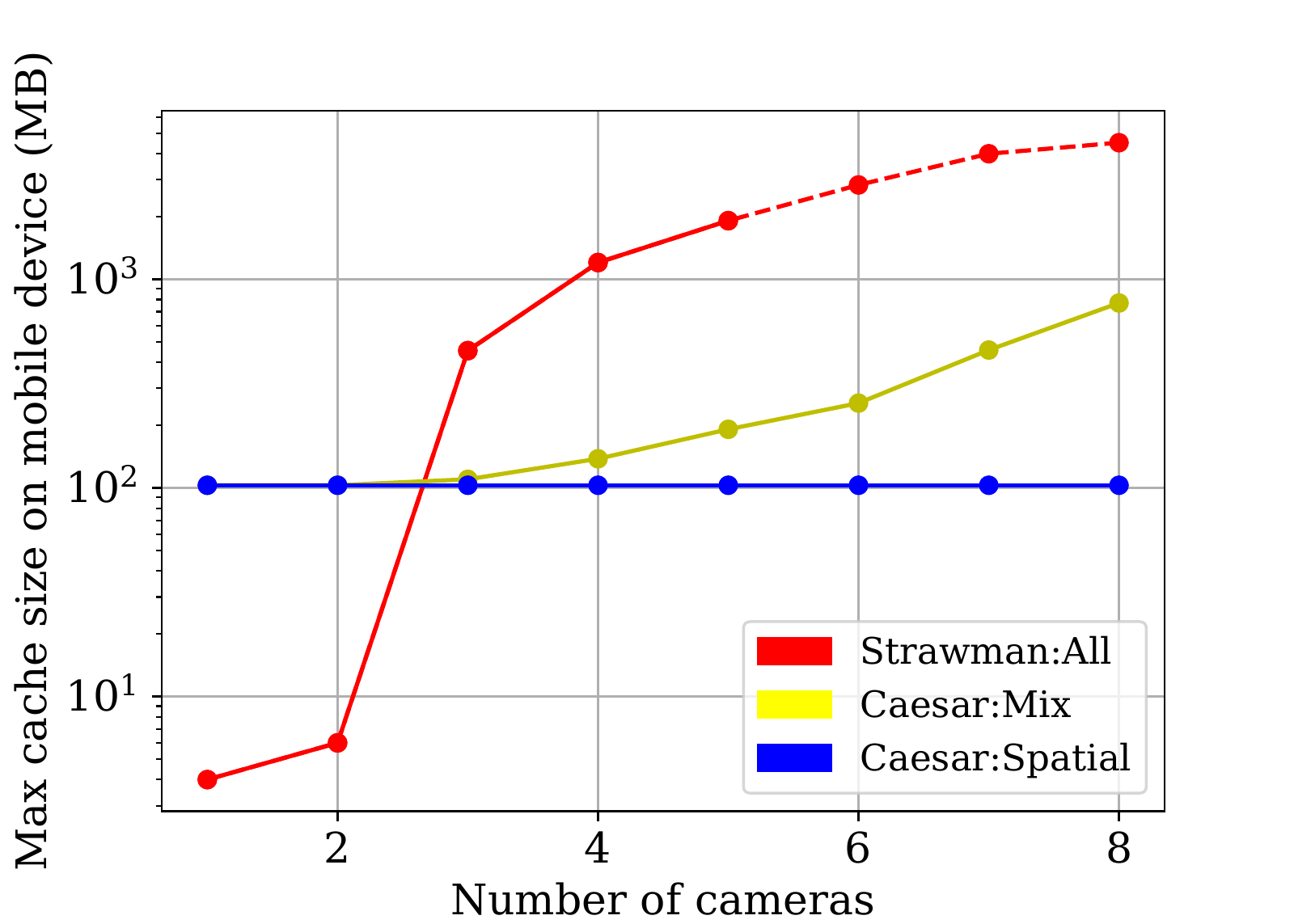}}
    \centerline{(c)}
    \label{fig:caesar_cache_camera}
  \end{minipage}
  \caption{\emph{(a) Latency of Caesar and the strawman solution with different number of inputs. (b) Throughput of Caesar and the strawman solution with different number of Inputs. (c) Maximum cache size needed for Caesar and the strawman solution to reach the best accuracy, with different number of Inputs.}}
  \label{fig:caesar_scalability}
\end{figure}

\parab{Caesar performance on more complex activities.} 
\added{Since the MTMC dataset has few complex activities, we
recorded another dataset to test Caesar on a variety of cross-camera complex activities Table~\ref{table:second_dataset}. We placed
three non-overlapping cameras in a plaza on our campus. Volunteers acted out activities not included in Table~\ref{table:comp_act_list}, such as
``shake hands'', ``eat'', and ``take a photo''; these invoke our action DNN. We observe (Table~\ref{table:second_dataset}) that
Caesar can capture these complex activities with precision $>$80\% (100\% for 5 of the 7 activities) and recall $\geq$75\% for 5 of the activities. All failures are caused by incorrect
re-ID due to varying lighting conditions. 
}

\begin{table}[]
\centering
\begin{tabular}{|l|l|l|}
\hline
\textbf{Complex Activity}                                                                                   & \textbf{\begin{tabular}[c]{@{}l@{}}\# of Samples\\ (Single/Multi)\end{tabular}} & \textbf{\begin{tabular}[c]{@{}l@{}}Recall(\%)\\ /Precision(\%)\end{tabular}} \\ \hline
Eat then drink                                                                                              & 5/0                                                                             & 80/80                                                                        \\ \hline
Shake hands then sit                                                                                          & 6/0                                                                             & 83.3/100                                                                       \\ \hline
Use laptop then take photo                                                                                  & 2/2                                                                             & 75/100                                                                       \\ \hline
Carry bag and sit then read                                                                                 & 2/4                                                                             & 66.7/80                                                                      \\ \hline
Use laptop, read then drink                                                                             & 0/4                                                                             & 75/100                                                                       \\ \hline
Read, walk then take photo                                                                              & 0/4                                                                             & 75/100                                                                       \\ \hline
\begin{tabular}[c]{@{}l@{}}Carry bag, sit, then eat, then \\ drink, then read, then take photo\end{tabular} & 0/4                                                                             & 50/100                                                                       \\ \hline
\end{tabular}
\caption{\emph{Activities in a three-camera dataset, and Caesar's accuracy.}}
\label{table:second_dataset}
\end{table}

\subsection{Scalability}

We measure the scalability of Caesar by the number of concurrent videos
it can support with fixed server-side hardware, assuming cameras have
on-board GPUs for running action detection.


\parab{Strawman approach for comparison.}
Caesar's lazy action detection is an important scaling optimization. To
demonstrate its effectiveness, we evaluate a strawman solution which
disables lazy action detection. The strawman retrieves images from the
camera for every bounding box, and runs action detection on
every tube.

Recall that lazy invocation of action detection does not help for
NN-only complex activities. Lazy invocation first matches spatial
clauses in the graph, then selectively invokes action detection. But,
NN-only activities do not contain spatial clauses, so Caesar invokes
action detection on all tubes, just like the strawman. However, for
Mixed or Spatial-only complex activities lazy invocation conserves the
use of GPU resources.

To highlight these differences, we perform this experiment by grouping
the complex activities into these three groups: Strawman (which is the
same as NN-only for the purposes of this experiment), Mixed, and
Spatial-Only. Thus, for example, in the Mixed group experiment, Caesar
attempts to detect all Mixed complex activities.

\parab{Latency.}
\caesarfig{scalability}(a) shows Caesar's \textit{detection latency} is
a function of the number of cameras for each of these three
alternatives. The detection latency is the time between when a complex
activity completes in a camera to when Caesar detects it.


The results demonstrate the impact of the scaling optimization in
Caesar. As the number of cameras increases, detection latency can
increase dramatically for the strawman, going up to nearly 1000
seconds with eight cameras. For Mixed complex activities, the latency
is an order of magnitude less at about 100 seconds; this illustrates
the importance of our optimization, without which Caesar's performance
would be similar to the Strawman, an order of magnitude worse. For
Spatial-only activities that do not involve action detection, the
detection latency is a few seconds; Caesar scales extremely well for
these types of complex activities.

Recall that in these experiments, we fix the number of GPU resources.
In practice, in an edge cluster, there is likely to be some elasticity
in the number of GPUs, so Caesar can target a certain detection latency
by dynamically scaling the number of GPUs assigned for action
detection. We have left this to future work.

Finally, we note that up to 2 cameras, all three approaches perform
the same; in this case, the bottleneck is the object detector on the
camera with a frame rate of 20~fps.

\parab{Frame rate.}
\caesarfig{scalability}(b) shows the average frame rate at which Caesar can
process these different workloads, as a function of the number of
cameras. This figure explains why Strawman's latency is high: its
frame rate drops precipitously down to just two frames per second with
eight concurrent cameras. Caesar scales much more gracefully for other
workloads: for both Mixed and Spatial-only workloads, it is able to
maintain over 15 frames per second even up to eight cameras.

These results highlight the importance of hybrid complex activity
descriptions. Detecting complex activities using neural networks can
be resource-intensive, so Caesar's ability to describe actions using
spatial and temporal relationships while using DNNs sparingly is the
key to enabling scalable complex activity detection system.


\parab{Cache size.}
Caesar maintains a cache of image contents at the camera. The
longer the detection latency, the larger the cache size. Thus,
another way to examine Caesar's scalability is to understand the cache
size required for different workloads with different number of
concurrent cameras. The camera's cache size limit is 4~GB.

\caesarfig{scalability}(c) plots the largest cache size observed during
an experiment for each workload, as a function of the number of
cameras. Especially for the Strawman, this cache size exceeded the
4~GB limit on the camera, so we re-did these experiments on a
desktop with more memory. The dotted-line segment of the Strawman
curve denotes these desktop experiments. \added{When Caesar exceeds memory on the device, frames can be saved on persistent storage on the mobile device, or be transmitted to the server for processing, at the expense of latency and bandwidth.}

Strawman has an order of magnitude higher cache size requirements
precisely because its latency is an order of magnitude higher than the
other schemes; Caesar needs to maintain images in the cache until
detection completes. In contrast, Caesar requires a modest and fixed
100~MB cache for Spatial-only workloads on the camera: this supports
lazy retrieval of frames or images for re-identification. The cache size for Mixed
workloads increases in proportion to the increasing detection latency
for these workloads.

\begin{figure}
\centering\includegraphics[width=0.55\columnwidth]{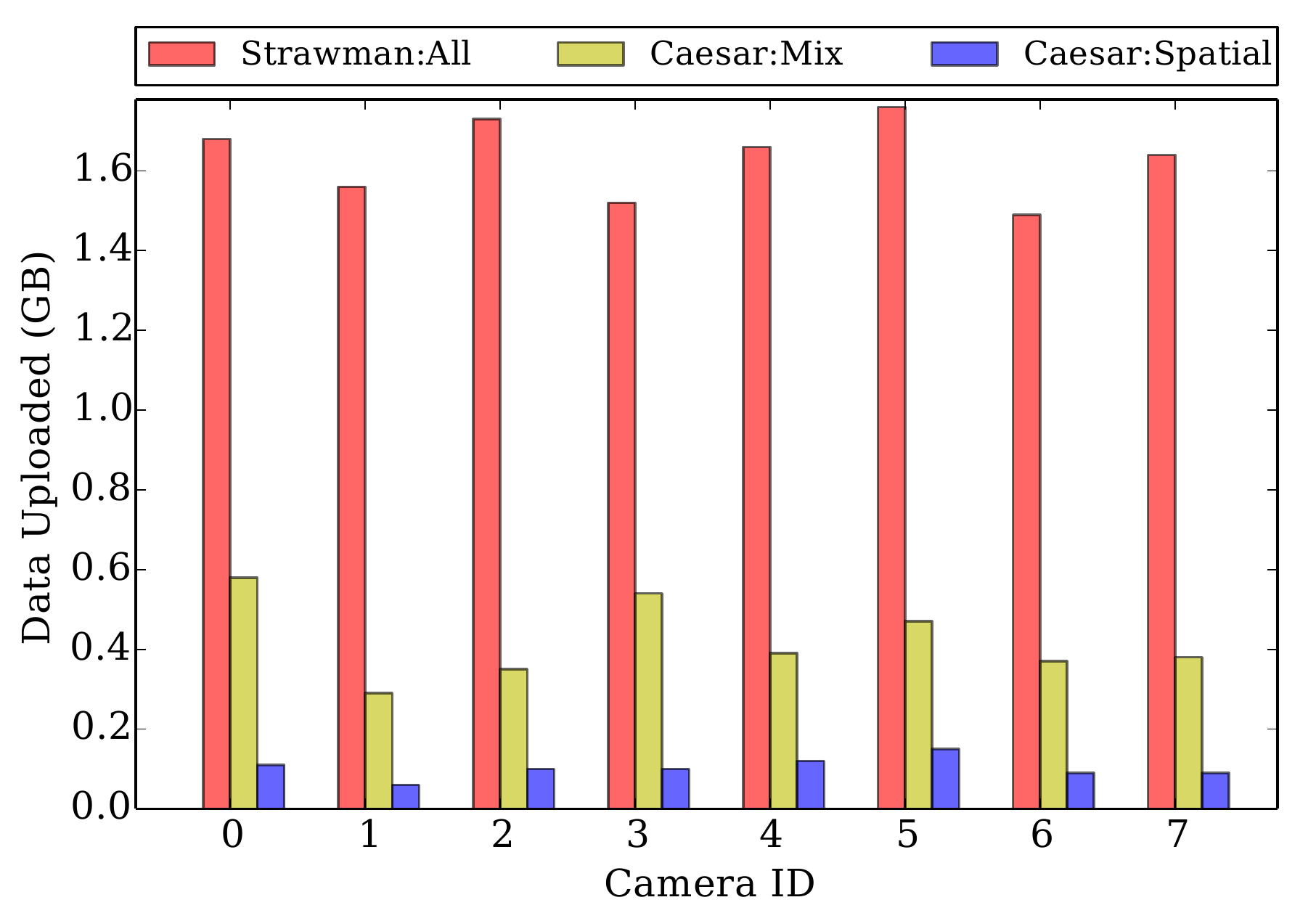}
\caption{\emph{The total amount of data to be uploaded from each camera, with different uploading schemes.}}
\label{fig:caesar_data_per_cam}
\end{figure}

\subsection{Data Transfer}

Caesar's lazy retrieval of images conserves wireless bandwidth, and to
quantify its benefits, we measure the total amount of data uploaded
from each camera (the edge cluster sends small control messages to
request image uploads; we ignore these). In \caesarfig{data_per_cam},
the strawman solution simply uploads the whole 30-min video with
metadata (more than 1.5~GB for each camera). Mixed transfers
$>$3$\times$ fewer data, and Spatial-only is an order of magnitude
more efficient than Strawman. Caesar's data transfer overhead can be
further optimized by transferring image deltas, leveraging the
redundancy in successive frames or images within successive bounding
boxes; we have left this to future work.

\begin{figure}
\centering\includegraphics[width=0.55\columnwidth]{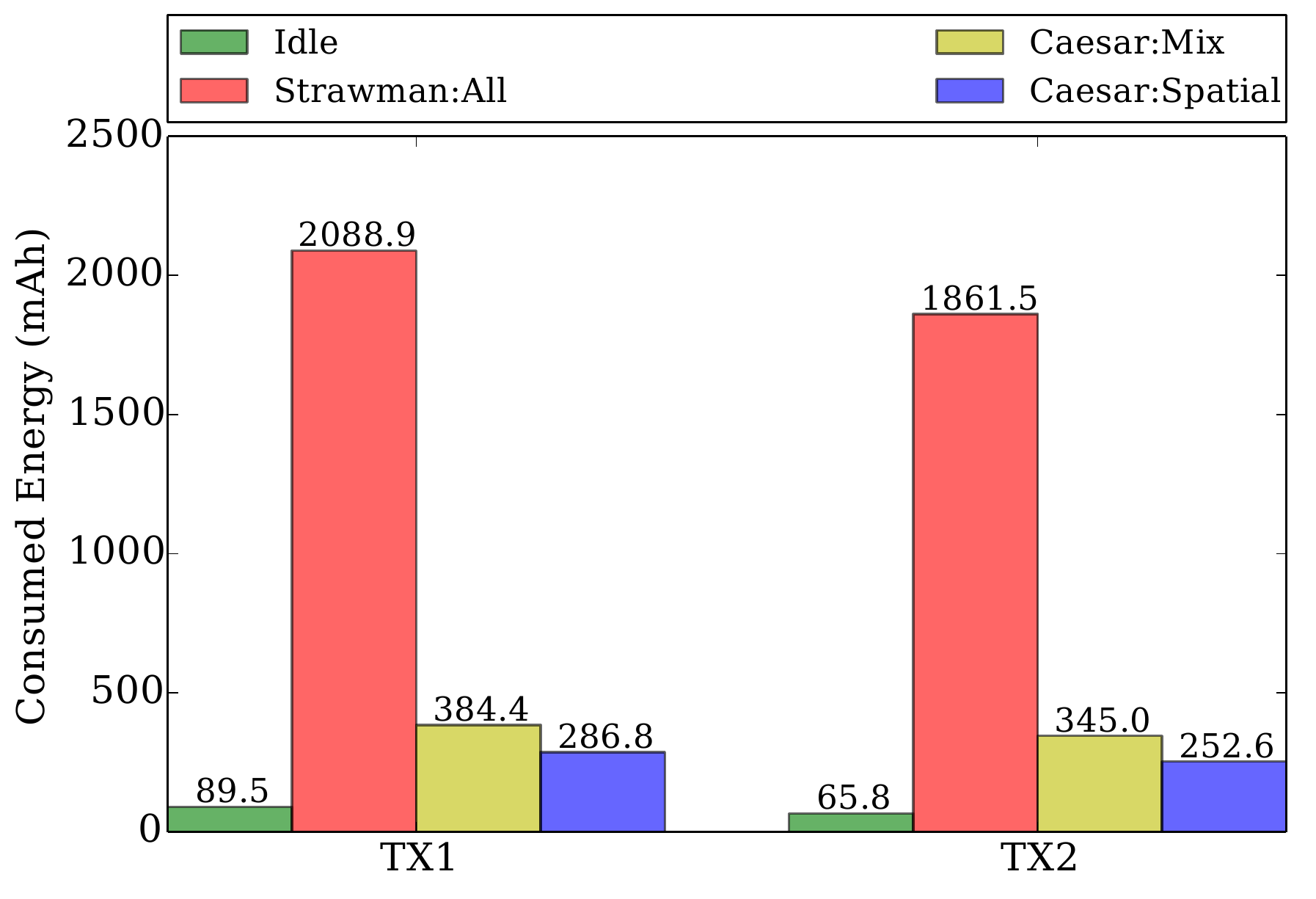}
\caption{\emph{The average energy consumption of cameras in Caesar, with different uploading scheme and action queries.}}
\label{fig:caesar_energy}
\end{figure}

\subsection{Energy Consumption}

Even for wireless surveillance cameras with power sources, it may be
important to quantify the energy required to run these workloads on
the camera. The TX2 GPUs' onboard power supply chipset
provides instantaneous power consumption readings at 20~Hz. We plot,
in \caesarfig{energy}, the total energy required for each workload by
integrating the power consumption readings across the duration of the
experiment. For context, we also plot the idle energy consumed by the
device when run for 30 mins.

Strawman consumes 1800~mAh for processing our dataset, comparable to
the battery capacity of modern smart phones. For Mixed and Spatial
workloads, energy consumption is, respectively, 6$\times$ to
10$\times$ lower, for two reasons: (a) these workloads upload fewer
images, reducing the energy consumed by the wireless network
interface; (b) Strawman needs to keep the device on for longer to serve
retrieval requests because its detection latency is high.

%% file: tex/paper_grab.tex
\chapter{Grab: A Cashier-Free Shopping System}\label{chap:grab}

\input{tex/grab/intro}
\input{tex/grab/design}
\input{tex/grab/eval}

%% file: tex/grab/intro.tex
\section{Introduction}

While electronic commerce continues to make great strides, in-store purchases are likely to continue to be important in the coming years: 91$\%$ of purchases are still made in physical stores \cite{forrester, phy_shopping} and 82$\%$ of millennials prefer to shop in these stores \cite{millennial}. However, a significant pain point for in-store shopping is the checkout queue: customer satisfaction drops significantly when queuing delays exceed more than four minutes \cite{checkout_time}. To address this, retailers have deployed self-checkout systems (which can increase instances of shoplifting \cite{selfcheckout_theft1, selfcheckout_theft2}), and expensive vending machines.

The most recent innovation is \textit{cashier-free shopping}, in which a networked sensing system automatically (a) \textit{identifies} a customer who enters the store, (b) \textit{tracks} the customer through the store, (c) and \textit{recognizes} what they purchase. Customers are then billed automatically for their purchases, and do not need to interact with a human cashier or a vending machine, or scan items by themselves. Over the past year, several large online retailers like Amazon and Alibaba~\cite{amazon, taobao} have piloted a few stores with this technology, and cashier-free stores are expected to take off in the coming years~\cite{cashierfreetrend,retaileradopt}. Besides addressing queue wait times, cashier-free shopping is expected to reduce instances of theft, and provide retailers with rich behavioral analytics.

\sepfootnotecontent{grab}{A shopper only needs to \textit{grab} items and go.}

Not much is publicly known about the technology behind cashier-free shopping, other than that stores need to be completely redesigned~\cite{amazon, taobao} which can require significant capital investment. In this chapter, we ask: Is cashier-free shopping viable without having to completely redesign stores? To this end, we observe that many stores already have, or will soon have, the hardware necessary to design a cashier-free shopping system: cameras deployed for in-store security, sensor-rich smart shelves~\cite{smart_shelves} that are being deployed by large retailers~\cite{smallbig} to simplify asset tracking, and RFID tags being deployed on expensive items to reduce theft. This chapter explores the design and implementation of a practical cashier-free shopping system called Grab\sepfootnote{grab} using this infrastructure, and quantifies its performance.

Grab needs to accurately identify and track customers, and associate each shopper with items he or she retrieves from shelves. It must be robust to visual occlusions resulting from multiple concurrent shoppers, and to concurrent item retrieval from shelves where different types of items might look similar, or weigh the same. It must also be robust to fraud, specifically to attempts by shoppers to confound identification, tracking, or association. Finally, it must be cost-effective and have good performance to achieve acceptable accuracy: specifically, we show that, for vision-based tasks, slower than 10 frames/sec processing can reduce accuracy significantly. 

\parab{Contributions}
An obvious way to architect Grab is to use deep neural networks (DNNs) for each individual task in cashier-free shopping, such as identification, pose tracking, gesture tracking, and action recognition. However, these DNNs are still relatively slow and many of them cannot process frames at faster than 5-8 fps. Moreover, even if they have high individual accuracy, their effective accuracy would be much lower if they were cascaded together.

Grab's architecture is based on the observation that, for cashier-free shopping, we can use a single vision capability (pose detection) as a building block to perform \textit{all} of these tasks. A recently developed DNN library, OpenPose~\cite{openpose} accurately estimates body "skeletons" in videos at high frame rates. 

Grab's first contribution is to develop a suite of lightweight identification and tracking algorithms built around these skeletons. Grab uses the skeletons to accurately determine the bounding boxes of faces to enable feature-based face detection. It uses skeletal matching, augmented with color matching, to accurately track shoppers even when their faces might not be visible, or even when the entire body might not be visible. It augments OpenPose's elbow-wrist association algorithm to improve the accuracy of tracking hand movements which are essential to determining when a shopper may pickup up items from a shelf.

Grab's second contribution is to develop fast sensor fusion algorithms to associate a shopper's hand with the item that he picks up. For this, Grab uses a  probabilistic assignment framework: from cameras, weight sensors and RFID receivers, it determines the likelihood that a given shopper picked up a given item. When multiple concurrent actions occur, it uses an optimization framework to associate hands with items.

Grab's third contribution is to improve the cost-effectiveness of the overall system by multiplexing multiple cameras on a single GPU. It achieves this by avoiding running OpenPose on every frame, and instead using a lightweight feature tracker to track the joints of the skeleton between successive frames.

Using data from a pilot deployment in a retail store, we show that Grab has  93\% precision and 91\% recall even when nearly 40\% of shopper actions were adversarial. Grab needs to process video data at 10 fps or faster, below which accuracy drops significantly: a DNN-only design cannot achieve this capability. Grab needs all three sensing modalities, and all of its optimizations: removing an optimization, or a sensor, can drop precision and recall by 10\% or more. Finally, Grab's design enables it to multiplex up to 4 cameras per GPU with negligible loss of precision.

%% file: tex/grab/design.tex
\section{Grab Design}

Grab addresses these challenges by building upon a vision-based keypoint-based pose tracker DNN for identification and tracking, together with a probabilistic sensor fusion algorithm for recognizing item pickup actions. These ensure a completely non-intrusive design where shoppers are not required to scan item codes or pass through checkout gates while shopping. Grab consists of four major components (\grabfig{sys_design}).

\begin{figure}
\centering\includegraphics[width=0.6\columnwidth]{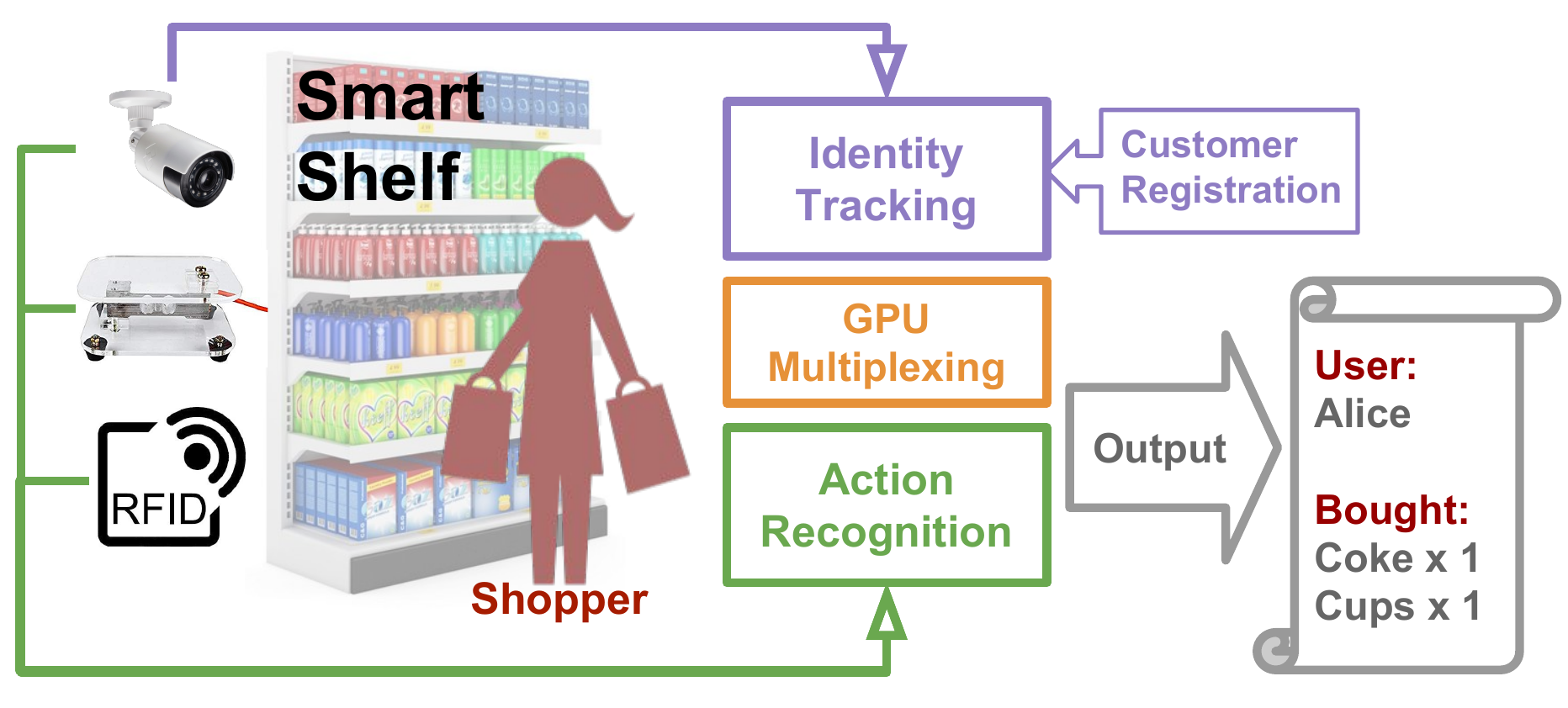}
\caption{\emph{Grab is a system for cashier-free shopping and has four components: registration, identity tracking, action recognition, and GPU multiplexing.}}
\label{fig:grab_sys_design}
\end{figure}

\textit{Identity tracking} recognizes shoppers' identities and tracks their movements within the store. \textit{Action recognition} uses a probabilistic algorithm to fuse vision, weight and RFID inputs to determine item pickup or dropoff actions by a customer.
\textit{GPU multiplexing} enables processing multiple cameras on a single GPU. Grab also has a fourth, offline component, \textit{registration}. Customers must register \textit{once} online before their first store visit. Registration involves taking a video of the customer to enable matching the customer subsequently.

\input{tex/grab/identity}
\input{tex/grab/behavior}

\input{tex/grab/scalability}

%% file: tex/grab/identity.tex
\subsection{Identity tracking}

Identity tracking consists of two related sub-components. Shopper identification determines \textit{who} the shopper is among registered users.  Shopper tracking, determines (a) \textit{where} the shopper is in the store at each instant of time, and (b) \textit{what} the shopper is doing at each instant.

\parab{Requirements and Challenges}
In designing Grab, we require first that customer registration be fast, even though it is performed only once: ideally, a customer should be able to register and immediately commence shopping. Identity tracking requires not just identifying the customer, but also detecting each person's \textit{pose}, such as hand position and head position. For each of these tasks, the computer vision community has developed DNNs. However, in our setting, computing efficiency is important for cost reasons: dedicating a GPU for each task can be prohibitively expensive. Running all of these DNNs on a single GPU can compromise accuracy.

\parab{Approach}
In this chapter, we make the following observation: we can build end-to-end identity tracking using a state-of-the-art fast (15~fps) pose tracker. Specifically, we use, as a building block, a \textit{keypoint} based body pose tracker, called OpenPose \cite{openpose}. Given an image frame, OpenPose detects keypoints for each human in the image. Keypoints identify distinct anatomical structures in the body (\grabfig{pose_combined}(a)) such as eyes, ears, nose, elbows, wrists, knees, hips \etc We can use these \textit{skeletons} for identification, tracking \textit{and} gesture recognition.

However, fundamentally, since OpenPose operates only on a single frame, Grab needs to add identification, tracking \textit{and} gesture recognition algorithms on top of OpenPose to continuously identify and tracks shoppers and their gestures. The rest of this section describes these algorithms.

\begin{figure}
\centering\includegraphics[width=0.5\columnwidth]{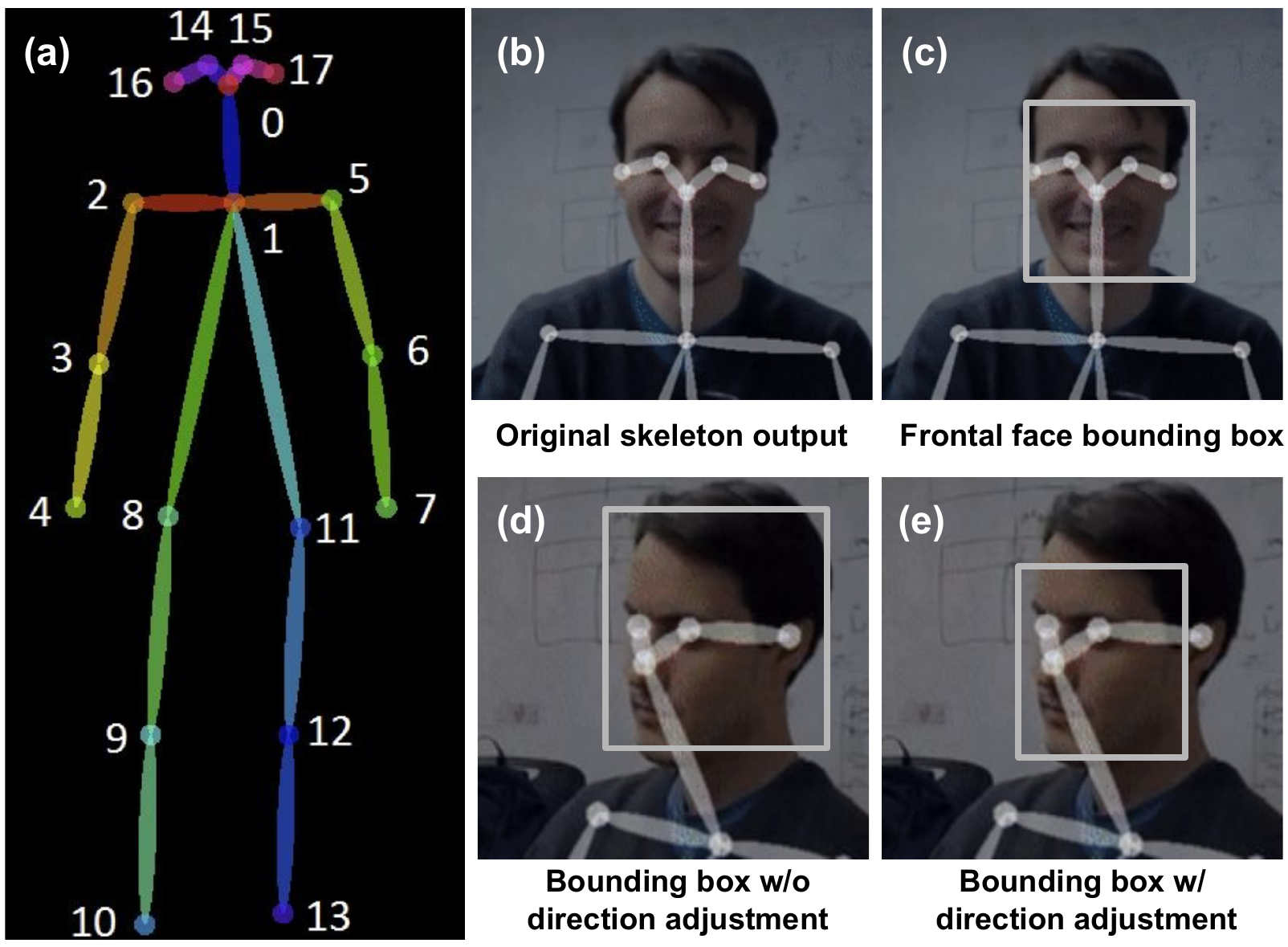}
\caption{\emph{(a) Sample OpenPose output. (b,c,d,e) Grab's approach adjusts the face's bounding box using the keypoints detected by OpenPose. (The face shown is selected from OpenPose project webpage~\cite{openpose})}}
\label{fig:grab_pose_combined}
\end{figure}

\parab{Shopper Identification}
Grab uses fast feature-based face recognition to identify shoppers. While prior work has explored other approaches to identification such as body features~\cite{bai2017scalable, chen2017beyond} or clothing color \cite{lu2001color}, we use faces because (a) face recognition has been well-studied by vision researchers and we are likely to see continued improvements, (b) faces are more robust for identification than clothing color, and (c) face features have the highest accuracy in large datasets.

\parae{Feature-based face recognition}
When a user registers, Grab takes a video of their face, extracts features, and builds a fast classifier using these features. To identify shoppers, Grab does not directly use a face detector on the entire image because non-DNN detectors~\cite{lienhart2002extended} can be inaccurate, and DNN-based face detectors such as MTCNN~\cite{zhang2016joint} can be slow. Instead, Grab \textit{identifies a face's bounding box using keypoints} from OpenPose, specifically, the five keypoints of the face from the nose, eyes, and ears (\grabfig{pose_combined}(b)). Then, it extracts features from within the bounding box and applies the trained classifier.

\parae{Fast Classification}
Registration is performed once for each customer, in which Grab extracts features from the customer's face. To do this, we evaluated several face feature extractors~\cite{baltruvsaitis2016openface, facefeature}, and finally selected ResNet-34's feature extractor~\cite{facefeature} which produces a 128-dimension feature vector, performs best in both speed and accuracy.

With these features, we can identify faces by comparing feature distances, build classifiers, or train a neural network. After experimenting with these options, we found that a $k$ nearest neighbor (kNN) classifier, in which each customer is trained as a new class, worked best among these choices. Grab builds one kNN-based classifier for all customers and uses it across all cameras.

\parae{Tightening the face bounding box}
During normal operation, Grab extracts facial features within a bounding box (derived from OpenPose keypoints) around each customer's face. Grab infers the face's bounding box width using the distance between two ears, and the height using the distance from nose to neck. This works well when the face points towards the camera (\grabfig{pose_combined}(c)), but can have inaccurate bounding box when customers face slightly away from the camera (\grabfig{pose_combined}(d)). This inaccuracy can degrade classification performance.

To obtain a tighter bounding box, we estimate head \textit{pitch} and \textit{yaw} using the keypoints. Consider the line between the nose and neck keypoints: the distance of each eye and ear keypoint to this axis can be used to estimate head \textit{yaw}. Similarly, the distance of the nose and neck keypoints to the axis between the ears can be used to estimate \textit{pitch}. Using these, we can tighten the bounding box significantly (\grabfig{pose_combined}(e)). To improve detection accuracy when a customer's face is not fully visible in the camera, we also use face alignment~\cite{baltruvsaitis2016openface}, which estimates the frontal view of the face.

\parab{Shopper Tracking}
A user's face may not always be visible in every frame, since customers may intentionally or otherwise turn their back to the camera. However, Grab needs to be able to identify the customer in frames where the customer's face is not visible, for which it uses \textit{tracking}.


\parae{Skeleton-based Tracking}
Existing human trackers use bounding box based approaches~\cite{tang2017multiple, wojke2017simpl}, which can perform poorly in in-store settings with partial or complete occlusions. 

Instead, we use the skeleton generated by OpenPose to develop a tracker that uses geometric properties of the body frame. We use the term \textit{track} to denote the movements of a distinct customer (whose face may or may not have been identified). Suppose OpenPose identifies a skeleton in a frame: the goal of the tracker is to associate the skeleton with an existing track if possible. Grab uses the following to track customers. It tries to align each keypoint in the skeleton with the corresponding keypoint in the last seen skeleton in each track, and selects that track whose skeleton is the closest match (the sum of match errors is smallest). Also, as soon as it is able to identify the face, Grab associates the customer's identity with the track (to be robust to noise, Grab requires that the customer's face is identified in 3 successive frames).

\parae{Dealing with Occlusions}
In some cases, a shopper may be obscured by another. Grab uses lazy tracking in this case (\grabfig{lazy_tracking}). When an existing track disappears in the current frame, Grab checks if the track was close to the edge of the image, in which case it assumes the customer has moved out of the camera's field of view and deletes the track. Otherwise, it marks the track as \textit{blocked}. When the customer reappears in a subsequent frame, it reactivates the blocked track.

\parab{Shopper Gesture Tracking}
Grab must recognize the arms of each shopper in order to determine which item he or she purchases. OpenPose has a built-in \textit{limb association} algorithm, which associates shoulder joints to elbows, and elbows to wrists. We have found that this algorithm is a little brittle in our setting: it can miss an association (\grabfig{openposefail}(a)), or mis-associate part of a limb of one shopper with another (\grabfig{openposefail}(b)). 

\begin{figure}
\centering\includegraphics[width=0.52\columnwidth]{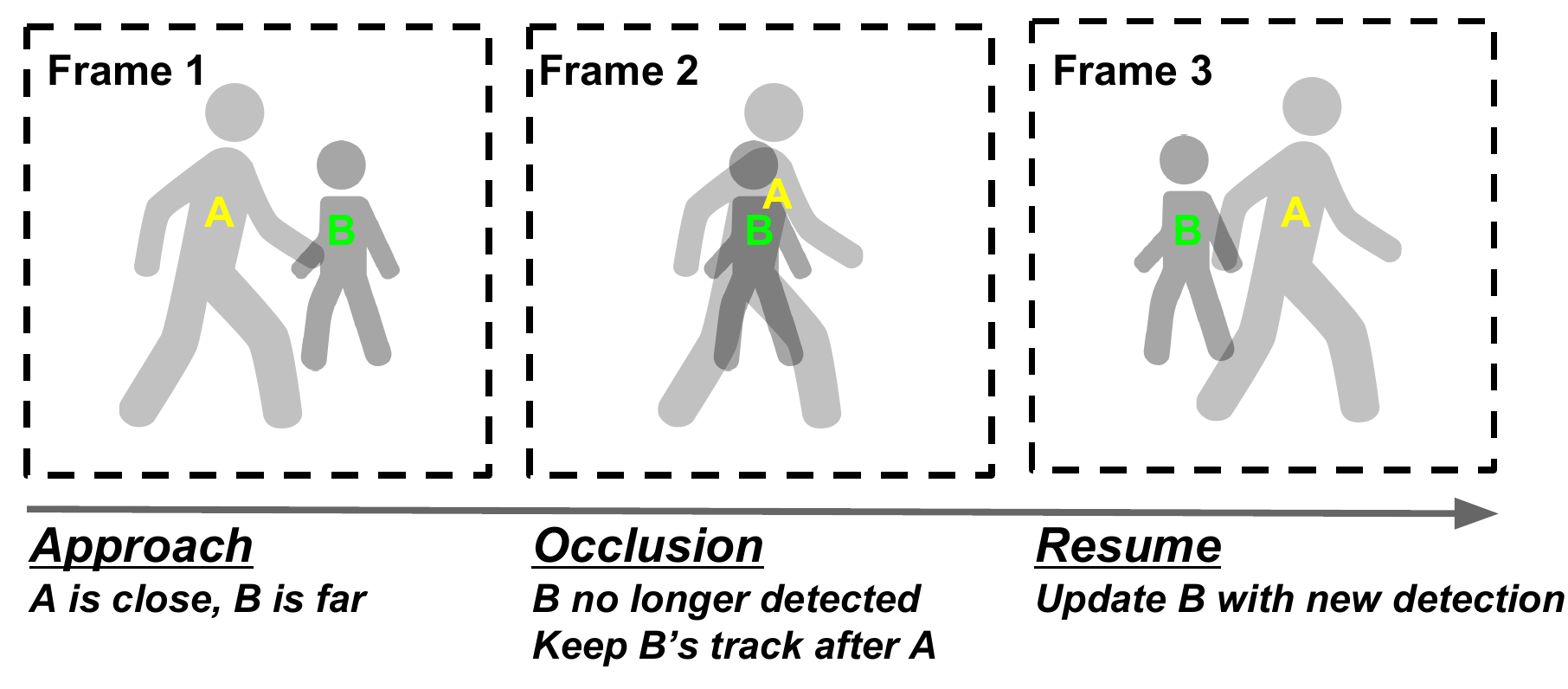}
\caption{\emph{When a shopper is occluded by another, Grab resumes tracking after the shopper re-appears (lazy tracking).}}
\label{fig:grab_lazy_tracking}
\end{figure}

\parae{How limb association in OpenPose works}
OpenPose first uses a DNN to associate with each pixel confidence value of it being part of an anatomical key point (\eg an elbow, or a wrist). During image analysis, OpenPose also generates vector fields (called \textit{part affinity fields}~\cite{cao2017realtime}) for upper-arms and forearms whose vectors are aligned in the direction of the arm. Having generated keypoints, OpenPose then estimates, for each pair of keypoints, a measure of alignment between an arm's part affinity field, and the line between the keypoints (\eg elbow and wrist). It then uses a bipartite matching algorithm to associate the keypoints.

\parae{Improving limb association robustness}
One challenge for OpenPose's limb association is that the pixels for the wrist keypoint are conflated with pixels in the hand (\grabfig{openposefail}(a)). This likely reduces the part affinity alignment, causing limb association to fail. To address this, for each keypoint, we filtered outlier pixels by removing pixels whose distance from the mediod~\cite{park2009simple} was greater than the 85th percentile.

\begin{figure}
\centering
\includegraphics[width=0.6\columnwidth]{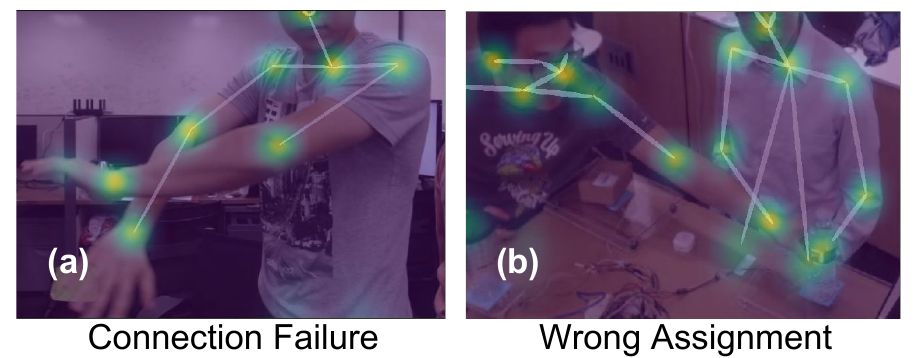}
\caption{\emph{OpenPose can (a) miss an assignment between elbow and wrist, or (b) wrongly assign one person's joint to another.}}
\label{fig:grab_openposefail}
\end{figure}

The second source of brittleness is that OpenPose's limb association treats each limb independently, resulting in cases where the key point from one person's elbow may get associated with another person's wrist (\grabfig{openposefail}(b)). To avoid this failure mode, we modify OpenPose's limb association algorithm to treat one person's forearms or upper-arms as a pair. 
To identify forearms (or upper-arms) as belonging to the same person, we measure the Euclidean distance $ED(.)$ between color histograms $F(.)$ belonging to the two forearms, and treat them as a pair if the distance is less than an empirically-determined  threshold $thresh$. We can formulate this as an optimization problem:

\vspace{-2ex}
{\small
\begin{maxi*}{i,j}{\sum_{i \in E}\sum_{j \in W}{A_{i,j}z_{i,j}}}{}{}
   \addConstraint{\sum_{j \in W}{z_{i,j}}}{\leq 1\quad}{\forall i \in E}
   \addConstraint{\sum_{i \in E}{z_{i,j}}}{\leq 1\quad}{\forall j \in W}
   \addConstraint{ED(F(i,j),F(i',j'))}{< thresh\quad}{\forall j, j' \in W \ i, i' \in E}
\end{maxi*}
}
\vspace{-2ex}

where $E$ and $W$ are the sets of elbow and wrist joints, and $A_{i,j}$ is the alignment measure between the $i$-th elbow and the $j$-th wrist, while $z_{i,j}$ is an indicator variable indicating connectivity between the elbow and the wrist. The third constraint models whether two elbows belong to the same body, using the Euclidean distance between the color histograms of the body color. This formulation reduces to a max-weight bipartite matching problem, and we solve it with the Hungarian algorithm~\cite{kuhn1955hungarian}. 

%% file: tex/grab/behavior.tex
\begin{figure}[htbp]
\centering
\includegraphics[width=0.5\columnwidth]{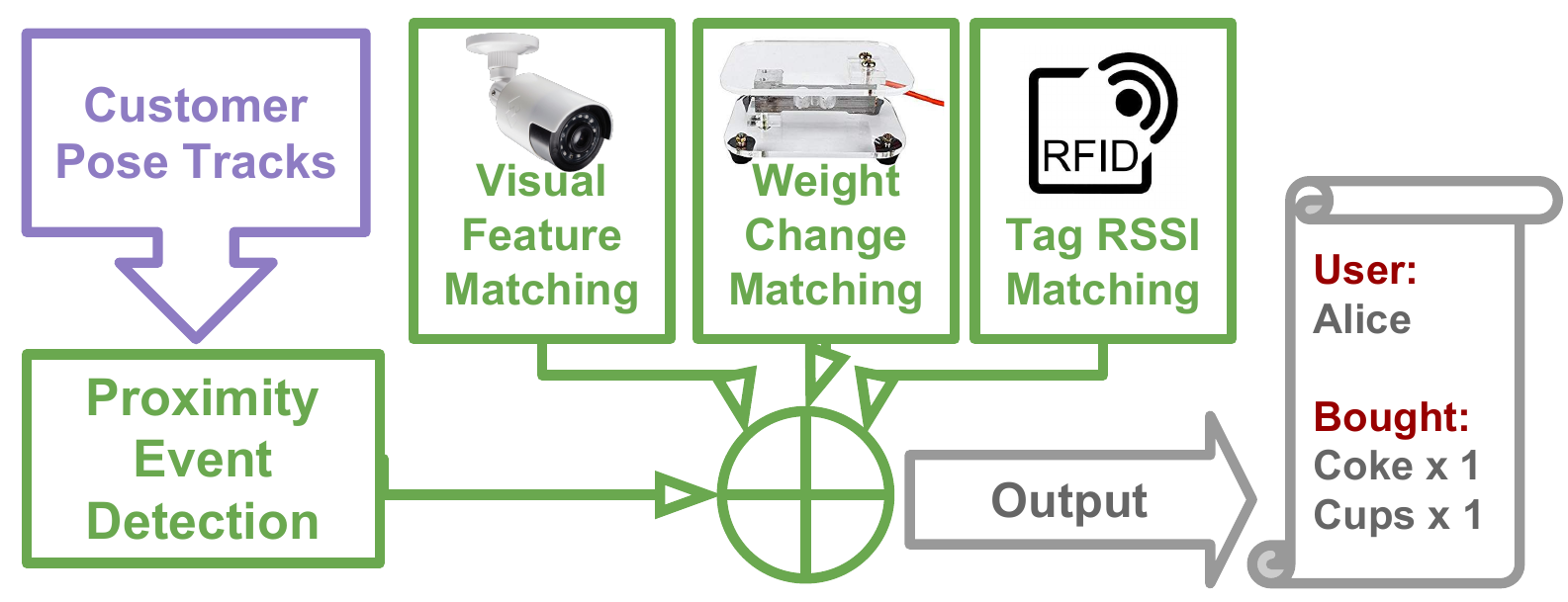}
\caption{\textit{Grab recognizes the items a shopper picks up by fusing vision with smart-shelf sensors including weight and RFID.}}
\label{fig:grab_behavior_workflow}
\end{figure}

\subsection{Shopper Action Recognition}

When a shopper is being continuously tracked, and their hand movements accurately detected, the next step is to recognize hand actions, specifically to identify item(s) which the shopper picks up from a shelf. Vision-based hand tracking alone is insufficient for this in the presence of multiple shoppers concurrently accessing items under variable lighting conditions. Grab leverages the fact that many retailers are installing smart shelves~\cite{smart_shelves,smart_shelves2} to deter theft, and commodity sensors are accurate enough to capture pick-up actions~\cite{lee2017secure}. These shelves have weight sensors and are equipped with RFID readers. Weight sensors cannot distinguish between items of similar weight, while not all items are likely to have RFID tags for cost reasons. So, rather than relying on any individual sensor, Grab fuses detections from cameras, weight sensors, and RFID tags to recognize hand actions.

\parab{Modeling the sensor fusion problem}
In a given camera view, at any instant, multiple shoppers might be reaching out to pick items from shelves. Our identity tracker tracks hand movement, the goal of the action recognition problem is to associate each shopper's hand with the item he or she picked up from the shelf. We model this association between shopper's hand $k$ and item $m$ as a probability $p_{k,m}$ derived from fusing cameras, weight sensors, and RFID tags (\grabfig{behavior_workflow}). $p_{k,m}$ is itself derived from \textit{association probabilities} for each of the devices, in a manner described below. Given these probabilities, we then solve the association problem using a maximum weight bipartite matching. In the following paragraphs, we discuss details of each of these steps.

\parab{Proximity event detection}
Before determining association probabilities, we need to determine when a shopper's hand approaches a shelf. This proximity event is determined using the identity tracker module's gesture tracking. Knowing where the hand is, Grab uses image analysis to determine when a hand is close to a shelf. For this, Grab requires an initial configuration step, where store administrators specify camera view parameters (mounting height, field of view, resolution \etc), and which shelf/shelves are where in the camera view. Grab uses a threshold pixel distance from hand to the shelf to define proximity, and its identity tracker reports \textit{start} and \textit{finish} times for when each hand is within the proximity of a given shelf (a \textit{proximity event}). (When the hand is obscured, Grab estimates proximity using the position of other skeletal keypoints, like the ankle joint).

\parab{Association probabilities from the camera}
When a proximity event starts, Grab starts tracking the hand and any item in the hand. It uses the color histogram of the item to classify the item. To ensure robust classification, Grab performs (\grabfig{behavior_figs}(a)) (a) background subtraction to remove other items that may be visible and (b) eliminates the hand itself from the item by filtering out pixels whose color matches typical skin colors. Grab extracts a 384 dimension color histogram from the remaining pixels.

During an initial configuration step, Grab requires store administrators to specify which objects are on which shelves. Grab then builds, for each shelf (a single shelf might contain 10-15 different types of items), builds a feature-based kNN classifier (chosen both for speed and accuracy). Then, during actual operation, when an item is detected, Grab runs this classifier on its features. The classifier outputs an ordered list of matching items, with associated match probabilities. Grab uses these as the association probabilities from the camera. Thus, for each hand $i$ and each item $j$, Grab outputs the camera-based association probability.

\begin{figure*}[htbp]
   \begin{minipage}{0.32\linewidth}
    \centerline{\includegraphics[width=0.95\columnwidth]{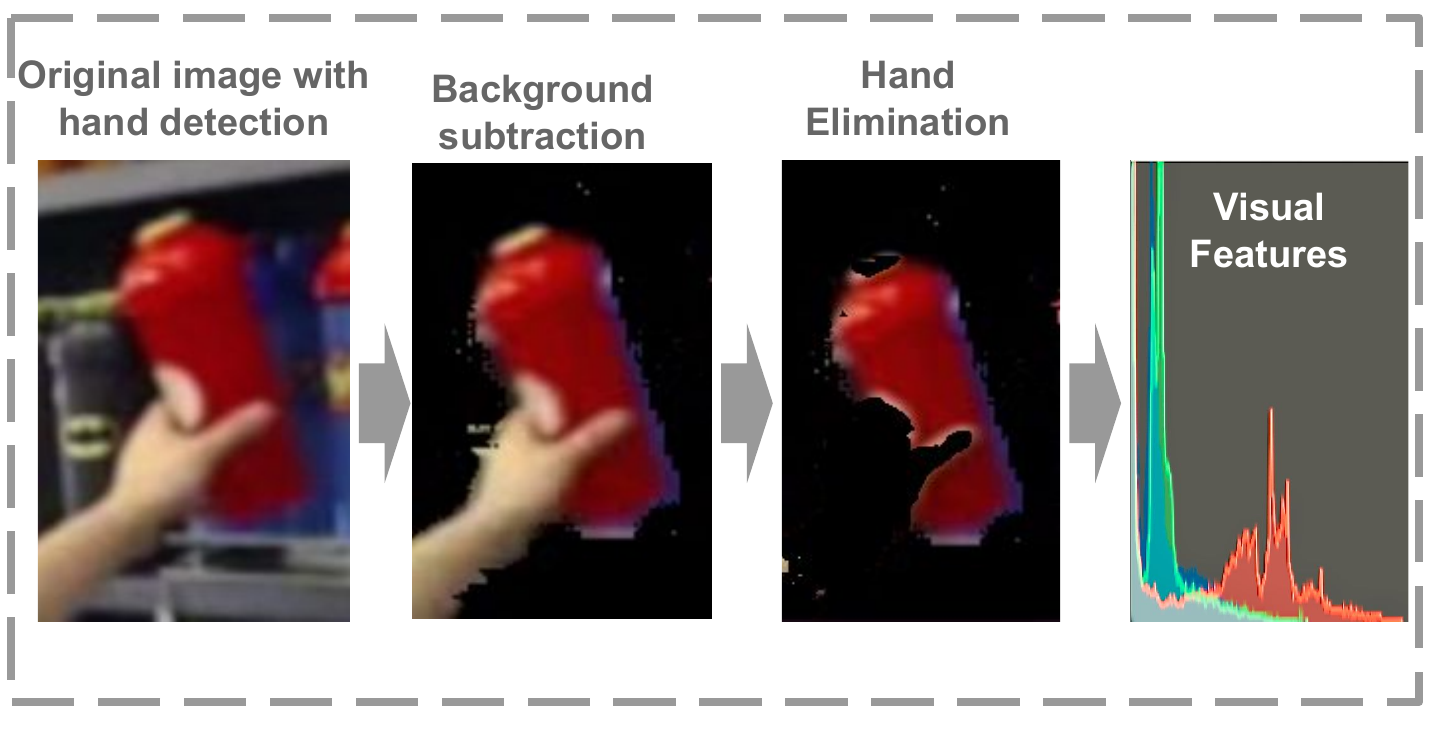}}
    \centerline{(a)}
    \label{fig:grab_vision_matching}
  \end{minipage}
  \begin{minipage}{0.32\linewidth}
    \centerline{\includegraphics[width=0.85\columnwidth]{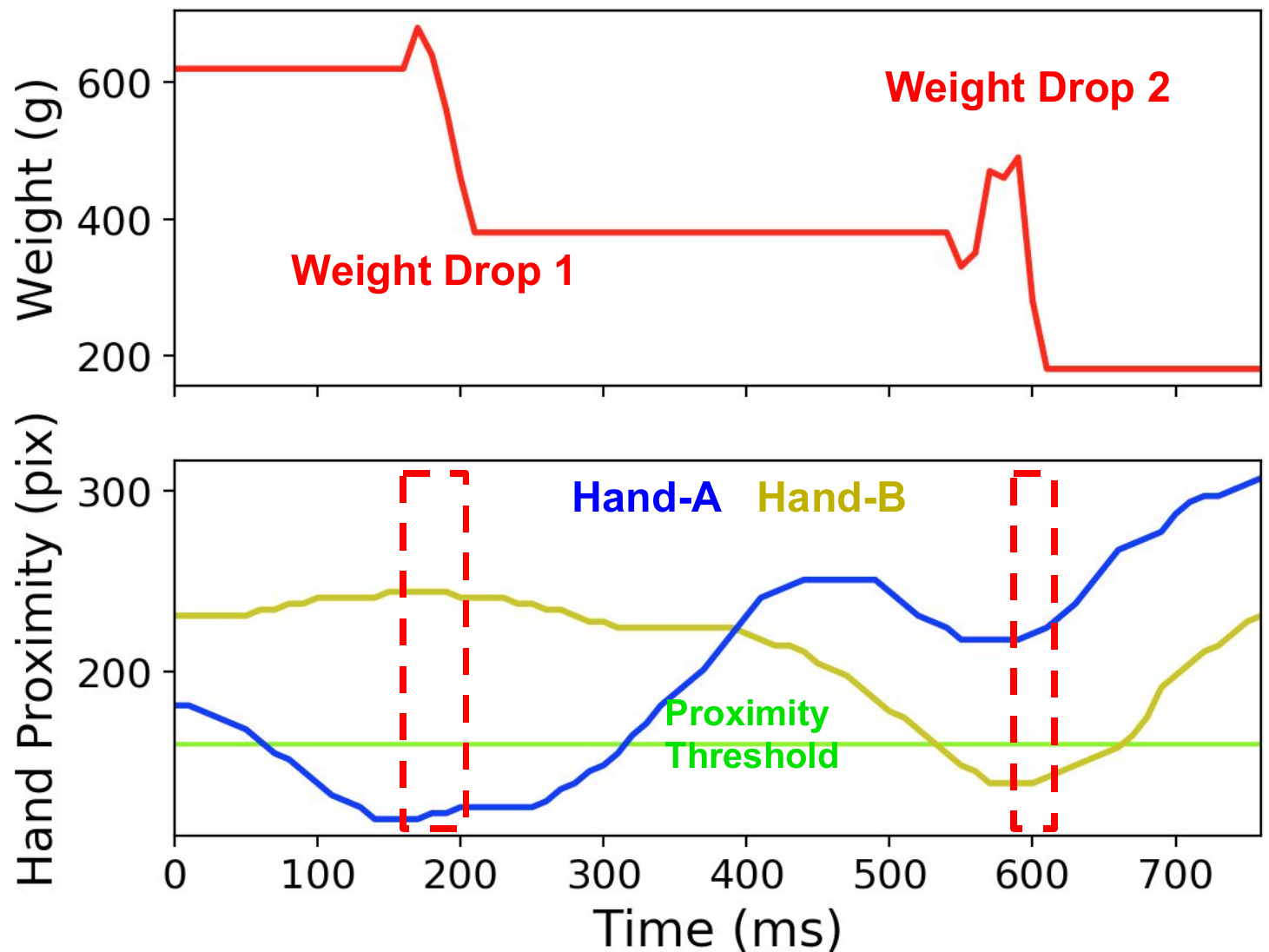}}
    \centerline{(b)}
    \label{fig:grab_weight_matching}
  \end{minipage}
    \begin{minipage}{0.32\linewidth}
    \centerline{\includegraphics[width=0.85\columnwidth]{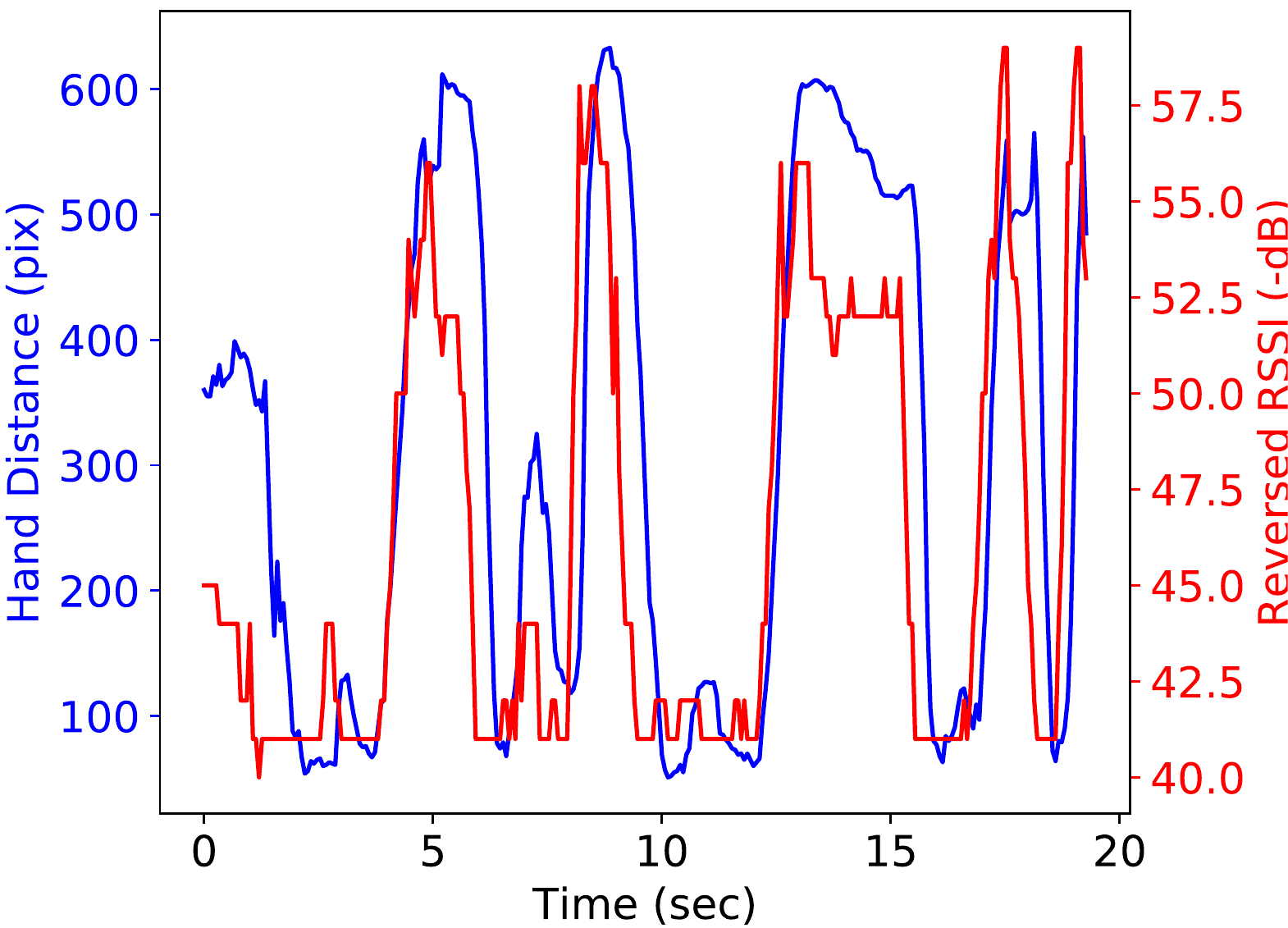}}
    \centerline{(c)}
    \label{fig:grab_rssi_distance}
  \end{minipage}
  \vspace{-2pt}
  \caption{\emph{(a) Vision based item detection does background subtraction and removes the hand outline. (b) Weight sensor readings are correlated with hand proximity events to assign association probabilities. (c) Tag RSSI and hand movements are correlated, which helps associate proximity events to tagged items.}}
  \label{fig:grab_behavior_figs}
\end{figure*}

\parab{Association probabilities from weight sensors} 
In principle, a weight sensor can determine the reduction in total weight when an item is removed from the shelf. Then, knowing which shopper's hand was closest to the shelf, we can associate the shopper with the item. In practice, this association needs to consider real-world behaviors. First, if two shoppers concurrently remove two items of different weights (say a can of Pepsi and a peanut butter jar), the algorithm must be able to identify which shopper took which item. Second, if two shoppers are near the shelf, and two cans of Pepsi were removed, the algorithm must be able to determine if a single shopper took both, or each shopper took one. To increase robustness to these, Grab breaks this problem down into two steps: (a) it associates a proximity event to \textit{dynamics in scale readings}, and (b) then associates scale dynamics to items by detecting weight changes.

\parae{Associating proximity events to scale dynamics}
Weight scales sample readings at 30 Hz. At these rates, we have observed that, when a shopper picks up an item or deposits an item on a shelf, there is a distinct "bounce" (a peak when an item is added, or a trough when removed) because of inertia (\grabfig{behavior_figs}(b)). If $d$ is the duration of this peak or trough, and $d'$ is the duration of the proximity event, we determine the association probability between the proximity event and the peak or trough as the ratio of the intersection of the two to the union of the two. As \grabfig{behavior_figs}(b) shows, if two shoppers pick up items at almost the same time, our algorithm is able to distinguish between them. Moreover, to prevent shoppers from attempting to confuse Grab by temporarily activating the weight scale with a finger or hand, Grab filters out scale dynamics where there is high frequency of weight change.

\parae{Associating scale dynamics to items}
The next challenge is to measure the weight of the item removed or deposited. Even when there are multiple concurrent events, the 30 Hz sampling rate ensures that the peaks and troughs of two concurrent actions are likely distinguishable (as in \grabfig{behavior_figs}(b)). In this case, we can estimate the weight of each item from the sensor reading at the beginning of the peak or trough $w_s$ and the reading at the end $w_e$. Thus $|w_s-w_e|$ is an estimate of the item weight $w$. Now, from the configuration phase, we know the weights of each type of item on the shelf. Define $\delta_j$ as $|w-w_j|$ where $w_j$ is the known weight of the $j$-th type of item in the shelf. Then, we say that the probability that the item removed or deposited was the $j$-th item is given by $\frac{1/{\delta}_{j}}{\sum_i (1/{\delta}_{i})}$. This definition accounts for noise in the scale (the estimates for $w$ might be slightly off) and for the fact that some items may be very similar in weight.

\parae{Combining these association probabilities} 
From these steps, we get two association probabilities: one associating a proximity event to a peak or trough, another associating the peak or trough to an item type. Grab multiplies these two to get the probability, according to the weight sensor, that hand $i$ picked item $j$.

\parab{Association probabilities from RFID tag}
For items which have an RFID tag, it is trivial to determine which item was taken (unlike with weight or vision sensors), but it is still challenging to associate proximity events with the corresponding items. For this, we leverage the fact that the tag's RSSI becomes weaker as it moves away from the RFID reader. \grabfig{behavior_figs}(c) illustrates an experiment where we moved an item repeatedly closer and further away from a reader; notice how the changes in the RSSI closely match the distance to the reader. In smart shelves, the RFID reader is mounted on the back of the shelf, so that when an object is removed, its tag's RSSI decreases. To determine the probability that a given hand caused this decrease, we use probability-based Dynamic Time Warping~\cite{bautista2013probability}, which matches the time series of hand movements with the RSSI time series and assigns a probability which measures the likelihood of association between the two. We use this as the association probability derived from the RFID tag.

\parab{Putting it all together} In the last step, Grab  formulates an assignment problem to determine which hand to associate with which item. First, it determines a time window consisting of a set of overlapping proximity events. Over this window, it first uses the association probabilities from each sensor to define a composite probability $p_{k,m}$ between the $k$-th hand and the $m$-th item: $p_{k,m}$ is a weighted sum of the three probabilities from each sensor (described above), with the weights being empirically determined.

Then, Grab formulates the assignment problem as an optimization problem:

\vspace{-2ex}
{\small
\begin{maxi*}
  {k,m}{\sum{p_{k,m}z_{k,m}}}{}{}
  \addConstraint{\sum_{k \in H}{z_{k,m}}}{\leq 1\quad}{\forall m \in I}
  \addConstraint{\sum_{l \in I_t}{z_{k,l}}}{\leq u_l}{\forall k \in H}
\end{maxi*}
}
\vspace{-1ex}

\noindent
where $H$ is the set of hands, $I$ is the set of items, and $I_t$ is the set of \textit{item types}, and $z_{k,m}$ is an indicator variable that determines if hand $k$ picked up item $m$. The first constraint models the fact that each item can be removed or deposited by one hand, and the second models the fact that sometimes shoppers can pick up more than one item with a single hand: $u_l$ is a statically determined upper bound on the number of items of the $l$-th item that a shopper can pick up using a single hand (\eg it may be physically impossible to pick up more than 3 bottles of a specific type of shampoo). This formulation is a max-weight bipartite matching problem, which we can optimally solve using the Hungarian~\cite{kuhn1955hungarian} algorithm.

%% file: tex/grab/scalability.tex
\subsection{GPU Multiplexing} 

Because retailer margins can be small, Grab needs to minimize overall costs. The computing infrastructure (specifically, GPUs) is an important component of this cost. In what we have described so far, each camera in the store needs a GPU. 

Grab actually enables multiple cameras to be multiplexed on one GPU. It does this by
avoiding running OpenPose on every frame. Instead, Grab uses a \textit{tracker} to track joint positions from frame to frame: these tracking algorithms are fast and do not require the use of the GPU. Specifically, suppose Grab runs OpenPose on frame $i$. On that frame, it computes ORB~\cite{ORB} features around every joint. 
ORB features can be computed faster than previously proposed features like SIFT and SURF. Then, for each joint, it identifies the position of the joint in frame $i+1$ by matching ORB features between the two frames. Using this it can reconstruct the skeleton in frame $i+1$ without running OpenPose on that frame.

Grab uses this to multiplex a GPU over $N$ different cameras. It runs OpenPose from a frame on each camera in a round-robin fashion. If a frame has been generated by the $k$-the camera, but Grab is processing a frame from another (say, the $m$-th) camera, then Grab runs feature-based tracking on the frame from the $k$ camera. Using this technique, we show that Grab is able to scale to using 4 cameras on one GPU without significant loss of accuracy.

%% file: tex/grab/eval.tex
\section{Evaluation}

We now evaluate the end-to-end accuracy of Grab and explore the impact of each optimization on overall performance.~\footnote{Demo video of Grab: https://vimeo.com/245274192}

\subsection{Grab Implementation}

\sepfootnotecontent{hx711}{The HX711 can sample at 80 Hz, but the Arduino MCU, when used with several weight scales, limits the sampling rate to 30 Hz.}

\parab{Weight-sensing Module}
To mimic weight scales on smart shelves, we built scales costing \$6, with fiberglass boards and 2~kg, 3~kg, 5~kg pressure sensors. The sensor output is converted by the SparkFun HX711 load cell amplifier~\cite{hx711} to digital serial signals. An Arduino Uno Micro Control Unit (MCU)~\cite{uno} (\grabfig{eva_figs}(a)-left) batches data from the ADCs and sends it to a server. The MCU has nine sets of serial Tx and Rx so it can collect data from up to nine sensors simultaneously. The sensors have a precision of around 5\-10~g, with an effective sampling rate of 30~Hz\sepfootnote{hx711}.  

\parab{RFID-sensing Module}
For RFID, we use the SparkFun RFID modules with antennas and multiple UHF passive RFID tags~\cite{sparkfun} (\grabfig{eva_figs}(a)-right). The module can read up to 150 tags per second and its maximum detection range is 4~m with and antenna. The RFID module interfaces with the Arduino MCU to read data from tags.

\parab{Video input}
We use IP cameras~\cite{ipcam} for video recording. In our experiments, the cameras are mounted on merchandise shelves and they stream 720p video using Ethernet. We also tried webcams and they achieved similar performance (detection recall and precision) as IP cameras.

\parab{Identity tracking and action recognition}
These modules are built on top of the OpenPose~\cite{openpose} library's skeleton detection algorithm (in C++). As discussed earlier, we use a modified limb association algorithm. Our other algorithms are implemented in Python, and interface with OpenPose using a boost.python wrapper. Our implementation has over 4K lines of code.

\begin{figure}[t]
  \begin{minipage}{0.52\linewidth}
    \centerline{\includegraphics[width=0.7\columnwidth]{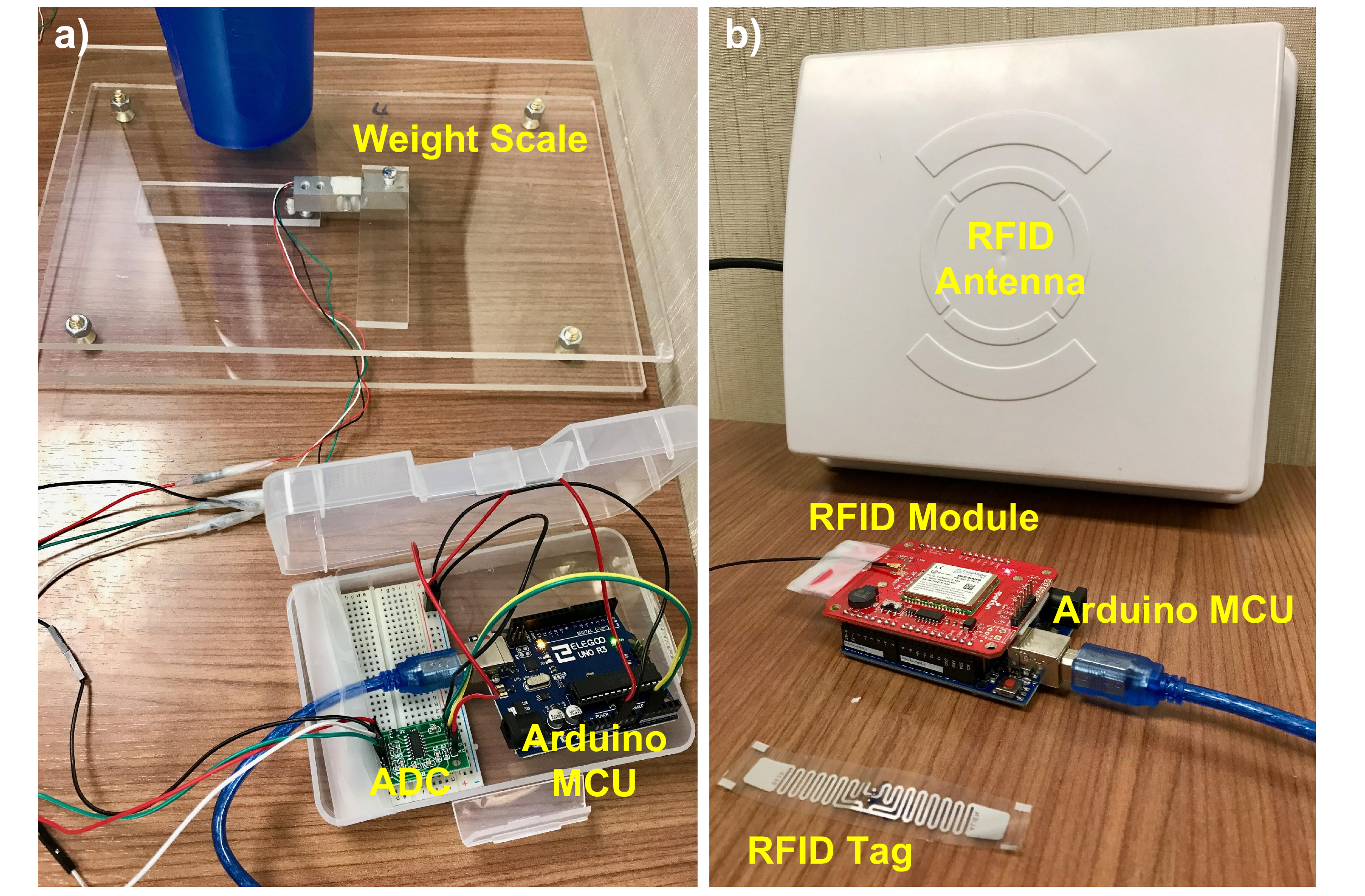}}
    \centerline{(a)}
    \label{fig:grab_sys_demo}
  \end{minipage}
    \begin{minipage}{0.47\linewidth}
    \centerline{\includegraphics[width=0.7\columnwidth]{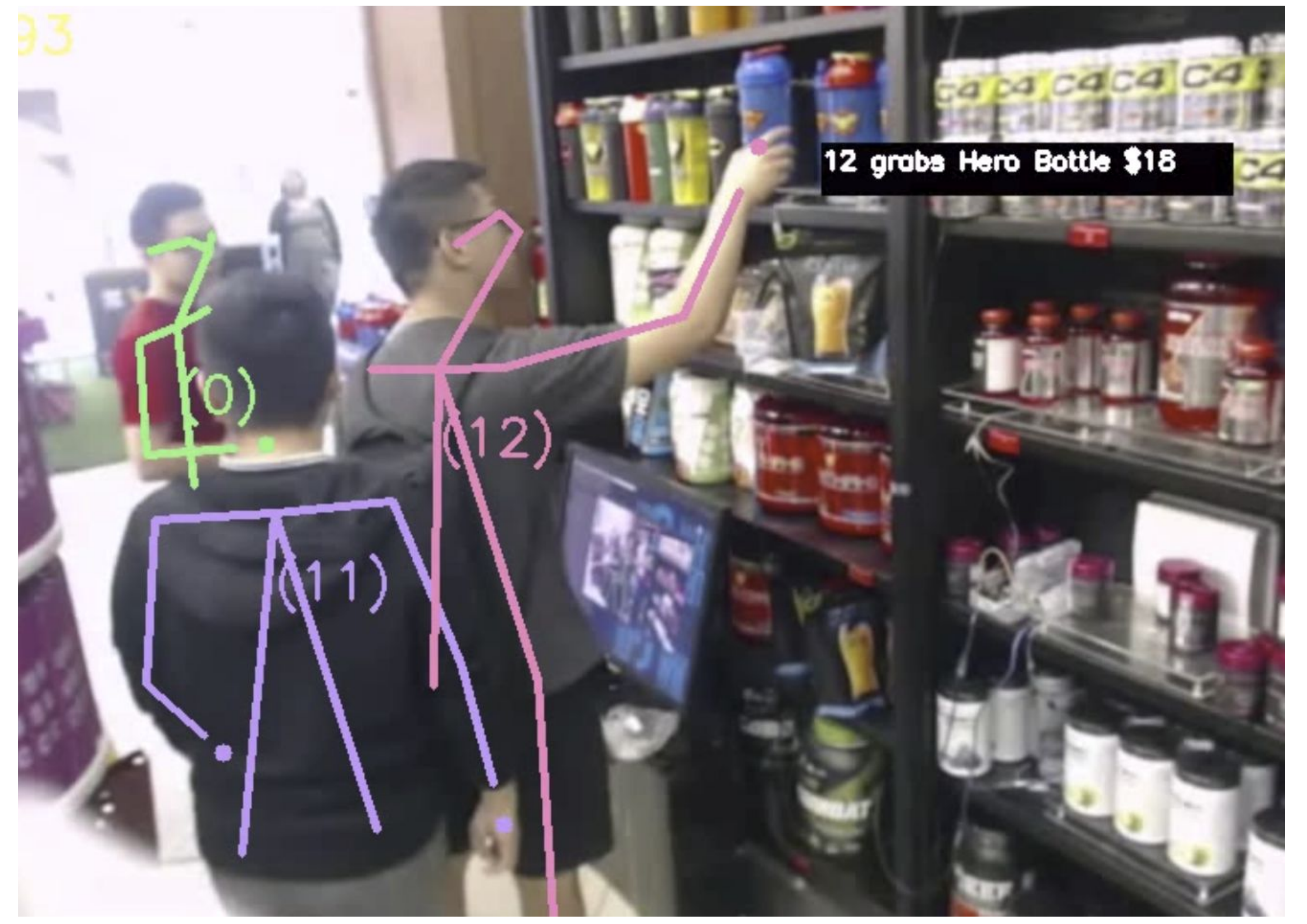}}
    \centerline{(b)}
    \label{fig:grab_tracking_eva}
  \end{minipage}
  \caption{\emph{(a) Left: Weight sensor hardware, Right: RFID hardware; (b) Grab sample output.}}
  \label{fig:grab_eva_figs}
\end{figure}

\subsection{Methodology, Metrics, and Datasets}
\label{sec:meth}

\parab{In-store deployment}
To evaluate Grab, we collected traces from an actual deployment in a retail store. For this trace collection, we installed the sensors described above in two shelves in the store. First, we placed two cameras at the ends of an aisle so that they could capture both the people's pose and the items on the shelves. Then, we installed weight scales on each shelf. Each shelf contains multiple types of items, and all instances of a single item were placed on a single shelf at the beginning of the experiment (during the experiment, we asked users to move items from one shelf to another to try to confuse the system, see below). In total, our shelves contained 19 different types of items. Finally, we placed the RFID reader's antenna behind the shelf, and we attached RFID tags to all instances of 8 types of items.

\parab{Trace collection}
We then recorded five hours worth of sensor data from 41 users who registered their faces with Grab. We asked these shoppers to test the system in whatever way they wished to (\grabfig{eva_figs}(b)). The shoppers selected from among the 19 different types of items, and interacted with the items (either removing or depositing them) a total of 307 times. Our cameras saw an average of 2.1 shoppers and a maximum of 8 shoppers in a given frame. In total, we collected over 10GB of video and sensor data, using which we analyze Grab' performance. 

\parab{Adversarial actions}
During the experiment, we also asked shoppers to perform three kinds of \textit{adversarial actions}. (1) \emph{Item-switching}: The shopper takes two items of similar color or similar weight then puts one back, or takes one item and puts it on a different scale; (2) \emph{Hand-hiding}: The shopper hides the hand from the camera and grabs the item; (3) \emph{Sensor-tampering}: The shopper presses the weight scale with hand. Nearly 40\% of the 307 recorded actions were adversarial: 53 item-switching, 34 hand-hiding, and 31 sensor-tampering actions.


\parab{Metrics}
To evaluate Grab's accuracy, we use \textit{precision} and \textit{recall}. In our context, precision is the ratio of true positives to the sum of true positives and false positives.  Recall is the ratio of true positives to the sum of true positives and false negatives. For example, suppose a shopper picks items A, B, and C, but Grab shows that she picks items A, B, D, and E. A and B are correctly detected so the true positives are 2, but C is missing and is a false negative. The customer is wrongly associated with D and E so there are 2 false positives. In this example, recall is 2/3 and precision is 2/4.

\subsection{Accuracy of Grab}
\label{sec:accuracy_eval}

\begin{figure*}[t]
   \begin{minipage}{0.99\linewidth}
    \centerline{\includegraphics[width=0.85\columnwidth]{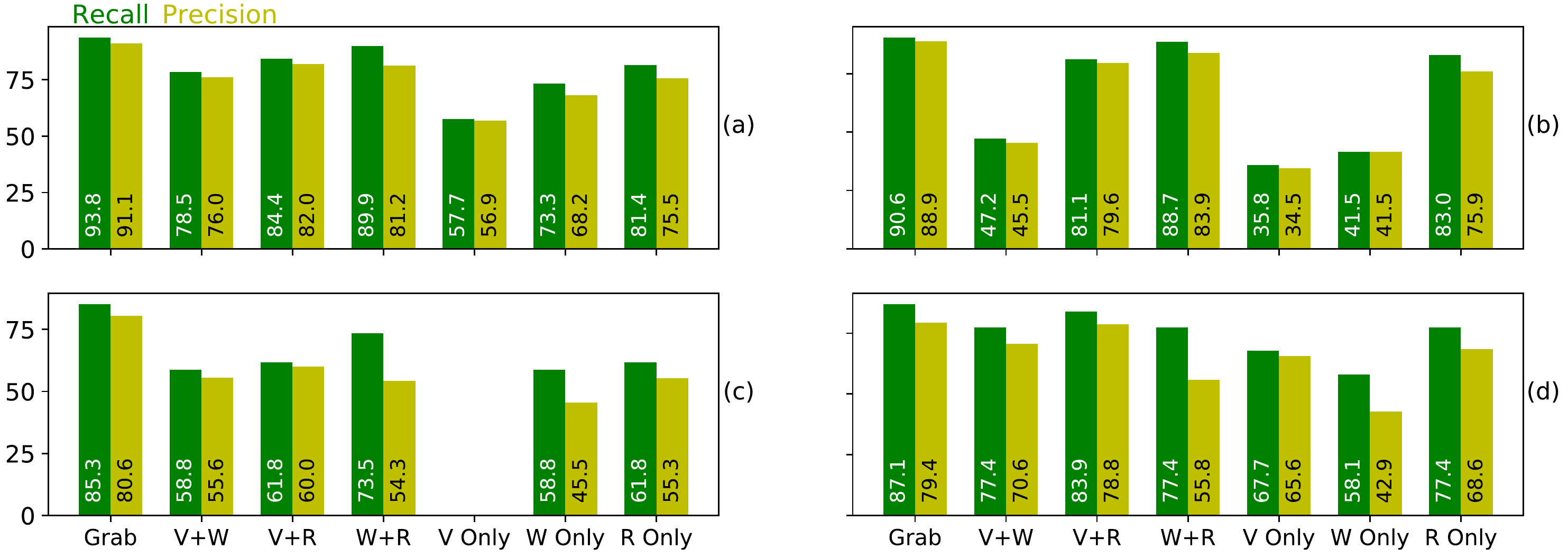}} 
  \end{minipage}
  \caption{\emph{Grab has high precision and recall across our entire trace (a), relative to other alternatives that only use a subset of sensors (\emph{W: Weight; V: Vision; R: RFID}), even under adversarial actions such as (b) Item-switching; (c) Hand-hiding; (d) Sensor-Tampering .}}
  \label{fig:grab_action_accuracy}
\end{figure*}

\parab{Overall precision and recall}
\grabfig{action_accuracy}(a) shows the precision and recall of Grab, and quantifies the impact of using different combinations of sensors: using vision only (\textit{V Only}), weight only (\textit{W only}), RFID only (\textit{R only}) or all possible combinations of two of these sensors. Across our entire trace, Grab achieves a recall of nearly 94\% and a precision of over 91\%. This is remarkable, because in our dataset nearly 40\% of the actions are adversarial. We dissect Grab failures below and show how these are within the loss margins that retailers face today due to theft or faulty equipment.

\sepfootnotecontent{singsens}{For computing the association probabilities. Cameras are still used for identity tracking and proximity event detection.}

\sepfootnotecontent{rfid}{In general, since RFID is expensive, not all objects in a store will have RFID tags. In our deployment, a little less than half of the item types were tagged, and these numbers are calculated only for tagged items.}

Using only a single sensor\sepfootnote{singsens} degrades recall by 12-37\% and precision by 16-36\% (\grabfig{action_accuracy}(a)). This illustrates the importance of fusing multiple sensors for associating proximity events with items. The biggest loss of accuracy comes from using only the vision sensors to detect items. RFID sensors perform the best, since RFID can accurately determine which item was selected\sepfootnote{rfid}. Even so, an RFID-only deployment has 12\% lower recall and 16\% lower precision. Of the sensor combinations, using weight and RFID sensors together comes closest to the recall performance of the complete system, losing only about 3\% in recall, but 10\% in precision.


\parab{Adversarial actions} \grabfig{action_accuracy}(b) shows precision and recall for only those actions in which users tried to switch items. In these cases, Grab is able to achieve nearly 90\% precision and recall, while the best single sensor (RFID) has 7\% lower recall and 13\% lower precision, and the best 2-sensor combination (weight and RFID) has 5\% lower precision and recall. As expected, using a vision sensor or weight sensor alone has unacceptable performance because the vision sensor cannot distinguish between items that look alike and the weight sensor cannot distinguish items of similar weight.

\grabfig{action_accuracy}(c) shows precision and recall for only those actions in which users tried to hide the hand from the camera when picking up items. In these cases, Grab estimates proximity events from the proximity of the ankle joint to the shelf and achieves a precision of 80\% and a recall of 85\%. In the future, we hope to explore cross-camera fusion to be more robust to these kinds of events. Of the single sensors, weight and RFID both have more than 24\% lower recall and precision than Grab. Even the best double sensor combination has 12\% lower recall and 20\% lower precision.

Finally, \grabfig{action_accuracy}(d) shows precision and recall only for those items in which the user trying to tamper with the weight sensors. In these cases, Grab is able to achieve nearly 87\% recall and 80\% precision. RFID, the best single sensor, has more than 10\% lower precision and recall, while predictably, vision and RFID have the best double sensor performance with 5\% lower recall and comparable precision to Grab.

In summary, Grab has slightly lower precision and recall for the adversarial cases and these can be improved with algorithmic improvements, its overall precision and recall on a trace with nearly 40\% adversarial actions is over 91\%. When we analyze only the non-adversarial actions, Grab has \textit{a precision of 95.8\% and a recall of 97.2\%}.


\parab{Taxonomy of Grab failures} 
Grab is unable to recall 19 of the 307 events in our trace. These failures fall into two categories: those caused by identity tracking, and those by action recognition. Five of the 19 failures are caused either by wrong face identification (2 in number), false pose detection (2 in number) (\grabfig{eva_figs}(c)), or errors in pose tracking (one). The remaining failures are all caused by inaccuracy in action recognition, and fall into three categories. First, Grab uses color histograms to detect items, but these can be sensitive to lighting conditions (\eg a shopper takes an item from one shelf and puts it in another when the lighting condition is slightly different) and occlusion (\eg a shopper deposits an item into a group of other items which partially occlude the items). Incomplete background subtraction can also reduce the accuracy of item detection. Second, our weight scales were robust to noise but sometimes could not distinguish between items of similar, but not identical, weight. Third, our RFID-to-proximity event association failed at times when the tag's RFID signal disappeared for a short time from the reader, possibly because the tag was temporarily occluded by other items. Each of these failure types indicates directions of the future work.

\parab{Contextualizing the results} 
From the precision/recall results, it is difficult to know if Grab is within the realm of feasibility for use in today's retail stores. Grab's failures fall into two categories: Grab associates the wrong item with a shopper, or it associates an item with the wrong shopper. The first can result in inventory loss, the second in overcharging a customer. A survey of retailers~\cite{nrf} estimates the \textit{inventory loss ratio} (if a store's total sales are \$100, but \$110 worth of goods were taken from the store, the inventory loss rate is 10\%) in today's stores to be 1.44\%. In our experiments, Grab's failures result in only 0.79\% inventory loss. Another study~\cite{shopper_cost_rate} suggests that faulty scanners can result in up to 3\% overcharges on average, per customer. In our experiments, we see a 2.8\% overcharge rate. These results are encouraging and suggest that Grab may be with the realm of feasibility, but larger scale experiments are needed to confirm this. Additional sensors and algorithm improvements, could further improve Grab's accuracy.

\subsection{The Importance of Efficiency}
\label{sec:efficiency_eval}

\begin{figure}
\centering\includegraphics[width=0.5\columnwidth]{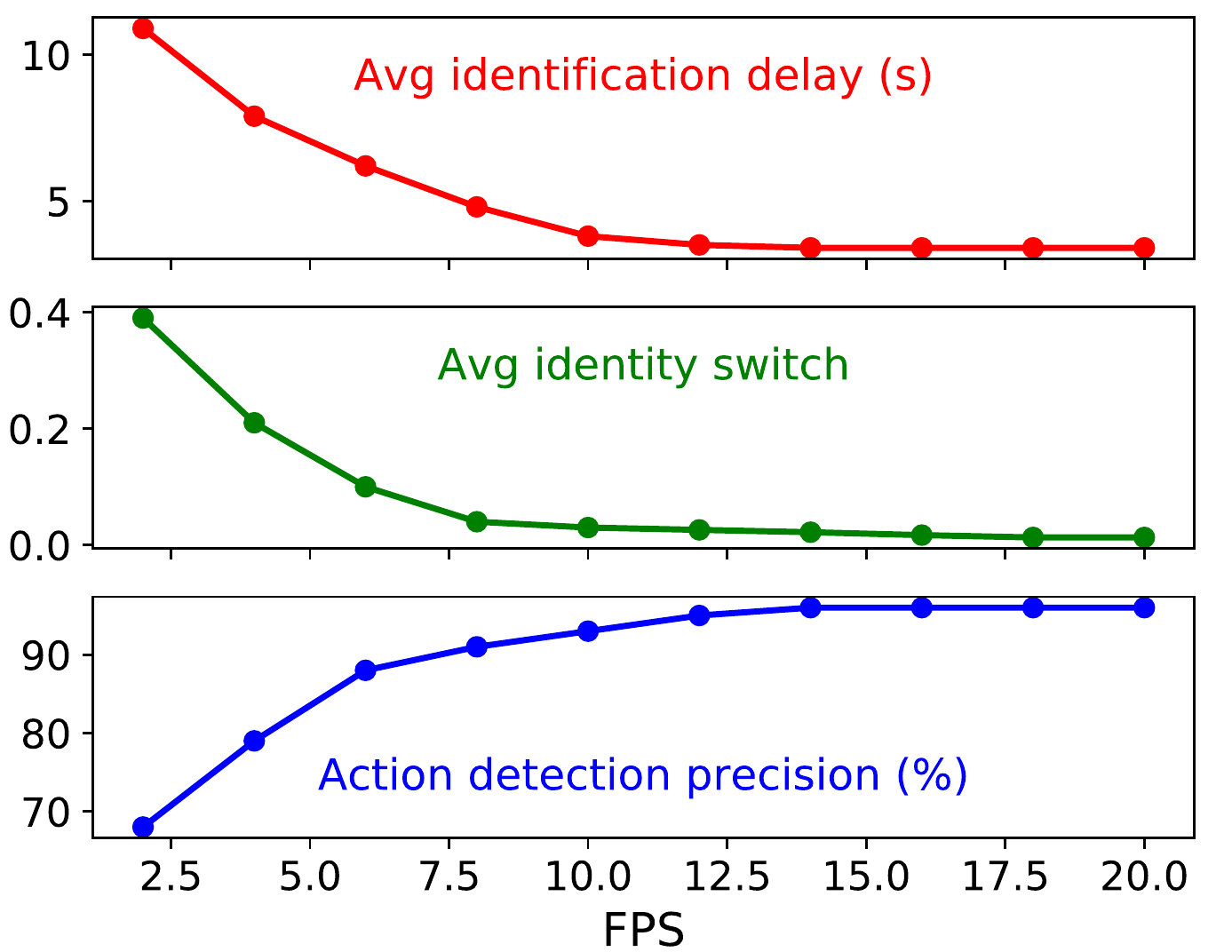}
\caption{\emph{Grab needs a frame rate of at least 10 fps for sufficient accuracy, reducing identity switches and identification delay.}}
\label{fig:grab_fps_relation}
\end{figure}

Grab is designed to process data in near real-time so that customers can be billed automatically as soon as they leave the store. For this, computational efficiency is important to lower cost, but also to achieve high processing rates in order to maintain accuracy. 

\sepfootnotecontent{precision}{In this and subsequent sections, we focus on precision, since it is lower than recall, and so provides a better bound on Grab performance.}

\parab{Impact of lower frame rates}
If Grab is unable to achieve a high enough frame rate for processing video frames, it can have significantly lower accuracy. At lower frame rates, Grab can fail in three ways. First, a customer's face may not be visible at the beginning of the track in one camera. It usually takes several seconds before the camera can capture and identify the face. At lower frame rates, Grab may not capture frames where the shopper's face is visible to the camera, so it might take longer for it to identify the shopper. \grabfig{fps_relation}(a) shows that this identification delay decreases with increasing frame rate approaching sub-second times at about 10 fps. Second, at lower frame rates, the shopper moves a greater distance between frames, increasing the likelihood of \textit{identity switches} when the tracking algorithm switches the identity of the shopper from one registered user to another. \grabfig{fps_relation}(b) shows that the ratio of identity switches approaches negligible values only after about 8~fps. Finally, at lower frame rates, Grab may not be able to capture the complete movement of the hand towards the shelf, resulting in incorrect determination of proximity events and therefore reduced overall accuracy. \grabfig{fps_relation}(c) shows precision\sepfootnote{precision} approaches 90\% only above 10~fps.

\parab{Infeasibility of a DNN-only architecture}
In we argued that, for efficiency, Grab could not use separate DNNs for different tasks such as identification, tracking, and action recognition. To validate this argument, we ran the state-of-the-art open-source DNNs for each of these tasks on our data set. These DNNs were at the top of the leader-boards for various recent vision challenge competitions~\cite{coco_challenge, mot_challenge, mpii}. We computed both the average frame rate and the precision achieved by these DNNs on our data (\tblref{other_options}).

For face detection, our accuracy measures the precision of face identification. The OpenFace~\cite{amos2016openface} DNN can process 15 fps and achieve the precision of 95\%. For people detection, our accuracy measures the recall of bounding boxes between different frames. Yolo~\cite{yolo} can process at a high frame rate but achieves only 91\% precision, while Mask-RCNN~\cite{he2017mask} achieves 97\% precision, but at an unacceptable 5 fps. The DNNs for people tracking showed much worse behavior than Grab, which can achieve an identity switch rate of about 0.027 at 10~fps, while the best existing system, DeepSORT~\cite{wojke2017simpl} has a higher frame rate but a much higher identity switch rate. The fastest gesture recognition DNN is OpenPose~\cite{cao2017realtime} (whose body frame capabilities we use), but its performance is unacceptable, with low (77\%) accuracy. The best gesture tracking DNN, PoseTrack~\cite{iqbal2016PoseTrack}, has a very low frame rate.

Thus, today's DNN technology either has very low frame rates or low accuracy for individual tasks. Of course, DNNs might improve over time along both of these dimensions. However, even if, for each of the four tasks, DNNs can achieve, say, 20~fps and 95\% accuracy, when we run these on a single GPU, we can at best achieve 5~fps, and an accuracy of $0.95^4~=~0.81$. By contrast, Grab is able to process a single camera on a single GPU at over 15 fps (\grabfig{scale_eval}), achieving over 90\% precision and recall (\grabfig{action_accuracy}(a)).

\begin{table}[htbp]
\small
\centering
\begin{tabular}{|l|l|l|}
\hline
\textit{\textbf{Face Detection}}              & \textbf{FPS} & \textbf{Accuracy}      \\ \hline
\textit{OpenFace~\cite{amos2016openface}}     & 15           & 95.1                   \\ \hline
\textit{RPN~\cite{hao2017scale}}              & 5.8          & 95.1                   \\ \hline
\textit{\textbf{People detection}}          & \textbf{FPS} & \textbf{Accuracy}      \\ \hline
\textit{YOLO-9000~\cite{yolo}}              & 35           & 91.0                   \\ \hline
\textit{Mask-RCNN~\cite{he2017mask}}        & 5            & 97.4                   \\ \hline
\textit{\textbf{People tracking}}           & \textbf{FPS} & \textbf{Avg ID switch} \\ \hline
\textit{MDP~\cite{HenschelLCR17}}           & 1.43         & 1.3                    \\ \hline
\textit{DeepSORT~\cite{wojke2017simpl}}     & 17           & 0.8                    \\ \hline
\textit{\textbf{Gesture Recognition}}               & \textbf{FPS} & \textbf{Accuracy*}     \\ \hline
\textit{OpenPose~\cite{cao2017realtime}}            & 15.5         & 77.3                   \\ \hline
\textit{DeeperCut~\cite{insafutdinov2016deepercut}} & 0.09         & 88                     \\ \hline
\textit{\textbf{Gesture Tracking}}              & \textbf{FPS} & \textbf{Avg ID switch} \\ \hline
\textit{PoseTrack~\cite{iqbal2016PoseTrack}}    & 1.6         & 1.8                    \\ \hline
\end{tabular}
\caption{\emph{State-of-the-art DNNs for many of Grab's tasks either have low frame rates or insufficient accuracy. (* Average pose precision on MPII Single Person Dataset)}}
\label{tbl:other_options}
\end{table}

\subsection{GPU multiplexing} 
\label{sec:scale_eval}

In the results presented so far, Grab processes each camera on a separate GPU. The bottleneck in Grab is pose detection, which requires about 63 ms per frame: our other components require less than 7 ms each.

In previous sections, we discussed an optimization that uses a fast feature tracker to multiplex multiple cameras on a single GPU. This technique can sacrifice some accuracy, and we are interested in determining the sweet spot between multiplexing and accuracy. \grabfig{scale_eval} quantifies the performance of our GPU multiplexing optimization. \grabfig{scale_eval}(a) shows that Grab can support up to 4 cameras with a frame rate of 10 fps or higher with fast feature tracking; without it, only a single camera can be supported on the GPU (the horizontal line in the figure represents 10 fps). Up to 4 cameras, \grabfig{scale_eval}(b) shows that the precision can be maintained at nearly 90\% (\ie negligible loss of precision). Without fast feature tracking, multiplexing multiple cameras on a single GPU reduces the effective frame rate at which each camera can be processed,  reducing accuracy for 4 cameras to under 60\%. Thus, with GPU multiplexing using fast feature tracking, Grab can reduce the investment in GPUs by 4$\times$.

\begin{figure}[htbp]
\centering
\includegraphics[width=0.5\columnwidth]{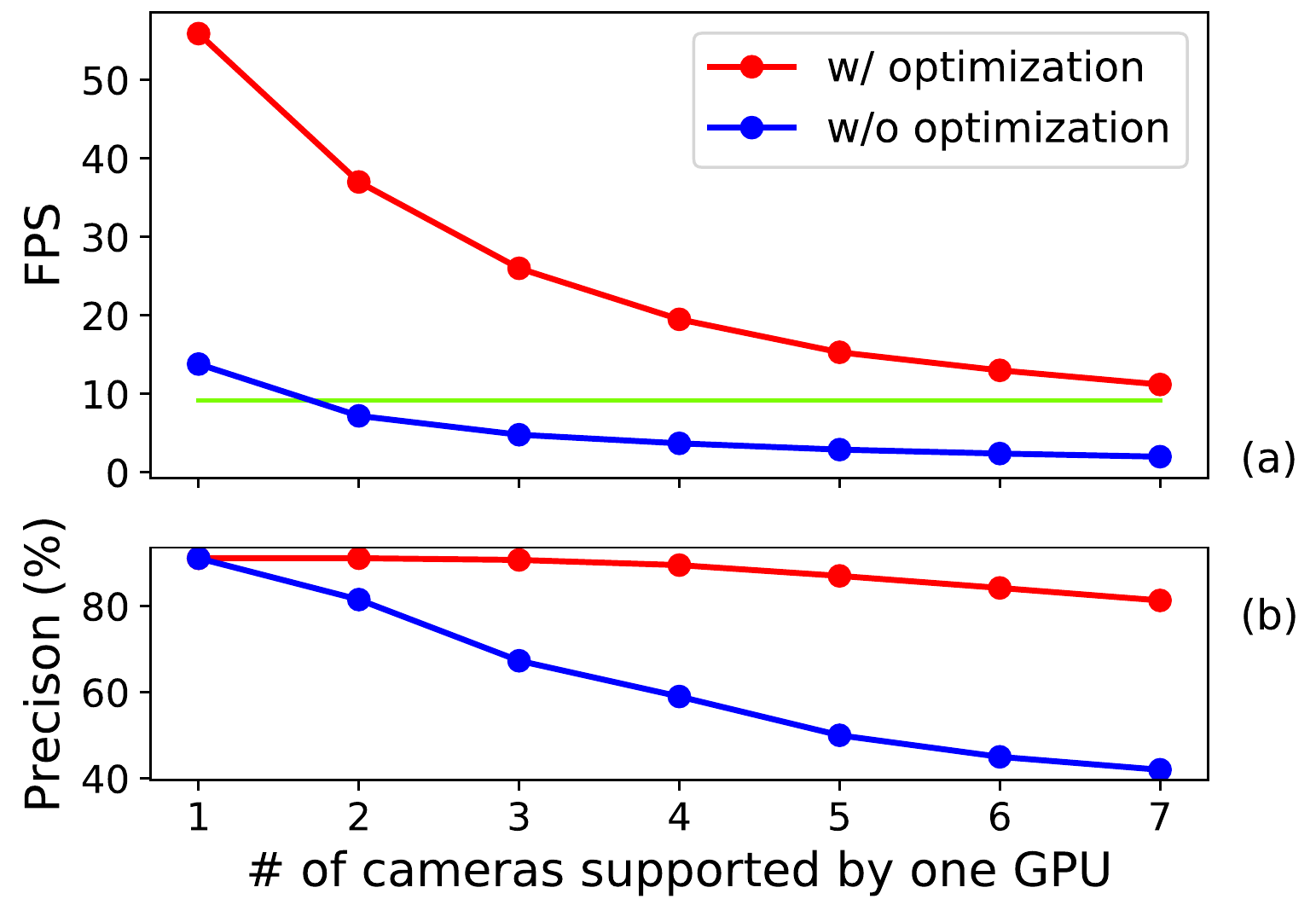}
\caption{\emph{GPU multiplexing can support up to 4 multiple cameras at frame rates of 10 fps or more, without noticeable lack of accuracy in action detection.}}
\label{fig:grab_scale_eval}
\end{figure}

%% file: tex/paper_tar.tex
\chapter{TAR: Enabling Fine-Grained Targeted Advertising in Retail Stores}\label{chap:tar}

\input{tex/tar/intro}
\input{tex/tar/motivation}
\input{tex/tar/design}
\input{tex/tar/eval}

%% file: tex/tar/intro.tex
\section{Introduction}

Digital interactions influence 49\% of in-store purchases, and over half of them take place 
on mobile devices~\cite{mobile_influence}. With this growing trend, brick-and-mortar retailers 
have been evolving their campaigns to effectively reach people with mobile devices, showcase products,
and ultimately, influence their in-store purchase. Among them, sending targeted advertisements (ads) to 
user's mobile devices has emerged as a frontrunner~\cite{mobileads}. 

\added{To send well-targeted information to the shopper, the retailers (and
advertisers) should correctly understand customers' interest. The key to learning the customer's interest is 
to accurately track and recognize the customer during her stay in the store.
Therefore, the retailers need a practical system for shopper tracking and identification 
with real-time performance and high accuracy. For example, the retailer's advertising system would 
require aisle-level or meter-level accuracy in tracking a shopper to infer customer's dwelling time
at a certain aisle. Moreover, the advertising should be able to reflect the customer's position 
change fast, because people usually stay at, or walk by, a specific shelf in just a few seconds.}

\added{Some sensor-based indoor tracking metrics are invented, such as Wi-Fi localization~\cite{euclidanalytics, ciscomeraki}, 
Bluetooth localization~\cite{swirl, inmarket}, stereo cameras~\cite{brickstream,xovis,hella,shoppertrak}, and thermal 
sensors~\cite{irisys}. However, such approaches are either expensive in hardware cost or inaccurate for retail scenarios. 
Some commercial solutions~\cite{fbad, twittermobile} send customers the entire store's information when they enter 
the store zone. Such promotions are coarse-grained and can hardly trigger customers' interests.}

\added{Recently, live video analytics has become a promising solution for accurate shopper tracking. 
Companies like Amazon Go~\cite{amazongo} and Standard Cognition~\cite{std_cog} use 
close-sourced algorithms to identify customers and track their in-store movement. The opensource community 
also has proposed many accurate metrics for people (customer) identification and tracking.} 

\added{For people (re)identification, there are two mainstream approaches: face recognition and body feature 
classification. Today's face recognition solutions (\cite{hayat2017joint, tran2017disentangled, masi2016pose}) 
can reach up to 95\% of precision on public datasets, thanks to the advance of deep neural networks (DNN). 
However, the customer's face is not always available in the camera, and the face image may be blurry 
and dark due to poor lighting and long distance. The body-feature-based solutions~\cite{bai2017scalable, chen2017beyond, zhao2017spindle} 
do not deliver high accuracy (< 85\%) and also suffer from bad video quality.}

\added{For people tracking, the retailer needs both single-camera tracking and cross-camera tracking 
to understand the walking path of each customer. 
Recent algorithms for single-camera tracking~\cite{wojke2017simple, bae2018confidence, milan2017online, fagot2016improving} 
leverage both the person's visual feature and past trajectory to track her positions in following frames. 
However, such algorithms cannot perform well in challenging environments, e.g., similar clothes, long occlusion, 
and crowded scene. Existing cross-camera tracking algorithms~\cite{ristani2018features, shiva2017inter, tesfaye2017multi, chen2017equalized} 
use the camera network's topology to estimate the cross-camera temporal-spatial similarity and match each customer's trace across cameras. 
Such solutions face challenges like unpredictable people movement (between the surveillance zones).}  

\added{In this chapter, we propose TAR to overcome the above limitations. As summarized above, 
existing indoor localization solutions are not accurate enough in practice and usually require the deployment of complicated 
and expensive equipment. Instead, this chapter proposes a practical end-to-end shopper tracking and identification system. 
TAR is based on two fundamental ideas: \emph{Bluetooth proximity sensing} and \emph{video analytics}.} 

To infer a shopper's identity, TAR looks into Bluetooth Low Energy (BLE) signal broadcasted from the user's device. 
BLE has recently gained popularity 
with numerous emerging applications in industrial Internet-of-Thing (IoT) and home automation. Proximity estimation is one of the most 
common use cases of BLE beacons~\cite{beaconImpact}. Apple iBeacon \cite{ibeacon}, Android EddyStone~\cite{androidBeacon}, and 
open-sourced AltBeacon~\cite{altBeacon} are available options. Several retail giants (e.g., Target, Macy's)
have already deployed them in stores to create a more engaging shopping experience by identifying items in proximity to customers
~\cite{beaconRetailMarketing, inmarket, swirl}. 

TAR takes a slightly different perspective from the above scenario in that shoppers carry BLE-equipped devices and TAR 
utilizes BLE signals to enhance tracking and identify shoppers. In a high level, TAR achieves identification by attaching 
the sensed BLE identity to a visually tracked shopper. TAR notices the pattern similarity between shopper's BLE proximity trace 
and her visual movement trajectories. Therefore, the identification problem converts to a trace matching problem. 

In solving this matching problem, TAR encounters four challenges. 
First, pattern matching in real-time is challenging due to different coordinate systems and noisy trace data. 
TAR transforms both traces into the same coordinates with camera homography projection and BLE signal processing. Then, 
TAR devises a probabilistic matching algorithm that based on Dynamic Time Warping (DTW)~\cite{dtw} to match the patterns. 
To enable the real-time matching, TAR applies a moving window to match trace segments and uses the cumulative confidence score 
to judge the matching result. 

Next, assigning the highest-ranked BLE identity to the visual trace is often incorrect. 
Factors like short visual traces could significantly increase the assignment uncertainty. To solve this problem, 
TAR uses a linear-assignment-based algorithm to correctly determine the BLE identity. Moreover, instead of focusing on a single trace, 
TAR looks at all visual-BLE pairs (i.e., a global view) and assigns IDs for all visual traces in a camera. 

Third, a single user's visual tracking trace can frequently break upon occlusions. To solve this issue, TAR implements a rule-based 
scheme to differentiate ambiguous visual tracks during the assignment process and connects broken tracks, regarding each BLE ID.   

Finally, it is non-trivial to track people across cameras with different camera positions and angles. 
\added{Existing works~\cite{solera2016tracking, xu2017cross, ristani2018features, shiva2017inter} either work offline or require 
overlapping camera coverage to handle a transition from one camera to the other}. 
However, overlapping coverage is not guaranteed in most shops. To overcome this issue, 
TAR proposes an adaptive probabilistic model that tracks and identifies shoppers across cameras with little constraint.

We have deployed TAR in an office and a retail store environment, and analyzed TAR's performance with various settings. 
Our evaluation results show that the system achieves 90\% accuracy, which is 20\% better than 
the state-of-the-art multi-camera people tracking algorithm. Meanwhile, TAR achieves a mean speed of 11 frame-per-second (FPS) 
for each camera, which enables the live video analytics in practice. 

The main contributions of our work are listed below:

\begin{itemize}
\setlength\itemsep{0.005em}
\item development of TAR, a system for multi-camera shopper tracking and identification. TAR can be seamlessly integrated 
with existing surveillance systems, incurring minimal deployment overhead;
\item introduction of four key elements to design TAR;
\item a novel vision and mobile device association algorithm with multi-camera support; and
\item implementation, deployment, and evaluation of TAR. TAR runs in real-time and achieves over 90\% accuracy.
 \end{itemize}

%% file: tex/tar/motivation.tex
\section{Motivation}

\parab{Retail trends:}
While the popularity of e-commerce continues to surge, offline in-store commerce still dominates in today's 
market.  Studies in \cite{forrester, deloitte} show that 91$\%$ of the purchases are made in physical 
shops. In addition, \cite{accenture} indicates that 82$\%$ of the Millennials prefer to shop in brick and mortar 
stores. As online shopping evolves rapidly, it is crucial for offline shops to change the form and offer 
better shopping experience. Therefore, it is essential for offline retailers to understand shoppers' demands for better service
given that today's customers are more informed about the items they want. 

\parab{The need for shopper tracking and identification:}
By observing where the shopper is and how long she visits each area, 
retailers can identify the customer's shopping interest, and hence, provide a customized shopping experience 
for each people. For example, many large retail stores (e.g., Nordstrom~\cite{nordstrom}, Family Dollar, Mothercare~\cite{familydollar}) are already adopting shopper tracking solutions (e.g., Wi-Fi localization). 
These retailers then use the gathered data to help implement store layouts, product placements, and product promotions.

\parab{Existing solutions:}
Several companies~\cite{skyrec, shoppertrak, retailnext} provide solutions for shopper behavior tracking by primarily using surveillance camera feeds. The solutions include features like shopper counting, the spatial-temporal distribution of customers, and shoppers' aggregated trajectory. However, they are not capable of understanding per-shopper insight (or identity). 
Services like Facebook~\cite{fbad, twittermobile} offer targeted advertisement for retail stores. They leverage coarse-grained location data and the shopper's online browsing history to identify the store-level information (which store is the customer visiting) and push relevant advertisements. Therefore, such solutions can hardly recognize the shopper's in-store behavior. 

Camera-based tracking with face recognition can be used to infer the shoppers' indoor activities, but it also introduces several concerns -- privacy, 
availability, and accuracy. First, the face is usually the privacy-sensitive information, and collecting such information might increase user's privacy concern (or even violation of law).  Second, the face in the surveillance camera is sometimes unavailable due to various camera angles and body poses. Moreover, face recognition algorithms are known to be vulnerable to factors like image quality. Finally, face recognition requires the user's face image to train the model, which adds overhead to shoppers; asking them to submit a good face image and verifying the photo authenticity (e.g., offline ID confirmation) are not easy. 

\parab{Our proposed approach:} TAR adopts a vision-based tracking metric but extends it to enable
shopper identification with BLE. We exploit the availability of BLE signals from the shopper's 
smartphone and combine the signal with vision-based technologies to achieve good tracking accuracy across cameras.

Modern smartphones equips with Bluetooth Low Energy (BLE) chip, and there are many BLE-based applications and hardware developed. 
A typical usage of BLE is to act as a beacon, which broadcasts BLE signal at a particular frequency. 
The beacon can serve as a unique identifier for the device and can be used to estimate the proximity to the receiver~\cite{proximities}. 
Our approach assumes the availability of BLE signals from shoppers, and this assumption becomes popular via incentive mechanism (e.g. mobile apps for coupons). 

\added{Therefore, in addition to our customized vision-based detection and tracking algorithms, 
 we carefully integrate them with BLE proximity information to achieve high accuracy for tracking and identification across cameras.}

In designing the system, we aim to achieve the following goals: 

\begin{itemize}

\item \textbf{Accurate}: TAR should outperform the accuracy of existing multi-camera tracking systems. 
It should also be precise in distinguishing people's identity. 

\item \textbf{Real-time}: TAR should recognize each customer's identity in a few seconds since a shopper
might be highly mobile across multiple cameras. Meanwhile, TAR should detect the appearance of people with high frame per second (FPS). 

\item \textbf{Practical}: TAR should not need any expensive hardware or complex deployment. It can leverage existing surveillance camera systems and the user's smartphone. 

\end{itemize}

%% file: tex/tar/design.tex
\section{TAR Design}

This section presents the design of TAR. We begin with an overview of the TAR's design 
and the motivating use cases of the design.  Then, we explain the detailed components that address 
technical challenges specific to the retail environment. 

\begin{figure}
\includegraphics[scale=0.5]{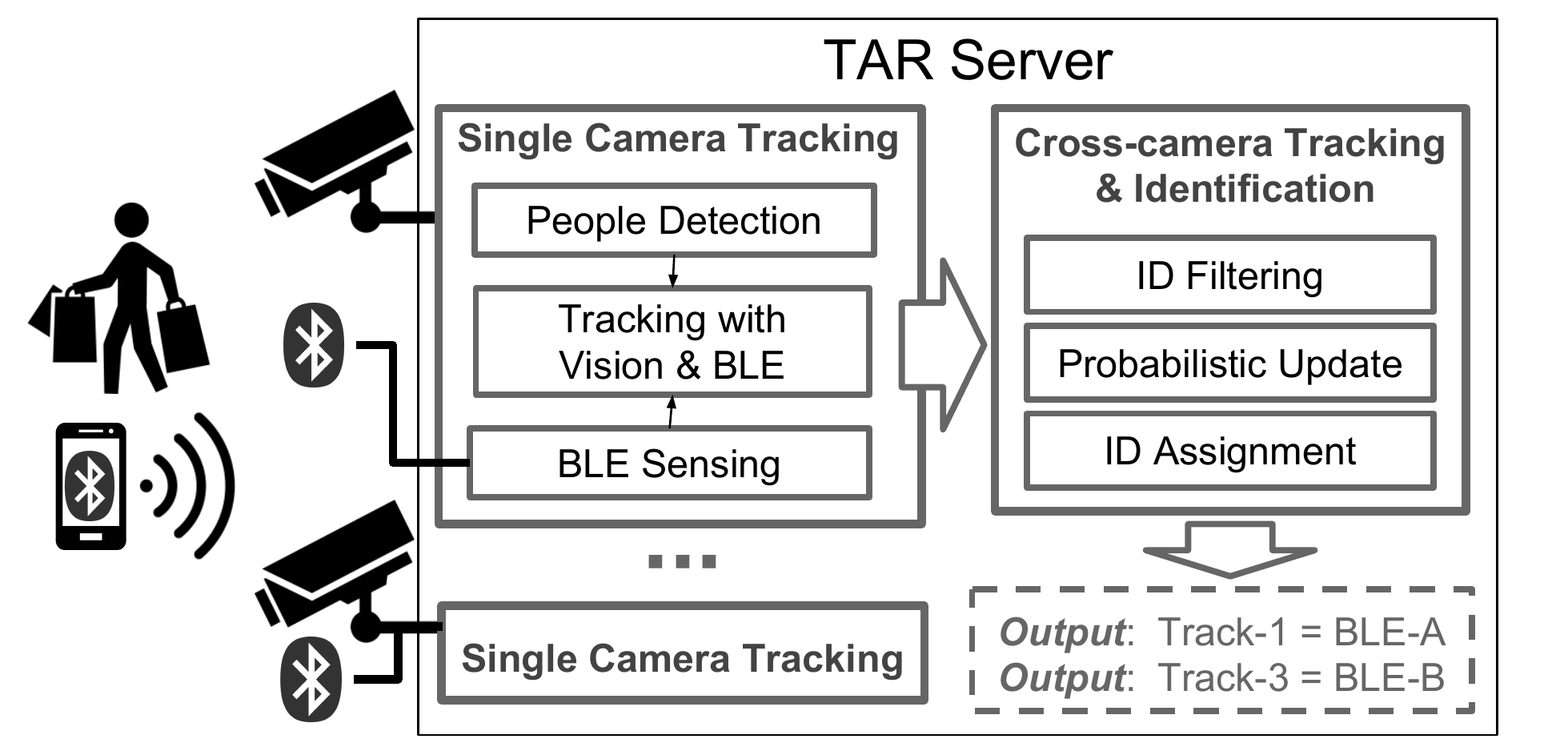}
\centering\caption{\emph{\small \added{System Overview for TAR}}}
\label{fig:tar_overview}\vspace{-0.2in}
\end{figure}


\subsection{Design Overview}
\label{sec:overview}

\tarfig{overview} depicts the design of TAR, and it consists of two major parts: 
1) mobile Bluetooth Low Energy (BLE) library that enables smart devices as BLE beacons 
in the background, and
2) server backend that collects the real-time BLE signals and video data as well as performs customer 
tracking and identification.  
First, we assume customers usually carry their smartphones with a store application 
installed~\cite{cartwheel}. The store app equips 
with TAR's mobile library that broadcasts BLE signal as a background thread. 
Note that the BLE protocol is designed with minimum battery overhead~\cite{blebroadcast}, and 
the broadcasting process does not require customer's intervention. 
Next, TAR's server backend includes several hardware and software components. 
We assume each surveillance camera equips with a BLE receiver for BLE sensing. 
Both the camera feed and the BLE sensing data are sent to TAR for real-time processing. 

TAR is composed of several key components to enable accurate tracking and identification. 
It has a deep neural network (DNN) based tracking algorithm to track 
users with vision trace, and then, incorporates a BLE proximity algorithm to estimate the user's  
movement. In addition, TAR adopts a probabilistic matching algorithm 
based on Dynamic Time Warping (DTW)~\cite{dtw} to associate both vision and BLE data and find out 
the user's identity. However, external factors such as people occlusion 
could harm the accuracy of sensed data and relying solely 
on the matching algorithm usually results in the error. To handle this issue, 
TAR uses a stepwise matching algorithm based on cumulative confidence score.  
After that, TAR devises an ID assignment algorithm to determine the correct identity
from the global view. As the vision-based trace might frequently break, 
sewing them together is essential to learning user interests. We propose 
a rule-based scheme to identify ambiguous user traces and properly connect them.  
Finally, the start of the probabilistic matching process 
will encounter more uncertainty due to the limited trace's length. Therefore, TAR 
considers each user's cross-camera temporal-spatial relationship and 
carefully initializes its initial confidence level to improve the identification accuracy.

\subsection{\added{A Use Case}}

\begin{figure}
\includegraphics[ scale=0.5]{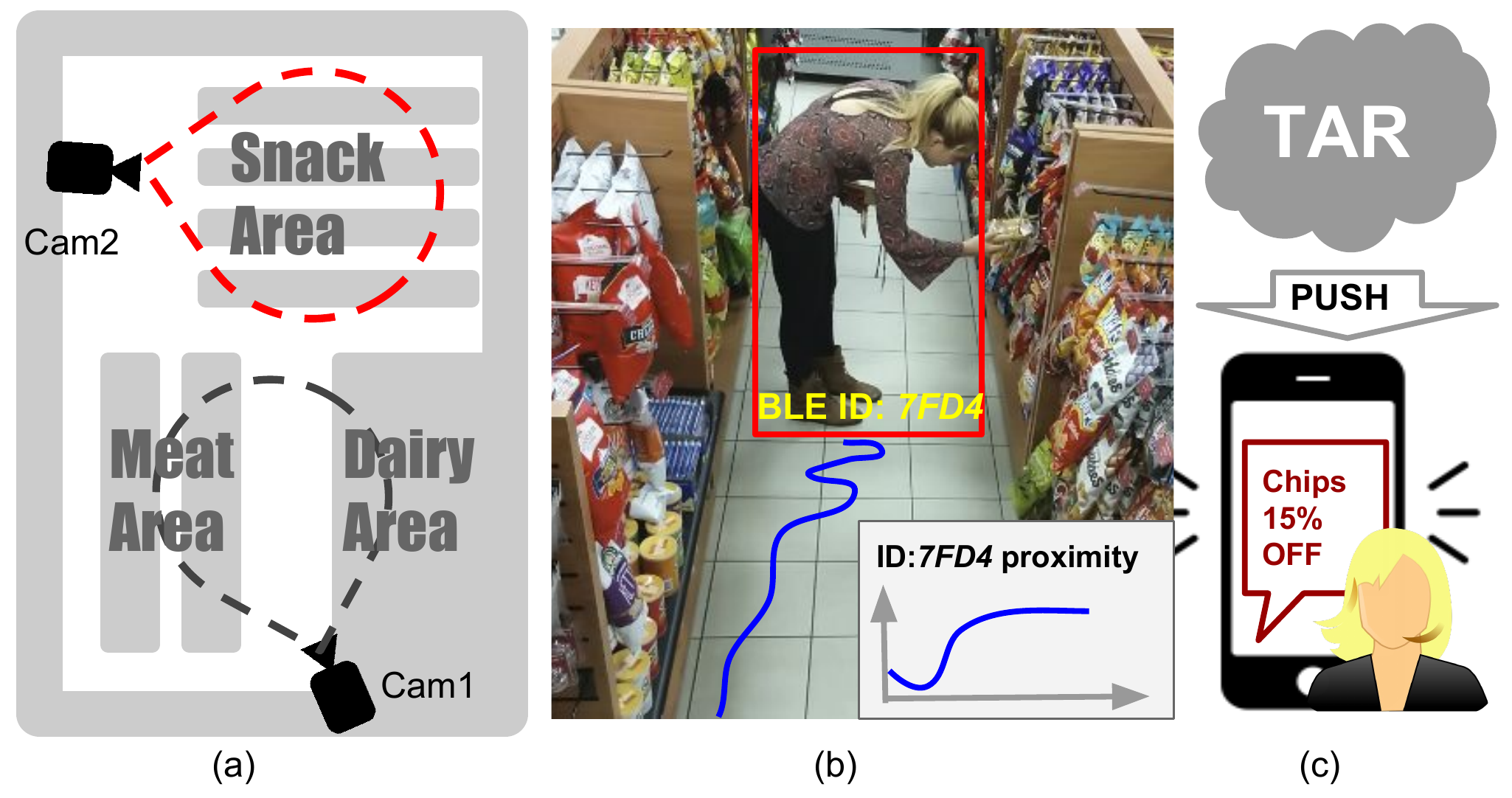}
\centering\caption{\emph{\small A Targeted Ad Working Example in Store}}
\label{fig:tar_example}\vspace{-0.2in}
\end{figure}

\tarfig{example} illustrates an example of how TAR works. A grocery store is 
equipped with two video cameras that cover different aisles, as shown in 
\tarfig{example}(a). Assume a customer with her smartphone enters the store 
and the app starts broadcasting BLE signal. The customer is looking for some snacks  
and finally finds the snack aisle. During her stay, two cameras can 
capture her trajectory. Briefly, camera-1 (bottom) sees the user at first and senses 
her BLE signals. Then the starts matching the user's visual trace to estimated 
BLE proximity trace.  TAR maintains a confidence score for the tracked customer's 
BLE identity. When the user exits the camera-1 zone and enters camera-2 region (top), 
TAR considers various factors including temporal-spatial relationship and visual feature 
similarity, and then, adjusts the initial confidence score for the customer in the
new zone. Then, camera-2 starts its own tracking and identification progress and 
concludes the customer's identity ($7FD4$ in ~\tarfig{example}(b)). TAR then 
continuously learns her dwell-time and fine-grained trajectory on each shelf.

In following sections, we detail core components of TAR to realize the features above
and other use cases of fine-grained tracking and identification. 

\subsection{Vision-based Tracking (VT)}
\label{sec:VT}

We design a novel vision-based tracking metric (VT) that 
consists of three components: people detection and visual feature
extraction, visual object tracking, and physical trajectory estimation.

\subsubsection{People Detection and Deep Visual Feature}
\label{sec:deepfeature}

Recent development in DNN provides us 
with accurate and fast people detector. It detects people in each 
frame and marks the detected positions with bounding boxes. Among various
proposals, we choose Faster-RCNN~\cite{ren2015faster} as TAR's people detector 
because it achieves high accuracy as well as a reasonable speed. 

In addition to the detection, TAR extracts and uses the visual feature of 
the detected bounding box to improve inter-frame people tracking. Briefly, 
once a person's bounding box is detected, TAR extracts its visual feature 
using DNN. The ideal visual feature could accurately classify each people
under different people poses and lighting conditions. 
Recently, DNN-based feature extractors have been proposed
and outperform other features (e.g., color histogram, SIFT~\cite{lowe2004distinctive}) 
regarding the classification accuracy. We have evaluated the state-of-art feature extractors, 
including CaffeNet, ResNet, VGG16, and GoogleNet~\cite{caffereid}, and have identified 
that the convolution neural network (CNN) version of 
GoogleNet~\cite{zheng2016discriminatively} delivers the best performance
in the tradeoff of speed and accuracy. After that, we have further trained the model 
with two large-scale people re\-identification datasets together (MARS~\cite{zheng2016mars} 
and DukeReID~\cite{gou2017dukemtmc4reid}), with over 1,100,000 images 
of 1,261 pedestrians.

\subsubsection{People Tracking in Consecutive Video Frames}
The tracking algorithm in TAR is inspired by DeepSort~\cite{wojke2017simple}, 
a state-of-the-art online tracker. 
In a high level, DeepSort combines each people's visual feature
with a standard Kalman filter, which matches objects 
based on squared Mahalanobis distance~\cite{de2000mahalanobis}. DeepSort 
optimizes the matching process by minimizing the cosine distance between deep features. 
However, it often fails when multiple people collide in a video. The cause 
is that the size of a detection bounding box, covering colliding people, 
becomes large, and the deep visual feature calculated from the bounding box 
cannot accurately represent the person inside. 

To overcome this problem, TAR leverages the geometric relationship 
between objects. When multiple people are close to each other and their 
bounding boxes have a large intersection-over-union (IOU) ratio, TAR 
will not differentiate those persons using DNN-generated visual features.
Instead, those people's visual traces will be regarded as "merged" until some people 
start leaving the group. When the bounding boxes' IOU values become lower than a certain
threshold (set to 0.3), they will be regarded as "departed" and 
TAR will resume the visual-feature-enabled tracking.

The hybrid metric above also faces some challenges. 
When two users with similar color clothes come across each other, 
the matching algorithm sometimes fails because the users' IDs (or tracking IDs) 
are switched. To avoid this error, we propose a \emph{kinematic verification} 
component for our matching algorithm. The idea is that 
people's movement is likely to be constant in a short period.
Therefore, we compute the velocity and the relative orientation 
of each detected object in the current frame, and then compare it to existing 
tracked objects' velocity and orientation. This component serves as 
a verification module that triggers the matching only for objects whose kinematic
conditions are similar. TAR avoids the confusion, as the two users 
above show different velocity and orientation. 

The people tracking algorithm in TAR synthesizes the temporal-spatial 
relationship and visual feature distance to track each person (her ID) 
accurately. First, it adopts a Kalman filter to predict the moving direction
and speed of each person (called, track), and then, predicts tracks' position 
in the next frame. In the next frame, TAR computes a distance between the predicted
position and each detection box's position. Second, TAR calculates each bounding box's 
intersection area with the last few positions of each track. 
Larger IOU ratio means higher matching probability. 
Third, TAR extracts a deep visual feature of the detected 
object, and then, compares the feature with the track's. Here, TAR can 
filter out tracks with the kinematic verification, and then apply all the three 
matching metrics. Finally, it assigns each detection to a track. 
If a detection cannot match any track with enough confidence, TAR will search one frame 
backward to find any matched track. On the other hand, if a track is not 
matched for a long time (a person moves out of a camera's view), 
it is regarded as ``missing'', and hence, will be deleted. 

\subsubsection{Physical Trajectory Estimation}
Once finishes the visual tracking, TAR then converts the 
results to physical trajectories by applying the homography 
transformation~\cite{homography}. Specifically, TAR infers people's
physical location by using both visual tracking results and several 
parameters of the camera.  
Assuming the surveillance cameras are stationary and 
well calibrated, TAR can estimate the height and the facing direction 
of detected objects in world coordinates. Moreover, these cameras
can provide information about their current resolution and angle-of-view. 
With these calibration properties, TAR calculates a projective 
transformation matrix~\cite{homography} $H$ that maps each pixel in the frame
to the ground location in the world coordinates. As a person (or track) 
moves, TAR can associate its distance change with a timestamp, 
yielding physical trajectory. 

However, the homography mapping process introduces a unique challenge;
it needs to project entire pixels in a detected bounding boxes 
(bbox) to estimate physical distance, but the bbox size 
may vary frame by frame. For example, a person's bbox may cover 
her entire body in one frame, and then, it might only include an 
upper body in the next. Moreover, transforming the whole pixels 
in the bbox imposes an extra burden on computation. To deal with 
this challenge, TAR chooses a single reference pixel 
for each detected person, while ensuring spatial consistency of 
the reference pixel even in changing bbox. Specifically, 
TAR picks a pixel that is crossing between the bbox and ground, 
i.e., a person's feet position. TAR uses this bottom-center pixel 
of the bbox to represent its reference pixel. One may argue
that the bbox's bottom may not always be a foot position (e.g., when the customers' 
lower body is blocked). TAR leverages the fact that a person's width 
and height show a ratio around 1:3. With this intuition, TAR checks whether
a detected bbox is "too short" -- blocked -- and, if so, TAR extends 
the bottom side of the bbox, based on the ratio. Our evaluation shows
that TAR's physical trajectory estimation achieves less than 10\% of 
an error, even in a crowded area. 


\subsection{People Tracking with BLE}
\label{sec:bletrack}

In addition to VT, TAR relies on BLE proximity to accurately
estimate people's trajectories. We first introduce BLE beacons and then 
explain TAR's proximity estimation algorithm. 

\parab{BLE background.}
BLE beacon represents a class of BLE devices. It periodically broadcasts
its identifier to nearby devices. A typical BLE beacon is powered by 
a coin cell battery and could have 1$-$3 years of lifetime. Today's
smartphones support Bluetooth 4.0 protocol so they can operate as a BLE 
beacon (transmitter). Similarly, any device that supports Bluetooth 4.0 
can be used as BLE receiver. TAR's mobile component enables a customer's 
smartphone as a BLE beacon.  This component is designed as a library, 
and other applications (e.g., store app) can easily integrate it and run
as a background process. 

\parab{Proximity Trace Estimation.}
The BLE proximity trace is estimated by collecting BLE beacons' time series proximity data. Through our extensive evaluation, we select the proximity algorithm in~\cite{altBeacon} to estimate the distance from BLE beacon to the receiver. \added{There are two ways to calculate the proximity using BLE Received Signal Strength (RSS): (1) $d\ =\ exp((E-RSS)\ /\ 10n)$ where $E$ is transmission power (dBm) and $n$ is the fading coefficient; (2) The beacon's transmission power $ts$ defines the expected RSS for a receiver that is one meter away from the beacon. We denote the actual RSS as $rs$. Then we get $rt=\frac{rs}{ts}$. The distance can be estimated with $rt < 1.0\ {?}\ rt^{10}\ :\ {c_1}rt^{c_2} + c_3$, in which $c_1$, $c_2$ and $c_3$ are coefficients from data regression. We implement both algorithms and compare their performance on the collected data. We find the second option is more sensitive to movement and therefore reflects the movement pattern more timely and accurately.} 

In practice, these coefficients depend on the receiver device manufacturers. For example, Nexus 4 and Nexus 5 use same Bluetooth chip from LG, so they have the same parameters. In TAR, we have full knowledge of our receivers, so we regress our coefficients accordingly. Since TAR also controls the beacon side, the transmission power of each beacon is known to TAR. Notice that the BLE RSS reading is inherently erroneous, so we apply the RSS noise filtering strategy similar to~\cite{chen2017locating} for the original signal and then calculate the current $rs$ with the above formula. TAR takes the time series of BLE proximity as the BLE trace for each device and its owner. We assume each customer has one device with TAR installed, while the case that one user carries multiple devices or other people's device is left for the future work. 

\begin{figure}
\centering\includegraphics[ scale=0.45]{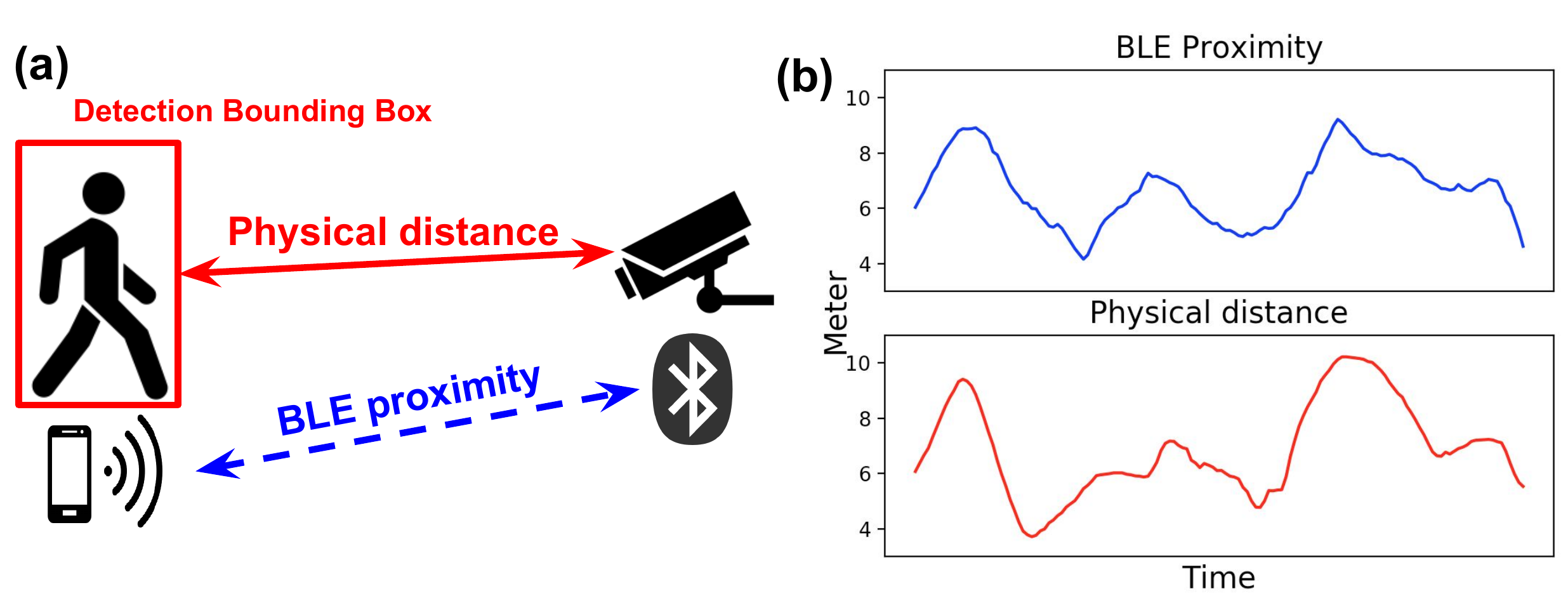}
\caption{\emph{\small Relationship between BLE proximity and physical distance}}
\label{fig:tar_ble_dist}\vspace{-0.2in}
\end{figure}

\begin{figure*}
   \begin{minipage}{0.24\linewidth}
    \centerline{\includegraphics[scale=0.26]{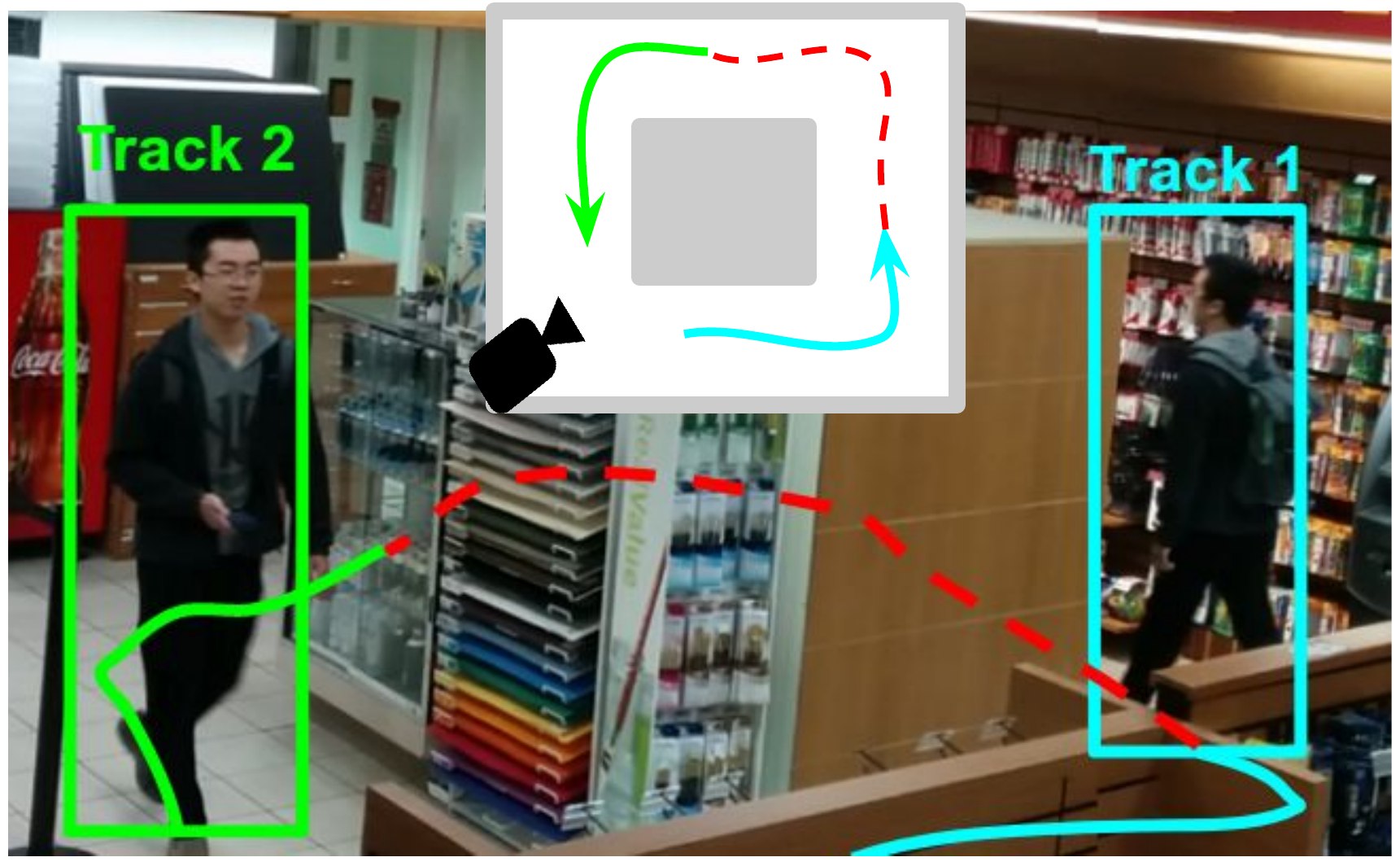}}
    \centerline{(a)}
  \end{minipage}
  \begin{minipage}{0.25\linewidth}
    \centerline{\includegraphics[scale=0.3]{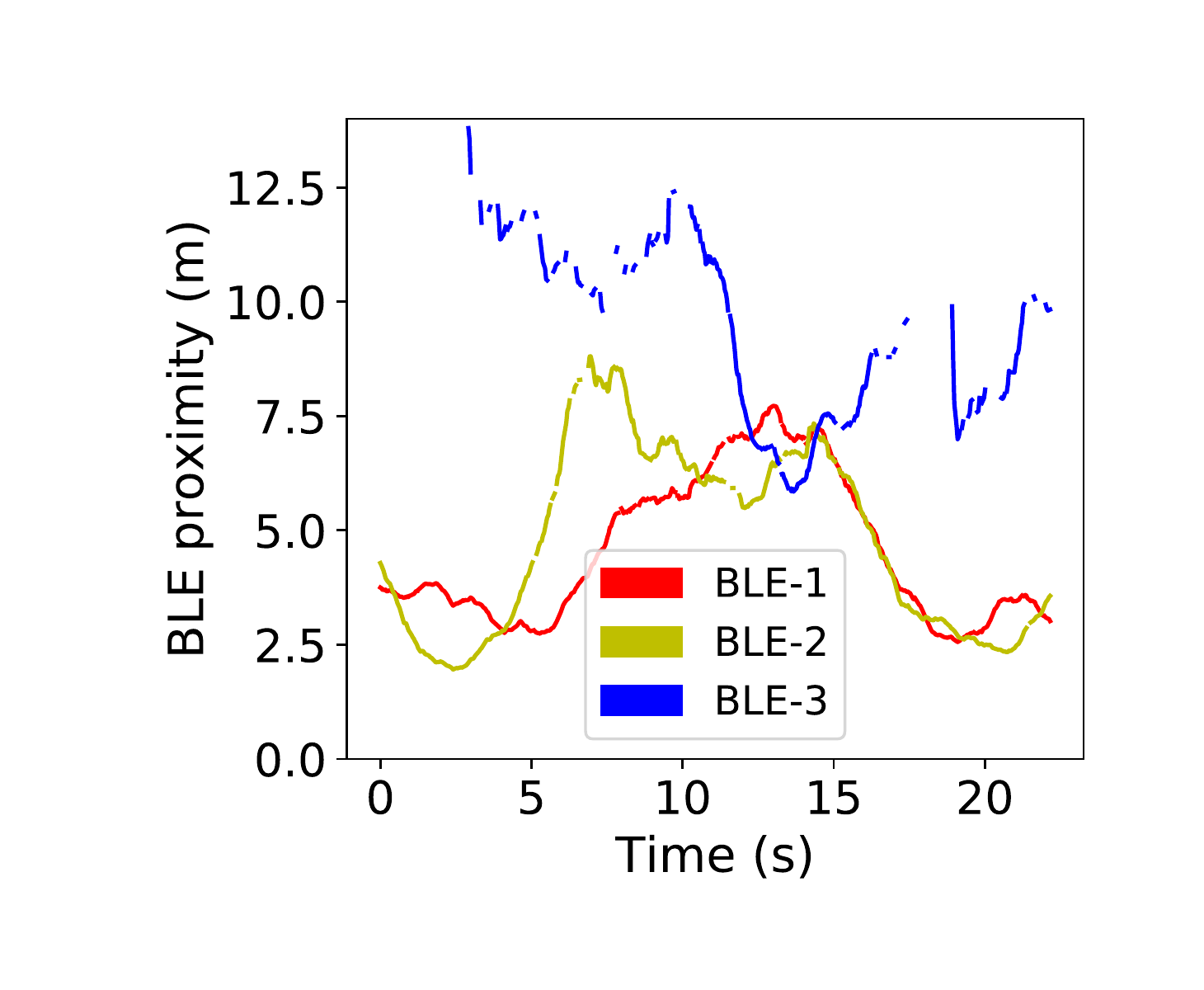}}
    \centerline{(b)}
  \end{minipage}
    \begin{minipage}{0.25\linewidth}
    \centerline{\includegraphics[scale=0.38]{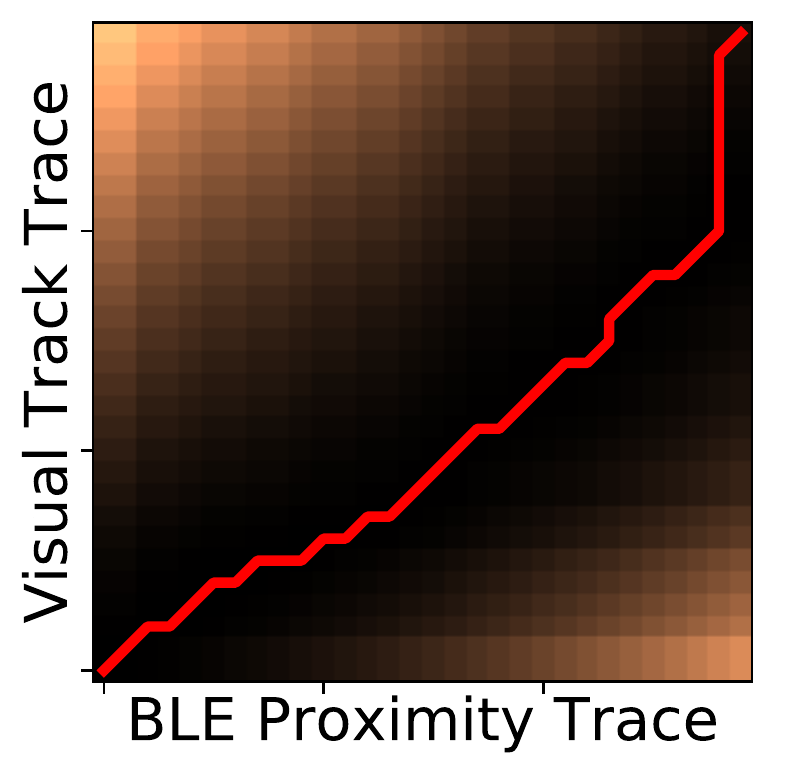}}
    \centerline{(c)}
  \end{minipage}
  \begin{minipage}{0.24\linewidth}
    \centerline{\includegraphics[scale=0.26]{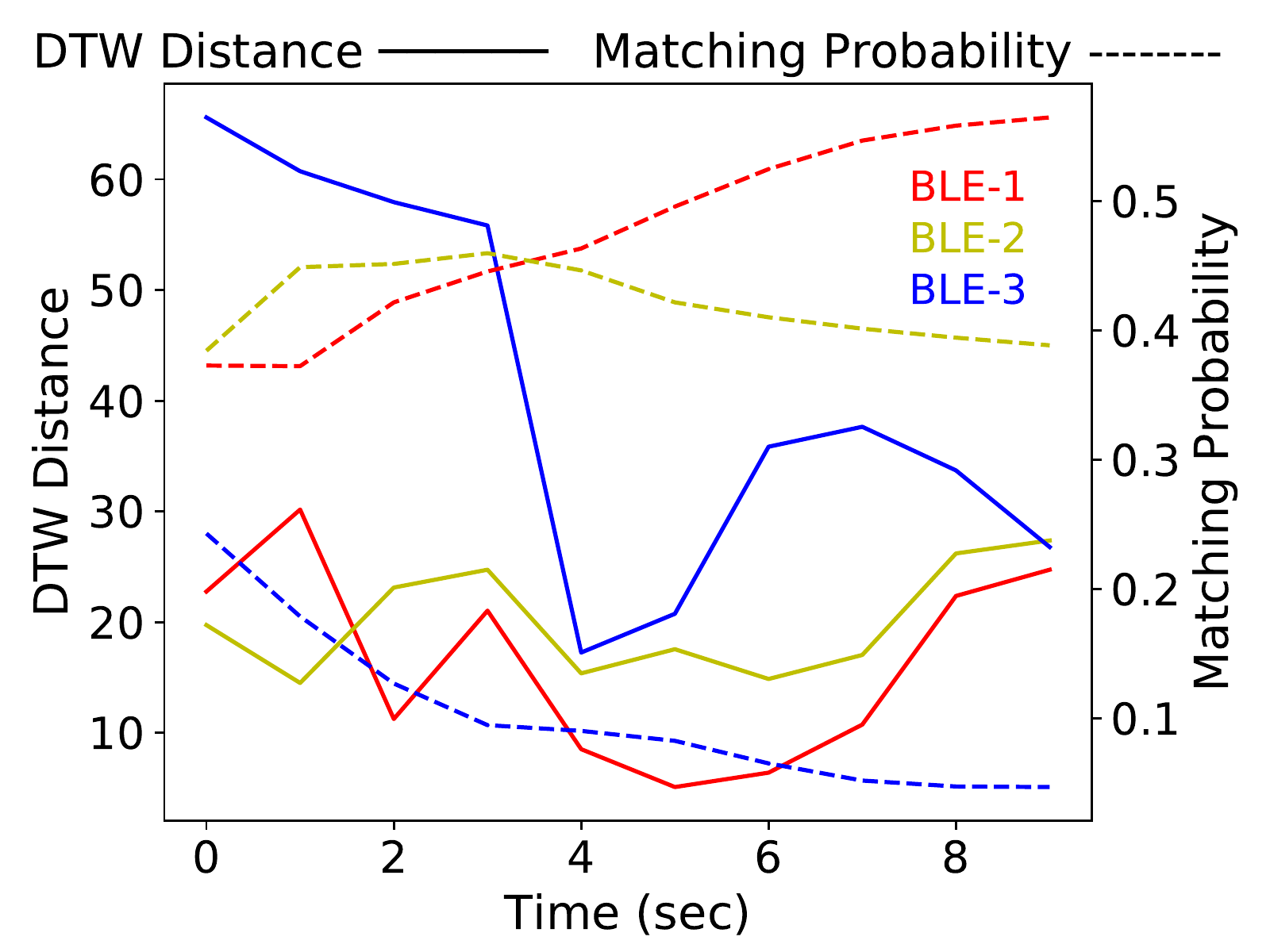}}
    \centerline{(d)}
  \end{minipage}
  \caption{\small \emph{\textbf{(a)} Example of a visual trace; \textbf{(b)} Sensed BLE proximity traces; 
  \textbf{(c)} DTW cost matrix for successful matching; \added{\textbf{(d)} Matching Process Illustration.}}}
  \label{fig:tar_match}
\end{figure*}

\subsection{Real-time Identity Matching}
\label{sec:match}

The key to learning the user's interest and pushing ads is accurate user tracking and identification. By tracking the customer, we know where she visits and what she's interested in. By identifying the user, we know who she is and whom to send the promotion. In practice, it is unnecessary to  know the user's real identity. Instead, recognizing the smart devices carried by users achieves the same goal. We find the BLE universally unique ID (UUID) can serve as the identifier for the device. If we associate the BLE UUID to the visually tracked user, we will successfully identify her and learn her specific interest by looking back at her trajectories. On the other hand, we notice that for a particular user, her BLE proximity trace usually correlates with her physical movement trajectory and her visual movement. ~\tarfig{ble_dist} shows the example traces of a customer and the illustration of the BLE proximity and the physical distance. Therefore, TAR aims to associate the user's visually tracked trace to the sensed BLE proximity trace. Inspired by the observation above, We propose a similarity-based association algorithm with movement pattern matching for TAR. 

\subsubsection{Stepwise Trace Matching}
\label{sec:timewin}
In the matching step, we first need to decide how the traces should be matched. We notice that the BLE proximity traces are usually continuous, but the visual tracks could easily break, especially in occlusion. With this observation, we use visual tracking trace to match BLE proximity traces. The BLE trace continuity, on the other hand, can help correct the real-time visual tracking. To match the time series data, we devised our algorithm based on Dynamic Time Warping (DTW). DTW matches each sample in one pattern to another using dynamic programming. If the two patterns change together, their matched sample points will have a shorter distance, and vice versa. Therefore, shorter DTW distance means higher similarity between two traces. Based on the DTW distance, we define \emph{confidence score} to quantify the similarity. Mathematically, assume $dt_{ij}$ is the DTW distance between visual track $v_i$ and BLE proximity trace $b_j$, the \emph{confidence score} is $f_{ij} = exp(\frac{-dt_{ij}}{100})$. There are some other ways to compare the trace similarity such as Euclidean distance, cosine distance, etc.

DTW is a category of algorithms for aligning and measuring the similarity between two time series. There are three challenges to apply DTW to synchronize the BLE proximity and visual traces. First, DTW normally processes the two traces offline. However, both traces are extending continuously in real-time in TAR. Second, DTW relies on computing a two-dimensional warping cost matrix which has size increasing quadratically with the number of samples. Considering the BLE data's high frequency and nearly 10 FPS video processing speed, the computation overhead can increase dramatically over time. Finally, DTW calculates the path with absolute values in two sequences, but the physical movement estimated from the BLE proximity and the vision-based tracking trace are inconsistent and inherently erroneous. Computing DTW directly on their absolute values will cause adverse effects in matching. 

First, to deal with the negative effect of absolute value input, TAR adopts the \emph{ data differential strategy} similar to \cite{chen2017locating, shu2015last}. We filter out the high-frequency points in the trace and calculate the differential of current data point by subtracting the prior with time divided. Through this operation, either data sequence is independent of the absolute value and can be compared directly. 

Second, a straightforward way to reduce the computation overhead is to minimize the input data size. TAR follows this path and designs a \emph{moving window algorithm} to prepare the input for DTW. More concretely, we set a sliding window of three seconds and update the windowed data every second. \added{We choose this window size for the balance between latency and accuracy. If the window is too short, the BLE trace and visual trace will be too short to be correctly matched. If the window is too long, we may miss some short tracks.} As the window moves, TAR performs the matching process in real-time, thus solves the DTW offline issue. The window triggers the computation once the current time window is updated. Although we get the confidence score with window basis,  connecting the matching windows for a specific visual track remains an issue. For example, a visual track $v_i$ has a higher confidence to match BLE ID-1 at window 1, but BLE ID-2 at window 2. To deal with this, TAR uses \emph{cumulative confidence} score to connect the windows for the visual track. TAR accumulates the confidence scores for consecutive windows of a visual trace and uses the cumulated the confidence score as the current confidence score for ID assignment. 

We use ~\tarfig{match} as one example to demonstrate the algorithm. In this case, a customer's moving trace is shown on the top of ~\tarfig{match}(a). Due to the aisle occlusion and pose change, our vision tracking algorithm obtains two visual tracks for him. ~\tarfig{match}(b) shows the sensed BLE proximity during this period. TAR tries to match the visual tracked trace to one of those BLE proximity traces. ~\tarfig{match}(c) shows the calculation process of DTW for a visual track and a BLE proximity trace, where the path goes almost diagonal. To illustrate our confidence score calculation process, we show the computation process for this example in ~\tarfig{match}(d). The x-axis shows the time, left y-axis shows the DTW score (solid lines) for each moving window, while the right y-axis shows the cumulative confidence score (dotted lines). We can see that BLE trace 2 has better confidence score at the beginning, but falls behind the correct BLE trace 1 after four seconds.

\begin{table}
\begin{center} 
\begin{tabular}{ |c|c|c|c|c| } 
\hline
    ID & 1 & 2 & ... & n \\ 
\hline
    $Track-{a}$ & $p_{a1}$ & $p_{a2}$ & ... & $p_{an}$\\ 
\hline
    $Track-{b}$ & $p_{b1}$ & $p_{b2}$ & ... & $p_{bn}$\\ 
\hline
    $Track-{c}$ & $p_{c1}$ & $p_{c2}$ & ... & $p_{cn}$\\ 
\hline
\end{tabular}
\end{center}
\caption{\small \emph{ID-matching matrix}}
\label{table:id_mat}
\end{table}

\subsubsection{Identity Assignment}
\label{sec:idassign}

To identify the user, TAR needs to match the BLE proximity trace to the correct visual track. Ideally, for each trace, the best cumulative confidence score decides the correct matching. However, there are two problems. First, as stated earlier, BLE proximity estimation is not accurate enough to differentiate some users. In practice, we sometimes see two BLE proximity traces are too similar to assign them to one user confidently. Second, visual tracks break easily in challenging scenarios, which often results in short tracks.  For example, the visual track of the user in ~\tarfig{match}(a) breaks in the middle, leading to two separate track traces. Although the deep feature similarity can help in some scenarios, it fails when the view angle or body pose changes. As TAR intends to learn the user interest, there needs a way to connect these intermittent visual track traces.

\parab{ID Assignment.} 
To tackle the first challenge, TAR proposes a global ID assignment algorithm based on linear assignment~\cite{jonker1987shortest}. TAR computes the confidence score for every track-BLE pair. At any time for one camera, all the visible tracks and their candidate BLE IDs will form a matrix called ID-matching matrix, where row $i$ stands for track $i$ and column $j$ is for BLE ID $j$. The element $(i,j)$ of the matrix is $Prob(BLE_{ij})$. Table~\ref{table:id_mat} shows the matrix structure. Note that each candidate ID only belongs to some of the tracks, so its matching probability is zero with other tracks. 

When the matrix is ready, TAR will assign one BLE ID for the track in each row. The goal of the assignment is to maximize the total sum of confidence score. We use Hungarian algorithm~\cite{kuhn1955hungarian} to solve the assignment problem in polynomial time. The assigned ID will be treated as the track's identity in the current time slot. As visual tracks and BLE proximity traces change with the time window,  TAR will update the assignment with updated matrix accordingly. If a track is not updated in the current window, it will be temporarily removed from the matrix as well as its candidate. When a track stops updating for a long time (> 20 sec), the system will treat the track as "terminated" and archives the last BLE ID assigned to the track.

\subsubsection{Visual Track Sewing}
\label{sec:connection}

The identity matching process is still insufficient for identity tracking in practice: the vision-based tracking technique is so vulnerable that one person's vision track may break into multiple segments. For example, upon a long period of occlusion, one person's trajectory in the camera may be split into several short tracks (see~\tarfig{match}(a)). Another case is that the customer may appear for a very short time in the camera (enters the view and quickly leaves). These short traces make the ID assignment result ambiguous as the physical distance pattern can be similar to many BLE proximity traces in that short period. 

TAR proposes a two-way strategy to handle this. First, TAR tries to recognize the ``ambiguous'' visual track in real-time. In our design, a track will be considered as ``ambiguous''  when it meets either of the two rules: 1) its duration has not reached three seconds; 2) its confidence score distinction among candidate BLE IDs is vague. Explicitly, the two candidates are considered similar when the rank 2's score is more than $\geq$ 80\% of the rank 1's. 

When there is an ambiguous track in assignment, TAR will first consider if the track belongs to an inactive track due to the occlusion. To verify this, TAR will search the inactive local tracks (not matched in the current window but is active within 20 seconds) and check if their assigned IDs are also top-ranked candidate IDs of the ambiguous track. If TAR cannot find such inactive tracks, that means the current track has no connection with previous tracks so the current one will be treated as a regular track to be identified with ID assignment process. 

When a qualified inactive track is found, TAR will check if the two tracks have spatial conflict. The spatial conflict means the two temporally-neighbored segments locate far from each other. For example, with the same assigned BLE ID, one track $v_1$ ends at position $P_1$ and the next track $v_2$ starts at position $P_2$. Suppose the gap time between two tracks is $t$, and the average moving speed of $T_1$ is $v$. In TAR, $T_1$ and $T_2$ will have a spacial conflict if $|P_1 - P_2| > 5v*t$. The intuition behind is that a person cannot travel too fast from one place to another. 

With the conflict check finished, TAR connects the inactive track with the ambiguous track. The trace during the gap time between two tracks is automatically fulfilled with the average speed. The system assumes that the people moves from the first track's endpoint to the second track's starting point with constant velocity during the occlusion. Then the combined track will replace the ambiguous track in the assignment matrix. After linear assignment, TAR will check if the combined track receives the same ID that is previously assigned to the inactive track. If yes, this means the track combination is successful and the ambiguous track is the extension of the inactive track. Otherwise, TAR will try to combine the ambiguous track with other qualified inactive tracks until successful ID assignment. If no combination wins, the ambiguous track will be treated as a regular track for the ID assignment process.

\subsubsection{Multi-camera Calibration}
\label{sec:crosscam}

In the discussion above, one problem with the matching process for the single camera is that the confidence score could be inaccurate when the tracks are short. This is due to limited amount of visual track data and the big size of candidate BLE IDs. For each visual track, we should try to minimize the number of its candidate BLE IDs. It is necessary because more candidates not only increase the processing time but also decrease the ID assignment accuracy. Therefore, TAR proposes \emph{Cross-camera ID Selection (CIS)} to prepare the list of valid BLE IDs for each camera. 

The task of CIS is to determine which BLE ID is currently visible in each camera. First, we observe that 15 meter is usually the maximum distance from the camera to a detectable device. Therefore TAR will ignore beacons with BLE proximity larger than 15 meters. However, the 15-meter scope can still cover more than 20 IDs in real scenarios. The reason is that the BLE receiver senses devices in all directions while the camera has fixed view angle. Therefore, some non-line-of-sight beacon IDs can pass the proximity filter. For example, two cameras are mounted on the two sides of a shelf (which is common in real shops). They will sense very similar BLE proximity to nearby customers while a customer can only show up in one of them. 

To solve the problem, TAR leverages the positions of the camera and the shop's floorplan to abstract the camera connectivity into an undirected graph. In the graph, a vertex represents a camera, and an edge means customers can travel from one camera to another. \added{~\tarfig{mct}(a) shows a sample topology where there are four cameras covering all possible paths within the area}. A customer ID must be sensed hop-by-hop. With this knowledge, TAR filters ID candidates with the following rules: 1) At any time, the same person's track cannot show up in different cameras if the cameras do not have the overlapping view. In this case, if an ID is already associated with a track in one camera with high confidence, it cannot be used as a candidate in other cameras (\tarfig{mct}(b)). 2) A customer's graph trajectory cannot "skip" node. For example, an unknown customer sensed by cam-2 must have shown up in cam-1 or cam-3, because cam-2 locates in the middle of the path from cam-1 to cam-3, and there's no other available path (\tarfig{mct}(c)). 3) The travel time between two neighbor cameras cannot be too short. We set the lower bound of travel time as 1 second (\tarfig{mct}(d)). 

CIS runs as a separate thread on the TAR server. In every moving window, it collects all cameras' BLE proximity traces and visual tracks. CIS checks each BLE ID in the camera's candidate list and removes the ID if it violates any of the rules above. The filtered ID list will be sent back to each camera module for ID assignment.

\begin{figure}
\centering
\includegraphics[scale=0.5]{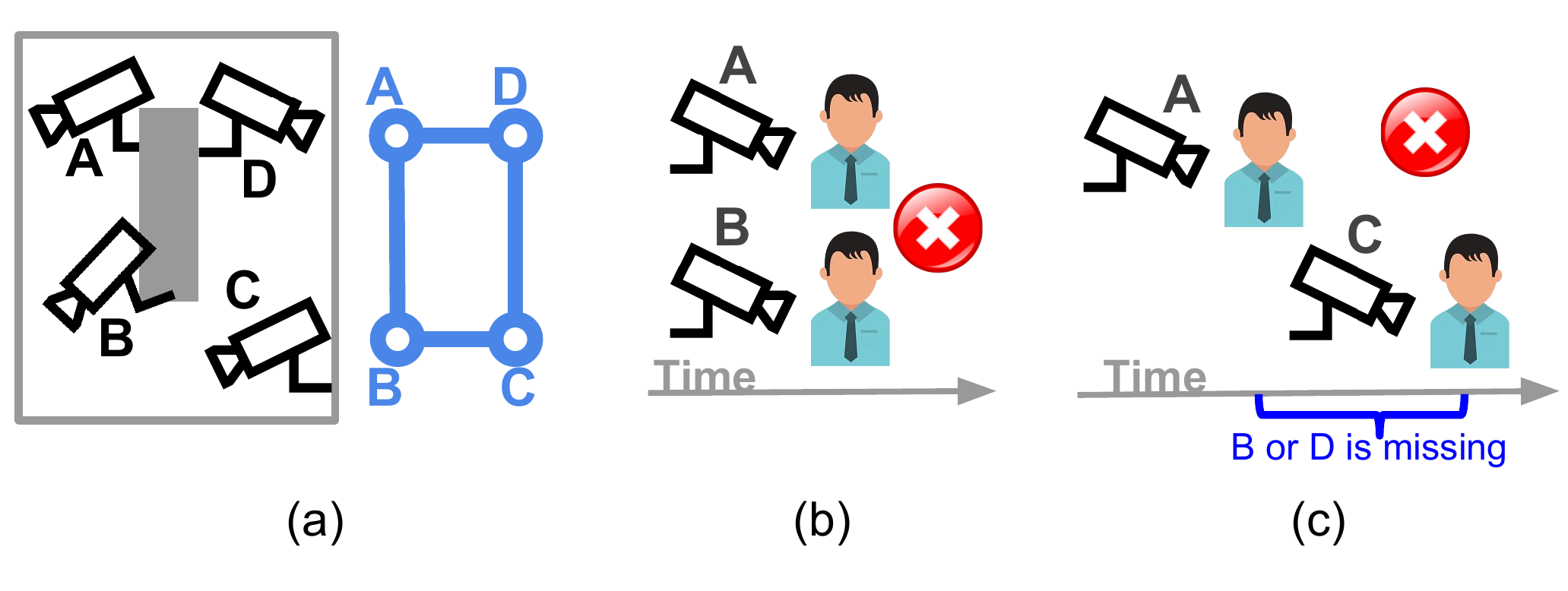}
\caption{\emph{\small \textbf{(a)} Camera Topology; \textbf{(b)} One ID cannot show in two cameras; \textbf{(c)} BLE ID must be sensed sequentially in the network; \textbf{(d)} It takes time to travel between cameras}}
\label{fig:tar_mct}
\end{figure}

%% file: tex/tar/eval.tex
\section{Evaluation}

In this section, we describe how TAR works in the real scenario and evaluate each of its components. 

\subsection{Methodology and Metrics}

\parab{TAR Implementation.}
Our implementation of TAR contains three parts: the BLE broadcasting and sensing, the live video detection, and tracking and the identity matching. 

BLE broadcasting is designed as a mobile library supporting both iOS and Android. TAR implements the broadcasting library with the \textit{CoreBluetooth} framework on iOS~\cite{corebt} and AltBeacon~\cite{altBeacon} on Android. The BLE RSS sensing module sits in the backend. In our experiments, we use Nexus 6 and iPhone 6 for BLE signal receiving. The Bluetooth data is pushed from the devices to the server through TCP socket. In TAR, we set both the broadcasting and sensing frequency at 10 Hz. 

The visual tracking module (VT) consists of a DNN-based people detector and a DNN feature extractor. One VT processes the video from one camera. TAR uses the Tensorflow version of Faster-RCNN~\cite{chen17implementation,ren2015faster} as people detector and our modified GoogleNet~\cite{caffereid} as the deep feature extractor. We train the Faster-RCNN model with VOC image dataset~\cite{pascal-voc-2012} and train the GoogleNet with two pedestrian datasets: Market-1501~\cite{zheng2015scalable} and DukeMTMC~\cite{zheng2017unlabeled}. The detector returns a list of bounding boxes (bbox), which are fed to the feature extractor. The extractor outputs 512-dim feature vector for each bbox. \added{We choose FastDTW~\cite{salvador2007toward} for DTW algorithm and its code can be downloaded from~\cite{fastdtw}.}

Since each VT needs to run two DNNs simultaneously, we cannot support multiple VTs on single GPU. To ensure performance, we dedicate one GPU for each VT instance in TAR, while leaving further scalability optimization to the future works. The tracking algorithm and identity matching algorithm is implemented with Python and C++. To ensure real-time process, all modules run in parallel through multi-threading. 

Our server equips with Intel Xeon E5-2610 v2 CPU and Nvidia Titan Xp GPU. In the runtime, TAR occupies 5.3GB of GPU memory and processes video at around 11FPS. Double VT instances on one GPU will not overflow the memory but will reduce the FPS by around half.

As cross-camera tracking and identification require collaboration among different cameras, TAR shares the data by having one machine as the master server and running Redis cache. Then each VT machine can access the cache to upload its local BLE proximity and tracking data. The server runs cross-camera ID selection with the cached data and writes filtred ID list to each VT's Redis space. 

 \begin{figure}
 \centering
 \includegraphics[scale=0.35]{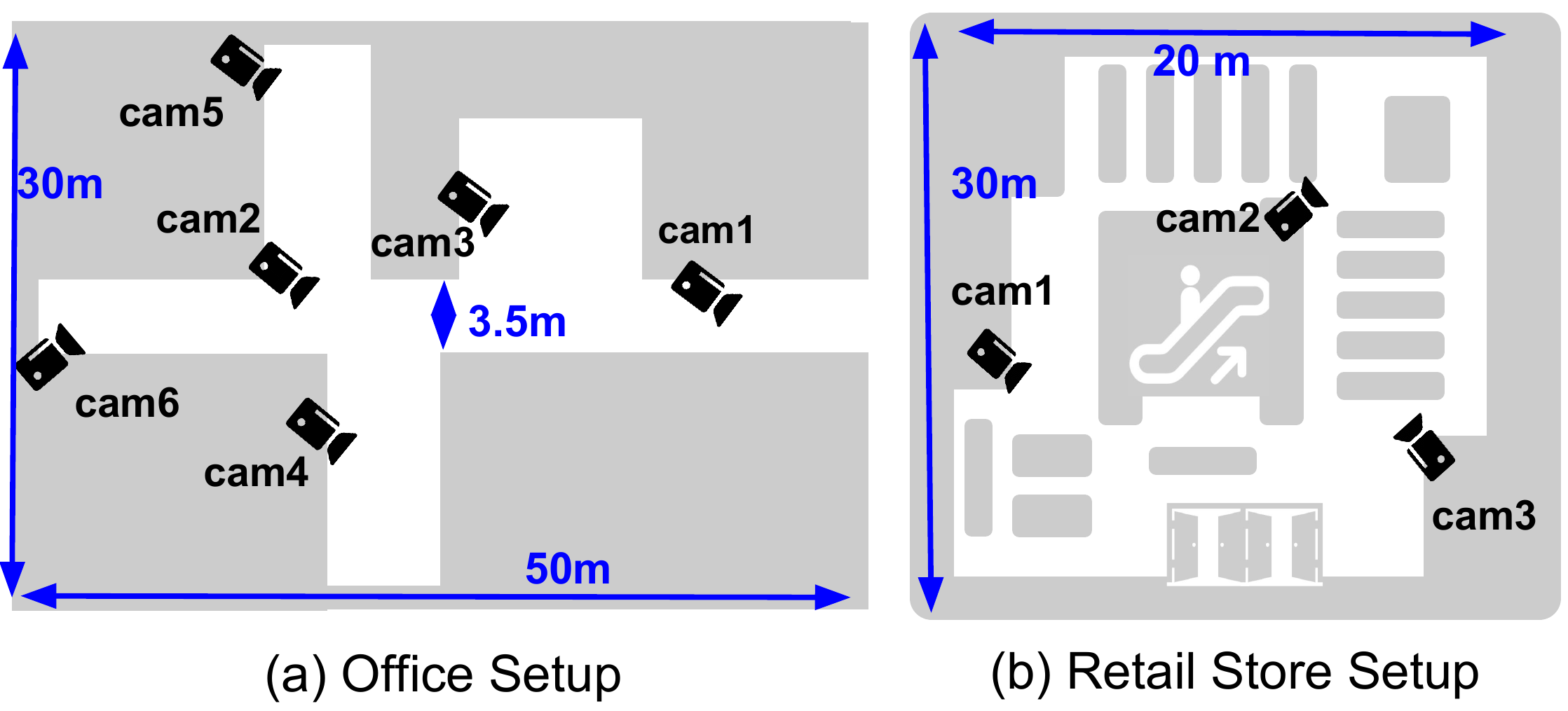}
 \vspace{-1mm}
 \caption{\emph{\small Experiment Deployment Layout: \textbf{(a)} Office; \textbf{(b)} Retail store.}}
 \vspace{-6mm}
 \label{fig:tar_explayout}
\end{figure}

 \begin{figure}
 \centering
 \includegraphics[scale=0.55]{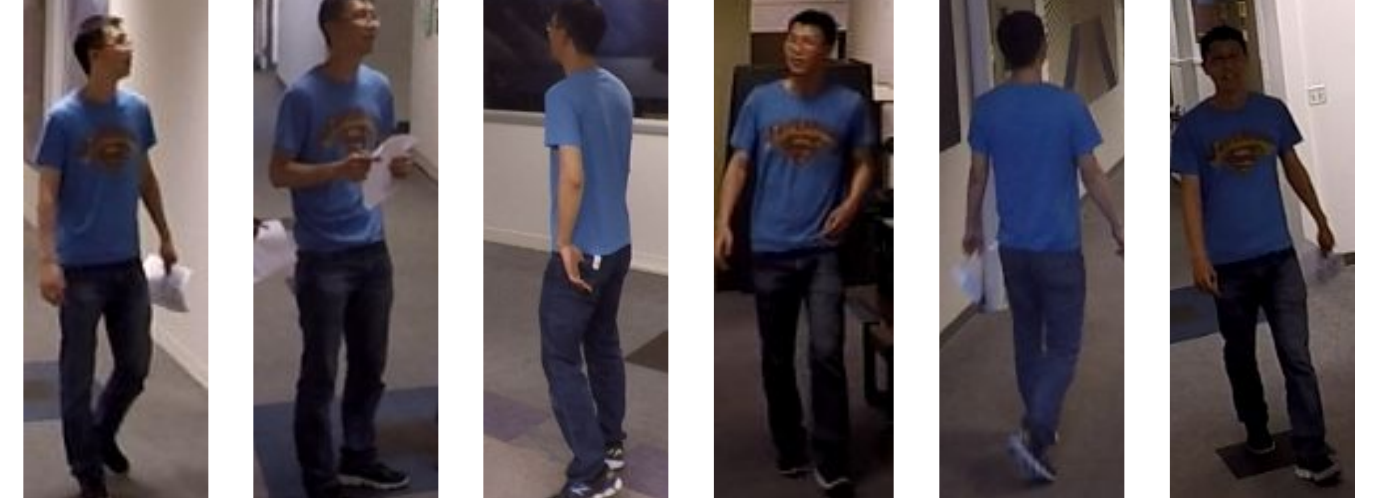}
 \caption{\added{\emph{\small Same person's figures under different camera views (office).}}}
 \vspace{-6mm}
 \label{fig:tar_people_snapshot}
 \vspace{3mm}
\end{figure}

\parab{TAR Experiment Setup.}
We evaluate TAR's performance by deploying the system in two different environments: \added{an office building (\tarfig{explayout}(a)) and a retail store (\tarfig{explayout}(b)).} We use Reolink IP camera (RLC-410S) in our setup. The test area for the office deployment is $50m \times 30m$ with the average path width of 3.5m, while the retail store is $20m \times 30m$. 

We deploy six cameras in the office building as shown in the layout, and three cameras in the retail store. \added{All the cameras are mounted at about 3m height, pointing \ang{20}-\ang{30} down to the sidewalk.} There are 20 different participants involved in the test, 12 in office deployment and 8 in retail store deployment. \added{Besides the recruited volunteers, TAR also records other pedestrians and it captures up to 27 people simultaneously in the cameras.} Each participant has TAR installed in their devices and walks around randomly based on their interest. To quantify the TAR performance, we record all the trace data in two deployment scenarios for later comparison. We've collected around 1-hour data for each deployment, including 30GB video data and 10MB BLE RSS logs. \added{\tarfig{people_snapshot} shows the same person's appearance in different cameras. We can see that some snapshots are dark and blurry, which makes it hard to identify people only with vision approach.}

For cross-camera tracking and identification, we mainly use IDF1 Score~\cite{ristani2016performance}, a standard metric to evaluate the performance of multi-camera tracking system. \added{IDF1 is the ratio of correctly identified detections over the average number of ground-truth, which equals (Correctly identified people in all frames) / (All identified people in all frames). For example, if one camera records three people A, B, and C. If an algorithm returns two traces: one on A with ID=A, and another on C with ID=B. In this case, we only have one person correctly detected, so the IDF1=33\%}. 

\subsection{TAR Runtime}
Before discussing our trace-based evaluation, we show the benefits of TAR's matching algorithm and optimization in the runtime.   

We first show TAR's ID assignment process~\footnote{https://vimeo.com/246368580}. In the beginning, with only detection and bbox tracking, we cannot tell the user identity. We consider the user movement estimated from the visual track and the BLE proximity traces and then apply our stepwise matching algorithm. After that, we use our ID assignment algorithm to report the user's possible identity. Although the user's identity is not correct at first, the real identity emerges as time window moves. This proves the effectiveness of our identity matching algorithm. 

We also demonstrate how TAR's track sewing works in runtime\footnote{https://vimeo.com/246388147}. As the first part of the video shows, in the case of broken visual tracks, the user may not get correctly identified after the break. By applying our track sewing algorithm, the user's tracks get correctly recognized much faster. Therefore, TAR's track sewing algorithm benefits those scenarios.

\begin{figure}
\centering\includegraphics[scale=0.6]{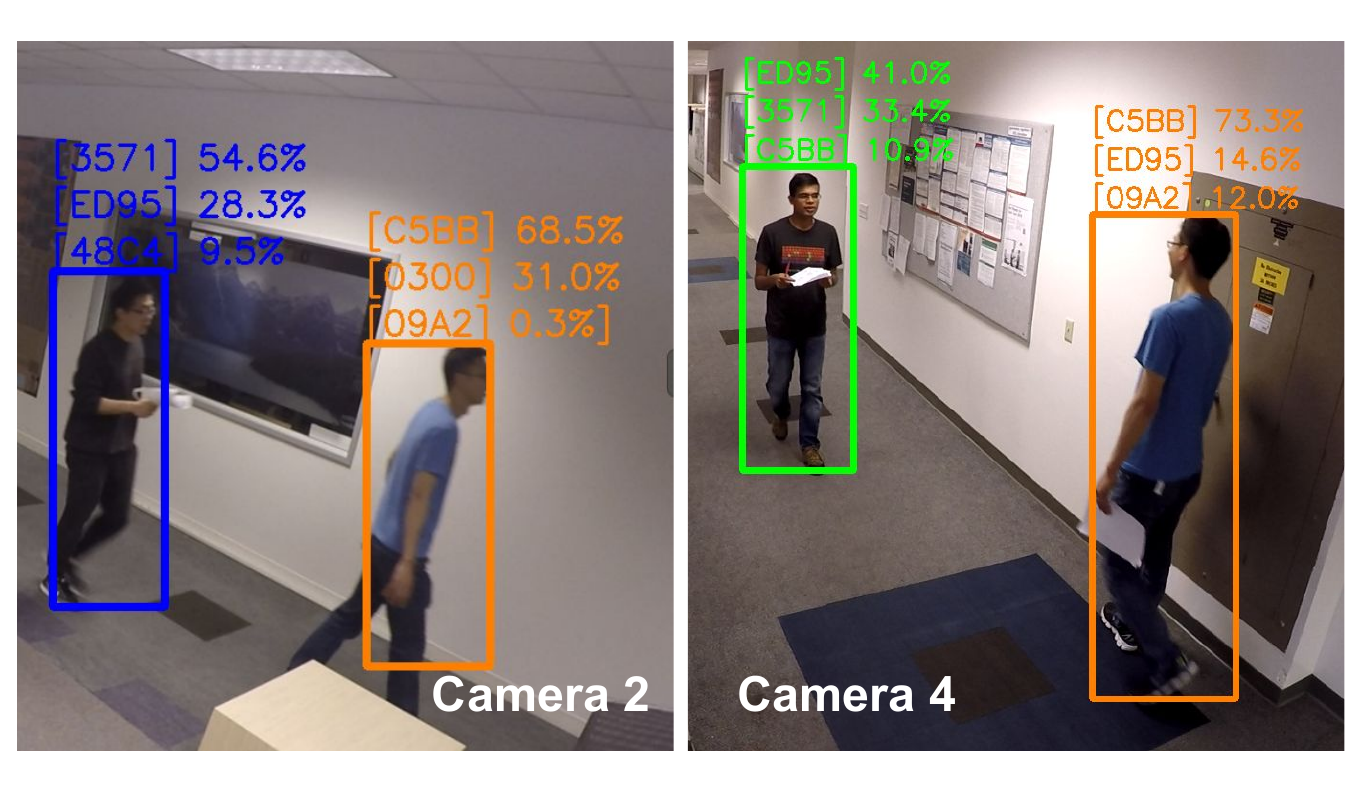}
\caption{\emph{\small Screenshots for Cross Camera Calibration}}
\label{fig:tar_identity}
\end{figure}

\tarfig{identity} shows two cameras' screenshots in the office settings at different time. In this trace, one user (orange bbox) walks from camera 2 to camera 4. Meanwhile, there are around 7 BLE IDs sensed. With the user enters camera 4, TAR uses temporal-spatial relationship and deep feature distance to filter out unqualified BLE IDs, and then assigns the highest-ranked identity to the user. As shown in camera 4's screenshot, the user is correctly identified.

\begin{figure*}
   \begin{minipage}{0.32\linewidth}
    \centerline{\includegraphics[scale=0.34]{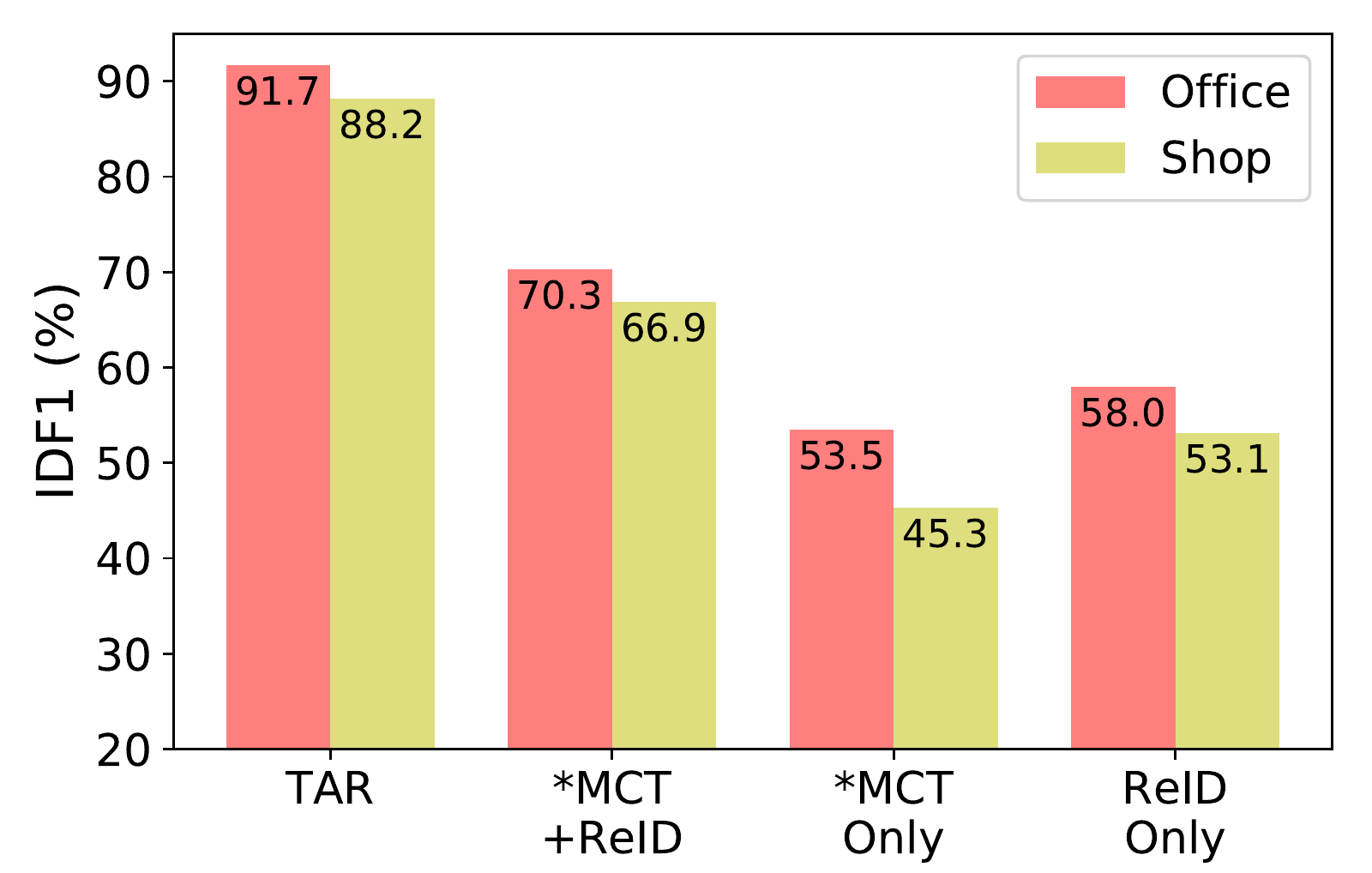}}
    \centerline{(a)}
  \end{minipage}
  \begin{minipage}{0.32\linewidth}
    \centerline{\includegraphics[scale=0.29]{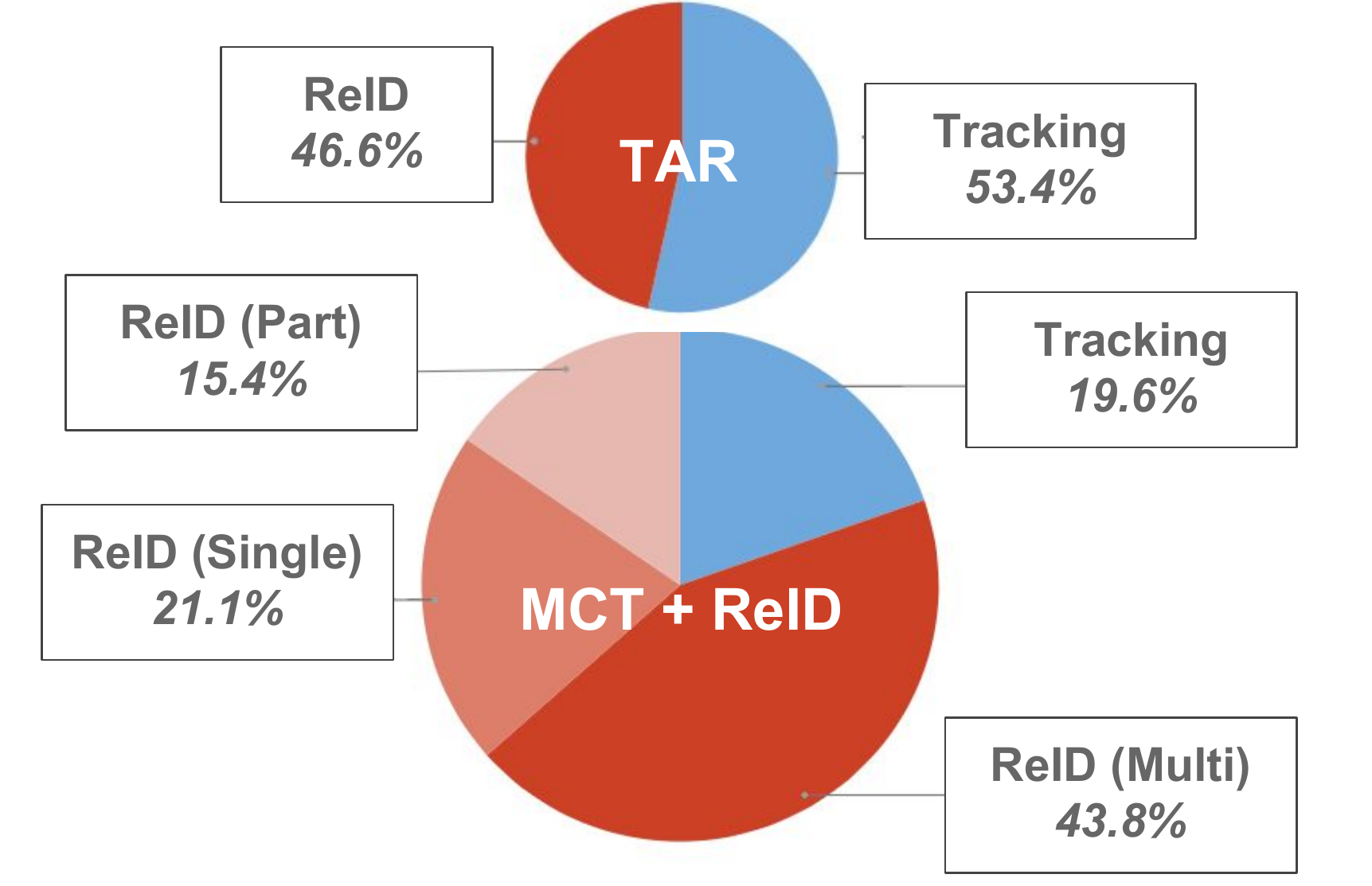}}
    \centerline{(b)}
  \end{minipage}
    \begin{minipage}{0.32\linewidth}
    \centerline{\includegraphics[scale=0.29]{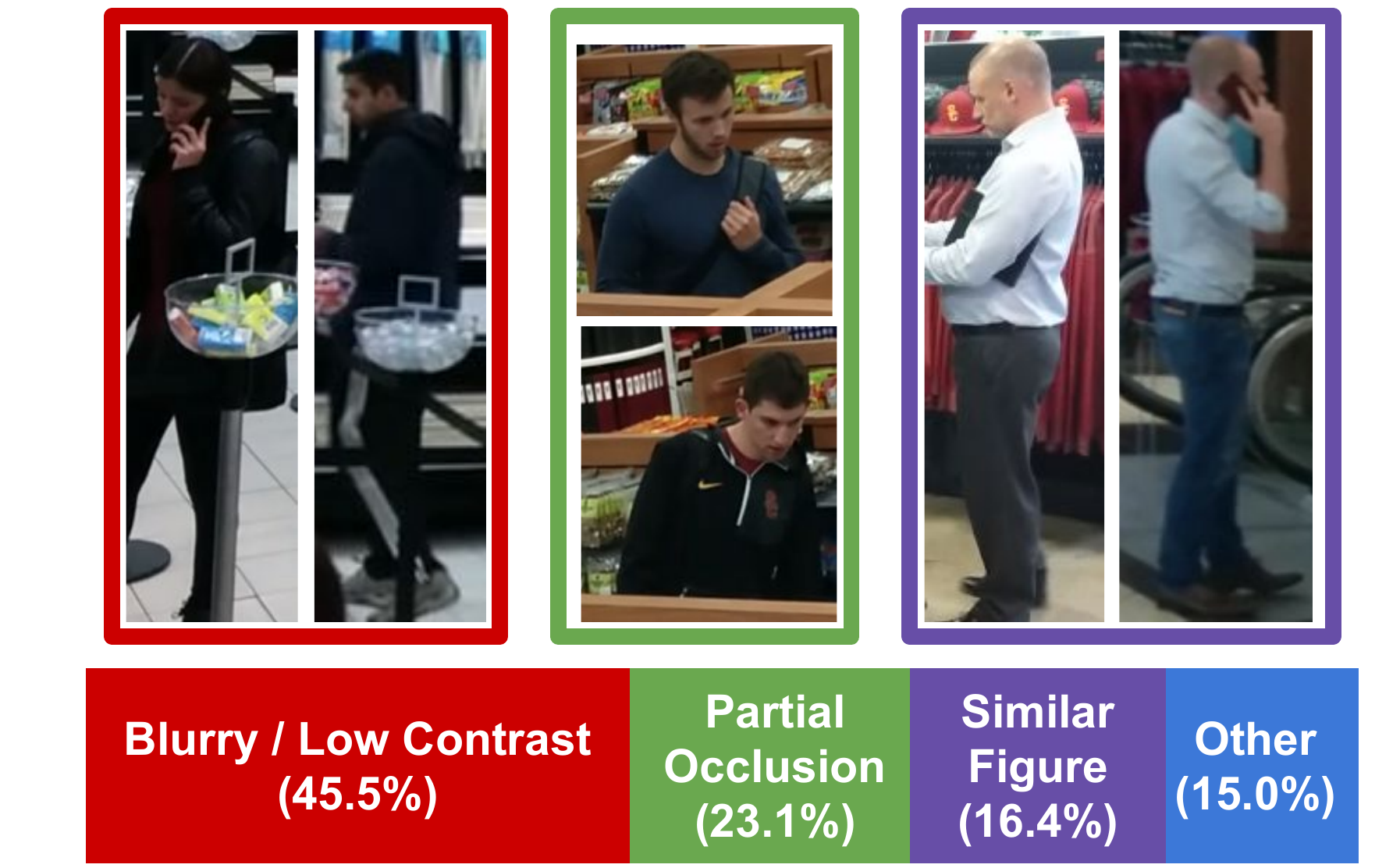}}
    \centerline{(c)}
  \end{minipage}
  \caption{\small \emph{\textbf{(a)} Multi-cam tracking comparison against state-of-the-art solutions (*offline solution); 
  \textbf{(b)} Error statistics of TAR and MCT+ReID; \textbf{(c)} Error statistics of re-identification in MCT+ReID and example images.}}
  \label{fig:tar_eval_combine}
\end{figure*}

\subsection{TAR Performance}

\subsubsection{Comparing with Existing Multi-cam Tracking Strategies}

\added{~\tarfig{eval_combine}(a) shows the accuracy of TAR. The y-axis represents IDF1 accuracy. As a comparison, we also evaluate the IDF1 of existing state-of-the-art algorithms from vision community:}

\added{\emph{(1) MCT+ReID}: We use the work from DeepCC~\cite{ristani2018features}, an open-sourced algorithm that reaches top accuracy in MOT Multi-Camera Tracking Challenge~\cite{mot_mtmc}. The solution uses DNN-generated visual features for people re-identification (ReID) and uses single-camera tracking and cross-camera association algorithms for identity tracking. The single-camera part of DeepCC runs a multi-cut algorithm for detections in recent frames and calculates best traces to minimize the assignment cost. For cross-camera identification, it not only considers visual feature similarity but also estimates the movement trajectory of each person in the camera topology to associate two tracks, which has the similar idea of TAR in cross-camera ID selection.}

\added{\emph{(2) MCT-Only}: We also tested MCMT~\cite{ristani2016performance}, the previous work of DeepCC~\cite{ristani2018features}, which shares similar logic for tracking as DeepCC (both single-camera and multi-camera) but does not have DNN for people re-identification.}

\added{\emph{(3) ReID-Only}: We directly run DeepCC's DNN to extract each people's visual feature in each frame and classify each person to be one of the registered users. This will show the accuracy of tracking with re-identification only.}

\parab{Analysis: } \changed{We can see that TAR outperforms existing best offline algorithm (MCT+ReID) by 20\%. Therefore, we analyze the failures in both TAR and MCT+ReID to understand why TAR gains much higher accuracy. There are two types of failures: erroneous single-camera tracking and wrong re-identification. Note that the re-identification is BLE-vision matching in TAR's case.}

\changed{As~\tarfig{eval_combine}(b) shows, the two failures have the similar contribution in TAR. In the vision-only scenario, most errors are from the re-identification process. We further break down the re-identification failures for MCT+ReID into three types: (1) multi-camera error: a person is constantly recognized as someone else in the cameras after his first appearance; (2) single-camera error: a customer is falsely identified in one camera; (3) part-of-track error: a person is wrongly recognized for part of her track in one camera. From~\tarfig{eval_combine}(b), we can see that more than half of the ReID problems are cross-camera type, which is due to the MCT module that optimizes identity assignment across cameras - if a person is assigned an ID, she will have a higher probability to get the same ID in following traces.} 

\changed{The root cause of the vision-based identification failure is the imperfect visual feature, which cannot accurately distinguish one person from another in some scenarios. From our observation, there are three cases that the feature extractor may easily fail: (1) blurry image; (2) partial occlusion; (3) similar appearance. ~\tarfig{eval_combine}(c) demonstrates each failure case where two persons are recognized as the same customer by TAR. The figure also shows the percentage of all failure cases in the test results. We can see that the blurry and low contrast figures lead to near half of errors and the other two types account for about 40\% of the failed cases.}

\begin{figure}
\centering
\includegraphics[scale=0.42]{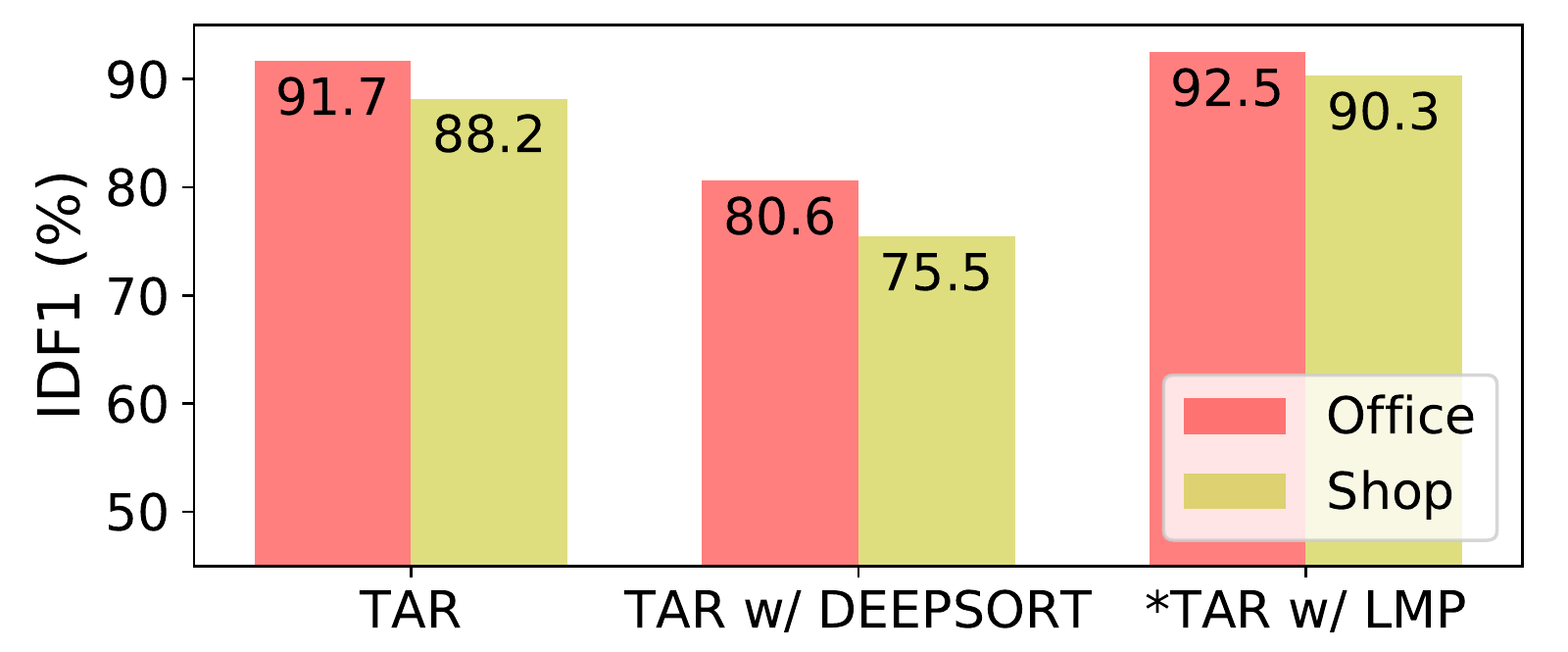}
\caption{\added{\emph{\small Importance of Tracking Components in TAR (*offline solution).}}}
\label{fig:tar_accuracy_vis}
\end{figure}

\begin{figure}
\centering
\includegraphics[scale=0.52]{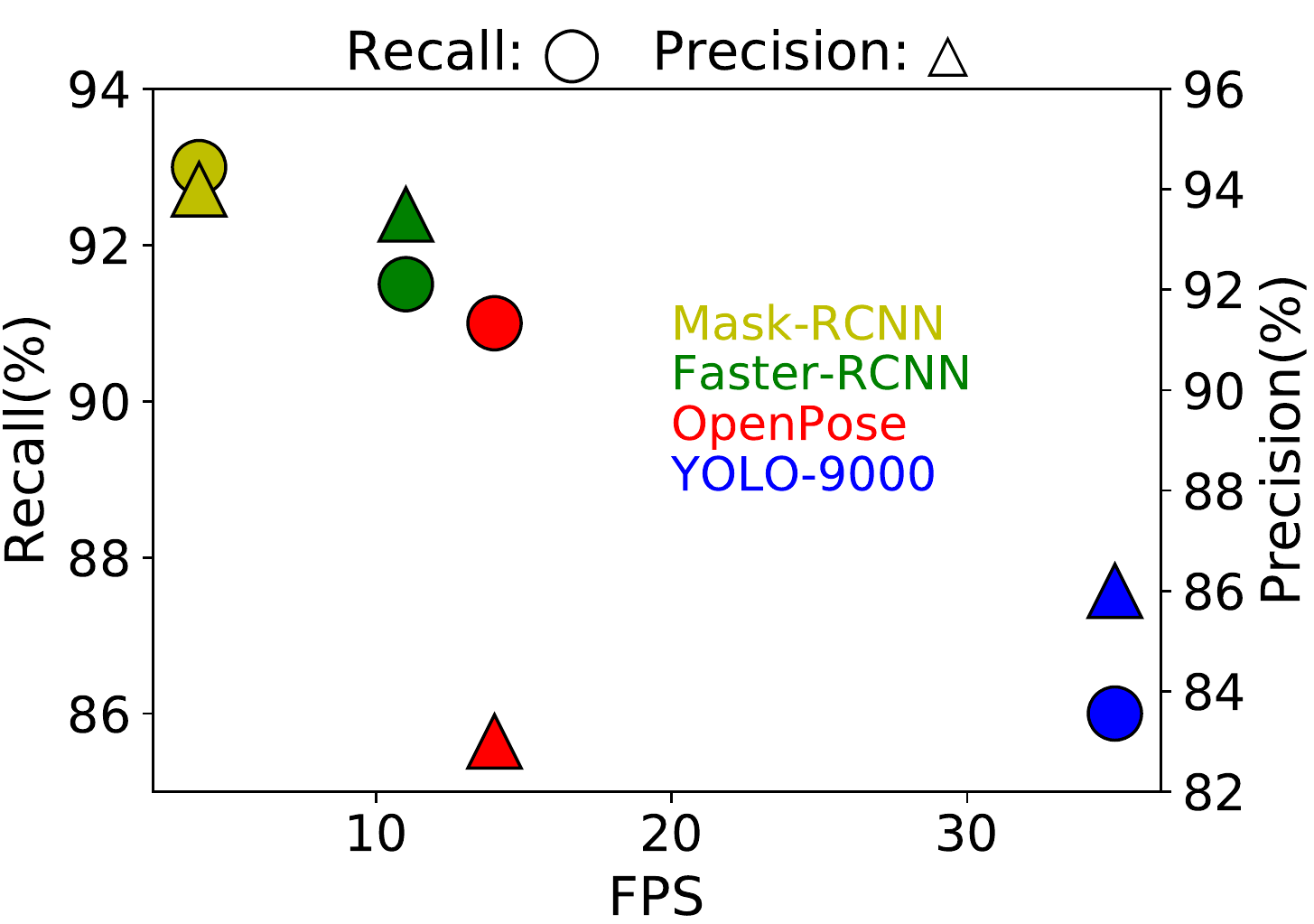}
\caption{\added{\emph{\small Recall, precision, and FPS of state-of-the-art people detectors.}}}
\label{fig:tar_eval_detector}
\end{figure}

\subsubsection{Importance of Different Components in TAR } Next, we analyze each component used in TAR. 

\parab{People Detection.} \added{The people detector may fail in two ways: false positive, which recognizes a non-person object as a people, and false negative, which fails to recognize a real person. For false positives, TAR could filter them out in the vision-BLE matching process. For false negatives, people occluded larger than > 80\% of their bodies usually will be hardly detected by the detection model. Such false negative cases can be handled by TAR's tracking algorithm and track sewing metric, which will also be evaluated. We evaluate the performance of current state-of-the-art open-sourced people detectors using our dataset and the results are shown in~\tarfig{eval_detector}. Besides Faster-RCNN (used by TAR), we also test Mask-RCNN~\cite{he2017mask}, YOLO-9000~\cite{redmon2017yolo9000}, and OpenPose~\cite{cao2017realtime}. We can see that YOLO and OpenPose have lower accuracy although they are fast. In contrast, Mask-RCNN is very accurate but works too slow to meet TAR's requirement.}

\begin{figure*}[t]
\begin{minipage}{0.95\linewidth}
\centering
\includegraphics[scale=0.48]{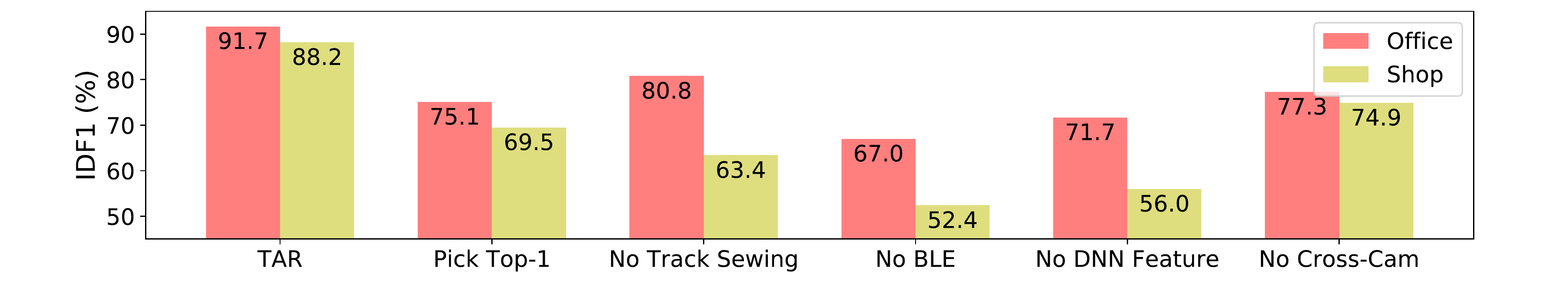}
\end{minipage}
\caption{\emph{\small Importance of Identification Components in TAR}}
\vspace{2mm}
\label{fig:tar_accuracy_multi}
\end{figure*}

\parab{Trace matching.}
\added{
DTW plays the key role in matching BLE traces to vision traces. Therefore, we should understand its effectiveness in TAR's scenario. In the experiment, we compute the similarity between one person's walking trace and all nearby BLE traces to find the one with the highest similarity. The association process succeeds if the ground truth trace is matched, otherwise, it fails. We calculate the number of correct matchings across the whole dataset and compute the successful linking ratio. Besides DTW, we also test other metrics including Euclidean distance, cosine distance, Pearson correlation coefficient~\cite{pearson_distance}, and Spearman's rank correlation~\cite{spearman_distance}. The average matching ratio of each method is shown in Table~\ref{table:trace_matching}, in which DTW gets the best accuracy. 
}

\begin{table}[]
\centering
\begin{tabular}{|l|l|}
\hline
\textit{\textbf{Similarity Metric}} & \textbf{Accuracy (\%)} \\ \hline
\textit{DTW (used in TAR)}          & 95.7                   \\ \hline
\textit{Euclidean Distance}         & 88.0                   \\ \hline
\textit{Cosine Distance}            & 84.9                   \\ \hline
\textit{Pearson Correlation}        & 66.4                   \\ \hline
\textit{Spearman Correlation}       & 72.5                   \\ \hline
\end{tabular}
\vspace{3mm}
\caption{\added{\emph{\small Accuracy (ratio of correct matches) of different trace similarity metrics}}}
\label{table:trace_matching}
\end{table}

\parab{Visual Tracking.} 
\added{Visual tracking is crucial for estimating visual traces. As TAR develops its visual-tracking algorithm based on DeepSORT~\cite{wojke2017simple}, we want to see TAR's performance improvement compared with existing state-of-the-art tracking algorithms. Towards this end, we replace our visual tracking algorithm with DeepSORT and LMP~\cite{tang2017multiple}, which achieves best tracking accuracy in MOT16 challenge. LMP uses DNN for people re-identification like DeepSORT and it works offline so it can leverage posteriori knowledge of people's movement and use lifted multi-cut algorithm to assign traces globally.}

\added{We calculate the IDF1 percentage of each choice in~\tarfig{accuracy_vis}. We can see from the first group of bars that TAR's visual tracking algorithm clearly outperforms DeepSORT by 10\%. This is because TAR's visual tracking algorithm considers several optimizations like kinematic verification, thus reduces ID switches. Moreover, TAR performs similarly with that with LMP as the tracker, which shows that our online tracking metric is comparable to the current state-of-the-art offline solution. LMP is not feasible for TAR since it works offline and slowly (0.5FPS) while our usage scenario needs real-time processing.}

\vspace{3mm}
\emph{We compare the following modules' performance by taking away each of them from TAR and show the system accuracy change in~\tarfig{accuracy_multi}.}

\parab{ID Assignment.}
An alternative solution for our ID assignment algorithm is to always choose the best (top-1) confident candidate for each track. Thus, we compare our ID assignment to the top-1 scheme and show the result in the second group of~\tarfig{accuracy_multi}. We can see that the top-1 scheme is almost 20\% worse than TAR. The reason is that the top-1 assignment usually has the conflict error, where different visual tracks get assigned to the same ID. TAR, on the other hand, ensures the one-on-one matching, which reduces such conflicts.

\parab{Track Sewing.}
If we remove the track sewing optimization, a person's fragmented tracks will need much longer time to be recognized, and some of them may be matched to wrong BLE IDs.~\tarfig{accuracy_multi}'s third group proves this point. Removing track sewing drops the accuracy for nearly 25\% in the retail store dataset, which has frequent occlusion. \added{In the evaluation, we find the average number of distinct tracks of the same person is 1.8, and the maximum number is 5.}

\parab{BLE Proximity.}
Incorporating BLE proximity is the fundamental part to help track and identify users. To quantify the effectiveness of BLE proximity, we calculate the accuracy with BLE matching components removed and TAR only relies on the cross-camera association and deep visual features to identify and track each user.~\tarfig{accuracy_multi}'s fourth group shows that the accuracy drops by 35\% at most without BLE's help.

\parab{Deep Feature.}
The deep feature is one of the core improvements in the visual tracking algorithm.~\tarfig{accuracy_multi}'s fifth group shows that the accuracy drops nearly 30\% because removing the deep feature will cause high-frequency ID switches in tracking. In this case, it is hard to compensate the errors even with our other optimizations.

\parab{Cross Camera Calibration.}
Our cross-camera calibration metric contains temporal-spatial relationship and deep feature similarity across cameras. To understand the impact of this optimization, we remove the component and evaluate TAR with the same dataset.~\tarfig{accuracy_multi}'s most right group shows a 10\% accuracy drop. Without cross camera calibration, we find that the matching algorithm struggles to differentiate BLE proximity traces. In some cases, these traces demonstrate similar pattern when they move around. For example, in the retail scenario, TAR tries to recognize one user seen in camera-1 and she's leaving the store. Meanwhile, another user is also moving out but with a different direction seen in camera-2. In this case, their BLE proximity traces are hard to distinguish only with camera-1's information.

\subsubsection{Robustness}
Robustness is essential for any surveillance or tracking system because some part of the system might fail, e.g., one or more cameras or BLE receivers stop working. This could happen in many situations due to battery outage, camera damaged, or the lighting condition is bad. Therefore, how will those failures affect the overall performance? We focus on the system accuracy under node failures.~\tarfig{node_failure} shows the performance change of TAR when failure happens. \added{Note that either the BLE failure or the video failure will cause the node failure because TAR needs both information for customer tracking. Therefore, we remove the affected nodes randomly from TAR's network to simulate the runtime failure.~\tarfig{node_failure} shows node failures and performance downgrades with the portion of failed nodes. We can see that TAR can still keep more than 80\% accuracy with half of the nodes down. The system is robust because each healthy node could identify and track the customer by itself. The only loss from the failed node is the cross-camera part, which uses the temporal-spatial relationship to filter out invalid BLE IDs.} 

\added{We also evaluate the relationship between the number of concurrent tracked people and the tracking accuracy (shown in~\tarfig{diff_people}). As the result shows, TAR accuracy drops as more people being tracked. The accuracy becomes stable around 85\% with 20 or more people. This is because that there is no "new" trace pattern since all possible paths in each camera view are fully occupied. Therefore, adding more people will not cause more uncertainty in trace matching.}

\begin{figure}
\centering\includegraphics[scale=0.52]{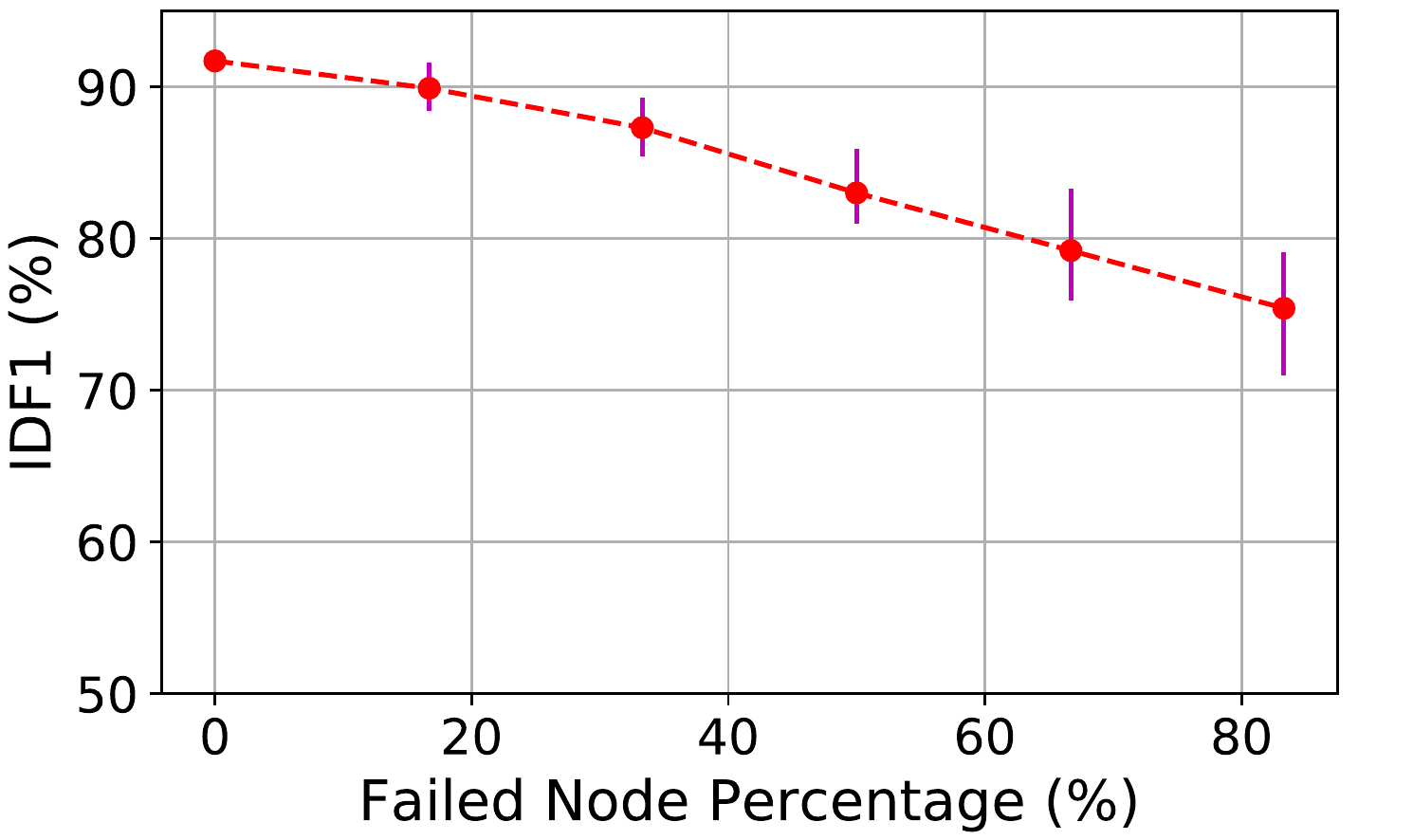}
\caption{\emph{\small Accuracy of TAR with different ratio of node failure (purple lines show the measured error).}}
\label{fig:tar_node_failure}
\end{figure}

\begin{figure}
\centering\includegraphics[scale=0.53]{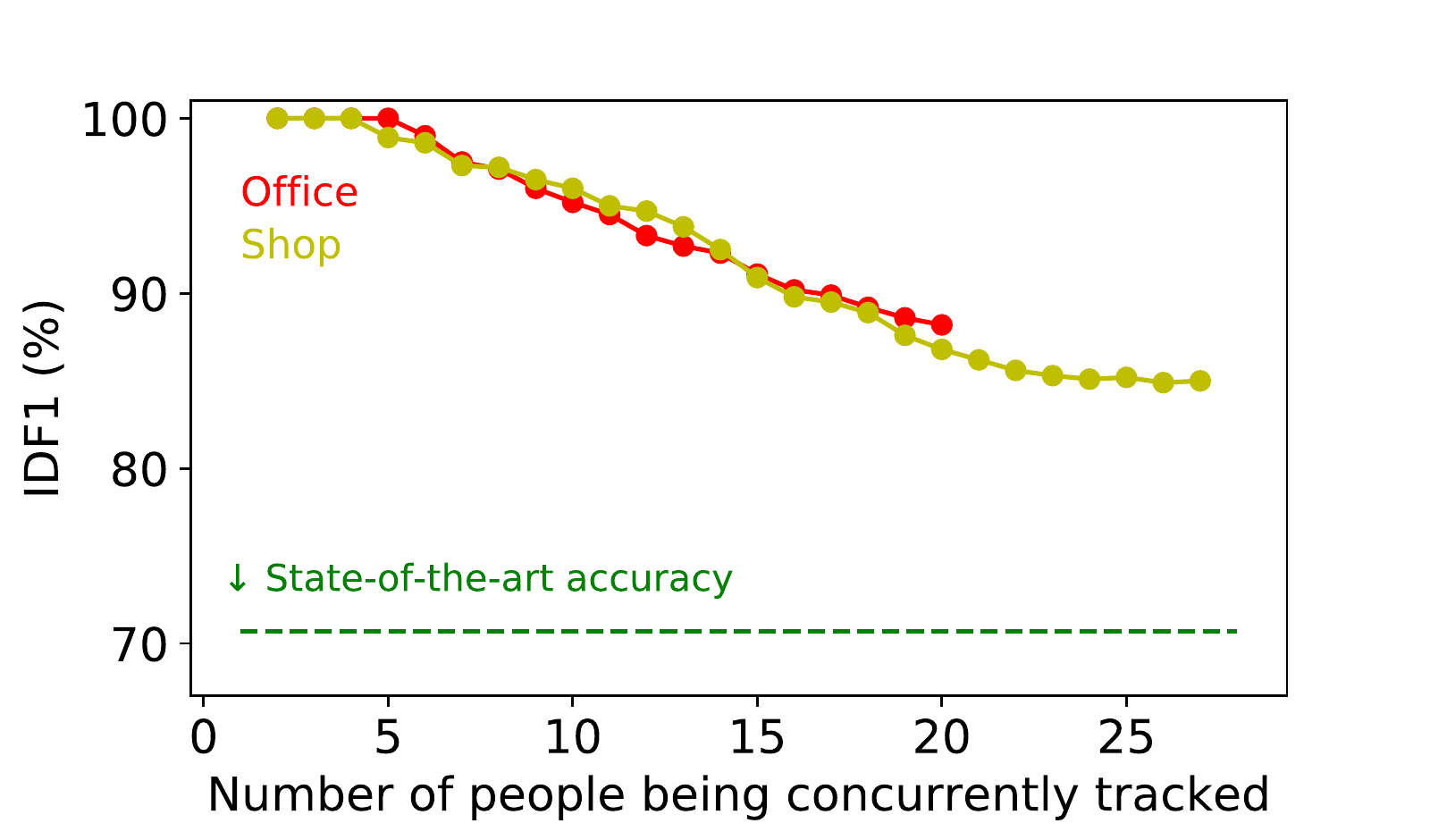}
\caption{\added{\emph{\small Relationship between the tracking accuracy and the number of concurrently tracked people.}}}
\label{fig:tar_diff_people}
\end{figure}

%% file: tex/related.tex
\chapter{Related Work}\label{chap:related}

Our prior work include several topics in the area of vision-based context sensing. In this chapter, we will review related work in each field. 

\section{Outdoor Localization and GPS Error Mitigation}

\paragraph{NLOS Mitigation} 
Prior work has explored NLOS mitigation. These all differ from Gnome along one or more of these dimensions: they either require specialized hardware, use simplistic or proprietary 3D models, or have not demonstrated scalability to smartphones. NLOS signals are known to be the major cause of GPS errors in urban canyons~\cite{GPSeBook, kaplan2005understanding}. Other work~\cite{ercek2006nlos} has shown that ray-tracing a single reflection generally works as well or better than ray-tracing multiple reflections. Early work~\cite{betaille2014enhance} uses the width of a street and the height of its buildings to build a simple model of an entire street as consisting of two reflective surfaces. This model deviates from reality in most modern downtown areas, where building heights vary significantly. To overcome this drawback, other work proposes specialized hardware to build models of reflective surfaces, including stereo fish-eye cameras~\cite{moreau2017fisheye}, LiDAR~\cite{mirrorSV}, or panoramic cameras~\cite{tay2013weighting}.

A long line of work has explored using proprietary 3D models to compute the path inflation, or simply to determine whether a satellite is within line of sight or not. One branch of this research uses 3D models to determine and filter out NLOS satellites~\cite{peyraud2013non, kumar2014identifying, drevelle2012igps}. However, as we show, in our dataset, nearly 90\% of the GPS readings see fewer than 4 satellites, and removing NLOS satellites would render those readings unusable. A complementary approach has explored, as Gnome does, correcting the NLOS path inflation. Closest to our work is the line of work~\cite{sahmoudi2014deep, miura2015gps, hsu20163d} that uses candidate positions like Gnome does, but estimates the ground truth position using path similarity. As we have shown earlier, this approach performs worse than ours. A second line  of work~\cite{groves2011shadow, wang2015smartphone, adjrad2015enhancing} assumes that NLOS satellites generally have lower carrier-to-noise density than LOS satellites. However, this may not hold in general and Gnome's evaluation shows that this approach also does not perform well.

\paragraph{Building height computation}
Several pieces of work have used techniques to build 3D models of buildings from a series of 2D images, using a technique called structure-from-motion (SfM~\cite{sfm}). Our approach uses 3D models made available from LiDAR devices mounted on Street View scanning vehicles. Other work has used complementary methods to obtain the height of buildings. Building height information is publicly available from government websites~\cite{la_data, ny_data} or 3rd party services~\cite{osm, property_shark}, and some work has explored building 3D models using images and building height information~\cite{guo2002snake, saito2016multiple}. However, these datasets have spotty coverage across the globe. Recognizing this, other work~\cite{soergel2009stereo, cellier2006building, brunnera2008building, brunner2008extraction} estimates building heights using Synthetic Aperture Radar (SAR) data generated with remote sensing technologies. This data also has uneven coverage. Finally, one line of work~\cite{shettigara1998height, shao2011shadow} analyzes the shape and size of building shadow to estimate building height. This approach needs the entire shadow to be visible on the ground with few obstructions. In downtown areas, the shadow of tall buildings fall on other buildings, and this approach cannot be used.

\paragraph{Complementary approaches to improving localization accuracy}
In recent years, the mobile computing community has explored several complementary ways to improve location accuracy: using the phone’s internal sensors to track the trajectory of a user~\cite{roy2014smartphone}; using cameras~\cite{mobieye} or fusing GPS readings with sensors, dead-reckoning, map matching, and landmarks to position vehicles~\cite{bo2013smartloc, jiang2015carloc}; using WiFi access point based localization~\cite{sen2012you, sen2013avoiding} as well as camera-based localization~\cite{xu2014topometric}; and crowdsourcing GPS readings~\cite{agadakos2017techu} to estimate the position of a target. Other work~\cite{hedgecock2013high, hedgecock2014accurate} has explored accurate differential GPS systems which require satellite signal correlation across large areas and don’t work well in downtown areas. GPS signals have also been used for indoor localization~\cite{nirjon2014coin}, and other work has explored improving trajectory estimation~\cite{wu2016only, wu2017clsters} by using map-matching to correct GPS readings. While map matching works well for streets, it is harder to use for pedestrians. In contrast to this body of work, Gnome attacks the fundamental problem in urban canyons: GPS error due to satellite signal reflections.

\section{Roadside Landmark Discovery and Localization}

Prior work in ubiquitous and pervasive computing deals with localizing objects within the environment, or humans. Some of these have explored localizing users using low-cost energy-efficient techniques on mobile devices: magnetic sensors \cite{MaLoc}, inertial and light sensors \cite{IndoorLoc}, inertial sensors together with wireless fingerprints \cite{calibration-free, graph-based}, RF signals \cite{transferring, LoCo, MobileRF}, mobility traces \cite{indoorALPS} and other activity fingerprints \cite{non-obstructive}. Other work has explored localizing a network of devices using low-cost RF powered cameras \cite{self-localizing}. Many of these techniques are largely complementary to ALPS, which relies on Street View images to localize common landmarks. Perhaps closest to our work is Argus \cite{Argus}, which complements WiFi fingerprints with visual cues from crowd-sourced photos to improve indoor localization. This work builds a 3-D model of an indoor setting using advanced computer vision techniques, and uses this to derive geometric constraints. By contrast, ALPS derives geometric constraints by detecting common landmarks using object detection techniques. Finally, several pieces of work have explored augmenting maps with place names and semantic meaning associated with places \cite{CheckInside, PinPlace}. ALPS derives place name to location mappings for common places with recognizable logos.

Computer vision research has considered variants of the following problem: given a GPS-tagged database of images, and a query image, how to estimate for the GPS position of the given image. This requires matching the image to the image(s) in the database, then deriving position from the geo-tags of the matched images. Work in this area has used Street View images~\cite{image_localization_eccv_10, image_localization_cvpr_12}, GIS databases~\cite{geospatial_localization_eccv_14}), images from Flickr~\cite{Crandall:2009}), or pre-uploaded images~\cite{ccode_tian}.
The general approach is to match features, such as SIFT in the query image with features in the database of images. ALPS is complementary to this line of work, since it focuses on enumerating common landmarks of a given type. Because these landmarks have distinctive structure, we are able to use object detectors, rather than feature matching.

Research has also considered another problem variant: given a set of images taken by a camera, finding the position of the camera itself. This line of work~\cite{agarwal15, liu12, ghosh2020localizing} attempts to match features in the images to a database of geo-tagged images, or to a 3-D model derived from the image database. This is the inverse of the our problem: given a set of geo-tagged images, to find the position of an object in these images. Finally, Baro~\textit{et al.}~\cite{baro09} propose efficient retrieval of images from an image database matching a given object. ALPS goes one step further and actually positions the common landmark.

\section{Cross-camera Person Re-identification and Tracking}
\paragraph{Tracking with other sensors}
There have been multiple tracking technologies available for person tracking.  \cite{brickstream,xovis,hella,shoppertrak} uses  stereo video system which utilizes camera pairs to sense 3D information of surroundings, however, the equipment are usually very expensive and hard to deploy.  \cite{irisys} uses thermal sensors to sense the existence and position of people, but its tracking accuracy can also be affected by occlusion, which makes it hard to distinguish people number. \cite{bea} uses laser and structured light to accurately infer the shape of people (usually in center-meter level), which makes them the most accurate solution for people counting. However, the short scanning range prevents the solution from continuous people tracking so other supporting solutions like cameras are needed. On the other hand,  Euclid Analytics~\cite{euclidanalytics} and Cisco Meraki ~\cite{ciscomeraki} have been relying on WiFi MAC Address to track the customer entry and exiting the stores. but this technology requires activation of customer WiFi and suffers from location accuracy. Swirl~\cite{swirl} and InMarket~\cite{inmarket} use pure BLE beacons to count customers, but the proximity based approach is far from the accuracy required to track people. Different from those approaches, TAR combines both vision and BLE proximity for not only actually tracking people in large scale, but also identifying them.

\paragraph{Single camera tracking.}
Tracking by detection has emerged as the leading strategy for multiple object tracking. 
Prior works use global optimization that processes the entire video batches to find object trajectories. For example, three popular frameworks, flow network formulations~\cite{zhang2008global,pirsiavash2011globally,berclaz2011multiple}, probabilistic graphical models~\cite{yang2012online,yang2012multi,andriyenko2012discrete,milan2013detection} and large temporal windows (e.g., \cite{fleuret2008multicamera,kuo2011does})
have been popular among them. However, due to the nature of batch processing, they are not applicable for real time processing where no future information is available.  

SORT~\cite{bewley2016simple} significantly improves the tracking speed by applying Kalman filter based tracking and achieves a fairly good accuracy. However, it performs  poorly during occlusions. DeepSORT~\cite{wojke2017simple} improves based on SORT by introducing the deep neural network feature of people. The algorithm works well for high quality video as deep features are more distinguishable. For the video with low light, the features become hard to distinguish, the performance degrades significantly. 
Different from SORT and DeepSORT, TAR takes all scope of information, including motion and appearance information, and train a more general convolutional neural network for feature extraction. These improvement increases the robustness against detection false negatives and occlusions.

\paragraph{Cross-camera people tracking.}
Traditional multi\-camera tracking algorithms like POM~\cite{fleuret2008multicamera} and KSP~\cite{ berclaz2011multiple} 
rely on homography transformation among different cameras, which requires  accurate camera angle and position measurement and camera overlapping.  However, for most cameras, the exact angle and position is not known, and scene overlapping is not satisfied due to various reasons. Moreover, those algorithms needs global trajectories for tracking, which is not suitable to online tracking. In contrast, TAR doesn't have these requirements, it utilizes various context information as well as Bluetooth signal to re-identify the objects across cameras.  

Several approaches~\cite{xu2017cross,
  nithin2017globality,solera2016tracking} track multiple people across
cameras but require overlapped cameras, which may be too restrictive
for most surveillance systems. In non-overlapping scenarios,
other approaches~\cite{ristani2018features, tesfaye2017multi,
  chen2017equalized} leverage the visual similarity between people's
traces in different cameras to match them. They also run a bipartite
matching algorithm globally to minimize the ID assignment error.
However, these approaches can only work offline for best performance
and are unsuitable for online processing purposes.

\section{Targeted Advertising and Cashier-free Shopping}

\paragraph{Commercial cashier-free systems}
We are not aware of published work on end-to-end design and evaluation of cashier-free shopping. Amazon Go was the first commercial solution for cashier-free shopping. Several other companies have deployed demo stores like Standard Cognition~\cite{std_cog}, Taobao~\cite{taobao}, and Bingobox~\cite{bingobox}. Amazon Go and Standard Cognition use computer vision to determine shopper-to-item association (\cite{amzn_no_rfid, go-geek, std_cog_forbes}). Amazon Go does not use RFID \cite{go-geek} but needs many ceiling-mounted cameras. Imagr\cite{imagr} uses a camera-equipped cart to recognize the items put into the cart by the user. Alibaba and Bingobox use RFID reader to scan all items held by the customer at a "checkout gate" (\cite{ali_tech, bingo_tech}). Grab incorporates many of these elements in its design, but uses a judicious combination of complementary sensors (vision, RFID, weight scales). Action detection is an alternative approach to identifying shopping actions. Existing state-of-the-art DNN-based solutions \cite{kalogeiton17iccv, sun2018optical} have not yet been trained for shopping actions, so their precision and recall in our setting is low.

\paragraph{Item detection and tracking}
Prior work has explored item identification using Google Glass~\cite{ha2014towards} but such devices are not widely deployed. RFID tag localization can be used for item tracking~\cite{shangguan2017, jiang2018orientation} but that line of work does not consider frequent tag movements, tag occlusion, or other adversarial actions. Vision-based object detectors~\cite{he2017mask, chen2018domain, redmon2018yolov3} can be used to detect items, but need to be trained for shopping items and can be ineffective under occlusions and poor lighting. Single-instance object detection scales better for training items but has low accuracy~\cite{karlinsky2017fine,held2016robust}.

\section{Complex Activity Detection}

Existing pipelines detect the person's location (bounding box) in the video,
extract representative features for the person's tube, and output the
probability of each action label. Recent
pipelines~\cite{ulutan2018actor, shou2018online, zhao2017temporal}
leverage DNNs to extract features from person tubes and predict
actions. Other work~\cite{gowsikhaa2012suspicious, liu2018tar} estimates human
behavior by detecting the head and hand positions and analyzing their
relative movement. Yet others analyze the moving trajectories of
objects near a person to predict the interaction between the person
and the object~\cite{mettes2017spatial, amor2016action}. By doing
this, the action detector can describe more complex actions. The above
approaches achieve high accuracy only with sufficient training
samples, which limits their applications for more complex activities
that involve multiple subjects and long duration.

Rather than analyzing a single actor's frames, other complementary
approaches~\cite{bagautdinov2017social, amer2014hirf} present their
solutions to detect group behavior such as ``walk in group'', ``stand
in queue'', and ``talk together''. The approach is to build a
monolithic model that takes input both the behavior feature of each
actor and the spatial-temporal relation (e.g. distance change), and
outputs the action label. The model could be an recurrent neural
network (\cite{bagautdinov2017social}) or a handcrafted
linear-programming model (\cite{amer2014hirf}). However, both models
require training videos to work properly because the models need
training, rendering these approaches unsuitable
for Caesar. A more recent approach~\cite{fu2019rekall} leverages face and object detection to extract meaningful objects in video clips, and then associate the objects with video's annotations using spatiotemporal relationship. The user could search for clips that contain specific object-annotation combinations. This system works offline and does not support mobile devices. It also lacks Caesar's extensible vocabulary which has object movement and atomic activities. 

Zero-shot action detection is closely related to Caesar, and targets near real-time detection
even when there are very few samples for training. Some
approaches~\cite{xu2016multi, qin2017zero} train a DNN-based feature
extractor with videos and labels. The feature extractor can generate
similar outputs for the actions that share similar attributes. When an unknown action tube arrives, these approaches
cluster it with existing labels, and evaluate its similarity with the
few positive samples. Another approach~\cite{gan2016recognizing} further decomposes an
action query sentence into meaningful keywords which have
corresponding features clusters, and waits for those clusters to be
matched together at runtime. However, these zero-shot detection
approaches suffer from limited vocabulary and low accuracy (<40\%).

\section{Wireless Camera Networks}
Wang \etal~\cite{wang2017networked} discuss an edge-computing based
approach in which a group of camera-equipped drones efficiently livestream a sporting event. However, they focus on
controlling drone movements and efficiently transmitting the video
frame over a shared wireless medium in order to maintain good quality
live video streaming with low end to end latency. Other
work~\cite{aziz2013energy} presents a new FPGA architecture and a
communication protocol for that architecture to  efficiently transmit images in a wireless camera network. San
Miguel \etal~\cite{sanmiguel2014self} present a vision of a smart
multi-camera network and the required optimization and properties, but
discuss no specific detection techniques.~\cite{hu2015data} proposes a real-time algorithm that schedules data retrieval from a wide range of wireless sensors given constraints.~\cite{rinner2012resource} proposes a method for re-configuring
the camera network over time based on the description of the camera
nodes, specifications of the area of interest and monitoring
activities, and a description of the analysis tasks. Finally,
MeerKats~\cite{boice2006meerkats} uses different image acquisition policies
with resource management and adaptive communication strategies.
No other prior work has focused on cross-camera complex activity
detection, as Caesar has.

\section{Scaling DNN Pipelines}
More and more applications rely
on a chain of DNNs running on edge clusters. This raises challenges
for scaling well with fixed number of computation resources. Recent
work~\cite{zhang2017live} addresses the problem by tuning different
performance settings (frame rate and resolution) for task queries to
maximize the server utilization while keeping the quality of service.
Downgrading frame rates and DNNs is not a good choice for Caesar because
both options will adversely impact accuracy.~\cite{hu2018olympian}
proposes a scheduler on top of TensorFlow Serving~\cite{tfserving} to
improve the GPU utilization with different DNNs on it. Caesar could
leverage such a model serving system, but is complementary to it.
Recent approaches~\cite{crankshaw2017clipper, lee2018pretzel} cache
the intermediate results to save GPU
cycles. Caesar goes one step further with lazily activating
the action DNN. \cite{crankshaw2017clipper} also
batches the input for higher per-image processing speed on GPU, which
Caesar also adopts to perform object detection on the mobile device.

%% file: tex/conclusion.tex
\chapter{Conclusions and Takeaways}\label{chap:conclusion}

In~\chapref{gnome}, we discuss Gnome as a practical and deployable method for correcting GPS errors resulting from non-line-of-sight satellites. Our approach uses publicly available 3D models, but augments them with the height of buildings estimated from panoramic images. We also develop a robust method to estimate the ground truth location from candidate positions, and an aggressive precomputation strategy and efficient search methods to enable our system to run efficiently entirely on a smartphone. Results from cities in North America, Europe, and Asia show 6-8m positioning error reductions over today’s highly optimized smartphone positioning systems. 

In~\chapref{alps}, we discuss ALPS, which achieves accurate, high coverage positioning of common landmarks at city-scales. It uses novel techniques for scaling (adaptive image retrieval) and accuracy (increasing confidence using zooming, disambiguating landmarks using clustering, and least-squares regression to deal with sensor error). ALPS discovers over 92\% of Subway restaurants in several large cities and over 95\% of hydrants in a single zip-code, while localizing 93\% of Subways and 87\% of hydrants with an error less than 10 meters. 

In~\chapref{caesar}, we discuss Caesar, a \emph{hybrid} multi-camera complex
activity detection system that combines traditional rule based
activity detection with DNN-based activity detection. Caesar supports an
extensible vocabulary of actions and spatiotemporal relationships and
users can specify complex activities using this vocabulary. To satisfy
the network bandwidth and low latency requirements for near real-time
activity detection with a set of non-overlapping (possibly wireless)
cameras, Caesar partitions activity detection between a camera and a
nearby edge cluster that lazily retrieves images and lazily invokes
DNNs. Through extensive evaluations on a public multi-camera dataset,
we show that Caesar can have high precision and recall rate with
accurate DNN models, while keeping the bandwidth and GPU usage an
orders of magnitude lower that a strawman solution that does not
incorporate its performance optimizations. Caesar also reduces the
energy consumption on the mobile nodes by 7$\times$.

In~\chapref{grab}, we discuss Grab, a cashier-free shopping system that uses a skeleton-based pose tracking DNN as a building block, but develops lightweight vision processing algorithms for shopper identification and tracking, and uses a probabilistic matching technique for associating shoppers with items they purchase. Grab achieves over 90\% precision and recall in a data set with up to 40\% adversarial actions, and its efficiency optimizations can reduce investment in computing infrastructure by up to 4$\times$. 

In~\chapref{tar}, we discuss TAR, a system that utilizes existing surveillance cameras and ubiquitous BLE signals to precisely identify and track people in retail stores. In TAR, we have first designed a single-camera tracking algorithm that accurately tracks people, and then extended it to the multi-camera scenario to recognize people across distributed cameras. TAR leverages BLE proximity information, cross-camera movement patterns, and single-camera tracking algorithm to achieve high accuracy of multi-camera multi-people tracking and identification. We have implemented and deployed TAR in two realistic retail-shop setting, and then conduct extensive experiments with more than 20 people. Our evaluation results demonstrated that TAR delivers high accuracy (90\%) and serves as a practical solution for people tracking and identification.

\section{Guidelines for Vision-Based Context Sensing Systems}

In this section, we will summarize all the lessons learned from prior work in designing efficient vision-based context sensing systems. At high level, a general process of designing such a system contains three steps: (1) Decide the data source and the hardware platforms used for the task; (2) Design the system architecture and major logic; (3) Design each algorithm and implement the system. 

\paragraph{Data Source and Platforms} It is costly to set up new sensors for collecting vision data at large scale with organized context. Our prior work show that leveraging existing sensing infrastructure and imagery data source is a good way to scale the data collection process. Specifically, people could use widely-deployed surveillance cameras for video collection, and use street-level imagery like Google Street View and Bing Streetside to collect images across the world. The surveillance cameras and street-level imagery usually contain location information. The images also contain accurate camera pose information like bearing and tilt. This could help many sensing tasks when it is analyzed together with the map information. For sensing platforms, we suggest that each available sensor should be evaluated in terms of its cost, sampling rate, data type, and data quality, and be selected to help cameras in sensing tasks. Cameras, as the major source of vision data, should be carefully analyzed and deployed because the camera placement, environment lighting, and image quality will strongly affect the end-to-end accuracy.

\paragraph{Architecture Design} Researchers should try to use a single context-extraction model in a system, instead of develop separated solutions for different parts of a task. For example, Grab buildings all person behavior detectors around the keypoint extraction model. Caesar uses and high-level action graph metric and supports it with various elements in the vocabulary. For the system's performance bottleneck, our prior work show that it is an effective way to spread the bottleneck workload among different hardware components: redundant CPU cycles could be used to handle tasks that originally require a GPU; the mobile devices and edge servers could split the computation according to their local hardware limitations.

\paragraph{Algorithm Choices} Complex neural-net-based vision algorithms are always applied for vision sensing tasks. However, they are heavy in processing time and their accuracy require sufficient training data. Our experience proves that, in many cases, researchers could apply much cheaper algorithms while keeping a good accuracy (e.g. joint tracking between key frames in Grab). Researcher could also use some task and data specific intuition to optimize the usage of expensive neural nets. For instance, Caesar lazily match action graphs to minimize GPU usage, and ALPS saves processing time with a two-stage candidate filtering metric.

\section{Potential Extensions}
For outdoor pedestrian localization, further effort could be used to optimize the energy usage and storage requirements of Gnome on Android phones, and test it more extensively in urban canyons in other major cities of the world.

For outdoor object positioning, potential future work includes documenting large cities with ALPS, and extending it to common landmarks that may be set back from the street yet visible in Street View, such as transmission or radio towers, and integrating Bing Streetside to increase coverage and accuracy.

For vision-only behavior sensing, the future work includes extending Caesar to re-identify more targets like cars and bags. Moreover, the activity graph could support extended vocabulary which includes time restriction, location, etc. 

For shopping behavior sensing (vision + sensor), much future work remains including obtaining results from longer-term deployments, improvements in robust sensing in the face of adversarial behavior, and exploration of cross-camera fusion to improve the accuracy of TAR and Grab even further.